\documentclass[
11pt, 
english, 
doublespacing, 
headsepline, 
]{MastersDoctoralThesis} 

\usepackage[utf8]{inputenc} 
\usepackage[T1]{fontenc} 
\usepackage{graphicx,kantlipsum,setspace}
\usepackage{mathpazo} 
\usepackage{multirow}
\usepackage{CJKutf8}
\usepackage[inkscapeformat=png]{svg}
\definecolor{DarkGreen}{RGB}{1,50,32}
  
\usepackage[backend=biber,style=authoryear,natbib=true,maxbibnames=10,uniquelist=false, url=false,doi=false,dashed=false]{biblatex} 
\usepackage[autostyle=true]{csquotes} 
\usepackage{soul}
\usepackage{datetime}
\usepackage{todonotes}
\usepackage{array}
\usepackage{subcaption}
\usepackage{textcomp}
\usepackage{amssymb}
\usepackage{pifont}
\newcommand{\cmark}{\ding{51}}%
\newcommand{\xmark}{\ding{55}}%

\newcommand{\add}[1]{#1}
\newcommand{\correction}[1]{#1}
\newcommand{\comment}[1]{}

\newcommand{\mytilde}{\raise.17ex\hbox{$\scriptstyle\mathtt{\sim}$}}
\newcommand{\webie}{\textsc{WebIE}}
\usepackage{silence}
\DeclareNameAlias{author}{given-family}

\DeclareFieldFormat{citehyperref}{%
  \DeclareFieldAlias{bibhyperref}{noformat}
  \bibhyperref{#1}}

\DeclareFieldFormat{textcitehyperref}{%
  \DeclareFieldAlias{bibhyperref}{noformat}
  \bibhyperref{%
    #1%
    \ifbool{cbx:parens}
      {\bibcloseparen\global\boolfalse{cbx:parens}}
      {}}}

\newbibmacro{string+doi}[1]{%
  \iffieldundef{doi}{\iffieldundef{url}{#1}{\href{\thefield{url}}{#1}}}{\href{http://dx.doi.org/\thefield{doi}}{#1}}}	
\DeclareFieldFormat*{title}{\usebibmacro{string+doi}{\mkbibemph{#1}}}

\savebibmacro{cite}
\savebibmacro{textcite}

\renewbibmacro*{cite}{%
  \printtext[citehyperref]{%
    \restorebibmacro{cite}%
    \usebibmacro{cite}}}

\renewbibmacro*{textcite}{%
  \ifboolexpr{
    ( not test {\iffieldundef{prenote}} and
      test {\ifnumequal{\value{citecount}}{1}} )
    or
    ( not test {\iffieldundef{postnote}} and
      test {\ifnumequal{\value{citecount}}{\value{citetotal}}} )
  }
    {\DeclareFieldAlias{textcitehyperref}{noformat}}
    {}%
  \printtext[textcitehyperref]{%
    \restorebibmacro{textcite}%
    \usebibmacro{textcite}}}

\thesistitle{Towards Knowledge-Grounded Natural Language Understanding and Generation} 
\supervisor{Dr Tillman \textsc{Weyde} \\ Dr Pranava \textsc{Madhyastha}} 
\examiner{} 
\degree{Doctor of Philosophy} 
\author{Chenxi \textsc{Whitehouse}} 
\addresses{} 

\subject{Biological Sciences} 
\university{City, University of London} 
\department{Department or Computer Science} 
\group{School of Science and Technology} 
\faculty{School of Science and Technology} 

\AtBeginDocument{
\hypersetup{pdftitle=\ttitle} 
\hypersetup{pdfauthor=\authorname} 
\hypersetup{pdfkeywords=\keywordnames} 
}

\begin{document}

\frontmatter 

\pagestyle{plain} 


\begin{titlepage}
\begin{center}

\begin{figure}[ht]
\centering
  \includegraphics[width=0.3\linewidth]{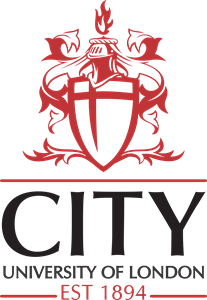}
\end{figure}

{\scshape\LARGE \univname\par}\vspace{3ex} 
\textsc{\Large Doctoral Thesis}\\[0.5cm] 

\HRule \\[0.4cm] 
{\LARGE \bfseries \ttitle\par}\vspace{0.3cm} 
\HRule \\[1.5cm] 
 
\begin{minipage}[t]{0.4\textwidth}
\begin{flushleft} \large
\emph{Author:}\\
{\authorname} 
\end{flushleft}
\end{minipage}
\begin{minipage}[t]{0.4\textwidth}
\begin{flushright} \large
\emph{Supervisors:} \\
{\supname} 
\end{flushright}
\end{minipage}\\[1cm]

\large \groupname\\\deptname\\[1cm] 

{\large  \monthname[\the\month], \the\year}\\[4cm]

\vfill
\end{center}
\end{titlepage}

\begin{abstract}
\addchaptertocentry{\abstractname} 

This thesis investigates how natural language understanding and generation with transformer models can benefit from grounding the models with knowledge representations. Currently, the most prevailing paradigm for training language models is 
through pre-training on abundant raw text data and fine-tuning on downstream tasks. Although language models continue to advance, especially the recent trend of Large Language Models (LLMs) such as ChatGPT, there seem to be limits to what can be achieved with text data alone and it is desirable to study the impact of applying and integrating rich forms of knowledge representation to improve model performance.

The most widely used form of knowledge for language modelling is structured knowledge in the form of triples consisting of entities and their relationships, often in English. 
This thesis explores beyond this conventional approach and aims to address several key questions: 
\begin{itemize}

  \item  Can knowledge of entities extend its benefits beyond entity-centric tasks such as entity linking? 
    \item How can we faithfully and effectively extract such structured knowledge from raw text, especially noisy web text? 
      \item How do other types of knowledge, beyond structured knowledge, contribute to improving NLP tasks?
    \end{itemize}

To this end, we study various tasks including multimodal and multilingual applications and consider a wide spectrum of knowledge, structured knowledge that is typically represented as triples with entities and their relations, and unstructured knowledge including parametric knowledge preserved in language models, knowledge distilled from Large Language Models, etc.

\subsubsection*{Knowledge-grounding with structured knowledge.}

We begin by investigating the integration of structured knowledge into language models. Knowledge of entities has shown benefits for entity-centric tasks such as entity linking and relation extraction, however, most studies have been limited to monolingual settings. We expand knowledge-grounding with structured knowledge, specifically entities, in two directions of research. 

Firstly, we study whether knowledge of entities can benefit real-world fake news detection. We hypothesise that the world knowledge embedded in entities can contribute to assessing the truthfulness of news statements. Evaluation of various knowledge integration approaches on distinct datasets reveals that knowledge-enhanced language models improve fake news detection when incorporated with a relevant and up-to-date knowledge base. 

The second direction expands beyond English and focuses on multilingual entities. We introduce \texttt{EntityCS}, where we first construct a code-switched (CS) training corpus from Wikipedia, by switching entities in English to their counterparts in other languages. Then we intermediate-train a pretrained multilingual model on this corpus for joint masked language modelling and entity prediction. Subsequent fine-tuning of the model on entity-centric downstream tasks consistently improves zero-shot cross-lingual transferability, demonstrating the benefit of integrating knowledge of multilingual entities.

\subsubsection*{Extracting structured knowledge from web text.}
We continue by studying effective, faithful, and robust extraction of structured knowledge from web text. Most existing information extraction (IE) datasets are constrained to Wikipedia articles, and models trained on such a rich factual text corpus show poor performance when applied to more noisy text from the web. To address these challenges, we introduce \texttt{WebIE}, a new dataset that takes raw sentences as input and structured triples as output. \texttt{WebIE} emphasises data quality by introducing negative examples and undergoing rigorous human annotation. We also propose faithful generative information extraction pipelines.  Our experiments with entity planning training and prefix-trie decoding show improvement in accurately extracting knowledge on the web. 

\subsubsection*{Knowledge-grounding beyond structured knowledge.}
To address our last research question, we study the impact of a broader sense of knowledge, including parametric knowledge (knowledge stored in the latent parameters of the models) derived from a model's self-explanations and knowledge distilled from LLMs via data augmentation.
 
We expand the application to multimodal language models and study knowledge-intensive visual question answering (VQA). We introduce a unified approach for fine-tuning multimodal models for jointly generating answers and explanations. Our experiments demonstrate enhancement in both answer accuracy and explanation quality. 

Lastly, as LLMs continue to advance in performance and size, we explore the utility of distilling commonsense knowledge from general-purpose LLMs to benefit smaller task-specific models. We prompt various LLMs to generate diverse examples on several challenging and scarce multilingual commonsense datasets. This augmentation shows consistent enhancements on fine-tuned smaller models, shedding light on data augmentation strategies for scenarios with limited training data.

In summary, this thesis explores the role of knowledge grounding in natural language understanding and generation across a broad spectrum of tasks. We found that incorporating relevant and up-to-date knowledge of entities benefits fake news detection, and entity-focused code-switching significantly enhances zero-shot cross-lingual transfer on entity-centric tasks. In terms of effective and faithful approaches to extracting structured knowledge, our study found that integrating negative examples and training with entity planning significantly improves performance. Additionally, we established that other general forms of knowledge, such as parametric and distilled knowledge, enhance multimodal and multilingual knowledge-intensive tasks. This research shows the tangible benefits of diverse knowledge integration and motivates further exploration in this direction.

\end{abstract}


\begin{acknowledgements}
\addchaptertocentry{\acknowledgementname} 

I want to express my sincere gratitude to my PhD supervisors for their consistent support and guidance throughout my doctoral journey. Special thanks to Dr Tillman Weyde and Dr Nikos Komninos for selecting me for the PhD studentship.
I am genuinely thankful to Dr Pranava Madhyastha, who joined my PhD supervision later on and inspired me to explore and gain a deeper understanding of the field of NLP.

I deeply appreciate all those who supported me during my research internships, which were an integral part of my PhD studies. It was an honour to intern at some of the world's leading research institutes and learn from experienced researchers. Warm thanks to Dr Fenia Christopoulou, my mentor during my first research internship at Huawei, whose dedication and hard work have consistently motivated me. I am grateful to Dr Clara Vania at Amazon for her encouragement and genuine support for my future endeavours.
Furthermore, I would also like to extend my gratitude to Dr Fantine Huot and Dr Jasmijn Bastings for the opportunity to intern at Google DeepMind, a dream come true, and to Dr Mostafa Dehghani and Prof. Mirella Lapata for their invaluable assistance and guidance throughout the journey.

I extend a warm thank you to Dr Alham Fikri Aji for his support, from Amazon to MBZUAI, and for encouraging me to expand my research collaboration.

Lastly, my sincere thanks go to Jaime, who has been a steadfast source of support in my life, believing in me and standing by my side through both the best and the most challenging times.

\end{acknowledgements}


\tableofcontents 

\listoffigures 

\listoftables 

\dedicatory{Dedicated to my parents and Jaime.} 


\mainmatter 

\pagestyle{thesis} 


\chapter{Introduction}

\label{Introduction} 

\begin{figure}[ht!]
\centering
\includegraphics[width=\linewidth]{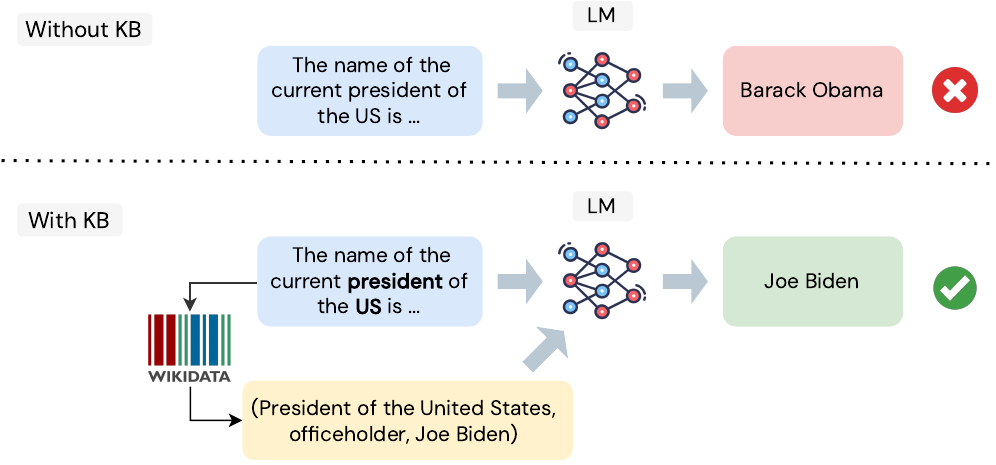}
    \caption{\add{Illustration of the role of knowledge bases in text generation: When the parametric knowledge stored in the language model becomes outdated (such as with GPT-2, trained in 2017), it is important to incorporate up-to-date knowledge bases for deriving correct answers, especially when using the frozen model at inference time.}}
\label{fig:kb-help}
\end{figure}

Natural Language Processing (NLP) has witnessed significant progress in recent years, particularly since the emergence of transformer-based~\citep{attention-2017} language models.
Most representative transformer models include BERT~\citep{devlin-etal-2019-bert}, RoBERTa~\citep{Liu2019RoBERTaAR}, BART~\citep{lewis-etal-2020-bart}, T5~\citep{2020t5}, to name a few, which are typically pre-trained on raw, unlabeled textual inputs with objectives such as Masked Language Modelling~\citep{devlin-etal-2019-bert}, and then fine-tuned on labelled task-specific downstream tasks.
Although this pre-training and fine-tuning paradigm has proven effective and achieved new start-of-the-art performance on various NLP tasks~\citep{devlin-etal-2019-bert}, it presents limitations in adapting to the ever-evolving world knowledge~\citep{zhang-etal-2019-ernie, Liu_Zhou_Zhao_Wang_Ju_Deng_Wang_2020}. \comment{[Chapter 1 - Correction point 1/1]~}\add{\autoref{fig:kb-help} shows an example where the knowledge base is crucial for the model generation when the parametric knowledge stored within the model parameters becomes stale.} 

To address this, a growing need has emerged to explore the integration of diverse knowledge representations and grounding into the language models to enhance their capabilities in tasks particularly requiring intensive knowledge~\citep{NEURIPS2020_6b493230, izacard2022few} or in-depth context understanding~\citep{chang-etal-2020-incorporating, bauer-bansal-2021-identify}.

The most commonly used knowledge for such integration is structured knowledge, typically represented in knowledge bases consisting of entities and their relationships~\citep{CHEN2020112948, zhu-etal-2022-knowledge}. Knowledge bases, such as Wikidata,\footnote{\url{https://www.wikidata.org/wiki/Wikidata:Main_Page}} provide a rich source of human-curated factual knowledge, which can be used as a complement to unlabelled raw text. 
Considerable research has focused on incorporating structured knowledge, especially entity-based knowledge (i.e., knowledge of entities), into language models \citep{zhang-etal-2019-ernie, peters-etal-2019-knowledge, wang-etal-2021-k, wang-etal-2021-kepler}. Although these prior works have proven the advantages of knowledge integration in specific tasks such as entity linking, named entity recognition, etc., the potential of a wider spectrum of knowledge representations (e.g., unstructured knowledge) and applications (e.g., beyond entity-centric tasks) remains relatively under-explored.
This motivates us to conduct a more extensive investigation into the following research questions.

\section{Research Questions}
This thesis aims to address the research questions outlined in the following three aspects.
\subsection*{Expanding the Utilisation of Structured Knowledge}
The first question we study is: \textit{Can the advantages of entity-based knowledge be extended to multilingual setups and beyond entity-centric tasks?}

Prior work such as KnowBert~\citep{peters-etal-2019-knowledge}, ERNIE~\citep{zhang-etal-2019-ernie}, K-BERT~\citep{Liu_Zhou_Zhao_Wang_Ju_Deng_Wang_2020}, etc. have demonstrated the success of integrating knowledge of entities into language model pre-training. However, their predominant focuses are limited to monolingual language models (specifically, English only), and downstream tasks that are exclusively related to entities, such as entity linking and named entity recognition.
This thesis broadens the application scope of structured knowledge in two ways: (i) we evaluate the effectiveness of knowledge-enhanced language models on the more complex task of fake news detection (Chapter~\ref{FakeNews}), and (ii) we propose 
the use of multilingual entity knowledge in an entity-centric code-switching method, \textsc{EntityCS}, to improve cross-lingual transferability on low-resource languages (Chapter~\ref{EntityCS}).

\subsection*{Effective Extraction of Structured Knowledge}
We continue to explore: \textit{How can structured knowledge be extracted effectively and accurately from diverse sources, particularly noisy web text?}

Prior work introduced above as well as our first studies~\citep{whitehouse2022evaluation, whitehouse-etal-2022-entitycs} demonstrate the benefit of structured knowledge in various NLP tasks. Yet the challenge lies in how to effectively obtain such knowledge, which is essential to adding new facts, keeping the relevance and accuracy of existing facts, etc. Hence, there is a compelling need to develop models capable of automatically extracting structured knowledge from vast textual sources~\citep{yao-etal-2019-docred, yao-etal-2021-codred, ormandi2021webred}, which requires high-quality information extraction datasets.
The majority of existing knowledge or information extraction datasets that are used to train such models are constructed based on clean and fact-rich resources such as Wikipedia~\citep{trisedya-etal-2019-neural, huguet-cabot-navigli-2021-rebel-relation, seganti-etal-2021-multilingual}. As a result, models trained on them may encounter difficulties when applied to noisy web text, as revealed in our preliminary studies.
Therefore, this thesis addresses the critical challenge by proposing a more generalised dataset, \textsc{WebIE}, as well as modelling approaches that are suitable for knowledge extraction from the web text (Chapter \ref{WebIE}).

\subsection*{Exploration of Diverse Knowledge Forms}
The final question explored in this thesis focuses on: \textit{How can broader forms of unstructured knowledge contribute to enhancing language models?}

The preceding two directions focus on the investigation of structured knowledge. However, there exists a diverse range of unstructured knowledge representations, such as those found in raw text~\citep{zhu-etal-2022-knowledge, pan2023large}, knowledge derived from the language models after pre-training (i.e., stored in the latent parameters) ~\citep{petroni-etal-2019-language, zhang-etal-2020-pretrain, ijcai2022p0318,neeman-etal-2023-disentqa}. In this part of the thesis, we aim to broaden the scope beyond structured knowledge and explore unstructured knowledge, including parametric knowledge \citep{neeman-etal-2023-disentqa} embedded within language models, through grounded answer and explanation generation in knowledge-intensive Visual Question Answering (VQA) (Chapter \ref{VQA}), and knowledge distilled from Large Language Models via data augmentation (Chapter \ref{LLM}). Emphasis is placed on the application of these studies in multimodal and multilingual use cases.

\section{Structure of the Thesis}

The thesis begins with an introduction that outlines the background of transformers, knowledge in NLP, and knowledge-enhanced language models in Chapter~\ref{Background}, which sets out the foundation for all the subsequent studies conducted. 
The main body of the thesis is structured around five chapters, each focusing on a published conference paper.

The first focus is on knowledge-grounding with entities and structured knowledge. Chapter~\ref{FakeNews} concentrates on knowledge-enhanced language models for the application of fake news detection. 
We evaluate the effectiveness of various approaches that incorporate entity knowledge in language models on fake news detection accuracy, covering news datasets in different domains (i.e., politics and COVID-19) with different linguistic features, where we find the benefit of incorporating knowledge that is relevant and up-to-date.

Chapter~\ref{EntityCS} focuses on studying the effectiveness of multilingual entity knowledge in zero-shot cross-lingual transfer. We propose \textsc{EntityCS}, where we utilise entity-centric code-switching with Wikipedia and Wikidata knowledge base to intermediate train a cross-lingual pre-trained language model. Subsequent fine-tuned models on entity-centric tasks, e.g., named entity recognition, slot filling, and fact retrieval, demonstrate strong performance compared to the baseline.

Chapter~\ref{WebIE} addresses the extraction of faithful and robust structured knowledge from the web domain. Information extraction is essential for knowledge base construction and population, which also provides reliable knowledge sources for training knowledge-enhanced models. 
In this chapter, we introduce the collection of a new dataset, \textsc{WebIE}, via crowdsourcing and propose various joint training strategies for mitigating the hallucination issues in generative information extraction.

The subsequent two chapters explore knowledge-grounding beyond structured knowledge. Chapter~\ref{VQA} expands the application to multimodal language models and studies knowledge-intensive VQA, utilising parametric knowledge derived from a model's self-explanations. We propose a novel direction of jointly generating answers and explanations with multimodal generative models making use of artificial special tokens, enabling the distinction of different tasks and datasets while learning shared semantics among different tasks.

As LLMs become ever more powerful, in Chapter~\ref{LLM}, we explore the utility of distilling commonsense knowledge from general-purpose LLMs to benefit smaller task-specific models. 
We prompt various LLMs to generate diverse examples on several challenging and scarce multilingual commonsense datasets. This augmentation shows consistent enhancements on fine-tuned smaller models, compared to those trained on limited human-created data.

In Chapter~\ref{Conclusion}, we provide a summary of the thesis with insights and key takeaways, as well as motivations for promising future research directions.

\section{Publications}
The main work proposed in this thesis has been published as conference papers. Chapter \ref{FakeNews}-\ref{LLM} contain existing, improved or extended results to the following publications:

\begin{enumerate}
\item Evaluation of Fake News Detection with Knowledge-Enhanced Language Models~\citep{whitehouse2022evaluation}
\begin{itemize}
\item Main content included in Chapter~\ref{FakeNews}
\item  \textit{Proceedings of the Sixteenth International AAAI Conference on Web and Social Media (\textbf{AAAI-ICWSM} 2022)}
\end{itemize}

\item EntityCS: Improving Zero-Shot Cross-lingual Transfer with Entity-Centric Code Switching~\citep{whitehouse-etal-2022-entitycs}
\begin{itemize}
\item Main content included in Chapter~\ref{EntityCS}
\item  \textit{Findings of the Association for Computational Linguistics: \textbf{EMNLP} 2022}
\item Work conducted as Research Intern at Huawei Noah’s Ark Lab, London, United Kingdom, 2021
\end{itemize}

\item Towards a Unified Model for Generating Answers and Explanations in Visual Question Answering~\citep{whitehouse-etal-2023-towards}
\begin{itemize}
\item Main content included in Chapter~\ref{VQA}
\item  \textit{Findings of the Association for Computational Linguistics: \textbf{EACL} 2023}
\end{itemize}

\item WebIE: Faithful and Robust Information Extraction on the Web~\citep{whitehouse-etal-2023-webie}
\begin{itemize}
\item Main content included in Chapter~\ref{WebIE}
\item  \textit{Proceedings of the 61st Annual Meeting of the Association for Computational Linguistics (\textbf{ACL} 2023)}
\item Work conducted as Research Intern at Amazon Alexa AI, Cambridge, United Kingdom, 2022
\end{itemize}

\item LLM-powered Data Augmentation for Enhanced Cross-lingual Performance~
\\\citep{whitehouse-etal-2023-llm}
\begin{itemize}
\item Main content included in Chapter~\ref{LLM}
\item  \textit{Proceedings of the 2023 Conference on Empirical Methods in Natural Language Processing (\textbf{EMNLP} 2023)}
\item Work conducted in collaboration with MBZUAI and Microsoft
\end{itemize}

\subsection*{Other Publications during the PhD}

Throughout the course of my doctoral studies, there are several additional papers that are not incorporated in the thesis, from my collaboration with researchers from Google DeepMind and MBZUAI.

\item Parameter-Efficient Multilingual Summarization: An Empirical Study~\citep{whitehouse2023lora}
\begin{itemize}
\item  \textit{Findings of the Association for Computational Linguistics: \textbf{NAACL} 2024}
\item Work conducted as Research Intern at Google DeepMind, Amsterdam, Netherlands, 2023
\end{itemize}

\item M4: Multi-generator, Multi-domain, and Multi-lingual Black-Box Machine-Generated Text Detection~\citep{wang-etal-2024-m4}
\begin{itemize}
\item  \textit{Proceedings of the 18th Conference of the European Chapter of the Association for Computational Linguistics (\textbf{EACL} 2024)}
\item Work conducted in collaboration with MBZUAI, 2023
\end{itemize}

\item Multitask Multilingual Model Adaptation with Featurised Low-Rank Mixtures~\citep{lin2024lora}
\begin{itemize}
\item  \textit{Preprint}
\item Work conducted as Research Intern at Google DeepMind, Amsterdam, Netherlands, 2023
\end{itemize}

\end{enumerate}

\chapter{Background: Transformer, Knowledge, Knowledge-Enhanced PLMs}

\label{Background} 

This section provides a background overview of three fundamental aspects: (i) the Transformer. All the modelling approaches in this thesis are based on the transformer architecture, (ii) knowledge sources and knowledge representation, and (iii) knowledge-enhanced Pre-trained Language Models (PLMs). These topics collectively establish the foundation for the core focus of this thesis: knowledge-grounding in language models.

\section{The Transformer}
\label{sec:transformer}
The transformer architecture, proposed by \citet{attention-2017}, has revolutionised the NLP field. 
Compared to its predecessors, including RNN and LSTM \citep{hochreiter1997long}, transformers distinguish themselves in the following key aspects. 
Firstly, transformers employ an attention mechanism to alleviate the limitations associated with long-term dependencies. 
Secondly, unlike previous sequential models, transformers can process input sequences in parallel, making them more efficient.
Transformers are also highly scalable. The stacked architecture can be easily scaled to be optimised for larger datasets and more complex tasks by adding more layers or units, which is a key factor in the development of recent powerful LLMs \citep{attention-2017, devlin-etal-2019-bert, brown2020language, anil2023palm}.

Thanks to their advantages, transformers have become the cornerstone of the current NLP field. All the models included in this thesis are based on transformers. We first provide an overview of the transformer architecture and then introduce several representative model categories, including encoder-only models, decoder-only models, and sequence-to-sequence models. Finally, we briefly touch upon multimodal vision-language models.

\subsection{Transformer Architecture}

\begin{figure}[th!]
\centering
\includegraphics[width=0.7\linewidth]{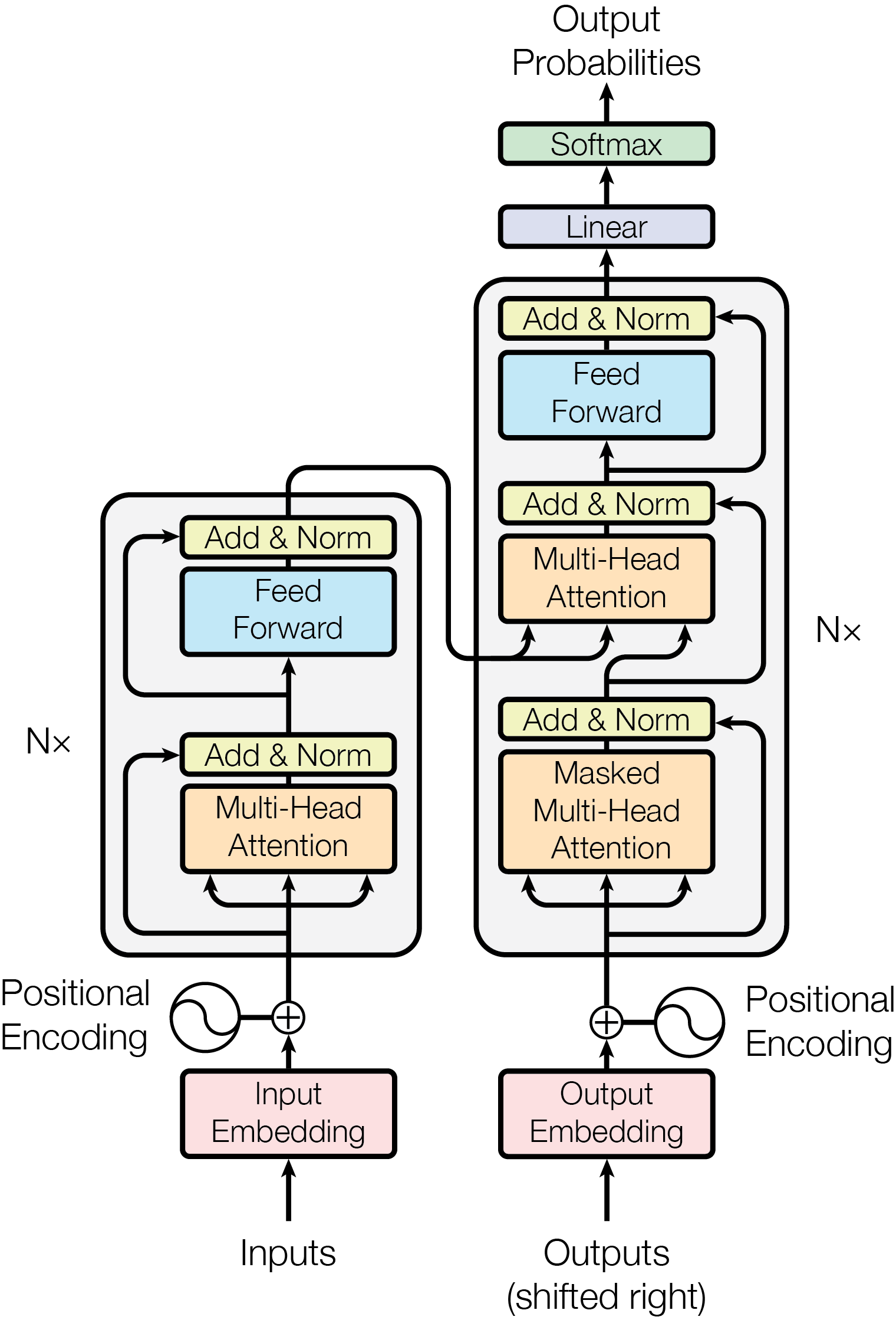}
    \caption{The Transformer Model Architecture. It is composed of a stack of 
\(N\) encoders (on the left) and \(N\) decoders (on the right). Each sublayer comprises a multi-head attention layer and a feed-forward layer. Layer Normalisation and Residual connections are applied after each layer. Source: \citet{attention-2017}.}
\label{fig:trasnsformer}
\end{figure}

The originally proposed transformer architecture, as illustrated in \autoref{fig:trasnsformer}, follows a sequence-to-sequence structure, which takes a sequence as input and outputs a new sequence. Its key innovation is the self-attention mechanism, which allows the model to weigh the relevance of each word in a sequence to every other word, enabling the capture of long-range dependencies and context in text data.

At its core, the transformer architecture contains a stack of encoder layers (left) and decoder layers (right).
Both the encoder stack and decoder stack have their corresponding Embedding layers to process the input, represented by the \texttt{Input Embedding} and \texttt{Output Embedding} layers in \autoref{fig:trasnsformer}.
Within each layer, \textit{Multi-Head Attention} is applied independently to all subwords or tokens in the input sequence, and the outputs are passed through \texttt{Feed Forward} neural network layers.
In the end, the output of the last layer of the decoder stack is fed to a \texttt{Linear} layer and a \texttt{Softmax} layer to obtain the final output sequence.

\subsubsection{Attention Mechanism}
Given a set of key-value pair vectors and a query vector, attention is a technique to compute a weighted sum of the values, dependent on the association of the query and the corresponding keys, i.e., the query attends to the values \citep{attention-2017}.

The core innovation of the Transformer is the incorporation of the attention mechanism. Unlike traditional models that linearly process sequences, the self-attention mechanism allows the model to weigh the significance of different words in a sequence when making predictions for a particular word. This mechanism enables the model to consider global context and dependencies within the input window.

\begin{figure}[th!]
\begin{minipage}[t]{0.5\textwidth}
  \centering
  Scaled Dot-Product Attention \\
  \vspace{0.7cm}
  \includegraphics[scale=0.75]{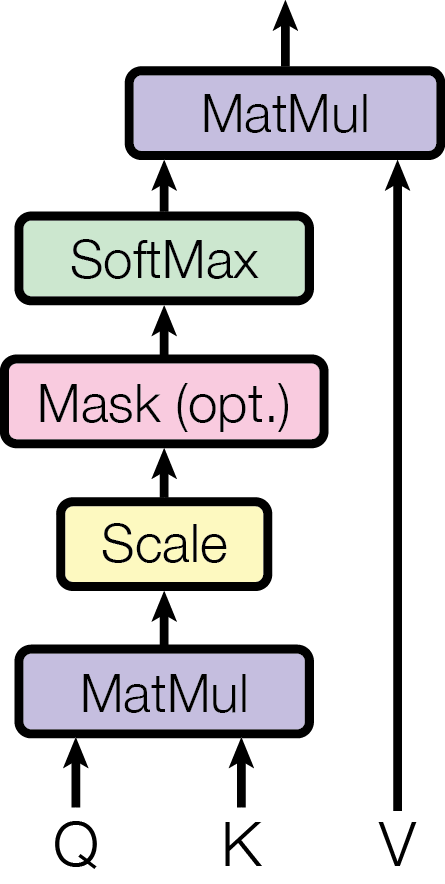} 
\end{minipage}
\begin{minipage}[t]{0.5\textwidth}
  \centering 
  Multi-Head Attention \\
  \vspace{0.3cm}
  \includegraphics[scale=0.75]{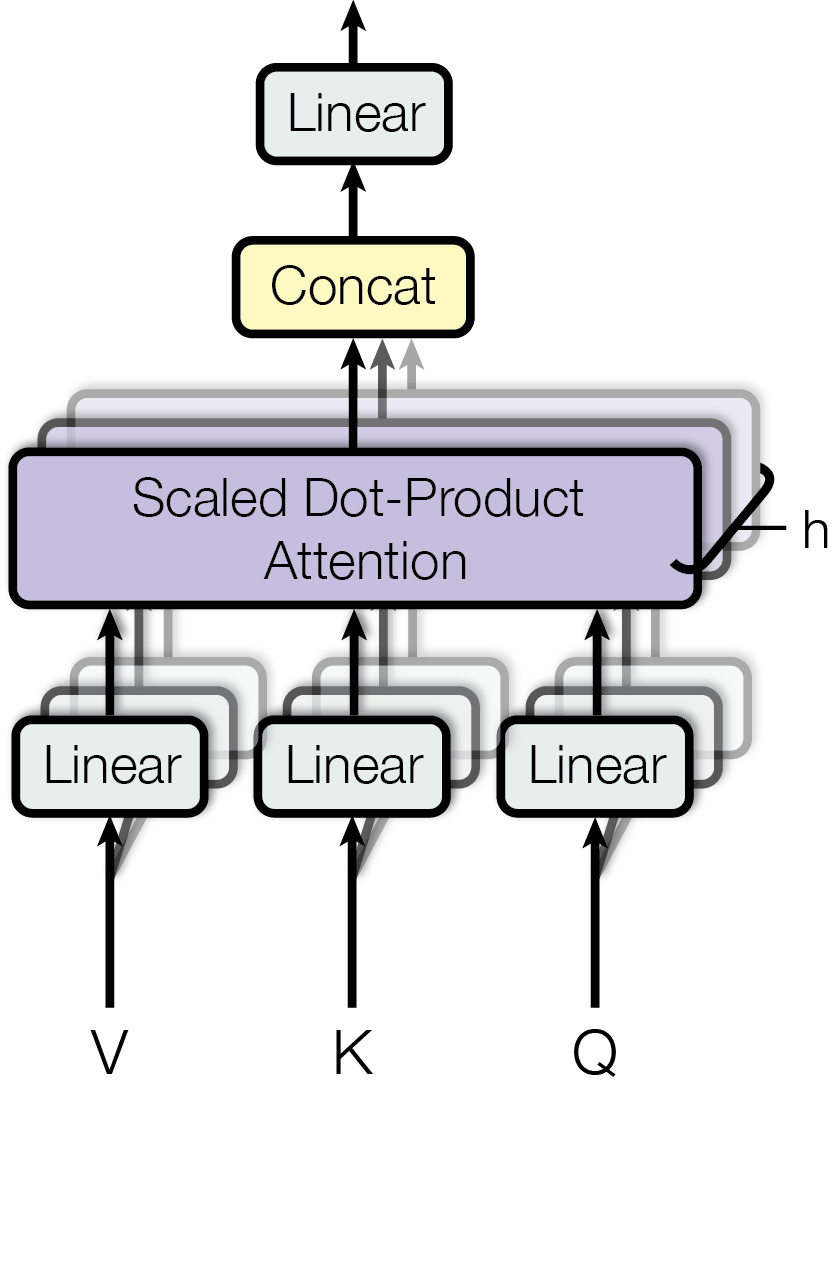}  
\end{minipage}

 \vspace{-1.2cm}
    \caption{Illustration of Multi-Head Attention. Input is passed through learnt \textit{query} (\(Q\)), \textit{key} (\(K\)), and \textit{value} (\(V\)) matrices to compute attention scores with respect to other tokens in the sequence. Multi-head attention (on the right) enables the calculation of attention individually and projects the input sequence into different subspaces. The outputs of all the heads are then concatenated and linearly transformed to produce the final output. Source: \citet{attention-2017}.}
  \label{fig:attention}
\end{figure}

\subsubsection*{Self-Attention}
\citet{attention-2017} introduce the \texttt{Scaled Dot-Product Attention} mechanism, illustrated in the left plot of \autoref{fig:attention}. 
Input vectors (input text embedding + positional embedding) are passed through three trainable matrices, \(Q\), \(K\), and \(V\), to compute the Query vector, Key vector, and Value vector. These vectors share the same dimensions as the input vector.

The initial step involves calculating a score that signifies the attention of one token, token\_\(i\), in a sequence to another token, token\_\(j\). This score is determined by the dot product of the Query vector of token\_\(i\) and the Key vector of token\_\(j\).
To prevent the dot products from becoming too large as the dimensionality increases, which can lead to gradients becoming too small during backpropagation, a scaling factor of \(\sqrt{d_k}\) is applied to the dot product, with \(d_k\) representing the dimension of \(K\), the Key matrix.

After computing the score for each token (including token\_\(i\) itself), the results are passed through a \texttt{Softmax} operation, which normalises the scores to positive numbers and a cumulative sum of 1. The normalised score at each position concerning the query token determines the level of attention that the query token allocates to each position. Finally, the output of the attention layer is computed as the sum of the Value vectors of each position weighted by the normalised scores.

In practice, a set of input vectors is packed into a matrix and simultaneously multiplied by the \(Q\),  \(K\) and \(V\) matrices. Therefore the attention mechanism can be represented as:
\[Attention(Q,K,V) = softmax(QK^T/\sqrt{d_k})V\]

\texttt{Self-attention} denotes applying the attention mechanism to tokens within the same sequence, and is employed in both encoder and decoder layers. The only distinction in the decoder layer lies in the masked attention mechanism, which only permits a token to attend to all the tokens that precede it but not those that follow.
This prevents the leakage of information from future tokens during training.

\subsubsection*{Cross-Attention}
In addition to the \texttt{self-attention} layers, \texttt{cross-attention} is also utilised in the transformer.
Here, Query vectors are from the decoder, while both Key and Value vectors are derived from the encoder output. Cross-attention enables each position in the decoder of a sequence-to-sequence model to effectively attend to all positions in the input sequence.

\subsubsection*{Multi-Head Attention} 
To enhance the expressive power of self-attention, the transformer uses multiple attention heads, i.e., \texttt{Multi-Head Attention}.
Instead of applying one attention operation with the entire Query, Key, Value vectors, \citet{attention-2017} find that dividing the vectors into multiple chunks with different learned linear projections achieves more efficient use of the model’s capacity: e.g., for English-to-German translation, the performance of the model with eight heads is almost 1 BLEU point higher than that of a model of the same size with single-head
attention \citep{attention-2017}.
As illustrated in the right plot of \autoref{fig:attention}, each head focuses on different parts of the input sequence, capturing diverse information. Afterwards, the outputs from multiple heads are concatenated and linearly transformed.

\subsubsection{Other Components in the Transformer}

Apart from the multi-head attention mechanism, the following components play important roles in the transformer architecture.

\subsubsection*{Input Embedding and Position Embedding}
The transformer utilises learned embeddings to represent the input. Input sequences are first tokenised into tokens or subwords and then converted into high-dimensional vectors through the input embedding layers.
The embedding layers are linear transformation layers at the bottom of both the encoder and the decoder stack (see \autoref{fig:trasnsformer}).

As the transformer does not inherently capture the sequential order of tokens, positional encoding is incorporated into the input embeddings to convey information about token positions. \citet{attention-2017} employ sine and cosine functions of varying frequencies to represent position information, with the positional embeddings typically sharing the same dimensions as the input embeddings.
In addition to absolute position embeddings, as seen in \citet{attention-2017}, various positional embeddings such as relative positional embeddings \citep{shaw-etal-2018-self} and fully learnable embeddings \citep{devlin-etal-2019-bert, Liu2019RoBERTaAR, radford2019language} have also been proposed in the literature. 

The two embeddings are often added before being fed to the encoder and decoder stack.
Input text embeddings encapsulate the semantic meaning of the input tokens, while positional embeddings capture the sequential order of the tokens in a high-dimensional vector space.

\subsubsection*{Position-wise Feed-Forward Networks}
Each encoder and decoder layer contains a fully connected feed-forward network (\texttt{FFN}) layer, which processes the output from the attention mechanism independently on each position in the sequence.
\texttt{FFN} layers typically consist of two linear transformations and a non-linear activation function such as ReLU \citep{attention-2017}.
\texttt{FFN} layers are crucial in transformers to capture complex, non-linear relationships within the input sequence.

\subsubsection*{Layer Normalisation and Residual Connections}
To enhance training stability, each sub-layer (\texttt{multi-head attention} and \texttt{FFN} layer) within both the encoder and decoder layers contains layer normalisation and a residual connection.
Layer normalisation \citep{ba2016layer} plays a crucial role in mitigating uninformative variations within the hidden vector values. It achieves this by normalising the values to a unit mean and standard deviation within each layer, facilitating more stable and faster training.
Residual connection \citep{He_2016_CVPR} is beneficial for easier backpropagation and contributes to a smoother loss curve. These components help prevent vanishing or exploding gradients, making it easier to train deep architectures.

\subsubsection*{Final Output Softmax Layer}
In the last output layer of the decoder stack, a learned linear transformation layer is used to convert the output vector into logits.
This linear layer at the output shares the same weight matrix with the two input embedding layers of the encoder and the decoder, with the logits representing the scores against each vocabulary item.
Subsequently, a \texttt{Softmax} function is applied to convert the logits to predict the probabilities of the next tokens. 
Different decoding strategies can then be employed to sample the tokens to produce the generated sequence.

\subsection{Transformer Models}

In this section, we provide an overview of three fundamental categories of the transformer models: encoder-only models, decoder-only models, and encoder-decoder or sequence-to-sequence models. Additionally, we briefly introduce multimodal transformers, specifically vision-language models.

\subsubsection{Encoder-Only Models}

The first category of the transformer models comprises only the encoder, with BERT \citep{devlin-etal-2019-bert} standing as one of the most widely adopted models to date.

BERT is trained on the English Wikipedia and Google's BooksCorpus \citep{Zhu_2015_ICCV}, focusing on two primary pre-training objectives: Masked Language Modelling (MLM) and Next Sentence Prediction. Next Sentence Prediction is a binary classification task that determines whether a given pair of consecutive sentences in a text corpus follow each other logically and semantically, while  MLM enables bidirectional learning from the text data. MLM is shown to be more potent than either a left-to-right model or a shallow concatenation of left-to-right and right-to-left models \citep{devlin-etal-2019-bert}.

Initially, the training text is tokenised using WordPiece \citep{johnson-etal-2017-googles}, encompassing a 30K token vocabulary. To prevent the models from directly seeing future words, 
\citet{devlin-etal-2019-bert} propose a masking strategy, which randomly masks 15\% of the tokens in a sentence and trains BERT to predict the original token.\
To enhance robustness, the randomly selected tokens to be masked are replaced by \texttt{[MASK]} 80\% of the time, by a random token 10\% of the time, and left unchanged the remaining 10\% of the time. The prediction is optimised using cross-entropy.

MLM has been widely adopted in the training of many models, including popular ones such as RoBERTa \citep{Liu2019RoBERTaAR}, ALBERT \citep{Lan2020ALBERT}, and ELECTRA \citep{clark2020electra}, among others.

The tokenised input to the models is first represented as the addition of the token embedding and the positional embedding, then fed to the stack of transformer layers, i.e., the output of the lower layer is treated as the input of the next layer, adopting the same attention mechanism as introduced in \citet{attention-2017}.
BERT is available in two different sizes: a base model with a total of 110M parameters across 12 transformer layers, and a large model with 340M parameters spanning 24 transformer layers.

After pre-training, BERT can then be fine-tuned on downstream tasks. The final representation vector in the last layer is used as the \textit{encoded} representation of the input sentence, which can later be applied to various tasks such as text classification.
With BERT setting the new state-of-the-art performance in various downstream NLP tasks, the pre-training and fine-tuning paradigm has proven to be highly effective and remains the predominant method for training language models in the current literature.

\subsubsection{Decoder-Only Models}
The second category of transformer models exclusively features the decoder, where the input data directly enters the decoder without being transformed into a higher, more abstract representation by an encoder. Representative models in this category include GPT (Generative Pre-trained Transformer) from OpenAI, along with its successive iterations like GPT-2 and GPT-3 \citep{radford2019language, NEURIPS2020_1457c0d6}.

For instance, GPT-2 uses Byte Pair Encoding (BPE) \citep{sennrich-etal-2016-neural} to tokenise the text and is trained on the objective of causal language modelling to predict the subsequent token in a sequence, typically with cross-entropy loss. The decoder in GPT models employs a specific type of attention mechanism known as masked self-attention, which allows a token to attend only to previous tokens, preventing access to future tokens during training.

The left-to-right decoder-only models are effective at generation tasks in an Auto-Regressive (AR) fashion. The process commences by introducing a start token \texttt{<s>} to the model, which traverses through a series of transformer decoder layers, generating token\_\(0\) conditioned on \texttt{<s>}. Subsequently, token\_\(0\) becomes part of the input to the model, alongside the start token, used for generating the subsequent tokens. This sequence continues until the end-of-sequence token \texttt{<e>} is generated, indicating the completion of the sequence. Fine-tuned GPT models have also demonstrated considerable potential in tasks such as translation, summarisation, and question answering \citep{radford2019language}.

The decoder-only architecture simplifies the model, enhancing its efficiency for specific tasks, such as language modelling. By eliminating the encoder, GPT models can process input data more directly and generate output more rapidly. This design also enables GPT models to be trained on a substantial amount of unlabelled data, a significant advantage in NLP where labelled data is often limited.
Thanks to the efficiency and 
generative capabilities, decoder-only models continue to be developed with improved size and performance.
Most recent LLMs adopt decoder-only transformer architectures, including GPT-3 (175B),\footnote{Numbers in the bracket show the model sizes in the number of parameters.} OPT \citep{zhang2022opt} (125M-175B), PaLM \citep{chowdhery2022palm} (8B-540B), LLaMA \citep{touvron2023llama} (7B-65B), LLaMA 2 \citep{touvron2023llama2} (7B-70B), to name a few.

\subsubsection{Sequence-to-Sequence Models}
The next category of the transformer model is the Sequence-to-Sequence (Seq2Seq) model, also referred to as the encoder-decoder transformer model, which contains both an encoder and a decoder. 
This architecture is highly versatile and has demonstrated effectiveness in various NLP applications.

Both the input and output of the Seq2Seq model are sequences.
Input sequences first undergo initial processing by the encoder.
Subsequently, the high-dimensional representation is passed on to the decoder through the cross-attention mechanism.
In each decoding step, the decoder attends not only to its previously generated tokens but also to the most relevant parts of the input sequence. This is achieved through a process similar to self-attention but with an additional step that combines information from both the history of the decoder and the encoded input, enabling the generation of the next token in the output sequence.

Popular models within this category include BART \citep{lewis-etal-2020-bart}, T5 (Text-to-Text Transfer Transformer) \citep{2020t5}, among others. BART is pre-trained on diverse tasks such as denoising auto-encoding, text infilling, and text generation. This pre-training allows it to excel in various NLP tasks as it learns a broad range of linguistic patterns.
T5, on the other hand, is designed with a \textit{text-to-text} framework, unifying diverse NLP tasks into a single system. It showcases flexibility by casting different tasks as text generation problems, allowing a seamless approach to different tasks with a unified mechanism.

Seq2Seq models,are preferred for handling diverse language-related tasks by effectively processing inputs, leveraging cross-attention mechanisms, and generating high-quality output sequences.
Compared to decoder-only models, fewer LLMs adopt the encoder-decoder transformers, examples include AlexaTM \citep{soltan2022alexatm} (20B), Flan-T5 \citep{chung2022scaling} (80M-11B).

\subsubsection{Multimodal Transformers}
\label{sec:multimodal}
Since the success of the transformer architecture in language processing, it has been adapted to other modalities such as image and audio processing.

\begin{figure}[ht!]
\centering
\includegraphics[width=\linewidth]{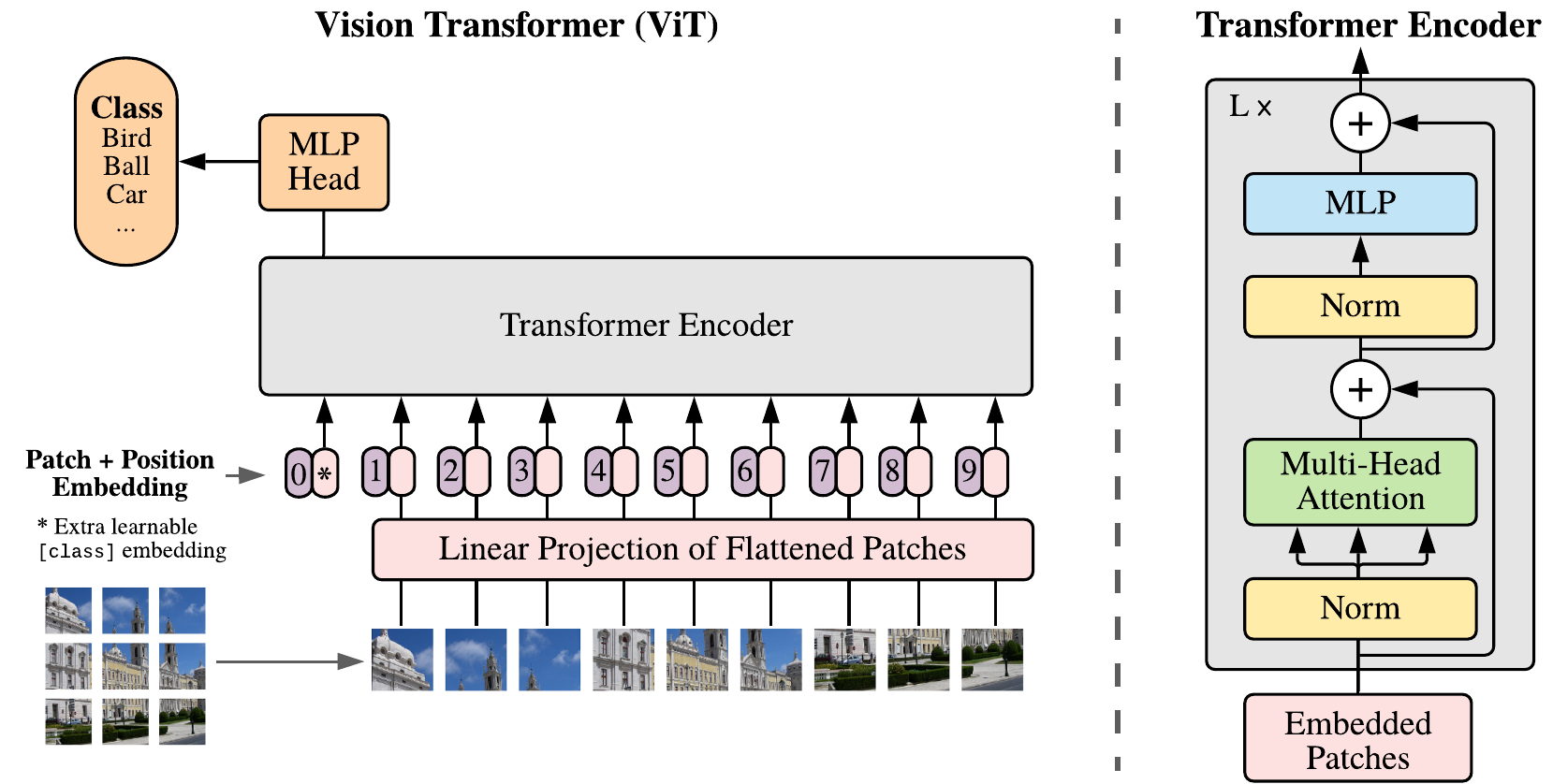}
    \caption{Vision Transformer. Images are split into fixed-size patches and linearly embedded as sequences. Then position embeddings are added and the resulting vectors are fed to a standard Transformer encoder. Source: \citet{dosovitskiy2021an}.
    }
\label{fig:vit}
\end{figure}

The Vision Transformer (ViT) model is introduced by \citet{dosovitskiy2021an}, closely following the design principles of the BERT model. 
An illustration of ViT is shown in \autoref{fig:vit}.
Images are split into a sequence of fixed-sized \(P \times P\) patches, where \(P\) is the patch size.
These flattened 2-D image patches are then linearly projected to match the transformer’s embedding dimension. 
ViT is pre-trained using supervised image classification such as ImageNet, which has demonstrated its capability to match or surpass previous stat-of-the-art CNN-based networks like ResNet \citep{he2016deep} on common downstream image classification benchmarks, particularly when ample data is available \citep{dosovitskiy2021an}.

Building on the success of vision transformers, the integration of multiple modalities, such as images and language, has become an active research area. 
Vision-language transformers, which are typically encoder-decoder transformers, fuse these diverse modalities. 
In the encoding phase, methods for vision-language models are categorised into two primary architectures: single-stream and dual-stream models.

The single-stream models encode both modalities within the same module. 
Embedded textual input and image features are either concatenated or combined via cross-modal cross-attention, leveraging Query vectors from one modality while Key and Value vectors come from another.
Although different vision-language tasks require different input formats, single-stream architecture is proven effective in models such as VisualBERT \citep{li2019visualbert}, V-L BERT \citep{Su2020VL-BERT}, OSCAR \citep{li2020oscar}, and OFA \citep{Wang2022UnifyingAT}, etc.

On the other hand, dual-stream models employ two single-modal encoders to independently process vision and text information. They then employ straightforward methods, such as shallow attention layers \citep{lee2018stacked} or dot products \citep{jia2021scaling} to project image and text embeddings into the same semantic space. 
Without the complex cross-attention mechanisms, the dual encoder strategy is often more efficient in modelling vision-language interactions.
Models such as ViLBERT \citep{lu2019vilbert}, and LXMERT \citep{tan-bansal-2019-lxmert} fall into this category.

\section{Knowledge Types and Sources}
\label{sec:knowledge}

The precedent sections introduced the advances in transformer-based language models that set the new state-of-the-art performance for a wide variety of NLP tasks, establishing the new paradigm of learning informative contextualised representations for text via training on the large-scale unlabelled corpus.
While existing research has showcased the implicit storage of certain knowledge within the latent parameters of language models \citep{petroni-etal-2019-language,roberts-etal-2020-much,wang-etal-2021-generative,cao-etal-2021-knowledgeable}, PLMs still face challenges in achieving reliable performance in knowledge-intensive tasks such as fact verification \citep{thorne-etal-2018-fever} and question answering \citep{kwiatkowski-etal-2019-natural}, etc.

Addressing this limitation, the research community has actively explored the integration of explicit knowledge, such as knowledge bases and Wikipedia articles, into large-scale PLMs.
In this section, we provide an overview of knowledge types and knowledge sources.

Knowledge comprises a wide spectrum of concepts, e.g., Wikipedia states that \textit{Knowledge is an awareness of facts, a familiarity with individuals and situations, or a practical skill}.\footnote{\url{https://en.wikipedia.org/wiki/Knowledge}} In the context of NLP, knowledge can be broadly classified into two categories: structured knowledge and unstructured knowledge \citep{zhu-etal-2022-knowledge}.

\textbf{Structured Knowledge} includes Knowledge Bases (KB) or knowledge graphs (KG), dictionaries \citep{yu-etal-2022-dict}, syntax-trees \citep{zhou-etal-2020-limit, bai-etal-2021-syntax}, etc. 
Knowledge Bases are the most common source of structured knowledge utilised in NLP \citep{zhu-etal-2022-knowledge}.
Widely-used KBs include Wikidata,\footnote{\url{https://www.wikidata.org/wiki/Wikidata:Main_Page}} DBpedia,\footnote{\url{https://www.dbpedia.org/}} WordNet \citep{miller-1993-wordnet}, ConceptNet,\footnote{\url{https://conceptnet.io/}} among others.
Both Wikidata and DBpedia are developed around Wikipedia, where structured knowledge is stored as triples, i.e., \texttt{<subject, relation, object>}, with subject and object being entities.
World knowledge and facts can effectively be represented by such triples, for example, \texttt{<United Kingdom, Capital, London>}.

\begin{figure}[th!]
\vspace{0.2cm}
\centering
\includegraphics[width=0.9\linewidth]{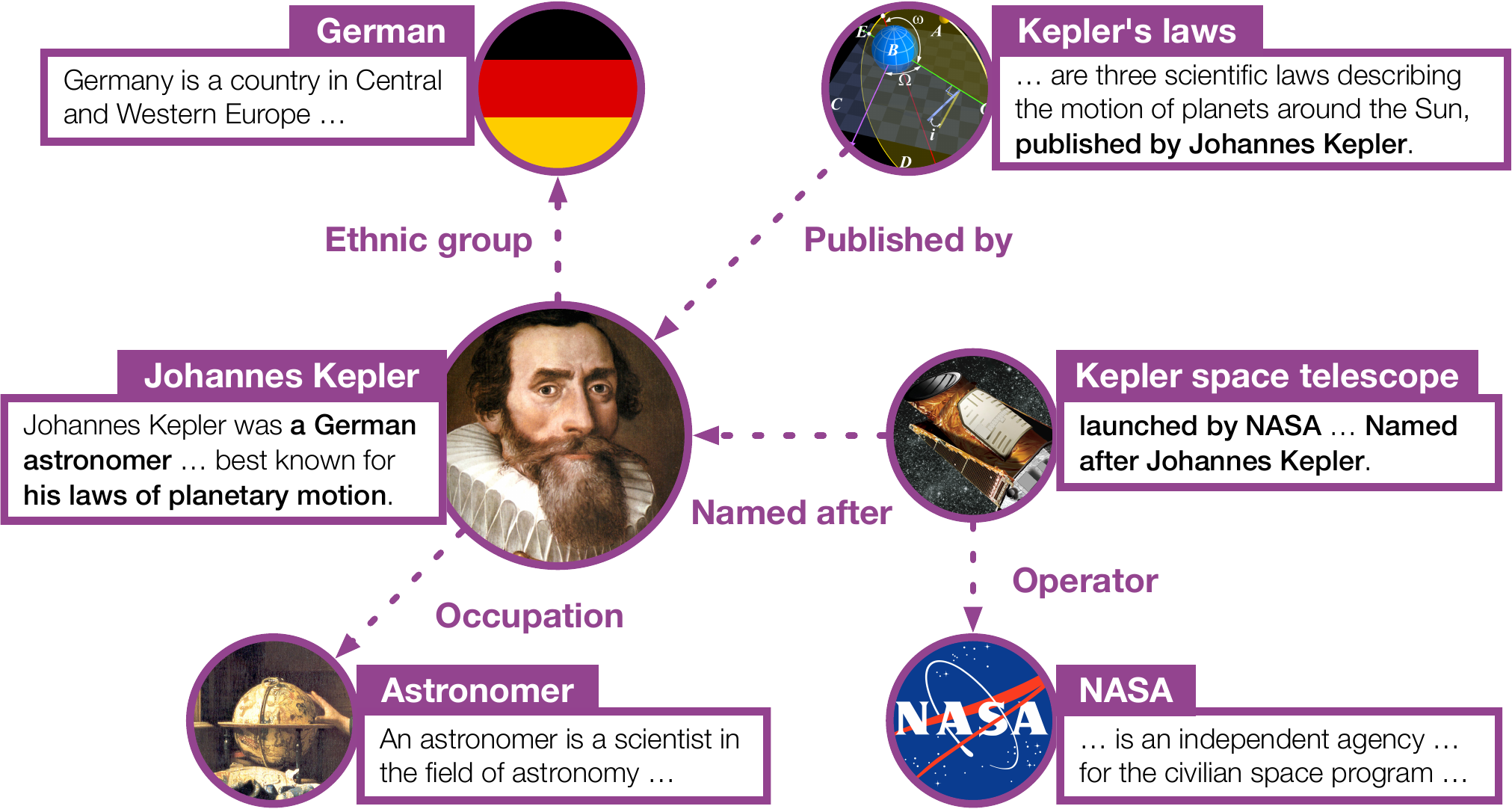}
    \caption{An example of a knowledge graph in Wikidata, which contains the information of entities, their relations, and the descriptions of the entities. Source: \citet{wang-etal-2021-kepler}.}
\label{fig:kb}
\end{figure}

Wikidata is a dynamic and collaborative knowledge base that represents and links entire Wikipedia articles. It provides document-oriented knowledge representation, making it a comprehensive source of diverse information. Wikidata also provides textual descriptions of entities, \autoref{fig:kb} shows an example (source from \citet{wang-etal-2021-kepler}). 
DBpedia aims at extracting structured knowledge from Wikipedia and transforming it into a machine-readable format.

WordNet is a lexical database of the English language, which categorises words into synsets, making it particularly useful for tasks such as word sense disambiguation.
ConceptNet is a multilingual knowledge base, representing words, phrases, and the commonsense relationships between them. The knowledge in ConceptNet is collected from a variety of resources including crowdsourcing.

\textbf{Unstructured knowledge} takes on various forms, with one prevalent example being text corpora utilised in language model pre-training. Prior research has shown that language models store or memorise knowledge after they are trained, hence can also be treated as an unstructured knowledge source \citep{petroni-etal-2019-language, heinzerling-inui-2021-language}. 

\citet{neeman-etal-2023-disentqa} categorise unstructured knowledge into two distinct sources: \textit{parametric knowledge}, which is encoded in or memorised by the model parameters, and \textit{contextual knowledge}, referring to knowledge encapsulated within external textual sources provided to the model at inference time of the downstream tasks, such as paragraphs retrieved based on the question query in question answering.

Knowledge is a theme throughout the thesis. In Chapter \ref{FakeNews}-\ref{WebIE}, we primarily consider \textit{Structured Knowledge}, especially the entity and the relations stored in knowledge bases, i.e., WikiData.
In Chapters \ref{VQA} and \ref{LLM}, the exploration of knowledge will be expanded to more flexible forms, including the unstructured \textit{Parametric Knowledge} via self-explanation (Chapter \ref{VQA}), and \textit{Distilled Knowledge} - extraction of knowledge from recent LLMs to power smaller task-specific models via data augmentation (Chapter \ref{LLM}).
Particularly in Chapter \ref{EntityCS}, \ref{WebIE}, and \ref{LLM}, we also study zero-shot cross-lingual transfer for multilingual tasks.

\section{Knowledge-Enhanced Language Models}
\label{sec:knowledge-plm}

Now that we have presented the fundamentals of transformer-based language models and knowledge representation, this section continues to introduce the techniques for combining them.
Specifically, we categorise knowledge integration methods for pre-trained language models into (i) the incorporation of entity embeddings, (ii) the utilisation of external memory, (iii) the addition of knowledge-related auxiliary pre-training tasks, (iv) the employment of adapters, and (v) the augmentation of retrieval components into the language models.

\subsection{Adding Entity Embeddings}
The first line of work involves the identification of entities in a sentence, enhancing the original input embeddings with entity knowledge represented in triples by incorporating entity embeddings.
Entity embeddings can take the form of pre-trained embedding vectors or share the embedding layer of the transformer model.

Diverse approaches have emerged to representing entities in knowledge bases via entity embeddings.
Knowledge graph embedding methods such as TransE \citep{NIPS2013_1cecc7a7} and TransR \citep{Lin_Liu_Sun_Liu_Zhu_2015}, employ score functions to measure the distance of two associated entities by their relation. 
The underlying intuition is that entities linked by a relation are proximal in the vector space, i.e., entity embeddings encapsulate relationship representations from the knowledge graph. Word-entity co-occurrence methods such as Wikipedia2Vec \citep{yamada-etal-2020-wikipedia2vec}, concurrently learn embeddings of words and entities by identifying co-occurring words around a given entity and placing similar words and entities in close proximity within a continuous embedding space. Another category of methods, like BLINK \citep{wu-etal-2020-scalable}, utilises entity descriptions encoded by transformer-based PLMs to represent entities.

Models that integrate knowledge by specific entity embeddings include ERNIE \citep{zhang-etal-2019-ernie}, KnowBert \citep{peters-etal-2019-knowledge}, K-BERT \citep{Liu_Zhou_Zhao_Wang_Ju_Deng_Wang_2020}, etc.
These models typically use an existing entity linker or jointly train an entity linker to identify mentions of entities in knowledge bases in the input text, and then retrieve the pre-trained entity embeddings or embed the corresponding entities.
Note that the dimensions of the pre-trained entity embeddings are generally much lower than those of the transformer encoders, where a fusion layer is typically added to effectively incorporate entity embeddings from a different embedding space and combine context and entity information.

\begin{figure}[t!]
\centering
\includegraphics[width=1\linewidth]{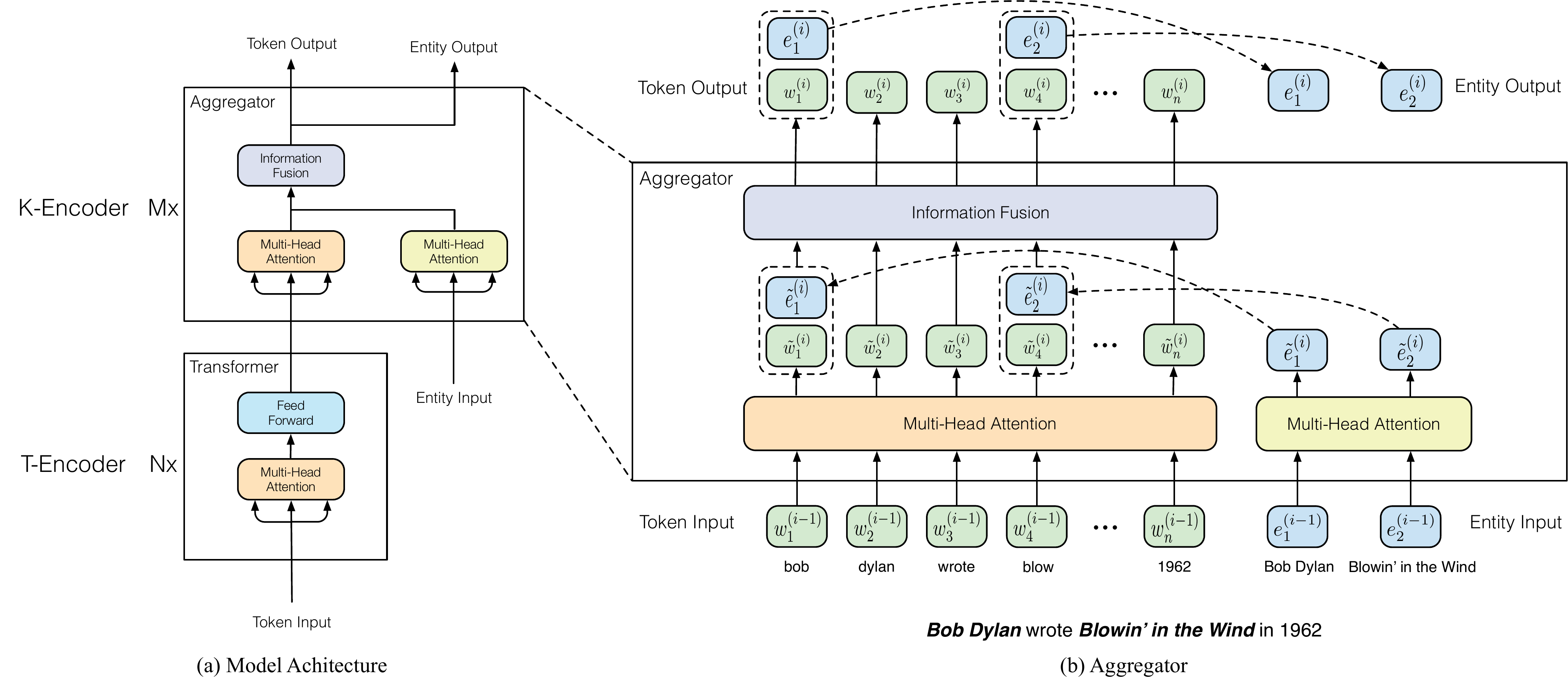}
    \caption{Illustration of ERNIE. ERNIE comprises a conventional text encoder and an additional knowledge encoder. An aggregator (on the right) is applied for the mutual integration of the input tokens and entities. The information fusion layer takes both token embedding and the concatenation of the token embedding and entity embedding as input, and outputs new token embeddings and entity embeddings for the next layer. Source: \citet{zhang-etal-2019-ernie}.}
\label{fig:ernie}
\end{figure}

ERNIE, for instance, employs TAGME \citep{Ferragina2010TAGMEOA} to link entities to Wikidata. TAMGE identifies entity mentions in the input text, links them to associated TransE entity embeddings, and fuses them into corresponding positions in the text, as illustrated in \autoref{fig:ernie}. The knowledge-based learning objective involves predicting correct token-entity alignments, enabling ERNIE to outperform BERT in tasks including entity typing and relation classification \citep{zhang-etal-2019-ernie}. However, challenges arise from the need for pre-annotated and linked entities, introducing potential noise through entity linkers like TAGME.

\begin{figure}[t!]
\centering
\includegraphics[width=1\linewidth]{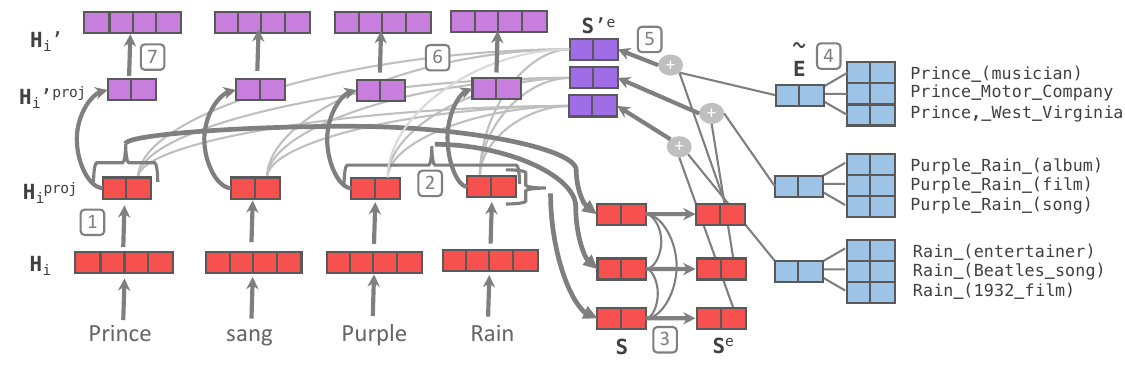}
    \caption{Illustration the Knowledge Attention and Recontextualisation (KAR) component in KnowBert.
BERT word piece representations ($\mathbf{H}_i$) are first projected to $\mathbf{H}^{\text{proj}}_i$ (1), then pooled over candidate mentions spans (2) to compute $\mathbf{S}$, and contextualized into $\mathbf{S}^e$ using mention-span self-attention (3).
An integrated entity linker computes weighted average entity embeddings $\tilde{\mathbf{E}}$ (4), which are used to enhance the span representations with knowledge from the KB (5), computing $\mathbf{S}'^e$.
Finally, the BERT word piece representations are recontextualised with word-to-entity-span attention (6) and projected back to the BERT dimension (7) resulting in $\mathbf{H}'_i$. Source: \citet{peters-etal-2019-knowledge}.}
\label{fig:knowbert}
\end{figure}

KnowBert, on the other hand, extends BERT by jointly training an entity linker using a knowledge attention and re-contextualisation mechanism, as illustrated in \autoref{fig:knowbert}. It identifies entity spans in the input text and incorporates an integrated entity linker to retrieve entity embeddings from a knowledge base. The entity linker is responsible for entity disambiguation, considering 30 entity candidates and using their weighted average embedding. Knowledge-enhanced entity-span representations are then re-contextualised with a word-to-entity attention technique. When entity-linking supervision is available, the model is trained with an additional knowledge-aware log-likelihood or max-margin objective. KnowBert has demonstrated improvements in relationship extraction, entity typing, and word sense disambiguation over BERT and ERNIE \citep{peters-etal-2019-knowledge}.

K-BERT expands language representation with adaptable knowledge bases via a knowledge layer. This layer detects and injects relevant triples from a knowledge base into the input sentence, converting it into a knowledge-rich sentence tree. The enriched sentence is then processed by the embeddings layer. Soft-position and visible matrix techniques are employed to control the utilisation of knowledge while maintaining the semantic meaning of the sentence close to the original input. K-BERT showcases advantages in knowledge-driven domain-specific tasks, such as question answering and named entity recognition in fields like law, finance, and medicine \citep{Liu_Zhou_Zhao_Wang_Ju_Deng_Wang_2020}. Moreover, K-BERT also offers flexibility in adapting to different knowledge bases without the need for re-training when changing the KB integrated.

\subsection{Using External Memory}
Another category of approaches involves the use of external memory to seamlessly integrate factual knowledge into language models. This external memory typically acts as a key-value store, providing access to knowledge base triples or contextual information. This methodology can better facilitate the incorporation and updating of knowledge from knowledge bases into PLMs, often without the need for extensive re-training.

Models that leverage external memory include KGLM \citep{logan-etal-2019-baracks}, KNN-LM \citep{Khandelwal2020Generalization}, and others.
The key idea of KGLM is to condition the language model on a knowledge graph.
Unlike most of the models discussed in this thesis, KGLM uses LSTM instead of transformer-based PLMs.
In addition to predicting the next word given the previous words in the sequence, KGLM also predicts the next entity given the previous words and entities in the sequence. 
As KGLM iterates over the sequence, it builds a local knowledge graph which is a subset of the full KG with only entities relevant to the sequence. 
LSTM is then used to predict the next word as well as its type: related entity, a new entity, or not an entity.
KGLM is found to outperform GPT-2 on fact completion tasks.
Qualitatively, KGLM tends to predict more specific tokens compared to GPT-2.

KNN-LM learns the similarities between text sequences and stores all representations of text sequences in a nearest neighbour data store. During inference, KNN-LM identifies the $k$ most similar sequences in the data store for a given input, retrieves the corresponding target (e.g., the next word) from these $k$ sequences, and combines the KNN probabilities with the probabilities computed by the language model for the final prediction.

\subsection{Adding Knowledge-Related Auxiliary Pre-training Tasks}
The next set of methods focuses on integrating knowledge by designing auxiliary knowledge-related pre-training tasks, extending beyond traditional training objectives like masked language modelling. These tasks encompass Masked Entity Prediction, Entity Prediction from Descriptions, Entity-Relation Discrimination, etc.

Many models in the literature fall within this category. Since the model architecture is typically not modified, these methods provide flexibility and entail no additional inference overhead during deployment. Below we review some representative models in this category.

\citet{Xiong2020Pretrained} train a model, WKLM, using Wikipedia articles. The mentions in the text are replaced with different entities of the same type from WikiData to create negative knowledge statements that are still linguistically correct. 
An entity replacement loss is introduced to train WKLM to distinguish between true and false knowledge, together with the MLM objective for pre-training.
\citet{Sun2019ERNIEER} adopt a masking strategy at both the phrase and entity levels during training. This introduces tasks that necessitate factual knowledge for model comprehension.
\citet{roberts-etal-2020-much} leverage salient span masking introduced by \citet{pmlr-v119-guu20a}, to mask out salient spans such as entities and dates. This approach demonstrates enhancements over T5, particularly in question answering.

\begin{figure}[t!]
\centering
\includegraphics[width=1\linewidth]{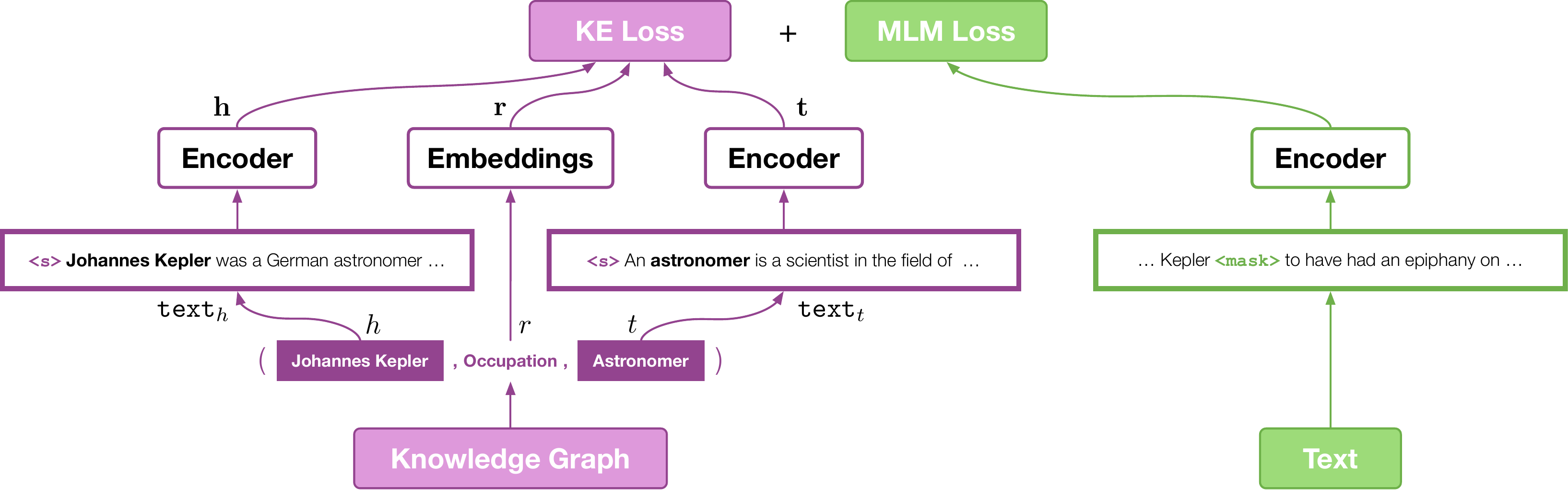}
    \caption{Illustration of the KEPLER framework. A language model is jointly trained on knowledge embedding (KE) and masked language modelling (MLM) objectives, where three types of entity embeddings are studied for the knowledge embedding task: entity descriptions as embeddings, entity and relation descriptions as embeddings, and entity embeddings conditioned on relations. Source: \citet{wang-etal-2021-kepler}.}
\label{fig:kepler}
\end{figure}

KEPLER \citep{wang-etal-2021-kepler}, illustrated in \autoref{fig:kepler}, introduces a knowledge embedding objective with supervision from a knowledge base and optimises jointly with language modelling objectives. KEPLER is specifically trained to encode entities from their contextual descriptions, enhancing the ability of PLMs to extract knowledge from text.
\citet{calixto-etal-2021-wikipedia} train a multilingual model using Wikipedia articles in 100 languages together with BabelNet \citep{BabelNet}, a multilingual sense-inventory for Word Sense Disambiguation, by predicting Wikipedia hyperlinks.

One of our works from this thesis, \texttt{EntityCS} \citep{whitehouse-etal-2022-entitycs}, detailed in Chapter \ref{EntityCS}, also utilises the entity prediction objectives on a code-switched training corpus constructed from Wikipedia and Wikidata.

\subsection{Adding Adapters}

Adapters are typically light-weighted or smaller transformer blocks with fewer layers.
Keeping the pre-trained language model parameters frozen, the adapter-based models provide efficiency in training and show advantages in mitigating the risk of catastrophic forgetting \citep{vu-etal-2022-overcoming}.
In addition to the common use case of adapter for parameter-efficient fine-tuning \citep{pmlr-v97-houlsby19a}, researchers have also developed models utilising adapters to integrate knowledge into PLMs. We briefly introduce several such models below.

\begin{figure}[t!]
\centering
\includegraphics[width=1\linewidth]{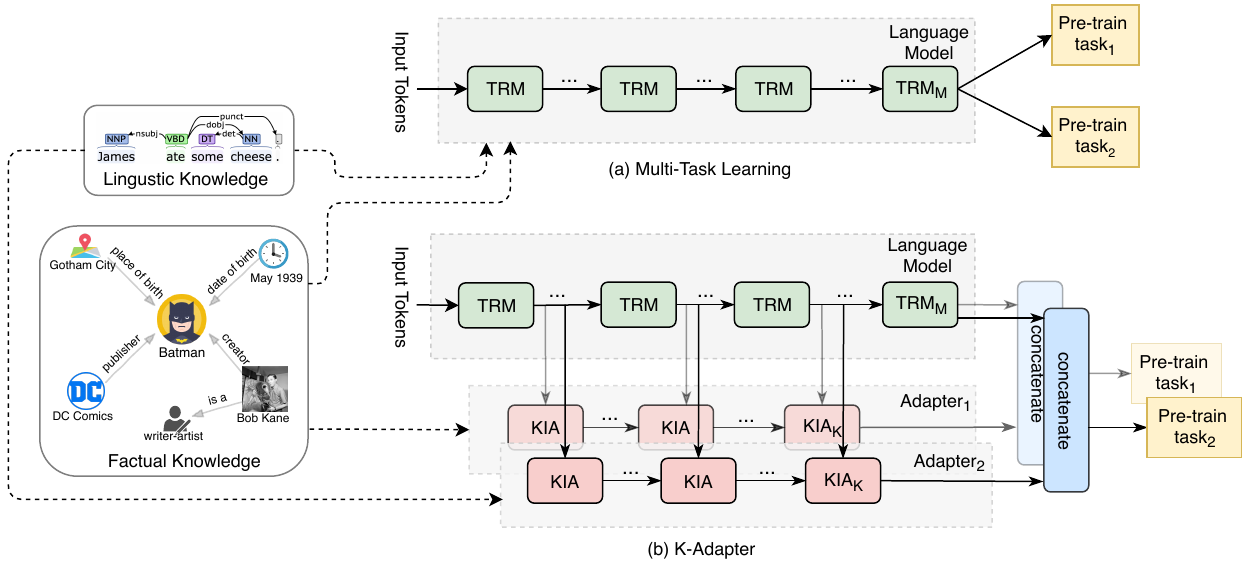}
    \caption{Illustration of K-ADAPTER. Compared to conventional multi-task learning for injecting knowledge (a) that may result in catastrophic forgetting, K-ADAPTER (b) injects knowledge by training adapters independently on different pre-train tasks, enabling continual knowledge infusion. When injecting new kinds of knowledge, the existing knowledge-specific adapters will not be affected. KIA represents the adapter layer and TRM represents the transformer layer. Source: \citet{wang-etal-2021-k}.}
\label{fig:kadapter}
\end{figure}

K-ADAPTER \citep{wang-etal-2021-k}, as illustrated in \autoref{fig:kadapter}, adds learnable adapters to RoBERTa that are trained in a multi-task setting on relation prediction and dependency-tree prediction. 
Two types of knowledge adapters are developed: factual knowledge obtained from automatically aligned text triples on Wikipedia and Wikidata, and linguistic knowledge obtained via dependency parsing. 
Both adapters have demonstrated effectiveness in improving relation classification, entity typing, and question answering \citep{wang-etal-2021-k}.
\citet{hou-etal-2022-adapters} introduce a lightweight adapter set to enhance multilingual PLMs with cross-lingual entity alignment and facts from multilingual KB for many languages. Their experiments showcase the benefits of incorporating multilingual factual knowledge, particularly for low-resource languages.
Many works focus on enhancing tasks requiring domain-specific knowledge with adapters, for instance, Biomedical NLP \citep{LAI2023104392}, Task-Oriented Dialogue Systems \citep{emelin-etal-2022-injecting}, as well as the exploration of the mixture of domain adapters \citep{diao-etal-2023-mixture}.

\subsection{Retrieval-Augmented Language Models}
Retrieval-augmented approaches have gained popularity for expanding contextual knowledge without the need for extensive model parametrisation. 
It involves training a retriever that dynamically retrieves relevant knowledge, such as passages from Wikipedia or sub-graphs from a KB at runtime. Without the necessity of storing vast amounts of knowledge within the model, retrieval-augmented language models enable more efficient and convenient updates of the evolving knowledge. 

Retrieval-augmented models are extensively applied to question-answering and text-generation tasks.
\citet{joshi2020contextualized} encode questions and passages alongside dynamically retrieved textual encyclopedic background knowledge from multiple documents, particularly focusing on entities mentioned in the text. This method exhibits effectiveness in tasks emphasising factual reasoning, such as reading comprehension.

\begin{figure}[t!]
\centering
\includegraphics[width=1\linewidth]{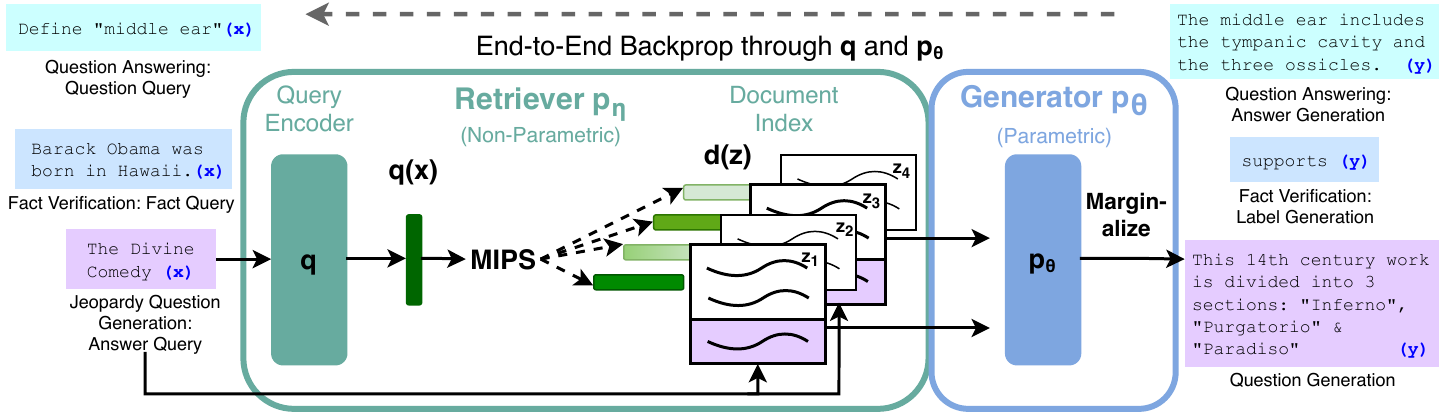}
    \caption{Illustration of RAG. RAG combines a pre-trained retriever (Query Encoder + Document Index) with a pre-trained Seq2Seq model (Generator) and fine-tune end-to-end. For query \(x\), Maximum Inner Product Search (MIPS) is utilised to find the top-K documents \(z\). For the final prediction \(y\), the retrieved documents are treated as latent variables and marginalised over Seq2Seq predictions given different documents. Source: \citet{NEURIPS2020_6b493230}.}
\label{fig:rag}
\end{figure}

RAG, Retrieval-Augmented Generation, proposed by \citet{NEURIPS2020_6b493230}, generates answers by retrieving relevant spans across external texts based on pre-trained sequence-to-sequence models. As illustrated in \autoref{fig:rag}, given a query, RAG leverages the input sequence to retrieve the top K relevant passages and generates output by conditioning on these latent documents together with the input.
REALM, an extension of RAG proposed by \citet{pmlr-v119-guu20a}, augments the language model by retrieving and attending over documents from a large corpus. It consists of two key components: a neural knowledge retriever implemented with the BERT framework, responsible for encoding input data and retrieving potentially helpful documents, and a knowledge-augmented encoder implemented with a Transformer, used to infuse entities in documents and predict words for question-answering tasks.

Retrieval augmentation has also shown advantages in enhancing the few-shot learning capability of language models. 
\citet{izacard2022few} introduce a retrieval-augmented language model, Atlas, which is designed to handle knowledge-intensive tasks with very few training examples. Atlas demonstrates robust few-shot performance across a diverse range of tasks, including KILT \citep{petroni-etal-2021-kilt}, Natural Questions \citep{kwiatkowski-etal-2019-natural}, etc.

Many new retrieval-augmented approaches have been proposed in the light of LLMs \citep{retro++, gao-etal-2023-enabling, asai2024selfrag}, among which, self-RAG \citep{asai2024selfrag} train a LM with reflection tokens. The reflection tokens determine if retrieval would be helpful and criticise its own output to choose the best generation path in terms
of factuality and overall quality.
Thanks to the effectiveness of extending non-parametric and dynamic knowledge, retrieval-augmented language models maintain their popularity even amid the latest trends of advanced Large Language Models \citep{liu2023reta, patil2023gorilla}, alleviating the need for frequent and expensive re-training of ever-larger language models.

\section{Summary}

The background section provides an overview of the transformer architecture, representative transformer-based models, knowledge types and sources, and various knowledge-enhanced language models. The subsequent chapters focus on distinct research aspects, with knowledge as a theme throughout, starting with the next chapter on the application of knowledge-enhanced language models for fake news detection.

\chapter{Knowledge-Enhanced Language Models for Fake News Detection} 

\label{FakeNews} 

This chapter focuses on evaluating the effectiveness of knowledge-enhanced language models on fake news detection tasks. 
Specifically, we consider structured knowledge, such as entity knowledge represented in knowledge bases.

The main content is an extended version of the paper ``Evaluation of Fake News Detection with Knowledge-Enhanced Language Models'' \citep{whitehouse2022evaluation}, published in \textit{the Sixteenth International AAAI Conference on Web and Social Media}.

\section{Background and Introduction}

This chapter studies the fake news detection task, which includes misinformation, disinformation, rumours, hoaxes, and other forms of rapid spread and factually inaccurate information \citep{Sharma2019CombatingFN}.
Due to the wide reach of social media, fake news has been observed to severely impact political processes  \citep{2016-Election}. 
Misinformation related to medical issues, such as the COVID-19 pandemic, can cost lives \citep{OConnor2020GoingVD}.
There is a growing desire to develop automated methods for fake news detection and mitigation, however, it remains a technically challenging problem \citep{thorne-vlachos-2018-automated}.

We focus on content-based fake news detection: methods that assess the truthfulness of news items based only on textual information without using metadata. 
State-of-the-art models for this task are driven by advances in 
large-scale pre-trained language models (PLMs) \citep{Liu2019ATM, kaliyar2021fakebert}, which are trained on vast amounts of raw web-based text using self-supervised methods \citep{rogers-etal-2020-primer}.  
As discussed in Chapter \ref{Background}, a major limitation of these models is the lack of explicit grounding to real-world entities and relations, which makes it difficult to recover factual knowledge \citep{Bender2021OnTD}. 
On the other hand, knowledge bases (KBs) provide a rich source of structured and human-curated factual knowledge, often complementary to what is found in raw text. 
This has recently led to the development of KB-augmented language models \citep{zhang-etal-2019-ernie, peters-etal-2019-knowledge}. 
We posit that fake news detection can particularly benefit from the integration of KBs, making such models less dependent and reliant on surface-level linguistic features. 

In this chapter, we empirically analyse the impact of recent state-of-the-art knowledge integration methods, which enhance PLMs with KBs, for content-based fake news detection tasks. 
We evaluate ERNIE \citep{zhang-etal-2019-ernie},  KnowBert \citep{peters-etal-2019-knowledge}, KEPLER \citep{wang-etal-2021-kepler}, and K-ADAPTER, \citep{wang-etal-2021-k} on two distinct publicly available datasets:  \texttt{LIAR} \citep{wang-2017-liar}, a politically oriented dataset, and \texttt{COVID-19} \citep{Sharma2019CombatingFN}, a dataset related to the COVID-19 pandemic. 
We find that integrating knowledge can improve fake news detection accuracy, given that the knowledge bases are relevant and up-to-date.
Our experiments are not designed to find new state-of-the-art models for these datasets, but to investigate the effect of knowledge base integration into PLMs.

The main contributions of this chapter are as follows: 
\begin{itemize}

    \item  We systematically assess various Knowledge Base integration methods for fake news detection. To the best of our knowledge, this is the first study towards the effectiveness of knowledge-enhanced language models on fake news detection tasks.
    \item We investigate both model and data aspects that may hinder the effectiveness of KB integration or pose challenges in its accurate measurement.
    \item We analyse and discuss the potential real-world applications of knowledge-enhanced language models for fake news detection, including dynamic adaptation, adversarial robustness, and the need for human verification in practical deployment scenarios.
 
\end{itemize}

In the following sections, we present a brief overview of four state-of-the-art methods that integrate KBs with the PLMs studied.
We then introduce and compare the datasets, the experiments with different knowledge-enhanced models, and the effectiveness of entity linking.
We discuss our findings with respect to the necessary conditions for KB integration to be effective and how to assess its effect in application scenarios.  
Finally, we discuss the challenges in fake news detection and promising future directions.

\section{Related Work}

A detailed review of related work on knowledge-enhanced language models is included in \autoref{sec:knowledge}. This section presents an overview of related work on fake news detection.

\subsection{Approaches to Fake News Detection}

Fake news, including disinformation, misinformation, rumours, hoaxes, etc. \citep{Combating}, poses significant risks to society. The widespread influence of social media can shape public opinion and manipulate political elections \citep{2016-Election}, and in the case of misinformation related to medical fields and health problems, such as the COVID-19 pandemic, even lead to direct loss of life \citep{OConnor2020GoingVD}.

Automated and accurate fake news detection and mitigation represent critical yet technically challenging problems \citep{Sharma2019CombatingFN, su2023adapting}. Over the past decade, various fake news detection methods using deep learning techniques have emerged. These methods can be categorised into three types: content-based approaches, user behaviour or propagation pattern analysis, and hybrid models combining both.

\subsubsection*{Content-based Methods}

Content-based methods focus on the textual statements of fake news \citep{oshikawa-etal-2020-survey} and analyse the language features of the content. For instance, fake news articles and posts often contain more negative and exaggerated words, which can be leveraged to assess the truthfulness of the news \citep{rubin-etal-2016-fake}.

The majority of current content-based methods employ neural networks for text analysis. For example, Convolutional Neural Networks, widely used in text classification tasks \citep{oshikawa-etal-2020-survey}, have demonstrated superior results on the LIAR dataset compared to traditional neural networks \citep{wang-2017-liar}.
In recent years, larger-scale pre-trained transformer-based models like BERT have significantly advanced the state-of-the-art for many NLP tasks, including content-based fake news detection \citep{kaliyar2021fakebert, farokhian2022fake, su2023adapting}. Most fake news detection approaches either combine text with metadata \citep{ding2020bert} or focus solely on the source of the text \citep{gruppi2022nela}.

In terms of the two specific datasets we study,  for \texttt{LIAR}, \textcite{alhindi-etal-2018-evidence} extend the data with evidence sentences in a new dataset LIAR-PLUS to enhance detection. \textcite{Chernyavskiy2020RecursiveNT} introduce a Deep Averaging Network to model the discursive structure of the text and use Siamese models on the extended text data. Additionally, \textcite{Liu2019ATM} predict labels at two levels of granularity. In the context of the \texttt{COVID-19} dataset, results from the CONSTRAINTS 2021 workshop \citep{constraint-2022-combating} showcase a variety of traditional and neural NLP models. Notably, none of these approaches incorporates external knowledge, suggesting potential benefits from knowledge base integration.

\subsubsection*{User Behaviour-based Methods}

In contrast to content-based fake news detection, user behaviour-based methods contend that algorithms focusing on clues from news content are generally less effective. This is because fake news is often intentionally crafted to mislead users by mimicking true news \citep{shu2029beyond}. Instead, these methods study the social context during the news dissemination process on social media, including user profiles and user behaviours such as likes or retweets.

\citet{shu2029beyond} propose a tri-relationship embedding framework called TriFN, which simultaneously models publisher-news relations and user-news interactions for fake news detection. Another approach by \citet{monti2019fake} involves an automatic fake news detection model based on geometric deep learning, offering language independence and improved resilience to adversarial attacks.

\subsubsection*{Hybrid Methods}

The third category of fake news detection methods involves hybrid models that combine content and user behaviour analysis during fake news propagation.

An example of a hybrid deep model is the CSI model \citep{ruchansky2017csi}. CSI comprises three modules: Capture, Score, and Integrate, correlating the characteristics of the news text, user response, and publisher behaviour. The Capture module captures temporal patterns of user responses to a specific news article and extracts latent features of the text using RNN. The Score module learns source characteristics based on user behaviour, scoring publishers based on user conduct. In the final Integrate module, the model combines response, text, and source information to classify each news item as fake or real.

\section{Models and Datasets}

After presenting the background and related work regarding fake news detection, we now introduce the models that we use in our experiments and provide details of the datasets and our experimental setup.

\subsection{Knowledge-Enhanced PLMs}
 
In Chapter \ref{Background}, we have comprehensively reviewed various methods for integrating knowledge into PLMs, and in this chapter, we specifically focus on the following four models: 

\paragraph*{ERNIE} enhances BERT \citep{devlin-etal-2019-bert} by introducing knowledge through pre-training on both extensive corpora and KBs. While retaining the text encoder of BERT, ERNIE adds an additional knowledge encoder. 
The knowledge encoder follows the standard transformer architecture, with each layer applying multi-head attention over entity embeddings and token embeddings. A fusion layer then combines the output of the attention heads.
ERNIE uses TAGME \citep{Ferragina2010TAGMEOA} to link entities to Wikidata.
TAMGE identifies entity mentions in the input text and links them to associated TransE \citep{NIPS2013_1cecc7a7} entity embeddings, which are then fused into the corresponding positions of the text.
Apart from Masked Language Modelling and Next Sentence Prediction pre-training tasks as in BERT, ERNIE also adopts a knowledge-based learning objective which predicts the correct token-entity alignment. 
ERNIE demonstrate superior performance over BERT in entity typing and relation classification \citep{zhang-etal-2019-ernie}. 

\paragraph*{KnowBert} incorporates KBs into BERT using knowledge attention and contextualisation mechanism.
It identifies entity spans in the input text and incorporates an integrated entity linker in the model to retrieve entity embeddings from a KB. 
The entity linker is responsible for entity disambiguation, which considers 30 entity candidates and uses their weighted average embedding. 
Knowledge-enhanced entity-span representations are then re-contextualised with a word-to-entity attention technique. 
When entity-linking supervision is available, the model is trained with an additional knowledge-aware log-likelihood or max-margin objective. 
KnowBert has shown improvement over BERT in relationship extraction, entity typing and word sense disambiguation \citep{peters-etal-2019-knowledge}.

\paragraph*{KEPLER} integrates factual knowledge into PLMs by adding a knowledge embedding objective with the supervision from a KB and optimising it jointly with language modelling objectives.
KEPLER is trained to encode the entities from their contextual descriptions, which enhances the ability of PLMs to extract knowledge from text.
By keeping the original structures of PLMs, KEPLER can be used in general downstream NLP tasks without additional inference overhead.
Specifically, after training on Wikidata5M,\footnote{\url{https://deepgraphlearning.github.io/project/wikidata5m}} a large KB-aligned dataset with entity descriptions, KEPLER shows improved performance over RoBERTa \citep{Liu2019RoBERTaAR} in relationship extraction, entity typing and link prediction \citep{wang-etal-2021-kepler}. 

\paragraph*{K-ADAPTER} retains the PLMs unchanged, 
but adds learnable adapter features that are trained in a multi-task setting on relation prediction and dependency-tree prediction. 
Two kinds of knowledge adapters have been developed by \textcite{wang-etal-2021-k}: factual knowledge obtained from automatically aligned text triples on Wikipedia and Wikidata, and linguistic knowledge obtained via dependency parsing. 
Both adapters have demonstrated effectiveness in improving relation classification, entity typing, and question answering \citep{wang-etal-2021-k}.

\subsection{Fake News Datasets}
In our experiments, we use \texttt{LIAR} and \texttt{COVID-19} to study fake news detection. 
Both datasets consist of short statements, however, they differ in content, collection timelines, and linguistic features.

\paragraph*{LIAR} was collected in 2017 from Politifact.\footnote{\url{https://www.politifact.com}}
It includes 12.8k human-labelled short statements about US politics from various contexts, i.e., news releases, TV interviews, campaign speeches, etc. 
Each statement has been rated for truthfulness by a Politifact editor using a six-grade scale: \textit{pants-fire}, \textit{false}, \textit{barely-true}, \textit{half-true}, \textit{mostly true}, and \textit{true}.
\texttt{LIAR} also provides metadata (e.g., speaker, context),
which we do not use in our experiments.
While \textcite{wang-2017-liar} has been widely cited, 
we only found three other results for our specific task (no metadata, six classes) \citep{alhindi-etal-2018-evidence, Liu2019ATM, Chernyavskiy2020RecursiveNT}, the latter has the best accuracy of 34.5\%.

\paragraph*{COVID-19} was collected in 2020 after the COVID-19 outbreak. 
It consists of 10.5k posts related to the pandemic which are obtained from different social media sites including Twitter, Facebook, and Instagram. 
The fake posts were collected from various fact-checking websites, i.e., Politifact and NewsChecker,\footnote{\url{https://newschecker.in/}} and the real posts were from Twitter using verified Twitter handles. 
Each post has a label, \textit{real} or \textit{fake}.
It was used as a shared task in the CONSTRAINT 2021 workshop \citep{constraint-2022-combating} with the best-reported accuracy of 98.69\%.

\subsubsection*{Linguistic Feature Analysis}
\label{sec:feature}

\begin{figure}[t!]
\centering
\scalebox{1}{%
\begin{subfigure}[b]{0.43\textwidth}
\includegraphics[width=\textwidth]{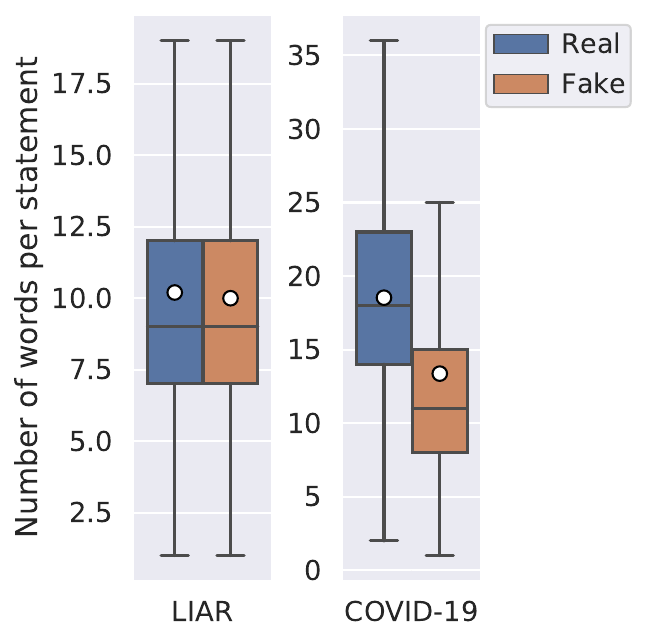}
\caption{Word count per statement}
\label{fig:len}
\end{subfigure}
\par\bigskip
\begin{subfigure}[b]{0.55\textwidth}
\centering
\includegraphics[width=\textwidth]{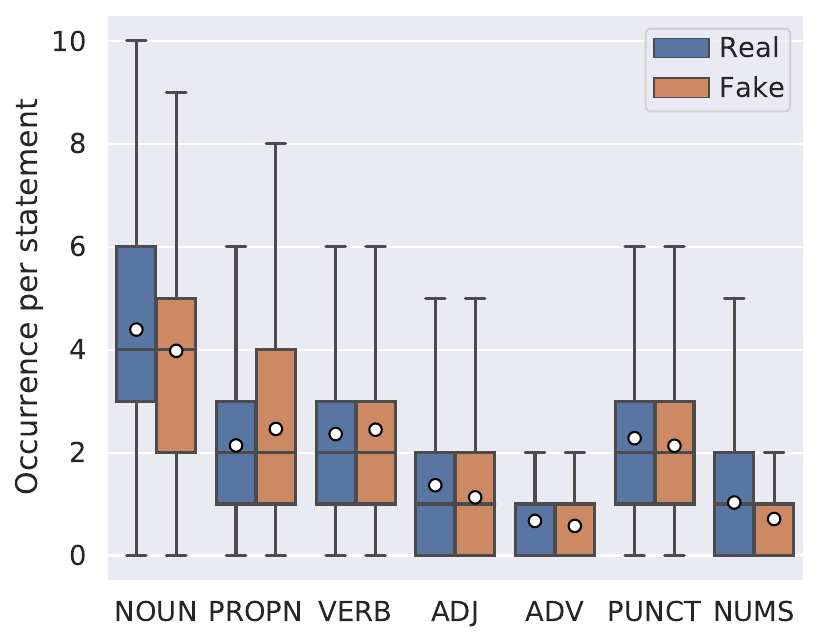}
\caption{POS, punctuation, numbers in \texttt{LIAR}}
\label{fig:liarPos}
\end{subfigure}
}
\par\bigskip
\vspace{1.5ex}
\begin{subfigure}[b]{0.6\textwidth}
\centering
\includegraphics[width=\textwidth]{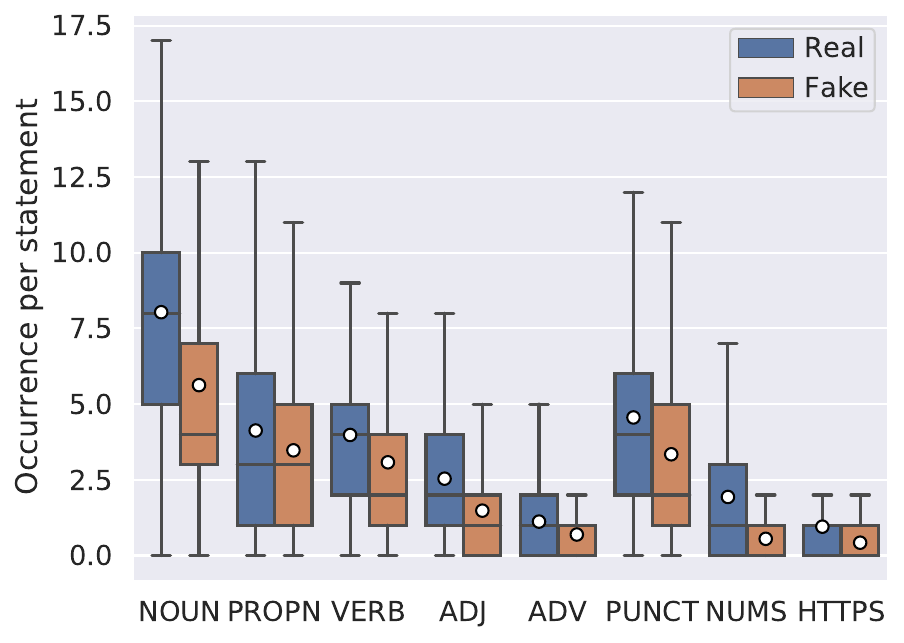}
\caption{POS, punctuation, numbers, https in \texttt{COVID-19}}
\label{fig:covidPos}
\end{subfigure}

\caption{Number of words, POS tags, punctuation, and number-like words per statement in \texttt{LIAR} and \texttt{COVID-19}, as well as
number of https-links per statement in \texttt{COVID-19}.
The mean values are shown as white-filled circles in the plots.}
\label{fig:box_plot}
\end{figure}

We perform a linguistic feature analysis following the work in 
\citet{Horne_Adali_2017} to investigate the stylistic difference between real and fake news in the datasets.
We use SpaCy\footnote{\url{https://spacy.io/}} to parse the statements and to obtain the Part-of-Speech (POS) tags. 
For \texttt{LIAR}, we group \textit{pants-fire}, \textit{false}, and \textit{barely-true} as fake and \textit{half-true}, \textit{mostly true}, and \textit{true} as real. 
We compare the distribution of various features including word count, POS tags (NOUN, PROPN, VERB, ADJ, ADV), punctuation, and number-like words, in each statement. The results are presented in \autoref{fig:box_plot}.

In terms of statement length, \texttt{COVID-19} exhibits notable disparities between real and fake classes, with averages of 32 and 22 words, respectively, as illustrated in \autoref{fig:len}. 
Conversely, \texttt{LIAR} demonstrates a more balanced distribution of statement lengths, with both real and fake statements comprising an average of 18 and 17 words, respectively.

In general, \texttt{COVID-19} presents distinct linguistic features between classes whereas \texttt{LIAR} shows more comparable features. 
Notably, \texttt{COVID-19} incorporates links, predominantly in the form of https links, forming a separate category with a markedly skewed distribution, as depicted in \autoref{fig:covidPos}.

\section{Experimental Setup}

To assess the influence of external knowledge, we compare the performance of each knowledge-enhanced PLM with the corresponding baseline model. It is noteworthy that ERNIE and KnowBert incorporate entity embeddings linked to the input, allowing us to visualise the entities contributing to fake news detection in ERNIE and design experiments to probe the impact of entity disambiguation in KnowBert.

For the evaluation of model performance in fake news detection, we fine-tune the knowledge-enhanced PLMs on the training set, employing consistent hyper-parameter settings. The input text undergoes processing by the PLM, followed by a dropout ($p=0.1$) and a linear layer. The output is then directed to a softmax layer for classification. 
The optimisation is carried out using the AdamW optimiser \citep{Loshchilov_Hutter_2019_Decoupled} with a learning rate of $5e^{-6}$, and cross-entropy serves as the loss function. 
We set the maximum input length to 128, and the batch size is fixed at 4. Training is conducted for 10 epochs, with convergence typically observed after five epochs. Each experiment is run five times, and the average accuracy, along with the standard deviation, is reported.

As for the models we use, ERNIE and KnowBert are built on BERT-base, whereas KEPLER and K-ADAPTER are enhanced from RoBERTa-base and RoBERTa-large, respectively.
In terms of hardware, a Nvidia GTX 1080 GPU with 12GB of VRAM is utilised for experiments involving BERT-base, RoBERTa-base, and models based on them. To maintain a consistent batch size for RoBERTa-large and K-ADAPTER models, a T4 with 16GB VRAM is employed.

\section{Results and Discussion}

We compare ERNIE, three pre-trained KnowBert models with different KBs (Wiki, WordNet, W+W),\footnote{They refer to Wikipedia, WordNet, and Wikipedia+WordNet as the knowledge base.} KEPLER, and K-ADAPTER with three adapters (F, L, F-L)\footnote{They refer to Factual, Linguistic, and Factual+Linguistic adapters.} in the published implementation, to the corresponding baselines.
This section presents the main results and discussion.

\subsection{Detection Accuracy}

\begin{table}[!ht]
\centering
\scalebox{1}{
\begin{tabular}{lclc}
\toprule
\sc {Model} & \sc {Base} & \multicolumn{1}{c}{ \sc {LIAR}} &\sc { COVID-19} \\
\midrule
 \textbf{B}ERT-\textbf{B}ase (BB) & - & 26.36 \textsubscript{$\pm$0.58} & 97.51 \textsubscript{$\pm$0.19} \\
\textbf{R}oBERTa-\textbf{B}ase (RB) & -  & 26.71 \textsubscript{$\pm$0.93} & 97.61 \textsubscript{$\pm$0.26}\\
\textbf{R}oBERTa-\textbf{L}arge (RL) & -  & \textbf{27.36} \textsubscript{$\pm$0.79} & \textbf{97.92} \textsubscript{$\pm$0.17}\\ 
 \midrule
ERNIE & BB &27.53 \textsubscript{$\pm$0.13} & 97.30 \textsubscript{$\pm$0.18} \\
 KnowBert-Wiki & BB & 27.64 \textsubscript{$\pm$0.09} & 97.37 \textsubscript{$\pm$0.09} \\ 
 KEPLER & RB & 26.77 \textsubscript{$\pm$1.15} & 97.58 \textsubscript{$\pm$0.15} \\
K-ADAPTER-F & RL &\textbf{28.63} \textsubscript{$\pm$0.90}$^*$ & \textbf{97.92} \textsubscript{$\pm$0.10}\\
\midrule
  KnowBert-WordNet & BB  & 26.95 \textsubscript{$\pm$0.45} & 97.00 \textsubscript {$\pm$0.06}\\
 KnowBert-W+W & BB  & \textbf{28.95} \textsubscript{$\pm$0.64}$^*$ & 97.56 \textsubscript{$\pm$0.15}\\ 
  K-ADAPTER-L & RL & 28.46 \textsubscript{$\pm$0.87}$^*$  & 98.07 \textsubscript{$\pm$0.09} \\
  K-ADAPTER-F-L & RL & 27.45 \textsubscript{$\pm$0.78} & \textbf{98.11} \textsubscript{$\pm$0.14} \\  
  \bottomrule
\end{tabular}
}
 \caption{Detection accuracy results (average of five runs).
 The first section corresponds to the baseline models. 
 Models in the second section use Wikidata KB. 
 The third section shows models using other KBs and features.  
 The best values within each section per dataset are marked in bold.
 The subscript numbers with $\pm$ show the standard deviation.
 Results with $^*$ indicate statistically significant improvements over the baseline, both for the mean (t-test, one-sided, $p<.05$) and median (Wilcoxon signed rank test, one-sided, $p<.05$). 
 }
 \label{tab:acc_table}
\end{table}
The detection accuracy of the knowledge-enhanced PLMs and their corresponding baselines is detailed in \autoref{tab:acc_table}. 

\comment{[Chapter 3 - Correction point 2/4]~}\add{To disentangle the effectiveness of knowledge integration methods, specific knowledge resources, and baseline model architectures, we categorise the results into three groups: baseline performance (top), models integrated with Wikidata knowledge base (middle), and models utilizing knowledge bases beyond Wikidata (bottom). 
Comparing models with Wikidata knowledge and their baselines, we observe consistent enhancements on the \texttt{LIAR} dataset. However, no improvement is observed on the \texttt{COVID-19} dataset. This discrepancy underscores the positive impact of relevant knowledge bases on transformer-based models, irrespective of the diverse integration approaches employed. Conversely, in scenarios where knowledge bases are outdated and datasets exhibit stylistic imbalances between classes (as outlined in \autoref{sec:feature}), external knowledge bases may even detrimentally affect performance.}

\comment{[Chapter 3 - Correction point 3/4]~}\add{Comparing models employing the same knowledge integration approach but utilising different knowledge sources reveals intriguing findings. Optimal performance on both \texttt{LIAR} and \texttt{COVID-19} datasets is achieved by models incorporating multiple knowledge sources. Specifically, KnowBERT-W+W, incorporating both Wikidata and WordNet knowledge bases, exhibits the most substantial overall improvement (a notable increase of $+2.59$ over BERT-base). Simultaneously, K-ADAPTER, incorporating linguistic adapters, demonstrates the most significant positive impact on the \texttt{COVID-19} dataset. This aligns with the observation that \texttt{COVID-19} exhibits distinct linguistic stylistic cues between different classes.}

Overall, across the \texttt{LIAR} dataset, all knowledge-enhanced methods demonstrate improvements over the baseline. In contrast, on the \texttt{COVID-19} dataset, only three out of eight models show improvement, and these improvements are marginal

The computational cost varies across different approaches. KEPLER maintains the baseline PLM architecture, incurring no additional overhead. For K-ADAPTER, the RoBERTa-large layers are frozen, resulting in a manageable overhead ranging from 9-23\% due to the adapters. However, for KnowBert, the overhead is more substantial, ranging from 40-87\%, and for ERNIE, it is even higher at 111-131\%.

\subsection{KB Linking}
\begin{figure}[h!]
\centering
\vspace{0.5em}
\begin{subfigure}[b]{0.8\textwidth}
\centering
\includegraphics[width=\textwidth]{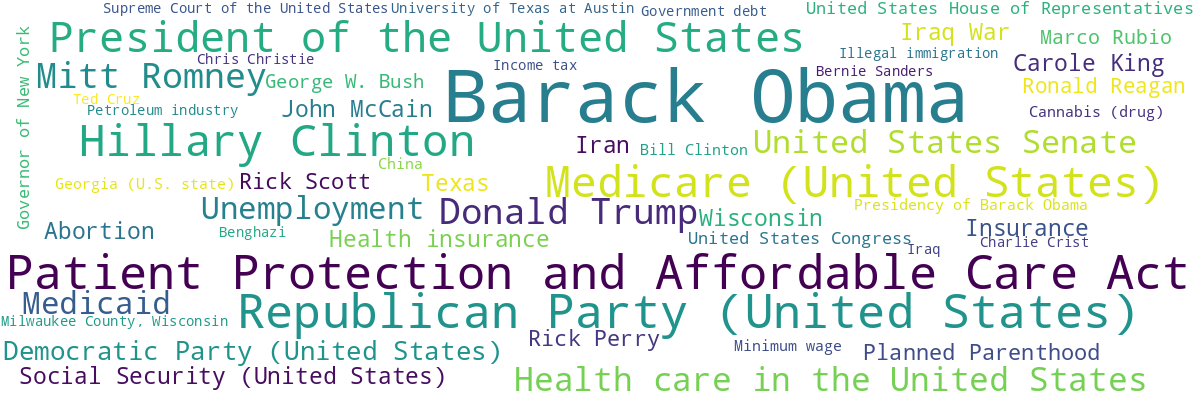}
\caption{Linked Entities in COVID-19}
\label{fig:e1}
\end{subfigure}

\begin{subfigure}[b]{0.8\textwidth}
\vspace{0.3cm}
\includegraphics[width=\textwidth]{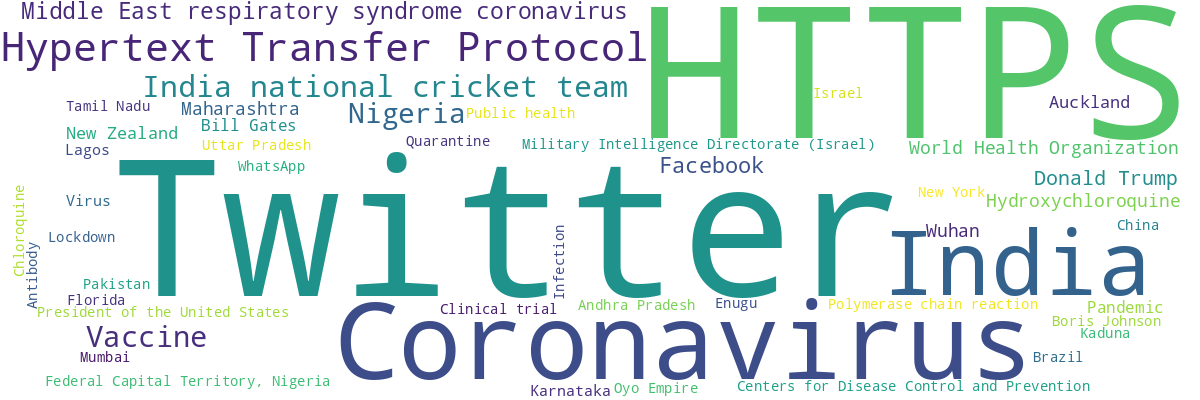}
\caption{Linked Entities in LIAR}
\label{fig:e2}
\end{subfigure}
\caption{Word cloud for the 50 most frequent entities linked by ERNIE in  \texttt{LIAR} and \texttt{COVID-19}.}
\label{fig:word_cloud_Ernie}
\end{figure}

ERNIE and KnowBert establish links between the text and KB entities at runtime, and the quality of this linking significantly impacts the output. ERNIE utilises TAGME and selects only one entity candidate per text span. In \autoref{fig:word_cloud_Ernie}, we present the 50 most frequently selected KB entities for each dataset.

In the case of \texttt{COVID-19}, the most frequently selected entities do not appear content-related, such as ``https'' and ``twitter'', while the highly relevant term ``COVID-19'' is notably absent in the linked entities. Conversely, for \texttt{LIAR}, the linked entities seem more relevant. This discrepancy may be attributed to the fact that \texttt{LIAR} was collected three years earlier, making it potentially better suited for the entity linker and the KB used.
Another potential factor influencing the effectiveness of KB integration is the number of linked entities. Unlike ERNIE, KnowBert selects the 30 most probable entities per text span. In a sensitivity study, we restrict KnowBert-W+W to only one entity, leading to a reduction in accuracy on \texttt{LIAR} from 28.95\% to 27.31\%, placing it below the accuracy of ERNIE (27.53\%).

\subsection{Discussion}

The consistent improvement in detection accuracy on \texttt{LIAR} achieved by integrating PLMs with the Wikidata KB demonstrates the potential of knowledge integration, surpassing results obtained by \textcite{wang-2017-liar} which integrated multiple types of metadata. However, the improvements, while notable, are not dramatic for \texttt{LIAR}, and they lack consistency for \texttt{COVID-19}. 
Two critical aspects contributing to these results are identified in the effective use of knowledge-enhanced models:
\begin{itemize}
 \item  Currentness and Relevance of the Knowledge Base: Since \texttt{COVID-19} was collected after most PLMs were trained, certain terms like ``COVID-19'' may not be present in the Knowledge Base.
 \item  Quality of the Dataset: 
The \texttt{COVID-19} dataset contains confounding factors that provide strong cues, potentially overshadowing the impact of the knowledge base. Notably, the prevalence of https links, occurring in 95.3\% of real posts but only 42.3\% of fake posts, can act as a \textit{shortcut} to deriving the correct prediction.
\end{itemize}

There is also potential to achieve more explainability and interpretability with direct KB integration at runtime. 
For instance, in the statement from \texttt{COVID-19}: \textit{``DNA Vaccine: injecting genetic material into the host so that host cells create proteins that are similar to those in the virus against which the host then creates antibodies''},
KnowBert-W+W correctly classifies it as ``real'', whereas BERT-base fails.
We observe that most mention spans in the statement, i.e.,  \textit{``DNA''}, \textit{``injecting''}, \textit{``genetic''}, \textit{``genetic material''}, \textit{``host''}, \textit{``cells''}, etc. are correctly linked to entities \textit{``DNA''}, \textit{``Injection\_(medicine)''}, \textit{``Genetics''}, \textit{``Genome''}, \textit{``Host\_(biology)''}, \textit{``Cell\_(biology)''}, respectively, 
suggesting that entity links may have contributed to KnowBert-W+W for this classification. However, the level of explainability remains limited. The entity linking in both models is generally of mixed quality as well, as illustrated in the \texttt{COVID-19} example. 

\subsubsection*{Application Aspects}
The application of automatic fake news detection in real-world scenarios introduces two dynamic aspects that are challenging to test with static datasets, as highlighted by our experiment on \texttt{COVID-19}:
\begin{itemize}
\item  Dynamic Adaptation: The system needs to adapt to the changing characteristics of real and fake news \citep{silva2021concept}. Knowledge-enhanced models that utilise Knowledge Bases at runtime provide an opportunity to update the KB independently of the model. This approach offers the advantage of recognising fake news as contradicting the KB even before specific examples of fake news emerge.

\item  Adversarial Robustness: Authors of fake news are likely to employ evasive strategies. Adapting the text style is relatively straightforward and can be automated, making the detection using stylistic features challenging \citep[see][]{NEURIPS2019_3e9f0fc9, schuster-etal-2020-limitations}.
\end{itemize}

The deployment of fake news detection in social media is likely to necessitate human verification, especially when users challenge actions taken against them. In this context, Knowledge Base integration can provide a valuable advantage by offering insights into the knowledge used in the detection process, thereby enhancing explainability. This transparency could be crucial in addressing user concerns and building trust in the fake news detection system.

\section{Conclusion and Future Work}

In this chapter, we study the effectiveness of enhancing PLMs with knowledge bases for fake news detection.
The findings underscore that the success of integrating knowledge with PLMs is contingent upon the availability of suitable KBs and the quality of the dataset. While better performance is observed on a \textit{static} dataset, there is room for improvement on both the modelling and application levels.
\comment{[Chapter 3 - Correction point 1/4]~}\add{To the best of our knowledge, our work is the first examination of knowledge-enhanced models in the context of fake news detection. The positive results on the \texttt{LIAR} dataset provide insights into the effectiveness of considering entity knowledge in areas beyond traditional tasks like entity linking.}

For practical application, more insight into the specific knowledge utilised during the detection process could contribute to more transparent and interpretable models. Furthermore, the potential for dynamic adaptation of models and KBs to the evolving characteristics of real and fake news is a promising avenue for exploration. The integration of KBs with PLMs presents an opportunity for more robust and timely fake news detection.

Future work could also investigate the development of a more reliable evaluation approach, for example involving testing scenarios that simulate dynamic knowledge updates as well as the challenges posed by adversarial and automatic fake news generators. 
Overall, as the landscape of misinformation evolves, the adaptability and resilience of knowledge-enhanced models demonstrate promise in their effectiveness in real-world applications.

\section{Limitations}

\comment{[Chapter 3 - Correction point 4/4]~}\add{We first discuss the general limitations of incorporating external knowledge resources for improving Transformer-based models, and then we describe the specific limitations of this chapter.
\subsection*{General Limitations of External KBs for
Improving Transformer Models}
\begin{itemize}
\item \textbf{Scalability and Efficiency}: Transformer models are already computationally intensive, especially for long contexts, due to the quadratic complexity of the attention calculation within the window length. Integrating external knowledge sources can further increase computational costs, especially when developing adaptive approaches that require up-to-date knowledge on demand.
\item \textbf{Domain Adaptation and Generalisation}: External knowledge resources often originate from specific domains or sources, which may not fully align with the target task or dataset. This misalignment poses a challenge, particularly as many knowledge integration approaches require linkage to pre-defined knowledge sources. As large language models evolve toward greater general-purpose utility, there arises a growing need for effective general-purpose knowledge fusion approaches.
\item \textbf{Selective Knowledge Incorporation}: Existing strategies for leveraging external knowledge in Transformer models often adopt a binary approach, either consistently integrating external knowledge or entirely disregarding it. However, as LLMs evolve and accumulate knowledge within their parameters, the benefit of external knowledge integration becomes context-dependent. There are instances where querying external knowledge may hinder model performance, given the limitations above (i.e., the computational overhead and potential mismatches between the external knowledge and the task at hand). A recent work, Self-RAG \citep{asai2024selfrag}, which trains the LM to reflect the necessity to retrieve external knowledge and critique its own generation conditioned on the external knowledge, shines a light into this direction.
\end{itemize}
}

\subsection*{Limitations of this Chapter}
\begin{itemize}
\item \textbf{Limited Datasets}: The experiments were conducted only on two fake news datasets, both consisting of short statements. Notably, the \texttt{COVID-19} dataset already exhibited a very high baseline accuracy. To gain more comprehensive insights into knowledge-enhanced language models, it would be beneficial to include more diverse and complex fake news detection datasets.

\item \textbf{Limited Knowledge Sources}: The evaluation in this chapter focuses on pre-trained models that integrate with knowledge bases primarily sourced from Wikidata and WordNet. However, for the specific use case of COVID-19, there is a potential value in conducting a comparison that involves integrating medical domain-specific knowledge bases.

\end{itemize}

\chapter{\correction{Entity-Centric Code-Switching for Enhanced Cross-lingual Transfer}}
\comment{[Chapter 4 - Correction point 1/6 (title change above)]}
\label{EntityCS}

In this chapter, we focus on utilising structured knowledge in a multilingual setup, for the improved cross-lingual transferability of pre-trained cross-lingual language models on entity-centric tasks.

The main content of this chapter is based on the paper ``\textsc{EntityCS}: Improving Zero-Shot Cross-lingual Transfer with Entity-Centric Code Switching'' \citep{whitehouse-etal-2022-entitycs} 
published in \textit{Findings of the Association for Computational Linguistics: EMNLP 2022}.

\section{Background and Introduction}
Cross-lingual pre-trained Language Models (XLMs), such as mBERT~\citep{devlin-etal-2019-bert} and XLM-R~\citep{conneau-etal-2020-unsupervised}, have achieved state-of-the-art zero-shot cross-lingual transferability across diverse Natural Language Understanding (NLU) tasks. These models have been notably enhanced through the incorporation of bilingual parallel sentences, along with alignment methods~\citep{Yang2020AlternatingLM, chi-etal-2021-infoxlm, hu-etal-2021-explicit, gritta-iacobacci-2021-xeroalign, feng-etal-2022-language}. However, acquiring high-quality parallel data is costly, especially for low-resource languages. Therefore, alternative data augmentation approaches have been proposed, one of which is Code Switching (CS).

Code Switching is a phenomenon in which multilingual speakers alternate between languages when they speak, a topic that has been studied for many years \citep{gumperz1977sociolinguistic, khanuja-etal-2020-gluecos, dogruoz-etal-2021-survey}. Code-switched sentences consist of words or phrases in different languages, capturing finer-grained cross-lingual expressions compared to parallel sentences. They have been utilised for multilingual intermediate training~\citep{Yang2020AlternatingLM} and fine-tuning~\citep{Qin2020CoSDAMLMC, krishnan-etal-2021-multilingual}. Nevertheless, manually creating large-scale CS datasets is expensive, and only a few natural CS texts exist~\citep{Lyu2015, barik-etal-2019-normalization, xiang-etal-2020-sina, chakravarthi-etal-2020-corpus, ASCEND-corpus}. As a result, research has turned to automatic CS data generation.

Some of these approaches generate CS data via dictionaries, often ignoring ambiguity~\citep{Qin2020CoSDAMLMC, conneau-etal-2020-emerging}. Others require parallel data and an alignment method to match words or phrases between languages \citep{Yang2020AlternatingLM, rizvi-etal-2021-gcm}. In both cases, what is switched is chosen randomly, potentially resulting in syntactically odd sentences or switching to words with little semantic content (e.g., conjunctions).
\comment{[Chapter 4 - Correction point 2/6]~}\add{This is in contrast to observations from prior work that have shown in real code-switched data, such as SEAME \citep{Lyu2015}, where nouns have the highest rate of code-switching \citep{cetinoglu-etal-2016-challenges}. They also find that people may switch grammar rules when they code-switch, making automatic code-switching by randomly replacing words in a sentence less feasible.}

On the other hand, entities contain external knowledge and do not alter sentence syntax when replaced with other entities, mitigating the need for parallel data or word alignment tools. 
Motivated by this, we propose \textsc{EntityCS}, a code-switching method that focuses on entities. Resources such as Wikipedia and Wikidata offer rich cross-lingual entity-level information and have shown benefits in XLMs pre-training~\citep{jiang-etal-2020-x, calixto-etal-2021-wikipedia, jiang-etal-2022-xlmk}. We use such resources to generate an entity-based CS corpus for the intermediate training of XLMs. Entities in wikilinks\footnote{\url{https://en.wikipedia.org/wiki/Help:Link\#Wikilinks_(internal_links)}} are switched to their counterparts in other languages retrieved from the Wikidata KB, thus alleviating ambiguity.

\comment{[Chapter 4 - Correction point 4/6]~}
\add{Training models on our synthetic entity-level code-switched data offers several advantages over using naturally occurring code-switched text. Firstly, our synthesised data allows for greater control over language diversity. While natural code-switched text often involves only one or two high-resource non-English languages, our dataset can incorporate a broader range of languages and combinations. Additionally, by focusing on entity-centric training objectives, the models are anticipated to capture finer-grained semantics shares across languages, which is particularly beneficial for downstream tasks that require accurate entity understanding.
However, it is important to note that synthetic data may not fully capture the complexities of naturally occurring code-switched text, which often involves subtle shifts in language use influenced by cultural, social, and contextual factors \citep{dogruoz-etal-2021-survey}.}

Using the \textsc{EntityCS} corpus, we propose a series of masking strategies that focus on enhancing Entity Prediction (EP) for better cross-lingual entity representations.
We evaluate the models on entity-centric downstream tasks, including Named Entity Recognition (NER), Fact Retrieval, Slot Filling (SF), and Word Sense Disambiguation (WSD). Extensive experiments demonstrate that our models outperform the baseline on zero-shot cross-lingual transfer, with a +2.8\% improvement on NER, surpassing the prior best result that uses large amounts of parallel data, +10.0\% on Fact Retrieval, +2.4\% on Slot Filling, and +1.3\% on WSD.

The main contributions of this chapter include:
\begin{itemize}
\item Construction of an entity-level CS corpus, \textsc{EntityCS}, based on the English Wikipedia and Wikidata, mitigating the need for parallel data, word-alignment methods, or dictionaries.
\item A series of intermediate training objectives, focusing on Entity Prediction.
\item Improvement of zero-shot performance on NER, Fact Retrieval, Slot Filling, and WSD.
\item Further analysis of model errors, the behaviour of different masking strategies throughout training, as well as the impact across languages, and demonstration of the particular benefit of non-Latin script languages.
\end{itemize}

\section{Related Work}

We introduce the related work of this chapter in the following three aspects:

\subsection{Cross-Lingual Pre-Training}
Most existing cross-lingual pre-trained language models employ parallel data to enhance multilingual contextualised word representations for different languages \citep{ouyang-etal-2021-ernie, luo-etal-2021-veco, chi-etal-2021-improving}. Adapters have also been applied to improve zero-shot and few-shot cross-lingual transfer by training only a small set of model parameters \citep{pfeiffer-etal-2020-mad, ansell-etal-2021-mad-g}. Additionally, meta-learning techniques \citep{nooralahzadeh-etal-2020-zero, tarunesh-etal-2021-meta} have proven highly effective for rapid adaptation to new languages \citep{dou-etal-2019-investigating}. In comparison to these approaches, our method aims to enhance cross-lingual transferability through intermediate training on an entity-based code-switching corpus created from Wikipedia wikilinks, without requiring parallel data.

\subsection{Code Switching}
Code Switching methods have shown success in cross-lingual model pre-training and fine-tuning across various NLU tasks, including NER \citep{ner-indian-2020, liu-etal-2021-mulda}, Part-of-Speech Tagging \citep{ball-garrette-2018-part}, Machine Translation~\citep{srivastava-singh-2020-phinc}, Intent Classification, and Slot Filling~\citep{krishnan-etal-2021-multilingual}. These methods have also been applied on code-switched datasets \citep{rizal-stymne-2020-evaluating, prasad-etal-2021-effectiveness}.

A significant challenge in studying Code Switching is the scarcity of training data \citep{gupta-etal-2020-semi}. Existing code-switched corpora mostly involve English and one other language (e.g., English-Chinese, English-Spanish, English-Hindi), extracted from social media platforms~\citep{barik-etal-2019-normalization, xiang-etal-2020-sina, chakravarthi-etal-2020-corpus, ASCEND-corpus}. 

Methods for generating code-switched data in multiple languages have been proposed.
\citet{Qin2020CoSDAMLMC} and \citet{conneau-etal-2020-emerging} create code-switched data from downstream task datasets by randomly switching individual words to a target language using translations from bilingual dictionaries. However, this introduces ambiguity errors and is prone to switching words without significant content. \citet{krishnan-etal-2021-multilingual} use Code Switching to improve Intent Classification and Slot Filling. Instead of switching individual words, they obtain phrase information from the slot labels and generate phrase-level code-switched sentences via automatic translations. \citet{Yang2020AlternatingLM} create code-switched sentences by randomly substituting source phrases with their target equivalents in parallel sentences after obtaining word alignments. \citet{jiang-etal-2020-x} select a subset of Wikipedia sentences in four languages that contain multilingual entities from X-FACTR and create code-switched sentences by switching entities from English to non-English entities \textit{and} vice versa, via Wikidata translations.

Our proposed \texttt{EntityCS} shares several similarities.
\comment{[Chapter 4 - Correction point 6/6]~}\add{The primary distinctions, particularly in contrast to \citet{jiang-etal-2020-x}, can be summarised in three key aspects: (i)\textit{Scale}: While \citet{jiang-etal-2020-x} focuses on a limited subset, we create a substantial corpus, magnifying the scale by a factor of 1000x, spanning across 93 languages rather than just four; (ii)  \textit{Diversity}: In contrast to the 30\% sentence code-switching and single non-English entity focus in \citet{jiang-etal-2020-x}, our approach code-switches every sentence to multiple candidate target languages. This modification enhances the model's ability to capture cross-lingual information comprehensively, resulting in improved cross-lingual transferability; (iii) \textit{Entity-Prediction Objective}: A pivotal difference lies in our incorporation of various entity-focused prediction objectives. This strategic design enhances the entity-awareness of the intermediate-trained model, which is further shown to contribute to entity-centric downstream tasks.}

\subsection{Knowledge Integration into Language Models}
As detailed in Chapter \ref{Background}, Pre-trained Language Models may lack explicit grounding to real-world entities and relations, making it challenging to recover factual knowledge \citep{Bender2021OnTD}. 

We refer the reader to \autoref{sec:knowledge-plm} for techniques of knowledge integration into PLMs that focus on monolingual models.
Integrating multilingual knowledge into XLMs has also been recently addressed. \citet{jiang-etal-2022-xlmk} train a model with two knowledge-related tasks: entity prediction and object entailment. They use WikiData description embeddings in one language (English and non-English) to predict an entity in a target language as a classification task, preserving an entity's vocabulary. \citet{calixto-etal-2021-wikipedia} use Wikipedia articles in 100 languages together with BabelNet~\citep{BabelNet}, a multilingual sense-inventory for WSD, by predicting the WikiData ID of each entity. Another work taking advantage of entities by \citet{ri-etal-2022-mluke} uses dedicated multilingual entity embeddings on 24 languages and outperforms word-based pre-trained models in various cross-lingual transfer tasks.

\section{Methodology}
In this section, we provide a comprehensive overview of the \textsc{EntityCS} corpus construction and detail various entity-oriented masking strategies employed in our experiments.

\subsection{\textsc{EntityCS} Corpus Construction} 
\label{corpus_construction}

Wikipedia is a multilingual online encyclopedia available in more than 300 languages.\footnote{\url{https://en.wikipedia.org/wiki/Wikipedia}}
Structured data of Wikipedia articles are stored in Wikidata, a multilingual document-oriented database.
With more than six million articles, English Wikipedia has the potential to serve as a rich resource for generating CS data.
We use English Wikipedia and leverage entity information from Wikidata to construct an entity-based CS corpus.

To achieve this, we make use of wikilinks in Wikipedia, i.e., links from one page to another.
We use the English Wikipedia dump\footnote{\url{https://dumps.wikimedia.org/enwiki/latest/} (Nov 2021 version)} and extract raw text with WikiExtractor\footnote{\url{https://github.com/attardi/wikiextractor}} while keeping track of wikilinks.
Wikilinks are typically surrounded by square brackets in Wikipedia dump, in the format of [[\textit{entity} | \textit{display text}]], where \textit{entity} is the title of the target Wikipedia page it links to, and \textit{display text} corresponds to what is displayed in the current article.
We then employ SpaCy\footnote{\url{https://spacy.io/}} for sentence segmentation.
To focus on entity-level code-switched instances, only sentences containing at least one wikilink are retained, and sentences exceeding 128 words are excluded from the dataset. This process results in a final \textsc{EntityCS} corpus comprising 54.5 million English sentences and 104 million entities.

\begin{figure}[th]
\centering
\includegraphics[width=0.9\linewidth]{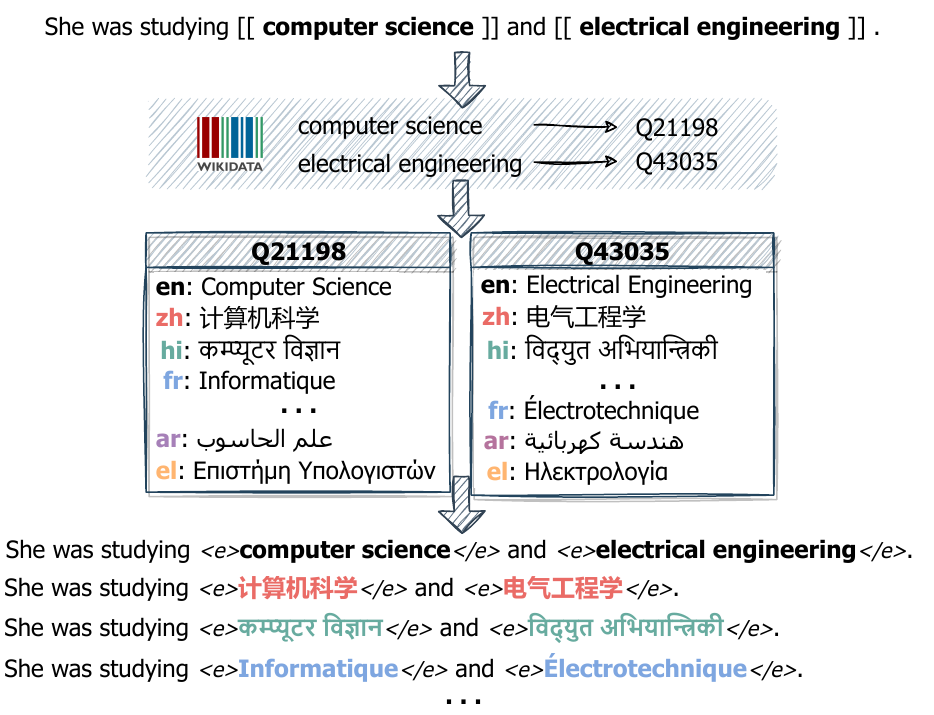}
    \caption{Illustration of generating \textsc{EntityCS} sentences from an English sentence extracted from Wikipedia. Entities in double square brackets indicate wikilinks.}
\label{fig:cs_example}
\end{figure}

As depicted in \autoref{fig:cs_example}, our Code Switching process begins with an English sentence containing wikilinks. Each entity within these links is mapped to its corresponding Wikidata ID, and translations for these entities are retrieved from Wikidata. The selection of target languages for Code Switching is based on the availability of translations for all entities within a given sentence.

We consider a set of 92 target languages (non-English), representing the overlap between languages available in Wikidata and those supported by XLM-R \citep{conneau-etal-2020-unsupervised}, the model utilised for intermediate training. To ensure coherence, all entities in a sentence are code-switched to the same target language, mitigating potential noise from introducing too many languages.

To manage the size of the corpus, we generate up to five entity code-switched sentences for each English sentence. Specifically, if fewer than five languages have translations available for all entities in a sentence, we create \textsc{EntityCS} instances with all available languages. Otherwise, we randomly select five target languages from the candidates. If no candidate languages are found, we retain the sentence in the English corpus without code-switching.

In the final step, we enclose each entity with entity indicators (\texttt{<e>}, \texttt{</e>}). This ensures clear identification of entities within the code-switched sentences.

\begin{figure}
\centering
\includegraphics[width=\textwidth]{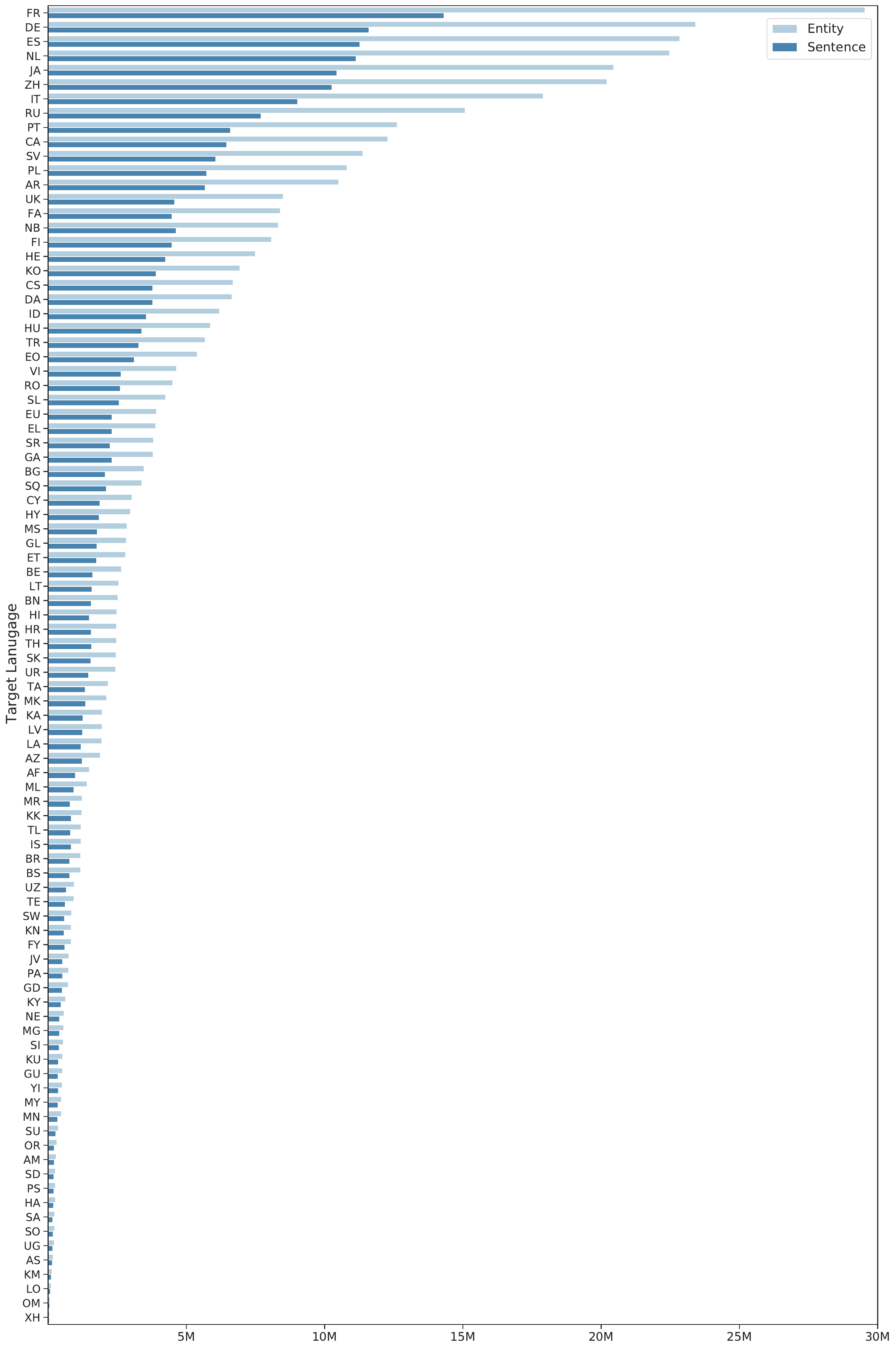}
\caption{Number of Code-Switched Entities and Sentences in the \textsc{EntityCS} corpus.}
\label{fig:corpus_stats}
\end{figure}

The statistical overview of the \textsc{EntityCS} corpus is presented in \autoref{tab:cs_stats}, and a histogram detailing the number of sentences and entities per language (excluding English) in the \textsc{EntityCS} corpus is illustrated in \autoref{fig:corpus_stats}.

\begin{table}[t!]
\centering
\scalebox{1}{
\begin{tabular}{lr}
\toprule
\textsc{Statistic} & \textsc{Count} \\ \midrule
Languages                     & 93             \\
English Sentences             & 54,469,214     \\
English Entities              & 104,593,076    \\
Average Sentence Length       & 23.37          \\
Average Entities per Sentence & 2  \\ \midrule
CS Sentences per EN Sentence     & $\leq$ 5 \\
CS Sentences       & 231,124,422    \\
CS Entities        & 420,907,878    \\

\bottomrule
\end{tabular}
}
\caption{Statistics of the \textsc{EntityCS} Corpus.}
 \label{tab:cs_stats}
\end{table}

\subsection{Masking Strategies}

\begin{figure}[!t]
    \begin{subfigure}{\linewidth}
        \centering
        \includegraphics[width=0.75\linewidth]{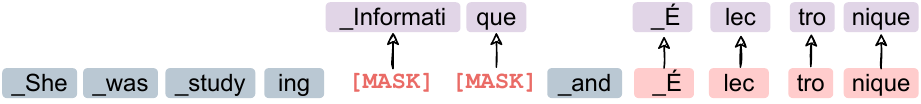}
        \caption{\textbf{W}hole \textbf{E}ntity \textbf{P}rediction (\sc {wep})}
        \label{fig:wep}
         \vspace{2ex}
    \end{subfigure}
 
    \begin{subfigure}{\linewidth}
        \centering
        \includegraphics[width=0.75\linewidth]{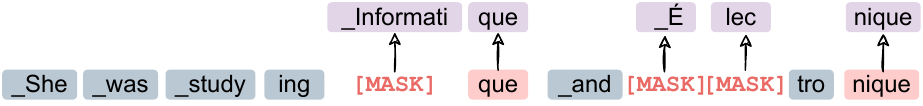}
        \caption{\textbf{P}artial \textbf{E}ntity \textbf{P}rediction (\sc {pep})}
        \label{fig:pep}
          \vspace{2ex}
    \end{subfigure}

    \begin{subfigure}{\linewidth}
        \centering
        \includegraphics[width=0.75\linewidth]{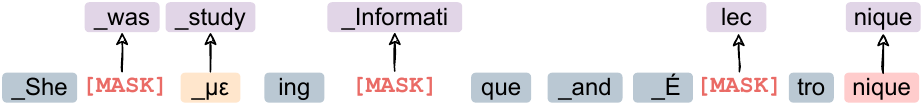}
        \caption{\textbf{P}artial \textbf{E}ntity \textbf{P}rediction with \textbf{MLM} (\sc {pep+mlm})}
        \label{fig:pep_mlm}
    \end{subfigure}
    \caption{Illustration of the proposed masking strategies.
    Random tokens are chosen from the entire vocabulary and thus can be from different languages. (c) shows a case where ``study'' is replaced with a token in Greek.}
    \label{fig:masking_strategies}
\end{figure}

To assess the efficacy of intermediate training on the generated \textsc{EntityCS} corpus, we experiment with various training objectives using an existing pre-trained language model. Initially, we adopt the conventional 80-10-10 Masked Language Modelling (MLM) objective, where 15\% of sentence subwords or tokens serve as masking candidates. Among these, we replace tokens with \texttt{[MASK]} 80\% of the time, with 10\% using random tokens (from the entire vocabulary), and the remaining 10\% left unchanged (Same).

To integrate entity-level cross-lingual knowledge into the model, we introduce Entity Prediction objectives, where we exclusively mask tokens belonging to an entity. By predicting the masked entities in \textsc{EntityCS} sentences, we anticipate the model capturing the semantics of the same entity in different languages. Two distinct masking strategies are proposed for predicting entities: Whole Entity Prediction (\textsc{WEP}) and Partial Entity Prediction (\textsc{PEP}).

In \textsc{WEP}, inspired by \citet{Sun2019ERNIEER} where whole-word masking is also employed, we consider all the words (and consequently subwords or tokens) inside an entity as masking candidates. Subsequently, 80\% of the time, we mask every token inside an entity, leaving 20\% unchanged. Notably, to predict the entire masked entity, we refrain from replacing it with random tokens, as it might introduce noise, leading to the model predicting incorrect entities. After masking entities, we remove the entity indicators \texttt{<e>}, \texttt{</e>} from the sentences before feeding them to the model. \autoref{fig:wep} provides an example of \textsc{WEP}.

For \textsc{PEP}, we also consider all entities as masking candidates. Unlike \textsc{WEP}, we do not enforce tokens belonging to one entity to be either all masked or all unmasked. Instead, each individual entity token is masked 80\% of the time. For the remaining 20\% of masking candidates, we experiment with three different replacements.

\textsc{PEP\textsubscript{mrs}} corresponds to the conventional 80-10-10 masking strategy, where 10\% of the remaining tokens are replaced with random tokens, and the other 10\% are left unchanged.
In \textsc{PEP\textsubscript{ms}}, we remove the 10\% random tokens substitution, predicting only the 80\% masked tokens and 10\% Same tokens from the masking candidates.
In \textsc{PEP\textsubscript{m}}, we further eliminate the 10\% Same tokens prediction, essentially predicting only the masked tokens. An example of \textsc{PEP} is illustrated in \autoref{fig:pep}.
Previous work has demonstrated the effectiveness of combining Entity Prediction with MLM for cross-lingual transfer \citep{jiang-etal-2020-x}. Therefore, we investigate the combination of Entity Prediction objectives with MLM on non-entity tokens. Specifically, when combined with MLM, we lower the entity masking probability ($p$) to 50\% to maintain roughly the same overall masking percentage. \autoref{fig:pep_mlm} illustrates an example of \textsc{PEP} combined with \textsc{MLM} on non-entity tokens.

A summary of the masking strategies is presented in \autoref{tab:masking_strategy}, along with the corresponding masking percentages.

\begin{table}[!t]
\centering
\scalebox{0.95}{
\begin{tabular}{llcccccccc}
\toprule
\multicolumn{2}{c}{\multirow{2}{*}{\sc \parbox{0.2\textwidth}{Masking Strategy}}} 
& \multicolumn{4}{c}{\sc Entity (\%)}  
& \multicolumn{4}{c}{\sc Non-Entity (\%)}  \\   
&& $p$ 
& \textsc{\textbf{m}ask}
& \textsc{\textbf{r}nd}
& \textsc{\textbf{s}ame}

& $p$ 
& \textsc{\textbf{m}ask} 
& \textsc{\textbf{r}nd} 
& \textsc{\textbf{s}ame} \\  \cmidrule(r){1-2}    \cmidrule(lr){3-6} \cmidrule(l){7-10}  \sc {mlm} & &15 & 80  & 10  & 10 & 15  & 80 & 10 & 10 \\
\midrule \sc {wep} &  &100  & 80 & 0 & 20  & 0
&--&--&--
\\
\sc {pep\textsubscript{mrs}}& & 100 & 80 & 10 & 10 &0  
&--&--&--
\\
 \sc {pep\textsubscript{ms}} & & 100 & 80 & 0 & 10 & 0 
&--&--&--
\\
 \sc {pep\textsubscript{m}} &  & 100 & 80 & 0 & 0 & 0 
&--&--&--
\\
\midrule

 \sc {wep} & \multirow{4}{*}{\sc {+ mlm}} & 50& 80 & 0 & 20 &  15  & 80 & 10 & 10 \\
\sc {pep\textsubscript{mrs}} & & 50 & 80 & 10 & 10 &  15  & 80 & 10 & 10 \\
 \sc {pep\textsubscript{ms}} && 50 & 80 & 0 & 10 &  15  & 80 & 10 & 10 \\
 \sc {pep\textsubscript{m}}  & &50 & 80 & 0 & 0 &  15  & 80 & 10 & 10 \\
\bottomrule
\end{tabular}
}
\caption{Summary of the proposed masking strategies. $p$ corresponds to the probability of choosing candidate items (entity/non-entity tokens) for masking.
\textsc{\textbf{m}ask}, \textsc{\textbf{r}nd}, \textsc{\textbf{s}ame} represent the percentage of replacing a candidate with Mask, Random or the Same item.
When combining \textsc{wep/pep} with \textsc{mlm (+mlm)}, we lower $p$ to 50\%.
} 
\label{tab:masking_strategy}
\end{table}

\section{Experimental Setup}

After constructing the \textsc{EntityCS} corpus, we proceed to further train an XLM model, utilising XLM-R-base\footnote{\url{https://huggingface.co/xlm-roberta-base}} for our experiments. We explore the effectiveness of \textsc{WEP}, \textsc{PEP}, \textsc{MLM}, and the joint objectives in intermediate training.

We adopt the sampling strategy proposed by \citet{XLM}, where we down-sample high-resource languages and increase the sampling frequency for low-resource languages. Recognising that semantic features are often emphasised in higher layers of pre-trained language encoders~\citep{tenney-etal-2019-bert, rogers-etal-2020-primer}, we restrict training to the embedding layer and the last two layers of the model. This approach helps prevent catastrophic forgetting, a phenomenon observed in preliminary experiments when updating the entire network. We randomly select 100 sentences from each language to serve as a validation set, measuring perplexity every 10K training steps.

\subsection{Datasets}
As the \textsc{EntityCS} corpus is constructed with Code Switching at the entity level, we expect our models to mostly improve entity-centric tasks.
Thus, we choose the following datasets: WikiAnn~\citep{pan-etal-2017-cross} for NER, X-FACTR~\citep{jiang-etal-2020-x} for Fact Retrieval, MultiATIS++~\citep{xu-etal-2020-end} and MTOP~\citep{li-etal-2021-mtop} for Slot Filling, and XL-WiC \citep{raganato-etal-2020-xl} for WSD.\footnote{The result reported on the XL-WiC for prior work is our re-implementation based on \url{https://github.com/pasinit/xlwic-runs}.}
The details of the datasets are introduced below.

\paragraph*{WikiAnn}\hspace*{-.3cm}~\citep{pan-etal-2017-cross} is a cross-lingual name tagging and linking dataset based on Wikipedia articles, where named entities are annotated as location (\texttt{LOC}), organisation (\texttt{ORG}) and person (\texttt{PER}) tags following the IOB2 format.
The original dataset contains 282 languages. We evaluate our models on the 40 languages from WikiAnn that are included in the XTREME benchmark~\citep{Hu2020XTREMEAM}.

\paragraph*{X-FACTR}\hspace*{-.3cm}~\citep{jiang-etal-2020-x} is a multilingual fact retrieval benchmark similar to LAMA~\citep{petroni-etal-2019-language}.
It probes factual knowledge stored in pre-trained language models by prompt-based fill-in-the-blank cloze queries, covering 23 languages.
X-FACTR includes both single- and multi-token entities, and two decoding methods (independent and confidence-based) are proposed.

\paragraph*{MultiATIS++}\hspace*{-.3cm}~\citep{xu-etal-2020-end} is an expansion of the Multilingual ATIS~\citep{MATIS} dataset, which includes nine languages (English, Spanish, German, French, Portuguese, Chinese, Japanese, Hindi and Turkish)  from four language families (Indo-European, Sino-Tibetan, Japonic and Altaic).
It contains dialogues in a single domain, Air Travel Information Services.
While processing the dataset, we noticed that 14 examples in the test set do not have a matching number of tokens and slot labels, which we ignored during the evaluation.

\paragraph*{MTOP}\hspace*{-.3cm}~\citep{li-etal-2021-mtop} is a Multilingual Task-Oriented Parsing dataset that includes six languages from 11 domains that are related to interactions with a personal assistant.
We use the standard flat labels as reported in \citet{li-etal-2021-mtop}.

\paragraph*{XL-WiC}\hspace*{-.3cm}~\citep{raganato-etal-2020-xl} is a cross-lingual word disambiguation dataset (Word in Context), formed as a binary classification problem. 
Given a target word and two contexts, the goal is to identify if the word is used in the same sense in both contexts. The dataset contains both nouns and verbs as target words, covers 12 languages and was created as an extension to the English WiC dataset~\citep{pilehvar-camacho-collados-2019-wic}.

\subsection{Hyper-Parameter Settings}
\label{sec:hypers}

We introduce below the detailed setting of our intermediate training and downstream tasks fine-tuning.

\subsubsection*{Intermediate Training} 

We use 8 Nvidia V100 32GB GPUs for training our models on the \textsc{EntityCS} corpus, with the Hugging Face library~\citep{wolf-etal-2020-transformers}. 
During fine-tuning, all models were run on a single Nvidia V100 32GB GPU.
We set the batch size to 16 and gradient accumulation steps to 2, resulting in an effective batch size of 256.
For speedup, we employ half-precision (fp16) in the experiments.
In each batch, we allow examples from multiple languages, based on the sampling strategy followed by \citet{XLM}.
We train for a single epoch with a maximum learning rate $5e^{-5}$ and linear decay scheduler, no warmup or weight decay, gradient clipping equal to 1.0, and early stopping if perplexity does not drop after 20 consecutive evaluations (we evaluate every 10K training steps).

\begin{table}[!t]
    \centering
    \scalebox{0.95}{
    \addtolength{\tabcolsep}{0pt}
    \begin{tabular}{l|cccccc}
    \toprule
    \multirow{2}{*}{\sc{Parameter}}  & 
    \multirow{2}{*}
    {\textsc{wiki-ann}}
    & \multicolumn{2}{c}{\textsc{multiatis++}}  & \multicolumn{2}{c}{\textsc{mtop}}
    & 
        \multirow{2}{*}{\textsc{xl-wic}} \\ 
        \cmidrule(lr){3-4} \cmidrule(lr){5-6}

    &     & \textit{SF} & \textit{Joint} &\textit{SF} & \textit{Joint} \\      
        \midrule
        \textsc{Learning Rate}     & $1e^{-5}$ & $3e^{-5}$ & $3e^{-5}$ & $2e^{-5}$ & $3e^{-5}$ &$1e^{-5}$ \\
        \textsc{Warmup Ratio}        &   0.1 & 0.0  &0.0 & 0.1 & 0.1 &0.0 \\
        \textsc{Batch Size}        & 8 & 8 & 8 & 8 & 8 &8 \\
    \bottomrule
    \end{tabular}
    }
    \caption{Best hyper-parameters used for the datasets.}
    \label{tab:best_hyper}
\end{table}

\subsubsection*{Downstream Tasks}

After intermediate training on the \textsc{EntityCS} corpus, we evaluate the zero-shot cross-lingual transfer of the models on downstream tasks by fine-tuning the model on task-specific English training data.
For downstream tasks, we evaluate models on the English validation set five times per epoch following \citet{dodge2020fine}.

For fine-tuning XLM-R-base on WikiAnn, MultiATIS++, MTOP and XL-WiC,
we fix the number of training epochs to 10, gradient clipping to $1.0$, and maximum sequence length to 128. We select the batch size from \(\{8, 32\)\}, learning rate from \(\{1e^{-5}, 2e^{-5}, 3e^{-5}, 4e^{-6}, 5e^{-6}, 6e^{-6}\}\), and warm up ratio from \(\{0, 0.1\}\).

The best hyper-parameters per task are reported in~\autoref{tab:best_hyper}.
We choose the checkpoints with the best performance on the English validation set.
For all experiments except X-FACTR, we fine-tune models with five random seeds and report average performance and standard deviation.

\subsection{Languages for Intermediate Training}
\label{sec:pretraining_langs}
Given the size of the \textsc{EntityCS} corpus, we primarily select a subset from the total 93 languages, that covers most of the languages used in the downstream tasks.
This subset contains 39 languages, from WikiAnn, excluding Yoruba.\footnote{Yoruba is not included in the \textsc{EntityCS} corpus, as we only consider languages that XLM-R is pre-trained on.}
We train XLM-R-base on this subset, and subsequently fine-tune the new checkpoints on the English training set of each dataset, with evaluations spanning all available languages.

\section{Main Results}

The main results are reported in~\autoref{tab:main_res} where we compare models trained on the \textsc{EntityCS} corpus with \textsc{MLM}, \textsc{WEP}, \textsc{PEP\textsubscript{ms}} and \textsc{PEP\textsubscript{ms}+MLM} masking strategies.
For MultiATIS++ and MTOP, we report results of training only Slot Filling (SF), as well as joint training of Slot Filling and Intent Classification (SF/Intent).
\begin{table}[t!]
\centering
\scalebox{0.75}{
\addtolength{\tabcolsep}{-2.5pt}
\begin{tabular}{l|lcccccccl}
\toprule
\multicolumn{1}{l|}{\multirow{3}{*}{\textbf{\textsc{model}}}} 
& {\textsc{\textbf{ner}} \small{(}\textsc{f}\small{1)}}
& \multicolumn{3}{c}{\sc\textbf{fact retr.} \small{(acc.)}}
& \multicolumn{4}{c}{\sc\textbf{slot filling} \small {(}f{\small 1}, f{\small 1/acc.)}} 
& \multicolumn{1}{c}{\sc\textbf{wsd} \small{(acc.)}}   
\\
                       
 & \multicolumn{1}{c}{\sc wikiann} 
& \multicolumn{3}{c}{\sc x-factr}   
& \multicolumn{2}{c}{\sc multiatis++} 
& \multicolumn{2}{c}{\sc mtop} 
& \multicolumn{1}{c}{\sc xl-wic} \\

 & &  \textsc{\textit{all}} & \textsc{\textit{single}} & \textsc{\textit{multi}} & \textsc{\textit{SF}} & \textsc{\textit{SF / Intent}} & \textsc{\textit{SF}} &  \textsc{\textit{SF / Intent}}  & 
\\
\cmidrule(r){1-1} \cmidrule(lr){2-2} \cmidrule(lr){3-5} \cmidrule(lr){6-7} \cmidrule(lr){8-9} \cmidrule(l){10-10}

 \textsc{xlm-r\textsuperscript{prior}} 
&  61.8  & 3.5 & ~~9.4 & 2.6  
&   -- & --
& -- &  -- 
& ~~~~~~58.0
\\

 \textsc{xlm-r\textsuperscript{ours}}
& 61.6~\textsubscript{0.3} 
& 3.5 & ~~9.4 & 2.6 
& 71.8~\textsubscript{2.0} & 73.0~\textsubscript{0.7} / 89.1~\textsubscript{1.0} 
& \textbf{73.2}~\textsubscript{0.9} & 72.5~\textsubscript{0.8} / 86.0~\textsubscript{0.7} 
& ~~~~~~59.1~\textsubscript{1.5}
\\
\midrule

 \textsc{mlm}
&  63.5~\textsubscript{0.5}
& 2.5 & ~~6.4 & 1.7
&  72.1~\textsubscript{2.3} & 74.0~\textsubscript{0.7} / 89.6~\textsubscript{1.4}  
&  72.8~\textsubscript{0.6} & 72.7~\textsubscript{0.3} / \textbf{86.3}~\textsubscript{0.4} 
& ~~~~~~59.3~\textsubscript{0.4}
\\

 \textsc{wep}
& 62.4~\textsubscript{0.7} 
& \textbf{6.1} & \textbf{19.4} & 3.0 
& 71.6~\textsubscript{1.2} & 71.7~\textsubscript{0.8} / 89.7~\textsubscript{1.3} 
& 72.2~\textsubscript{0.6} & \textbf{73.0}~\textsubscript{0.5} / 86.0~\textsubscript{0.4} 
& ~~~~~~\textbf{60.4}~\textsubscript{1.0} 
\\

 \textsc{pep\textsubscript{ms}}
& 63.3~\textsubscript{0.7} 
& 6.0 & 15.0 & \textbf{4.3} 
&  73.4~\textsubscript{1.7}  & \textbf{74.4}~\textsubscript{0.7} / \textbf{90.0}~\textsubscript{0.9} 
&  71.5~\textsubscript{0.7}  & 72.7~\textsubscript{0.6} / 86.1~\textsubscript{0.5} 
& ~~~~~~60.2~\textsubscript{0.9}
\\

 \textsc{pep\textsubscript{ms}+mlm}  
& \textbf{\textbf{64.4}}~\textsubscript{0.5}  
&  5.7 & 13.9 & 3.9 
& \textbf{74.2}~\textsubscript{0.4} & 74.3~\textsubscript{0.8} / 89.0~\textsubscript{0.9} 
& 73.0~\textsubscript{0.3} & 72.5~\textsubscript{0.6} / 85.8~\textsubscript{0.8} 
&  ~~~~~~59.8~\textsubscript{0.8}
\\ 

\bottomrule
\end{tabular}
}
\caption{Average performance across languages on the test set of downstream tasks.
\textsc{xlm-r\textsuperscript{prior}} corresponds to previous reported results with XLM-R-base, referring to \citet{chi-etal-2021-improving} for WikiAnn, \citet{jiang-etal-2020-x} for X-FACTR and \citet{raganato-etal-2020-xl} for XL-WiC. 
\textsc{xlm-r\textsuperscript{ours}} shows our re-implemented results with XLM-R-base.
Results (excluding X-FACTR) are averaged across five seeds with standard deviation reported as a subscript.
}
\label{tab:main_res}
\end{table}

\subsection{Named Entity Recognition} 
In NER, models with CS intermediate training consistently demonstrate improvement on WikiAnn over the baseline, with \textsc{PEP\textsubscript{ms}+MLM} exhibiting a substantial +2.8\% absolute improvement. This outperformance extends to XLM-Align\footnote{\url{https://huggingface.co/microsoft/xlm-align-base}} \citep{chi-etal-2021-improving}, which employs a significant amount of parallel data (see \autoref{tab:xlms-compare}). The conventional MLM objective proves similarly effective with \textsc{PEP}, potentially due to the overlap in entities chosen as masking candidates. However, \textsc{WEP} yields lower performance, suggesting that predicting entire entities from the surrounding context poses greater challenges.

\begin{table}[!t]
\centering
\scalebox{0.85}{
\addtolength{\tabcolsep}{-1.5pt}
\begin{tabular}{l|cccccccccccccc}
\toprule
\sc\textbf{model} & \sc{ar} & \sc{he} & \sc{vi} & \sc{id} & \sc{jv} & \sc{ms} & \sc{tl} & \sc{eu} & \sc{ml} & \sc{ta} & \sc{te} & \sc{af} & \sc{nl} & \sc{en}  \\ 
\midrule
\sc {xlm-r\textsuperscript{ours}} & 44.6 & 51.9 & 68.3 & 48.6 & 59.6 & 63.3 & 72.5 & \textbf{61.2} & 63.2 & 54.3 & 49.3 & 76.3 & 80.7 & 83.4  \\
\sc {pep\textsubscript{ms}}   & 49.6 & 53.0 & 70.0 & 58.5 & \textbf{62.0} & 64.9 & \textbf{75.7} & 59.8 & 63.3 & \textbf{57.7} & 52.1 & 76.4 & 80.9 & 83.8\\
\sc {pep\textsubscript{ms}+mlm} & \textbf{51.5} & \textbf{54.0} & \textbf{70.9} & \textbf{61.1} & 59.3 & \textbf{69.9} & 74.6 & 59.3 & \textbf{66.3} & 57.6 & \textbf{54.8} & \textbf{77.9} & \textbf{81.5} & \textbf{84.2} \\

 \midrule
& \sc{de} & \sc{el} & \sc{bn} & \sc{hi} & \sc{mr} & \sc{ur}  & \sc{fa} & \sc{fr} & \sc{it} & \sc{pt} & \sc{es} & \sc{bg} & \sc{ru} & \sc{ja} 
\\ \midrule
\sc {xlm-r\textsuperscript{ours}} & 75.4 & 74.2 & 67.9 & 68.3 & 61.8 & 55.8  & 47.6 & 78.0 & 78.2 & 78.9 & 76.2 & 77.3 & 63.9 & 22.9  \\ 
\sc {pep\textsubscript{ms}}  & 75.1 & 76.3 & 72.5 & 70.1 & \textbf{66.8} & 61.5 & \textbf{55.6} & 78.8 & 78.5 & 78.6 & 75.8 & 78.0 & 66.4 & 21.3    \\
\sc {pep\textsubscript{ms}+mlm} & \textbf{75.5} & \textbf{77.1} & \textbf{74.6} & \textbf{70.7} & 66.3 & \textbf{65.9} & 54.2 & \textbf{79.5} & \textbf{78.9} & \textbf{80.1} & \textbf{78.2} & \textbf{79.6} & \textbf{67.7} & \textbf{23.2} \\

 \midrule
 & \sc{ka} & \sc{ko} & \sc{th} & \sc{sw} & \sc{yo} & \sc{my} & \sc{zh} & \sc{kk} & \sc{tr} & \sc{et} & \sc{fi} & \sc{hu} \\ \midrule
\sc {xlm-r\textsuperscript{ours}}& 66.4 & 48.8 & 4.3 & \textbf{68.3} & 45.4 & 52.7 & 27.7 & 44.2 & 76.9 & 72.4 & 75.6 & 76.9 \\ 
\sc {pep\textsubscript{ms}}  & 67.0 & 50.2 & \textbf{4.6} & 66.9 & 44.7 & 55.2 & 26.9 & 48.9 & 77.4 & \textbf{73.4} & 76.6 & 77.8 \\
\sc {pep\textsubscript{ms}+mlm}   & \textbf{68.2} & \textbf{52.1} & 4.0 & 66.4 & \textbf{48.4} & \textbf{56.1} & \textbf{29.8} & \textbf{52.0} & \textbf{78.6} & 71.9 & \textbf{76.8} & \textbf{78.8}\\

\bottomrule
\end{tabular}
}
\caption{F1-score per language on the WikiAnn test set. 
Results are averaged across five seeds.
}
 \label{tab:ner_all_langs}
\end{table}

Performance per language for \textsc{PEP\textsubscript{ms}} and \textsc{PEP\textsubscript{ms}+MLM} is detailed in~\autoref{tab:ner_all_langs}.\footnote{Per-language results on WikiAnn for other models are reported in \autoref{tab:ner_all_langs_others} in \autoref{sec:entitycs_additional_results}.} Notably, nearly all languages benefit from training on the \textsc{EntityCS} corpus. In the optimal setting, \textsc{PEP\textsubscript{ms}+MLM}, languages AR, ID, and UR exhibit the most substantial improvement (approximately +10\% in AR and ID). However, EU and SW result in lower performance compared to the baseline.

\begin{figure}[!t]
    \centering
    \includegraphics[width=0.9\linewidth]{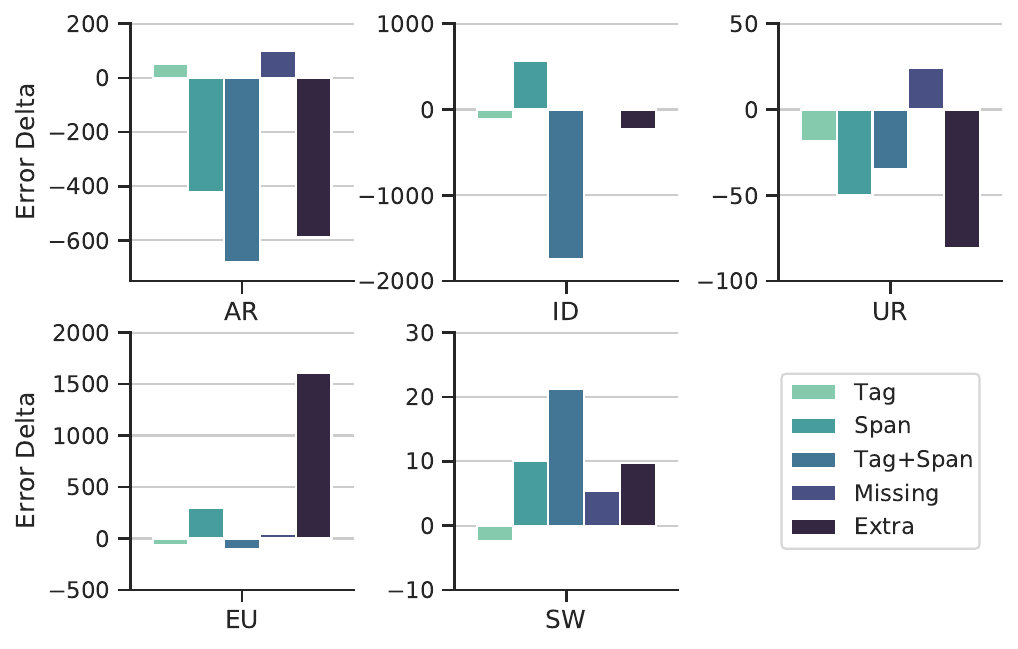}
    \caption{Error Delta (lower is better) for different types of errors in the WikiAnn test set between vanilla XLM-R-base and \textsc{pep\textsubscript{ms}+mlm}.
We show error count differences for AR, ID and UR, the three languages with the largest F1-score improvement, as well as EU and SW, the two languages that underperform the baseline.}
\label{fig:error_types}
\end{figure}

Given this, we conduct a detailed analysis of the \textsc{PEP\textsubscript{ms}+MLM} model, specifically focusing on NER errors categorised into five types: Tag, Span, Tag+Span, Missing Extraction, and Extra Extraction. In this context, Missing Extraction denotes instances where the model fails to identify an entity, while Extra Extraction refers to errors where a non-entity is incorrectly predicted as an entity. For this analysis, we select EU and SW (with lower F1-scores than the baseline),\footnote{Thai is excluded due to problematic tokenisation in WikiAnn, where everything is tokenised into individual \textit{characters}.} as well as AR, ID, and UR (languages with the most significant improvement). The delta bar plot in~\autoref{fig:error_types} illustrates the comparative analysis.

Compared to the baseline, AR, ID, and UR consistently improve in Tag+Span and Extra Extraction categories. All except ID show enhancement in Span detection, while most languages exhibit poorer performance in Missing Extraction. On the flip side, EU and SW result in slightly worse Span and Extra Extraction errors.

We further investigate the reasons for this behaviour. 
In ID, we observe that around 80\% of the Span errors are due to additionally identifying the token \textit{``ALIH''} (means ``moving'', ``changing'' in English) as the start of an entity.
For example, for the input [\textit{``ALIH''}, \textit{``Indofood''}, \textit{``Sukses''}, \textit{``Makmur''}], the gold entity is \textit{``ORG: Indofood Sukses Makmur''}, whereas the model predicts \textit{``ORG: ALIH Indofood Sukses Makmur''}.
This pattern is also observed in XLM-R, accounting for 68\% of the span errors.
Consultation with a native speaker reveals this may be an inaccuracy of the dataset (\textit{``ALIH''} should not appear before the actual entities in WikiAnn).

As for EU, a lower overlap between WikiAnn entities and Wikipedia (only 47\% compared to the average of 57\% across all WikiAnn languages) might explain the prediction of additional entities not present in the dataset.

\subsection{Fact Retrieval}
For X-FACTR, models trained with Entity Prediction consistently outperform the baseline, while \textsc{MLM} performs worse than vanilla XLM-R, as anticipated. Notably, \textsc{WEP} achieves the best results for single-token classification, showcasing a remarkable +10\% gain over XLM-R. On the other hand, \textsc{PEP\textsubscript{ms}}, which involves masking part of the entity tokens for prediction, excels when dealing with multi-token entities. Interestingly, models trained on large parallel data, such as InfoXLM\footnote{\url{https://huggingface.co/microsoft/infoxlm-base}} \citep{chi-etal-2021-infoxlm} and XLM-Align~\citep{chi-etal-2021-improving}, demonstrate poor performance with single-token accuracy of 3\% and 5\%, respectively, and less than 1\% multi-token accuracy. This discrepancy can be attributed to their focus on sentence-level alignment. Results per language for X-FACTR are available in~\autoref{tab:xfactr_results} of~\autoref{sec:entitycs_additional_results}.

\subsection{Slot Filling}
In the case of SF-only training, the most effective model is \textsc{PEP\textsubscript{ms}+MLM}, showcasing a +2.4\% gain over XLM-R on MultiATIS++. This performance is competitive with the best result from XLM-Align (74.4, see \autoref{sec:entitycs_additional_results}). Conversely, no improvements are observed in MTOP over the baseline. A manual inspection of the dataset reveals that this disparity can be attributed to domain differences. MultiATIS++ contains entities such as city names, whereas MTOP consists of dialogues with a personal assistant, involving tasks like setting up reminders, where fewer entities occur, limiting the benefits of entity-centric CS training.

When jointly optimising SF and Intent Classification, models also demonstrate improvements over the baseline on SF (+1.4\% for MultiATIS++ and +0.5\% for MTOP), albeit with lower gains. We speculate that the additional intent labels offer complementary information to the task, mitigating the impact of external information.

\begin{table}[!t]
\centering
\scalebox{0.95}{
\addtolength{\tabcolsep}{0pt}
\begin{tabular}{l|ccccc|c||ccc|c}
\toprule
\multirow{2}{*}{\sc\textbf{model}}   

& \multicolumn{6}{c||}{\sc {Latin Script}} 
& \multicolumn{4}{c}{\sc {Non Latin Script}} 
  \\
  \cmidrule(lr){2-7} \cmidrule(l){8-11} 
                
  & \textsc{es} & \textsc{de} 
& \textsc{fr}  & \textsc{pt} & \textsc{tr}  &  \textsc{\textit{avg}}  
& \textsc{zh}   & \textsc{ja} 
& \textsc{hi}
& \textsc{\textit{avg}}    
\\
\midrule

\sc {xlm-r\textsuperscript{ours}}
 & {81.5} & 79.8 
& 74.8 & {76.5} & 43.0   & 71.1
& 77.2 & 56.8 
& 50.6 
& 61.5
\\
\midrule
\sc {mlm} 
 & 78.8 & 78.0 & 74.4 & 74.6 & 39.7 & 69.1 & 76.4 & 70.3 & 61.5 & 69.4

\\
\sc {pep\textsubscript{ms}}
 & 79.3 & 79.7 & 75.3 & 76.2 & 45.3 & 71.1 &77.8 & 69.0 & 62.9 & 69.9
 \\
\sc {pep\textsubscript{ms}+mlm} 
& 81.3 & 81.4 & 78.2 & 76.1 & 42.1 & \textbf{71.8} & 78.8 & 68.8 & 65.8 & \textbf{71.1}
\\
\bottomrule
\end{tabular}}
\caption{F1-score (average across five seeds) for languages with Latin and Non-Latin script on MultiATIS++ test set when using SF-only training.}
\label{tab:multiatis_lang}
\end{table}

We subsequently categorise languages in MultiATIS++ based on whether they share the same script as English (Latin) and scrutinise their performance in SF-only training. As evident in \autoref{tab:multiatis_lang}, models trained on the \textsc{EntityCS} corpus exhibit significant improvements in languages with non-Latin scripts, achieving an average gain of +9.6\%. This suggests that entity-focused training enables models to capture information that proves particularly beneficial for languages featuring scripts different from English. Language-specific results on MultiATIS++ can be found in \autoref{tab:multiatis_main} of \autoref{sec:entitycs_additional_results}.

\subsection{Word Sense Disambiguation}

In the case of XL-WiC, we note the modest improvement across tasks, with \textsc{WEP} yielding the best performance at +1.3\% over the baseline. This trend can be attributed to the task's nature, where our entity-based training objectives assume that disambiguation has already been addressed and is treated as implicit external information. Notably, in the evaluation of XLM-Align, which leverages parallel data, we observe no enhancement in disambiguating word-level semantics across languages (57\% accuracy).
Per-language performance on WSD is included in \autoref{tab:xlwic_results} of \autoref{sec:entitycs_additional_results}.

\subsection{Comparison with Models trained with Parallel Data}
We summarise the comparison with InfoXLM and XLM-Align in \autoref{tab:xlms-compare}. Although these results are not fair comparisons for \textsc{EntityCS}, given that InfoXLM and XLM-Align utilise parallel data, it is evident that \textsc{EntityCS} consistently demonstrates competitive or superior performance across the board.

\begin{table}[th]
\centering
\vspace{1.5ex}
\scalebox{0.95}{
\addtolength{\tabcolsep}{-2pt}
\begin{tabular}{l|lcccccc}
\toprule
\multicolumn{1}{l|}{\multirow{3}{*}{\textbf{\textsc{model}}}} 

& {\sc\textbf{ner} \sc(f{\small 1})} 
& \multicolumn{3}{c}{\sc\textbf{fact retr.} (acc.)}
& \multicolumn{2}{c}{\sc\textbf{slot filling} (f{\small 1})} 
& \multicolumn{1}{c}{\sc\textbf{wsd} (acc.)}   
\\
                       
 & \multicolumn{1}{c}{\sc wikiann} 
& \multicolumn{3}{c}{\sc x-factr}   
& \multicolumn{1}{c}{\sc multiatis++} 
& \multicolumn{1}{c}{\sc mtop} 
& \multicolumn{1}{c}{\sc xl-wic} \\

 & &  \textsc{\textit{all}} & \textsc{\textit{single}} & \textsc{\textit{multi}} & \textsc{\textit{SF}} & \textsc{\textit{SF}}  & 
\\
\cmidrule(r){1-1} \cmidrule(lr){2-2} \cmidrule(lr){3-5} \cmidrule(lr){6-6} \cmidrule(lr){7-7} \cmidrule(l){8-8}
\textsc{xlm-r\textsuperscript{ours}}
& 61.6~\textsubscript{0.28} 
& 3.5 & ~~9.4 & 2.6 
& 70.6~\textsubscript{1.55} 
& 72.3~\textsubscript{0.98}
& 59.1~\textsubscript{1.52}
\\
 \textsc{infoxlm} *
& 62.8 &  1.1  & ~~3.3 &  0.6
&  73.9 \textsubscript{1.95} 
& 74.7 \textsubscript{0.30}   
& 56.9~\textsubscript{0.81} 

\\
 \textsc{xlm-align} **
& 63.7 &  1.5  & ~~5.0 & 1.0
&  \textbf{74.4} \textsubscript{0.29} 
& \textbf{74.9} \textsubscript{0.36}   
& 56.9~\textsubscript{1.22} 
\\

\midrule

\textsc{wep}
& 62.4~\textsubscript{0.68} 
& \textbf{6.1} & \textbf{19.4} & 3.0 
& 71.6~\textsubscript{1.20}
& 73.2~\textsubscript{0.89}
& \textbf{60.4}~\textsubscript{0.97} 
\\

\textsc{pep\textsubscript{ms}}
& 63.3~\textsubscript{0.70} 
& 6.0 & 15.0 & \textbf{4.3}
& 73.4~\textsubscript{1.70}  
& 71.5~\textsubscript{0.67}  
& 60.2~\textsubscript{0.85}
\\

 \textsc{pep\textsubscript{ms}+mlm}  
& \textbf{64.4}~\textsubscript{0.50}  
& 5.7 & 13.9 & 3.9 
& 74.2~\textsubscript{0.43}
& 73.0~\textsubscript{0.33} 
& 59.8~\textsubscript{0.75}
\\ 

\bottomrule
\end{tabular}
}
\caption{Comparison with models using parallel data. Results for InfoXLM and XLM-Align are obtained from ~\citet{chi-etal-2021-infoxlm} and~\citet{chi-etal-2021-improving}, respectively.}
\label{tab:xlms-compare}
\end{table}

\section{Analysis}
We conduct further analysis on different masking strategies, examining their impact across languages and training steps during intermediate training on the \textsc{EntityCS} corpus. Our primary focus is on WikiAnn, which encompasses the largest number of languages from the datasets we evaluate.

\subsection{Performance vs Languages in Intermediate Training}

For WikiAnn, X-FACTR, and MultiATIS++, we conduct additional experiments by training \textsc{MLM}, \textsc{WEP}, and \textsc{PEP\textsubscript{ms}+MLM} with varying numbers of languages in the \textsc{EntityCS} corpus. We explore three scenarios: using English only (no Code Switching), employing the subset of 39 languages (as mentioned in \autoref{sec:pretraining_langs}), and incorporating all 93 languages.

\begin{table}[t!]
\centering
\scalebox{0.95}{
\addtolength{\tabcolsep}{0pt}
\begin{tabular}{ll|ccrcrr}
\toprule
\multicolumn{2}{l|}{\multirow{2}{*}{\textbf{\textsc{model}}}} &\multirow{2}{*}{\textsc{wikiann}} 
 & \multicolumn{3}{c}{\textsc{x-factr}}&
\multicolumn{2}{c}{\textsc{multiatis++}} \\ 
   \cmidrule(lr){4-6}  \cmidrule(l){7-8}
\multicolumn{2}{l|}{}         &          &  \textsc{\textit{all}} & \textsc{\textit{single}} & \textsc{\textit{multi}}    &  \textsc{\textit{SF}} &  \textsc{\textit{SF / Intent}}  \\ \midrule
\multicolumn{2}{l|}{\sc {xlm-r\textsuperscript{ours}}}
& 61.6 & 3.5 & 9.4 & 2.6
& 70.6
& 73.0 / 88.9

\\ \midrule
\multirow{3}{*}{\textsc{mlm}}            
& \textsc{en}        
&    61.0
&   1.1  & 2.7  & 0.7     
&  71.5 &   72.1 / 89.6    \\

& \textsc{39}
&   \textbf{63.5}   
& 2.6 & 6.4 &1.7       
& 72.5 & \textbf{73.8} / \textbf{90.2}        \\
 
& \textsc{93}  
&  63.3    
&  \textbf{2.7} & \textbf{6.8}  &  \textbf{1.8} 
& \textbf{72.7}     &    73.4 / 89.6      
\\ \midrule

\multirow{3}{*}{\textsc{wep}}            
& \textsc{en}        
& 61.9   &     
3.3 & 8.5 & 1.6    
&    \textbf{71.8}   &    72.2 / \textbf{91.1}     \\

& \textsc{39}  &  \textbf{62.4}   
&  \textbf{6.1} & \textbf{19.4} &  \textbf{3.0}      
&    71.1  & 71.7 / 89.7       \\
 
& \textsc{93}   &  59.4    
&  5.8 & 18.6 & 2.7
&  70.4          
&    \textbf{72.9} / 90.3    \\ \midrule

\multirow{3}{*}{\sc \parbox{0.05\textwidth}{\textsc{pep\textsubscript{ms}
+mlm}}}
& \textsc{en}      
&     61.2 & 
2.7 & 6.6 & 1.6   
&  71.3       
&    72.3 / 90.7    \\

& \textsc{39}      
&   \textbf{64.4} &  
\textbf{5.7}   &   \textbf{13.9}  &     \textbf{3.9}
&  \textbf{73.4}      
&   \textbf{74.4} / 90.0       \\
 
& \textsc{93}      
&     63.6
&  5.5  &   13.2  &  3.8   
&   72.8   &   72.7 / \textbf{90.8}      \\ 
\bottomrule
\end{tabular}
}
\caption{Results (average over five seeds) with a different number of pre-training languages.}
 \label{tab:pre-training_langs}
\end{table}

As shown in \autoref{tab:pre-training_langs}, models trained solely on English sentences do not exhibit a noticeable improvement in average performance across languages (except for Intent Classification accuracy and WikiAnn with \textsc{WEP} over the baseline). However, English-only training proves beneficial for English performance, with an average gain of +23.1\% for single-token and +5.6\% for multi-token predictions in X-FACTR over the baseline XLM-R, using \textsc{WEP}.

When trained with all 93 languages, models employing all masking strategies demonstrate improved performance compared to XLM-R. However, the prevailing trend indicates that training on a more restricted set of languages generally results in better performance. This suggests that incorporating a broader array of languages may not necessarily contribute to superior results, underscoring the non-trivial challenge of scaling to too many languages. Notably, the subset of 39 languages already covers most languages in the downstream tasks. In scenarios where additional languages are introduced, an increase in the number of pre-training languages could potentially lead to improved performance.

\subsection{Performance vs Training Steps}

\begin{figure}[t!]
\centering
    \includegraphics[width=\linewidth]{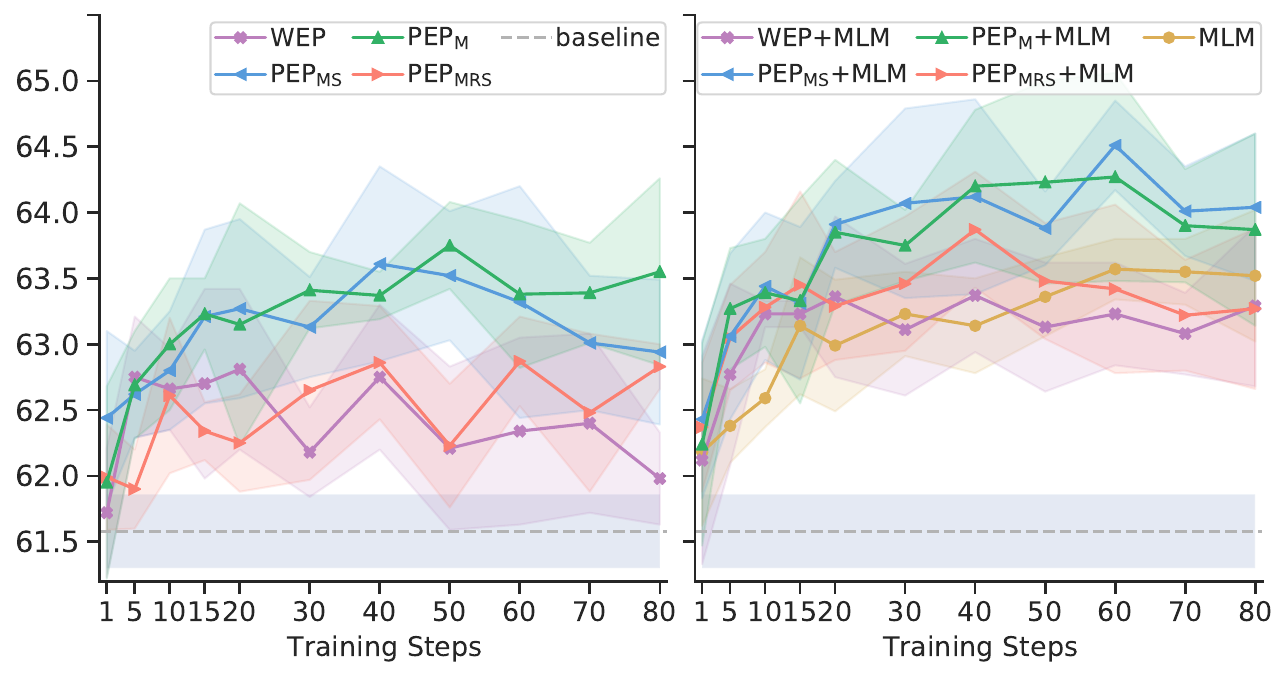}
    \caption{F1-score comparison on WikiAnn test set (average across five seeds) as a function of the number of training steps (in \textit{ten} thousands) with various masking objectives. EP-only strategies are on the left, and EP + MLM strategies are on the right.}
\label{fig:ner39}
\end{figure}

\autoref{fig:ner39} provides a comparative analysis of different training objectives on the \textsc{EntityCS} corpus, showcasing their performance across the number of training steps on the WikiAnn test set. The figure reveals that most masking strategies reach a plateau after the middle of training. Notably, objectives involving MLM training exhibit a clear performance increase across the board, underscoring the benefits of joint training for entities and non-entities, not only in terms of performance but also in terms of smoother learning curves. Importantly, all objectives consistently outperform the baseline throughout the training process.

It's worth noting that the observed gains from Code Switching intermediate training do not stem from additional training data. This is substantiated by \autoref{tab:pre-training_langs}, where the WikiAnn F1-score trained on English-only sentences (without Code Switching) shows that additional English-only training data does not improve upon the XLM-R baseline (61.6). This observation is further supported by \autoref{fig:ner39} at step 200K, corresponding to the steps required for training on the English-only sentences. At this point, all models exhibit an F1 score above 62.3, indicating that the NER performance gain can be attributed to the design of the \textsc{EntityCS} corpus and the associated training objectives.

\subsection{Random and Same Token Prediction}

We further investigate the influence of Random token substitution and Same token prediction when implementing \textsc{PEP}. By comparing \textsc{PEP\textsubscript{mrs}}, \textsc{PEP\textsubscript{ms}}, and \textsc{PEP\textsubscript{m}} in \autoref{fig:ner39}, a clear trend emerges: incorporating Random token substitution in \textsc{PEP} leads to inferior performance while omitting the prediction of Same tokens has a negligible effect.

Intuitively, Random token substitution introduces the risk of predicting an incorrect entity, potentially undermining the model's learning by steering it towards inaccurate predictions from the randomly replaced tokens. On the contrary, predicting the Same token involves a more straightforward task of replicating the input entity to the output. Consequently, models appear to neither gain nor lose performance with or without the Same token prediction.

These observations align with the findings in \citet{wettig-etal-2023-mask}, albeit their focus on monolingual settings. A similar pattern can also be observed when combining \textsc{PEP} with \textsc{MLM}.

\subsection{Entity Masking Percentage}

To assess the impact of the percentage of the entity masking candidates during training, we increment the masking probability ($p$) from 50\% to 80\% and 100\% for the best model, \textsc{PEP\textsubscript{ms}+MLM}, and evaluate its effects on the WikiAnn test set.

Notably, we observe that a further increase in the masking percentage leads to a performance decline, from a 64.4\textsubscript{$\pm{0.5}$} F1-score to 63.8\textsubscript{$\pm{0.7}$} at 80\%, and 64.0\textsubscript{$\pm{0.4}$} at 100\%. We posit that masking too many tokens renders the task of entity prediction from the remaining context more challenging. However, the minimal performance difference between percentages is noteworthy, likely stemming from the observation that, on average, only two entities exist per sentence, as demonstrated in \autoref{tab:cs_stats}.

\section{Conclusion and Future Work}

In summary, this chapter underscores the valuable role of multilingual knowledge sourced from Wikipedia and Wikidata in enhancing zero-shot transfer learning for entity-oriented tasks through entity-level Code Switching. The creation of the \textsc{EntityCS} corpus, leveraging English Wikipedia and Wikidata, facilitates the substitution of entities in wikilinks with their counterparts in multiple languages. By introducing entity-oriented training objectives, we consistently improve performance across various datasets, including Named Entity Recognition, Fact Retrieval, and Word Sense Disambiguation, surpassing baseline models and outperforming previous methods relying on extensive parallel data. Notably, our approach demonstrates competitive performance in Slot Filling.

Our findings highlight the task-specific optimal nature of different masking strategies, revealing that Whole Entity Prediction excels when emphasising single-token factual knowledge, while Partial Entity Prediction is particularly beneficial for entity typing and multi-token factual retrieval. Simultaneously predicting non-entity and entity tokens proves advantageous for tasks where the entire input context plays a crucial role, with a notable impact on languages with non-Latin scripts.

The generic nature of our corpus construction process allows scalability to a more extensive range of languages, underlining the broader applicability of our methodology. Further research could explore Code Switching beyond entities, encompassing verbs and phrases.
\comment{[Chapter 4 - Correction point 3/6]~}\add{However, it is important to acknowledge the potential challenges associated with automatically code-switching parts of sentences beyond entities, particularly in the absence of readily available aligned corpora. We hypothesise that a specifically trained module is likely required to discern suitable target candidates, especially when multiple translations are available. This module would need to understand the context of the sentence and ensure the selection of the most appropriate option for code-switching that does not alter the semantic meaning or the syntax of the original sentence.}
On the other hand, future works may consider incorporating additional sources beyond Wikipedia and Wikidata (which requires an entity linking module), which provide insights into the approach on the more general domains. Overall, our study reinforces the potential of leveraging multilingual knowledge sources in advancing cross-lingual transfer learning.

\section{Limitations}

The limitations of this chapter are outlined as follows:

\begin{itemize}
\item \textbf{Morphological Inflexion Oversight}: Before code-switching an entity, its morphological inflexion is not verified. This can potentially introduce errors, as the form of the code-switched entity might not align with the surrounding context (e.g., plural form). Addressing this issue is crucial for future versions of the corpus.
\item \textbf{Language Diversity Constraint}: The diversity of languages in the \textsc{EntityCS} corpus is constrained to the overlap between WikiData and XLM-R pre-training languages. While this decision facilitates a more robust model comparison, there is potential to enrich the corpus with additional languages not covered by XLM-R.

\item \textbf{Task Specificity in Evaluation}: The proposed approach's primary evaluation is centred on entity-centric tasks. The exploration of broader natural language understanding tasks, such as Natural Language Inference, is feasible, and investigating the impact on such tasks may provide more insights into the generalisability of the models.

\item \textbf{Unidirectional Code-Switching}: \comment{[Chapter 4 - Correction point 5/6]~}\add{Code-switching is performed only from English to other languages while maintaining the context in English. Although the choice is based on the abundance of English resources for studying cross-lingual transfer, exploring bidirectional code-switching, particularly from non-English articles to English, is a promising direction for future research. We anticipate that a model trained using bidirectional code-switching could yield similar enhancements, as observed in prior work by \citet{jiang-etal-2020-x}, however, one limiting factor may be the availability of Wikipedia articles in very low-resource languages.}

\item \textbf{Model Size}: The experiments were conducted solely with base-sized models for speed considerations. Extending these experiments to larger models is a logical progression that could provide valuable insights into scalability and performance improvements.

\end{itemize}
\chapter{Faithful and Robust Knowledge Extraction on the Web}

\label{WebIE} 

In the previous two chapters, we explored how structured knowledge enhances NLP applications, including fake news detection and multilingual entity-centric tasks. This chapter focuses on investigating faithful and robust methods for extracting information or structured knowledge from web text.

The main content of the chapter is based on the paper ``\textsc{WebIE}: Faithful and Robust Information Extraction on the Web'' 
\citep{whitehouse-etal-2023-webie} 
published in \textit{Proceedings of the 61st Annual Meeting of the Association for Computational Linguistics (ACL 2023)}.

\section{Background and Introduction}

Information Extraction (IE) is the task of extracting structured information from unstructured text, typically presented in the form of triples \textit{<subject, relation, object>}. 
It is essential for many Natural Language Processing applications such as knowledge base population, question answering, faithful summarisation, and fake news detection \citep{trisedya-etal-2019-neural,huguet-cabot-navigli-2021-rebel-relation, narayan-etal-2021-planning, whitehouse2022evaluation}.

Closed IE systems, specifically those extracting triples with predefined entities and relations from a knowledge base (KB), require two essential pieces of information: (i) the entities mentioned in the text and (ii) the relations between each pair of entities. Due to the expense of annotations, most existing IE datasets, like WikiNRE \citep{trisedya-etal-2019-neural} or REBEL \citep{huguet-cabot-navigli-2021-rebel-relation}, are constructed using Wikipedia. Entities are linked through wikilinks, and relations are automatically extracted using a distant supervision (DS) approach \citep{mintz-etal-2009-distant} based on a KB such as Wikidata, which assumes that if two entities share a relation in a KB, sentences mentioning both entities express that relation.

While models trained only on this fact-rich domain\footnote{We use the term \textit{domain} to refer to the URL domain.} have shown to be useful for IE applications, they have limited capacity when applied to extracting information in other web domains, which often contains noisy text or text without any factual information.
For example, AllenAI's C4 dataset,\footnote{We use the dataset from \url{https://huggingface.co/datasets/allenai/c4.}} an open-sourced version of Google's C4 \citep{JMLR:v21:20-074} dataset based on Common Crawl, demonstrate this challenge.
Our analysis using the DS approach reveals that fewer than 15\% of the sentences contain triples (refer to \autoref{sec:auto}). \comment{[Chapter 5 - Correction point 1/4]~}\add{In other words, the remaining sentences extracted from C4 either do not contain entities or do not have entities that share a relation. In such cases, faithful IE models are expected \textit{not} to output triples.} However, a state-of-the-art generative IE model, GenIE \citep{josifoski-etal-2022-genie}, trained on REBEL, generates triples for nearly every sentence, leading to a high false positive rate and hallucination issues.

To address these challenges and facilitate future work on IE on the web, we present \textsc{WebIE}, the first large-scale, entity-linked closed IE dataset collected from web sources.
The \textsc{WebIE} dataset is collected from the 200 most frequent URL domains from the C4 dataset. First, we use ReFinED \citep{ayoola-etal-2022-refined}, a state-of-the-art Entity Linking (EL) model to identify mention spans of the entities and link them to Wikidata.
We then apply the DS approach to extract triples and use a Natural Language Inference (NLI) model to filter out triples not expressed by the sentence.
We also include negative examples, i.e., sentences without any factual information, to better reflect the data on the web.
Our final dataset consists of 1.6M sentences, and we annotate a subset of \mytilde21K triples through crowdsourcing.
The annotated set is exclusively used as part of the test set to allow more reliable evaluation.
Finally, we introduce m\textsc{WebIE}, which contains human-corrected translations of the annotated version of \textsc{WebIE} in four languages: French, Spanish, Portuguese, and Hindi.

Previous studies have highlighted the superiority of generative models over discriminative pipelines, which often suffer from accumulative errors due to separate Entity Linking and Relation Extraction (RE) steps \citep{mesquita-etal-2019-knowledgenet, trisedya-etal-2019-neural, josifoski-etal-2022-genie}. Therefore, we primarily benchmark \textsc{WebIE} with generative, transformer-based encoder-decoder models, BART \citep{lewis-etal-2020-bart} and mBART \citep{tang-etal-2021-multilingual}. The latter is used for zero-shot cross-lingual transfer performance evaluation on m\textsc{WebIE}.

\begin{figure}[!t]
    \begin{subfigure}{\linewidth}
        \centering
        \includegraphics[width=0.95\linewidth]{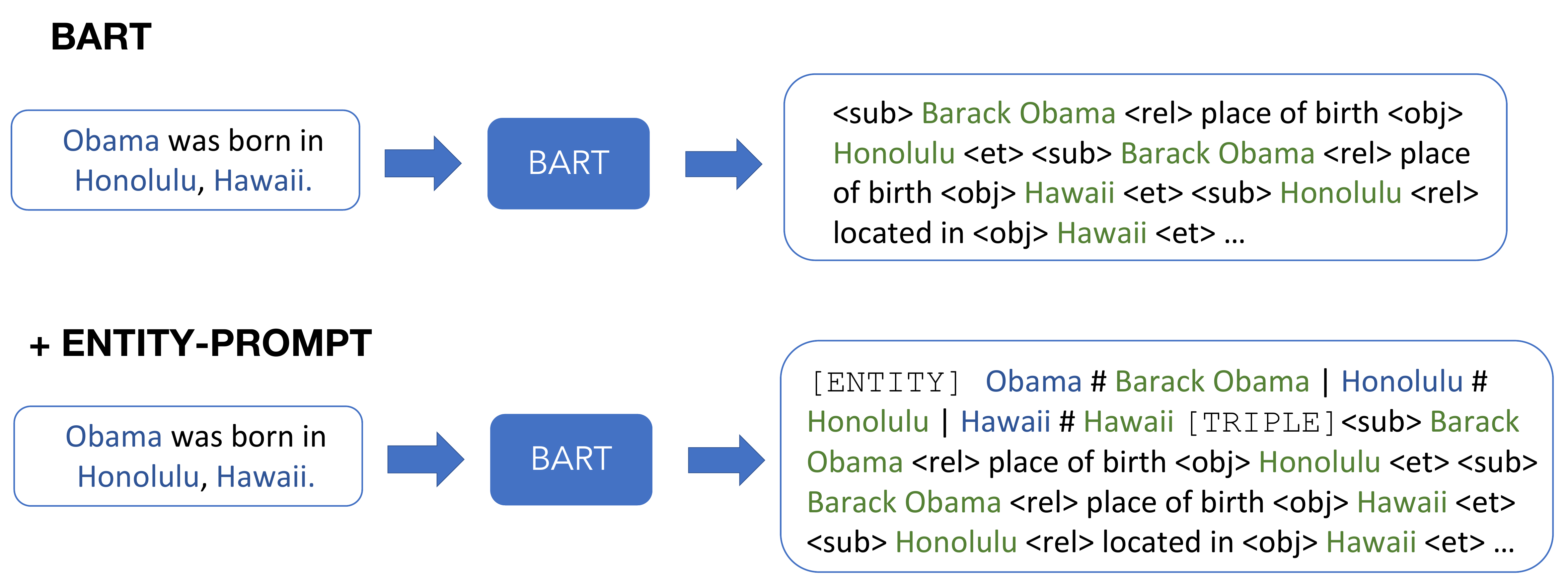}

        \label{fig:m1}
    \end{subfigure}
 
    \begin{subfigure}{\linewidth}
        \centering
        \includegraphics[width=0.95\linewidth]{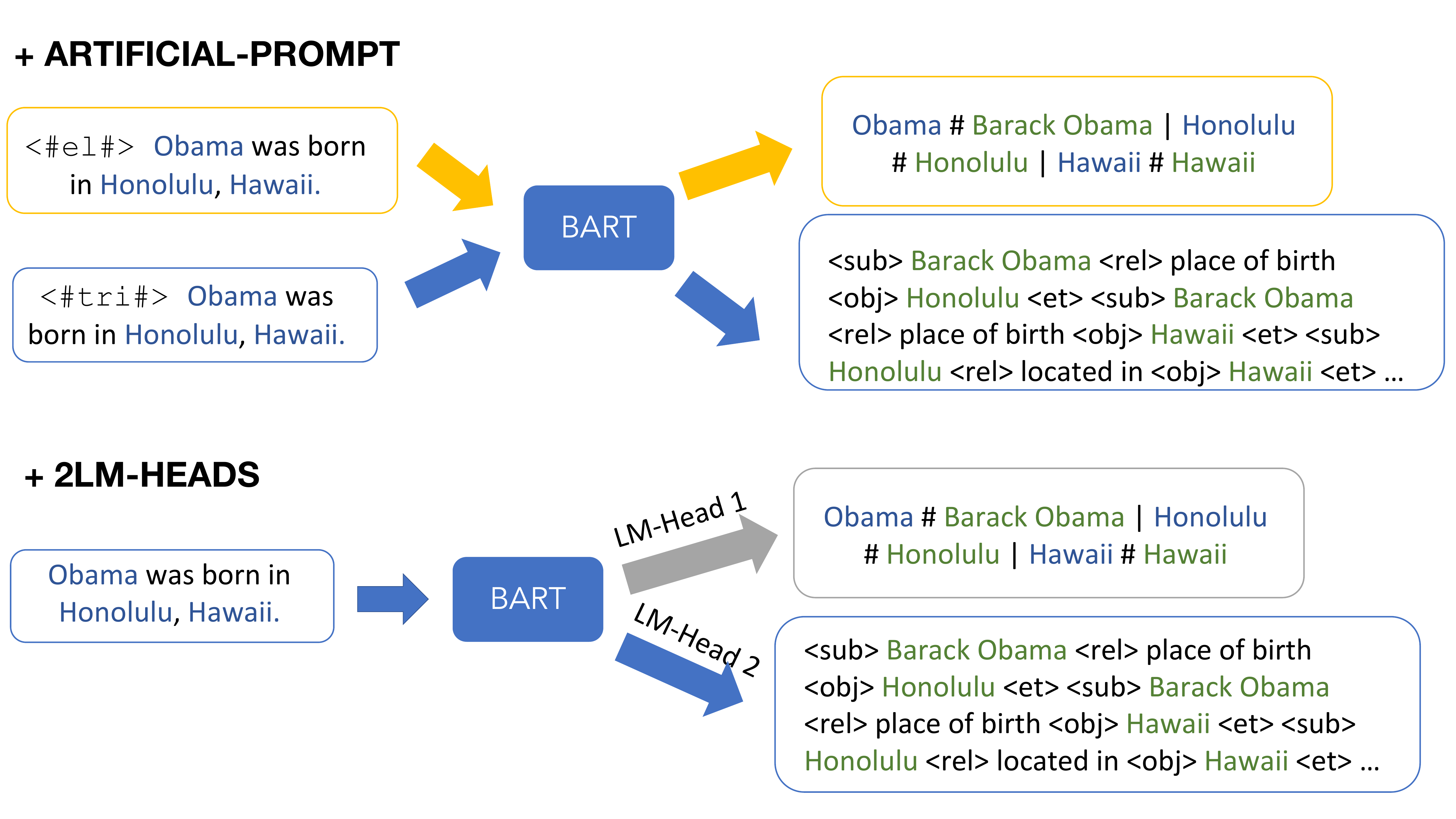}
        \label{fig:m2}
    \end{subfigure}
    \caption{Illustration of the training strategies.
    The blue and green text refer to \textit{mention span} and its corresponding \textit{Wikipedia title} (used as entity labels).
    For standard BART training, the target output is the linearised triples (\autoref{sec:linearised}).
    For \textsc{Entity-Prompt}, the target is the EL output (\autoref{sec:joint}) concatenated with the linearised triples.
    In \textsc{Artificial-Prompt}, we prepend an artificial token to the input to indicate the desired output, EL (yellow box) or linearised triples. 
    For \textsc{2LM-Heads}, we add an additional task-specific LM head to the decoder for the EL task (grey box).}
\label{fig:webie_example}
\end{figure}

Additionally, we propose three training strategies (\autoref{sec:joint}) that utilise entity linking as an auxiliary task for generative IE: joint generation with the linked-entity prompt (\textsc{Entity-Prompt}), multi-task learning with distinguished artificial prompt tokens (\textsc{Artificial-Prompt}), and training with an additional task-specific language model (LM) head (\textsc{2LM-Heads}). Our experiments demonstrate that incorporating entity-linking objectives leads to better and more faithful IE results. An illustration of these training strategies is provided in \autoref{fig:webie_example}.

Our experiments reveal that models trained on \textsc{WebIE} are more robust and generalisable compared to models trained solely on Wikipedia datasets. These models achieve a new state-of-the-art performance on REBEL (\autoref{sec:results}) and competitive zero-shot performance on WikiNRE. We demonstrate that \textsc{WebIE} serves as a complementary dataset to existing Wikipedia-based datasets and emphasise the significance of including negative examples to address false positives in generative IE.

Our main contributions are as follows:
\begin{itemize}
 \item  We introduce (m)\textsc{WebIE}, the first large-scale, entity-linked IE dataset from the web, with a subset annotated by humans and translated into four other languages.
 \item  We propose and assess the effectiveness of using entity linking as an auxiliary task for generative IE through various training strategies.
 \item  Comprehensive experiments demonstrate that models trained on \textsc{WebIE} exhibit enhanced generalisability in Information Extraction from the web domain, including competitive zero-shot performance on IE tasks on Wikipedia.
\end{itemize}

\section{Related Work}
We review the related work from the following two aspects: datasets and approaches for Information Extraction.
\subsection{Information Extraction Datasets}
The term Information Extraction has been used for different tasks in the literature.
Most existing IE datasets are collected from Wikipedia articles aligned with Wikidata, including sentence-level IE datasets such as REBEL, WikiNRE, FewRel \citep{han-etal-2018-fewrel}, T-REx \citep{elsahar-etal-2018-rex}; document-level Relation Extraction\footnote{We consider RE dataset as the ones that focus on extracting relations but without entity spans and/or linking information.} datasets, e.g., DocRED \citep{yao-etal-2019-docred}, CodRED \citep{yao-etal-2021-codred}.   
SMiLER \citep{seganti-etal-2021-multilingual} is a multilingual sentence-level IE dataset that is also based on Wikipedia, covering 14 languages and 36 relations. 
These sentence-level IE datasets typically do not contain negative examples. \comment{[Chapter 5 - Correction point 2/4]~}\add{Although it is possible to augment these datasets with negative instances sourced from Wikipedia, they do not adequately reflect the true data distribution on the web. To enable more general information extraction models beyond solely relying on Wikipedia articles, we argue for the necessity of datasets that accurately mirror web data -- a point later confirmed in our experiments (refer to \autoref{sec:results}).}

Datasets such as TACRED \citep{zhang-etal-2017-position}, RE-TACRED \citep{Stoica_Platanios_Poczos_2021}, and WebRED \citep{ormandi2021webred} contain negative relation examples, but they lack linkage to knowledge bases. \comment{[Chapter 5 - Correction point 3/4]~}\add{Furthermore, applying entity linking to a knowledge base in a post-hoc fashion is not trivial, as these relation extraction datasets like WebRED primarily focus on \textit{relation}, where the subject and object in a triple can be a name such as \textit{Alice}, which may not be linkable to a KB.} Therefore, our proposed dataset \textsc{WebIE} stands out from existing datasets in that it pertains to the web domain, is entity-linked, and includes negative examples.

\subsection{Information Extraction Approaches}
IE approaches can be classified into two categories: pipeline systems with discriminative models, and sequence-to-sequence systems with generative models. 
Pipeline models typically include separate modules for Named Entity Recognition (NER), Entity Linking and Relation Extraction \citep{chaganty-etal-2017-importance, yamada-etal-2020-luke}.
Systems that jointly train NER, EL, and RE, have also been explored, taking advantage of the information shared among the tasks \citep{ji-etal-2020-span, span2020}.

In recent years, generative IE has gained a lot of attention.
\citet{Nayak_Ng_2020} utilise an LTSM model and propose a pointer network-based decoding.
More recent approaches, e.g., as introduced in REBEL and GenIE, train a transformer-based encoder-decoder model with standard maximum-likelihood objectives to convert sentences to linearised output.
KnowGL \citep{knowgl-aaai_2023_demo} improves upon REBEL with additional entity type information added to the linearised output.
Our work extends GenIE and experiments with three different approaches where we incorporate explicit EL information as an auxiliary task with adapted constraint Trie decoding.

\subsection{\textsc{WebIE} Collection}
\label{sec:auto}

In this section, we provide a detailed explanation of the dataset collection process for (m)\textsc{WebIE}.

\paragraph*{Data Pre-processing} 
\mbox{} \\
We begin the process with the English segment of AllenAI's C4 dataset, retaining the most frequent 200 URL domains. Subsequently, one million documents are randomly sampled, and SpaCy\footnote{\url{https://spacy.io/}} is employed for sentence segmentation. Sentences containing fewer than 10 words are removed, resulting in approximately 20 million sentences.

\paragraph*{Entity Linking and DS Dataset} 
\mbox{} \\
Next, we run ReFinED \citep{ayoola-etal-2022-refined}, a state-of-the-art EL model on the sentences to identify entity spans and link them to their corresponding Wikidata ID. 
Besides named entities, ReFinED also extracts \textit{numerical} entities that do not have Wikidata ID.
In this work, we only consider numerical entities that express dates and map them to the corresponding \textit{year} for simplicity.\footnote{For example, ``October 10, 2018'' will be mapped to ``2018''.}
Some examples of ReFinED processed output are included in \autoref{tab:refined-examples} in \autoref{appendix-webie}.

After obtaining the entity-linked sentences, we apply the DS paradigm to retrieve the set of relations that exist between each pair of entities in each sentence using Wikidata (September 2022 dump) as our KB and build a DS dataset.
After the above steps, we obtain \webie\ DS dataset consisting of 21.2M entities and 4.8M triples.

\paragraph*{Entailment Filtering} 
\mbox{} \\
One major drawback of the DS approach is that the triples extracted may or may not be expressed by the source sentence \citep{riedel2010}. Following previous work on obtaining a cleaner version of the DS dataset \citep{huguet-cabot-navigli-2021-rebel-relation,vania-lee-and-andrea-pierleoni-2022-improving}, we apply an NLI model, \texttt{nli-deberta-v3-large},\footnote{\url{https://huggingface.co/cross-encoder/nli-deberta-v3-large} achieved superior results among the models we evaluated in our preliminary experiments.} that is trained on SNLI \citep{bowman-etal-2015-large} and MultiNLI \citep{williams-etal-2018-broad},
to filter out triples that do not entail the sentence.
Each source sentence is treated as the \textit{premise} and we use manually created templates (similar to \citet{vania-lee-and-andrea-pierleoni-2022-improving}) to convert a DS triple to one or more \textit{hypotheses}.

We then obtain the entailment probability score for each premise-hypothesis pair and take the maximum score for cases with multiple converted hypotheses.
We set the threshold to be 0.7, similar to \citet{huguet-cabot-navigli-2021-rebel-relation}, and only keep triples with an entailment score above the threshold. We retain 2.1M triples (44\% of the previous DS triples, see \autoref{tab:WebIE}) after this filtering process.

\paragraph*{Negative Examples} 
\mbox{} \\
After the DS creation and NLI filtering steps, only less than 10\% of the original sentences contain triples.  
To train models for extracting facts from the web and alleviate false positives, we include two kinds of negative examples in \textsc{WebIE}:
(i) sentences with one or zero entities, and 
(ii) sentences with two or more entities, but without any factual information (i.e., no relation between the entities). 
We randomly sample negative instances covering both cases evenly and add them to \textsc{WebIE}.
In the end, \textsc{WebIE} consists of 1.6M sentences, where 50\% are negative examples.
A summary of the statistics of \textsc{WebIE} with a comparison with other datasets is shown in \autoref{tab:WebIE}.
The dataset is randomly split into train/validation/test sets using a 90/5/5 split.

\begin{table}[!t]
\centering
\scalebox{0.68}{
\addtolength{\tabcolsep}{-2pt}
\begin{tabular}{ p{1.55cm}|p{1.5cm}p{1cm}p{1.3cm}p{1.6cm}p{1.3cm}p{1.2cm}p{1.2cm}p{1.2cm}p{1.4cm}p{1.6cm}p{1.6cm}}
\toprule
\multicolumn{1}{l|}{Dataset} &{Domain} &  {Entity Linked} &  {Relation Type}  & \multicolumn{1}{r}{Sentence} & \multicolumn{1}{c}{Train\textsuperscript{$\dagger$}} & \multicolumn{1}{c}{Val\textsuperscript{$\dagger$}} &\multicolumn{1}{c}{Test\textsuperscript{$\dagger$}} & \multicolumn{1}{c}{Triple} &
{Anno. Triple}   &

Negative Instance
&
Language (Test Set)\\

\midrule
TACRED & Web
&\multicolumn{1}{c}{\xmark}
& \multicolumn{1}{c}{~~~~42} 
& \multicolumn{1}{r}{106,264}
& \multicolumn{1}{r}{68,124}
& \multicolumn{1}{r}{22,631}
& \multicolumn{1}{r}{15,509}
&\multicolumn{1}{r}{106,264}
&  \multicolumn{1}{c}{106,264}
&  \multicolumn{1}{c}{79.5\%} 
&  \multicolumn{1}{c}{1}\\
WebRED & Web
& \multicolumn{1}{c}{\xmark}
& \multicolumn{1}{c}{~~523}
& \multicolumn{1}{r}{117,717}
& \multicolumn{1}{c}{~~--}
& \multicolumn{1}{c}{--}
& \multicolumn{1}{c}{--}
& \multicolumn{1}{r}{117,717}
&  \multicolumn{1}{c}{117,717} 
& \multicolumn{1}{c}{65\%}
&  \multicolumn{1}{c}{1}\\
\midrule
WikiNRE & Wikipedia
&\multicolumn{1}{c}{\cmark}
& \multicolumn{1}{c}{~~158} 
& \multicolumn{1}{r}{255,654}
& \multicolumn{1}{r}{224,881}
& \multicolumn{1}{r}{988}
& \multicolumn{1}{r}{29,785}
& \multicolumn{1}{r}{330,005}
&  \multicolumn{1}{c}{~~~~0}
&  \multicolumn{1}{c}{0} 
&  \multicolumn{1}{c}{1}\\
REBEL & Wikipedia 
&\multicolumn{1}{c}{\cmark}
& \multicolumn{1}{c}{1146}
&\multicolumn{1}{r}{3,059,894}
&  \multicolumn{1}{r}{2,754,387} 
& \multicolumn{1}{r}{152,672} 
& \multicolumn{1}{r}{152,835}  
& \multicolumn{1}{c}{10,311,293}
&  \multicolumn{1}{c}{~~~~0}
&  \multicolumn{1}{c}{0}
&  \multicolumn{1}{c}{1}\\
WebIE & Web
& \multicolumn{1}{c}{\cmark} & 
\multicolumn{1}{c}{~~661}
& \multicolumn{1}{r}{1,649,167}
& \multicolumn{1}{r}{1,480,223}
& \multicolumn{1}{r}{82,914}
& \multicolumn{1}{r}{86,030}
& \multicolumn{1}{r}{1,905,205}
& \multicolumn{1}{c}{~~21,113}
& \multicolumn{1}{c}{50\%}
&  \multicolumn{1}{c}{5}\\
\bottomrule
\end{tabular}
}
\caption{Statistics of WebIE and comparison with other sentence-level RE (top two rows) and IE datasets. 
Sentences correspond to the number of distinct sentences.
We report the publicly available version of WebRED.
$\dagger$ shows the number of examples in each split. 
Anno. Triple represents the number of human-annotated triples.
}
 \label{tab:WebIE}
\end{table}

\subsection{Human Annotation}
Existing IE datasets, such as REBEL, are often automatically annotated using the DS approach, hence the labels can be noisy.
To allow a more reliable evaluation of \webie, we randomly sample \mytilde21K triples from the most frequent 200 relations and annotate them with MTurk.
Given a sentence, each HIT (Human Intelligence Task) is designed to verify if a DS triple is correctly expressed in the sentence.\footnote{We ensure \textit{all} DS triples in a selected sentence are annotated.}
First, the annotators are asked to verify if the head entity (subject) and tail entity (object) are linked correctly.
For each entity, we provide its Wikipedia title and link to its Wikidata page as additional context. 
After that, the annotators are asked to verify if the triple relation is correctly inferred from the sentence. Here, we provide the relation descriptions and example use cases of each relation.
We ask three MTurk workers to annotate each DS triple and take the majority vote as the final label for each triple.
A triple is considered valid if both entities are linked to the correct Wikidata entities and the relation is inferred\footnote{We ask for \textit{inferred} instead of explicit expression since some relations may not be explicitly expressed in the sentence, e.g., ``located in'' (London, UK) or ``date of birth''  XX (1986-2022).} by the sentence.
An annotation interface is shown in \autoref{appendix-webie}.

To ensure the annotation quality, we set qualifications with additional requirements for MTurk workers (see \autoref{appendix-webie} for details).
The agreement among the three annotators is high: 99.4\% for the head entities, 99.2\% for the tail entities, and 76.1\% for the relations have all three annotators agreeing on the same label.
After the majority vote, 92.1\% of the triples are labelled as inferred and therefore kept as valid triples.

\subsection{Multilingual \textsc{WebIE}}
\label{sec:mwebie}
To enable zero-shot cross-lingual transfer evaluation on \webie, we further extend the annotated subset, with additional negative examples, to four other languages: French, Spanish, Portuguese, and Hindi.
First, we use a neural machine translation model, the distilled 1.3B variant,\footnote{\url{https://huggingface.co/facebook/nllb-200-distilled-1.3B}} of NLLB-200 \citep{costa2022no} to translate the English sentences into the target languages.
We then use MTurk to verify the translation and add entity span information in the translated sentences. 
We provide the English sentence (with the entity spans highlighted) and its translation, and first, ask the annotators to correct the translation. After that, MTurk workers are asked to mark the corresponding entity spans in the target language.
We ask two annotators to complete the aforementioned HIT, and an additional worker to select the better translation, which is used in our final dataset.
To obtain translations with higher quality, we restrict the region of the workers to countries where the target language is the official language.\footnote{See details for m\textsc{WebIE} annotations in \autoref{sec:mturk}.}
The final m\webie\ consists of 9K instances in each language, which corresponds to roughly 90\% of the 21K annotated triples.

\section{Generative Information Extraction}

This section describes the training strategies we use for benchmarking (m)\textsc{WebIE}. 

\subsection{Sentence-to-Triples Generation}
\label{sec:linearised}
We use BART and mBART for all of our experiments.
Given a sentence $s$ as input, we train the model to autoregressively generate the linearised triples $t$ as an output.
Following the practice from \citet{huguet-cabot-navigli-2021-rebel-relation} and  \citet{josifoski-etal-2022-genie}, we linearise a triple $t_i$ by converting it into ``\texttt{<sub> head entity label <rel> relation <obj> tail entity label <et>}'', where the tags in brackets represent \textbf{sub}ject, \textbf{rel}ation, \textbf{obj}ect, and the \textbf{e}nd of \textbf{t}riple, respectively. 
Head/tail entity label refers to the Wikipedia title that the mention span in the sentence is mapped to, which also has a one-to-one correspondence with the Wikidata ID.\footnote{For example, a mention span of ``UK'' is linked to Wikipedia title ``United Kingdom'' and mapped to Q145 in Wikidata.}

For each sentence, we order its linearised triples according to the order in which they appear in the input sentence; first by the order of the appearance of the head entity, and then by the order of the tail entity (for cases when the head entities are the same).
The conditional probability of generating $t$ is formulated as $p(t|s) = \prod_{t=0}^{N} p(t_i |t_{<i}, s)$. 
We use the standard cross-entropy loss and maximise the output sequence likelihood with teacher forcing \citep{NIPS2014_a14ac55a}.
An example of input and output can be seen at the top of \autoref{fig:webie_example}.

\subsection{Entity-Linking as an Auxiliary Task}
\label{sec:joint}

The standard linearised triples output only contains the label of the entity and not the span. As a result, it may be difficult to trace back from which input span an entity is generated, especially in the case when the model hallucinates (e.g., by generating an entity that is not mentioned in the sentence).
To encourage models to generate faithful and interpretable output, we also experiment with models that are jointly optimised for generating triples and EL.
The goal of the EL task is to identify and extract entity spans from the input sentence and link them to their corresponding KB entities.
We posit that adding the EL task as an additional training objective will teach the model to attend to the input spans when generating the output.

We experiment with the following three approaches.

\subsubsection*{\textsc{Entity-Prompt}}
\citet{narayan-etal-2021-planning, narayan-etal-2022-well} have shown that generation with entity-chain planning, i.e., generating the desired entities first before the actual output, is effective in improving the faithfulness and controlling hallucinations in text generation tasks such as abstractive summarisation.
For generative IE tasks, EL can be used as an intermediate plan to ground the generation of the linearised triples.
We define the Entity-Linking target in the format of ``\texttt{Mention Span\textsubscript{1} \# Entity Label\textsubscript{1} | Mention Span\textsubscript{2} \# Entity Label\textsubscript{2} | ...}'', where we order the entity spans as they appear in the text.
We then prepend the Entity-Linking target to the linearised triples target, using special symbols as separators, i.e., ``\texttt{[ENTITY] Entity-Linking target [TRIPLE] Linearised Triples Target}''. Here ``\texttt{[ENTITY]}'' is the start symbol before generating the EL output, and ``\texttt{[TRIPLE]}'' is the start symbol before generating the linearised triples. 
Given an input sentence, we essentially train the decoder to first generate the EL chain and then generate the triples, conditioned on both the input sentence and the EL output.\footnote{The EL target only includes mention spans that contribute to valid triples, consistent with the triples that are later generated conditioned on the linked entities.}

\subsubsection*{\textsc{Artificial-Prompt}} 
Artificial Prompt tokens are symbols placed in front of the input sequence, which has previously been explored in areas such as neural machine translation to distinguish the language of the target output translation \citep{johnson-etal-2017-googles}.
We adapt this approach for jointly training our models for Entity Linking and generative IE.
Specifically, we use an artificial prompt token \texttt{<\#el\#>} at the beginning of the input sentence when training for the Entity-Linking target, and use \texttt{<\#tri\#>}\footnote{Both artificial prompt tokens are added as the special tokens to the tokenizer to avoid bias from pre-trained embeddings, but are intended to be biased to the associated task.} for the linearised output target.
Training instances for both tasks are mixed and randomly shuffled for training.

\subsubsection*{\textsc{2LM-Heads}}
Finally, inspired by \citet{gontier2022does}, the third approach that we experiment with is the addition of a second language model (LM) head in the decoder, which is initialised with the same weights as the first (standard) LM head.
The first LM head is optimised for generating the linearised triples while the second LM head is optimised for the EL task, thus each training instance has two different target outputs.
During training, the input sentence is fed to the encoder once, and different target outputs are given to the \textit{same} decoder. Each task-specific LM head is then responsible for generating output targeted for it.
The training loss is then formulated as a weighted sum of the losses from both tasks: $\mathcal{L}=\alpha \mathcal{L}\textsubscript{IE} + (1-\alpha) \mathcal{L}\textsubscript{EL}$.

\subsection{Inference with a Constraint Trie}
In addition to standard beam search decoding, we experiment with constraint decoding by restricting the generated output to be valid Wikipedia titles and Wikidata relations using a prefix Trie, following the ideas proposed in GENRE \citep{cao2021autoregressive} and GenIE \citep{josifoski-etal-2022-genie}. 
The constraint Trie improves generative IE by ensuring that the generated strings are valid entities or relations.
We construct two constraint Tries: an entity Trie and a relation Trie, following the same method as introduced by \citet{cao2021autoregressive}. The entity Trie is built using all Wikipedia titles (as the entity labels),
and the relation Trie is built using all Wikidata relation property labels.

We use four special symbols, \texttt{<sub>}, \texttt{<rel>}, \texttt{<obj>} and \texttt{<et>} to define the state of the generation. 
We apply both constraint Tries as follows. 
We adopt the constraint Trie so that, in the very first decoding state, the model is allowed to either (i) return an empty string for a negative example, or (ii) generate \texttt{<sub>}, which is the start symbol for generating a triple.
If the \texttt{<sub>} symbol is generated, then we generate the head entity using the entity Trie, i.e., only valid entities will be considered.
Once the generation of the head entity is completed, the model proceeds to generate \texttt{<rel>} (i.e., the start symbol for generating relation string) and then subsequently generate allowed tokens from the relation Trie which is built from the relations in Wikidata.
After that, the model generates \texttt{<obj>} and the tail entity, in the same manner, using the entity Trie.
After generating the full triple (indicated by \texttt{<et>} generated after the tail entity), the decoder can either stop the generation or start a new iteration for generating the next triple.

For the \textsc{Entity-Prompt} models, since the entity mention spans are text from the input sentences and usually are not the same as the entity labels in Wikidata, we propose a \textit{partial} constraint generation approach. 
Specifically, we start the standard beam search for the EL target output and only activate the Trie constraints after that when generating the linearised triples.

\section{Experimental Setup}

In this section, we explain the datasets used in the experiments and the detailed modelling setup.

\subsection{Dataset}
\label{sec:dataset}
In addition to our proposed \textsc{WebIE} dataset, we also use the following datasets for our experiments.

\paragraph*{WikiNRE}\hspace*{-.3cm}~\citep{trisedya-etal-2019-neural} is an IE dataset based on Wikipedia which is automatically constructed by aligning Wikipedia sentences to Wikidata triples using the DS approach.
The authors apply a coreference resolution model \citep{clark-manning-2016-improving} to obtain sentences with implicit entity names and use a paraphrase detection model \citep{ganitkevitch-etal-2013-ppdb,grycner-weikum-2016-poly} to filter out sentences that do not express the DS triples. In our experiments, we only use WikiNRE for zero-shot evaluation.

\paragraph*{REBEL}\hspace*{-.3cm}~\citep{huguet-cabot-navigli-2021-rebel-relation} is a large-scale IE dataset constructed automatically from Wikipedia abstracts. 
Using the Wikipedia hyperlinks in the abstracts, as well as numerical values and dates, they map the entity spans to their corresponding Wikidata entities. They then use the DS approach to identify triples in each sentence.
To filter out false positives, the authors use an NLI model by \textit{concatenating} the entities and the relation as the hypothesis.
In our experiment, we use the REBEL dataset that is sub-sampled by \citet{josifoski-etal-2022-genie}, where 857 relations are considered. Both WikiNRE and REBEL do not contain negative examples and are not annotated by humans.

\subsection{Models}
We experiment with BART using two settings: \textsc{Bart\textsubscript{plm}} with the pre-trained weights from \citet{lewis-etal-2020-bart},\footnote{\url{https://huggingface.co/facebook/bart-large}} and \textsc{Bart\textsubscript{rand}}, using the same configuration and architecture but randomly initialised weights.
Across the two settings, \citet{josifoski-etal-2022-genie} find that \textsc{Bart\textsubscript{rand}} generates better results than \textsc{Bart\textsubscript{plm}} on REBEL.
For m\textsc{WebIE}, we experiment with the mBART-50\footnote{\url{https://huggingface.co/facebook/mbart-large-50}} model (for simplicity we refer to it as mBART in this thesis).

To compare models trained on different datasets, we train both \textsc{Bart\textsubscript{plm}} and \textsc{Bart\textsubscript{rand}} on REBEL \textsc{(r)}, \textsc{WebIE} \textsc{(w)}, and both datasets together \textsc{(r+w)}.
We evaluate the performance of the generated triples by parsing the linearised output to a list of triples and comparing it to the gold label to calculate precision, recall, and F1 scores.
For \textsc{WebIE}, we also calculate the accuracy of the prediction of negative instances, where a prediction is considered correct if the model accurately generates empty strings rather than hallucinating triples.

For training with EL as an auxiliary task, we primarily experiment with the \textsc{Bart\textsubscript{rand}}. 
We prepare the training instances as described in \autoref{sec:joint}, and train separate models on REBEL and on \textsc{WebIE}.
For the \textsc{2LM-Heads}, we conduct experiments with different values of the $\alpha$ parameter in the combined loss function, specifically, we set it to 0.5 and 0.75.

We use 8 GPUs, each with 32G VRAM, for all experiments.
We set the batch size to 8 and the accumulated gradient batches to 32.
We follow the hyper-parameters settings from \citet{josifoski-etal-2022-genie} and set the learning rate to $3e^{-5}$, weight decay to 0.01, and warmup steps to 5K.\footnote{For \textsc{Bart\textsubscript{plm}(w)} we find it is necessary to use a lower learning rate $5e^{-6}$ for more stable training.}
We train for up to 30 epochs with early stopping (patience 10), validate twice per epoch, and take the last checkpoint for evaluation. 
Training one epoch takes approximately 1.5 hours for BART and 2 hours for mBART.

\section{Results and Analysis}
\label{sec:results}
This section presents the main results of (m)\textsc{WebIE} and compare different training strategies.

\begin{table}[t!]
\centering
\scalebox{0.7}{
\addtolength{\tabcolsep}{-2.4pt}
\begin{tabular}{ll|cccrcccrcccccccccc}
\toprule
&{\multirow{2}{*}{\sc Model}}  &\multicolumn{4}{c} {\sc WebIE (all test)}   &\multicolumn{4}{c}  {\sc WebIE (anno. test)}& \multicolumn{3}{c} {\sc REBEL}   & \multicolumn{3}{c} {\sc Wiki-NRE}
\\
&& \textit{P} &\textit{R} & \textit{F1}  & \textit{Acc-N}& \textit{P} &\textit{R} & \textit{F1}  & \textit{Acc-N} & \textit{P} &\textit{R} & \textit{F1 }& \textit{P} & \textit{R} &\textit{F1}  \\
\cmidrule(r){1-2} 
\cmidrule(lr){3-6} \cmidrule(lr){7-10} \cmidrule(lr){11-13}  \cmidrule(l){14-16} 
\multirow{7}{*}{\sc {\rotatebox[origin=c]{90}{\small{\textsc{Unconstrained}}}}} &
\sc Bart\textsubscript{rand} \small{(r)}  
&  \cellcolor[HTML]{DDEBF7}10.83
& \cellcolor[HTML]{DDEBF7}16.00
& \cellcolor[HTML]{DDEBF7}12.92
& \cellcolor[HTML]{DDEBF7}0.00
&  \cellcolor[HTML]{DDEBF7}10.70		
&  \cellcolor[HTML]{DDEBF7}13.26
& \cellcolor[HTML]{DDEBF7}11.84  
& \cellcolor[HTML]{DDEBF7}0.00
& 64.34	&67.90&	66.07 		
&  \cellcolor[HTML]{DDEBF7}15.83
& \cellcolor[HTML]{DDEBF7}52.09 
& \cellcolor[HTML]{DDEBF7}24.28\\
&\sc Bart\textsubscript{plm} \small{(r)}  
& \cellcolor[HTML]{DDEBF7}17.58
& \cellcolor[HTML]{DDEBF7}34.20
&  \cellcolor[HTML]{DDEBF7}23.23
&	 \cellcolor[HTML]{DDEBF7}2.28
&  \cellcolor[HTML]{DDEBF7}17.95		
&  \cellcolor[HTML]{DDEBF7}30.02
& \cellcolor[HTML]{DDEBF7}22.47  
& \cellcolor[HTML]{DDEBF7}1.97
& 63.83	&76.66&	69.66
&  \cellcolor[HTML]{DDEBF7}18.34		
& \cellcolor[HTML]{DDEBF7}65.04
& \cellcolor[HTML]{DDEBF7}28.62
\\
\cmidrule(l){2-16}
&\sc Bart\textsubscript{rand} \small{(w)} 
& 55.06&	54.90&	54.98&89.67
&51.64&44.46&47.78&94.74
&  \cellcolor[HTML]{DDEBF7}22.45	
& \cellcolor[HTML]{DDEBF7}20.42 
& \cellcolor[HTML]{DDEBF7}21.39
&  \cellcolor[HTML]{DDEBF7}10.95
& \cellcolor[HTML]{DDEBF7}31.49
& \cellcolor[HTML]{DDEBF7}16.25\\
&\sc Bart\textsubscript{plm} \small{(w)}   
& 54.81 & 70.29 & 61.59 & 87.59 
& 53.40 & 62.36 & 57.53 & 93.58
&\cellcolor[HTML]{DDEBF7} 28.05 
&\cellcolor[HTML]{DDEBF7} 37.28 
&\cellcolor[HTML]{DDEBF7} 32.01
&\cellcolor[HTML]{DDEBF7} 15.55 
&\cellcolor[HTML]{DDEBF7} 60.45 
&\cellcolor[HTML]{DDEBF7} 24.73
\\
\cmidrule(l){2-16}
&\sc Bart\textsubscript{rand} \small{(r+w)}   
&51.34 & 61.22 & 55.85 & 86.80
&49.64 &51.62 &50.61& 93.15
&64.38 &69.57 & 66.87
&  \cellcolor[HTML]{DDEBF7}17.68	
& \cellcolor[HTML]{DDEBF7}65.96
& \cellcolor[HTML]{DDEBF7}27.89
\\
&\sc Bart\textsubscript{plm} \small{(r+w)} 
& 53.04	&75.29&	62.23 & 76.66
&  53.18&	68.41&	59.84& 82.96
 &63.49 &75.30	&	68.89
&  \cellcolor[HTML]{DDEBF7}18.93
& \cellcolor[HTML]{DDEBF7}73.52
& \cellcolor[HTML]{DDEBF7}30.11
\\

\midrule
\multirow{7}{*}{\sc {\rotatebox[origin=c]{90}{\small{\textsc{Constraint Trie}}}}} &
\sc Bart\textsubscript{rand} \small{(r)}  		
&  \cellcolor[HTML]{DDEBF7}11.93
&  \cellcolor[HTML]{DDEBF7}18.91
& \cellcolor[HTML]{DDEBF7}14.63 
& \cellcolor[HTML]{DDEBF7}0.00
&  \cellcolor[HTML]{DDEBF7}11.82		
&  \cellcolor[HTML]{DDEBF7}15.63
& \cellcolor[HTML]{DDEBF7}13.46  
& \cellcolor[HTML]{DDEBF7}0.00
& 66.89&70.37&	68.58 
&\cellcolor[HTML]{DDEBF7}27.61
&\cellcolor[HTML]{DDEBF7}66.73
&\cellcolor[HTML]{DDEBF7}39.06\\
&\sc Bart\textsubscript{plm} \small{(r)}  		
&  \cellcolor[HTML]{DDEBF7}15.24
&  \cellcolor[HTML]{DDEBF7}39.30 
& \cellcolor[HTML]{DDEBF7}21.96
&  \cellcolor[HTML]{DDEBF7}0.00
&  \cellcolor[HTML]{DDEBF7}15.98		
&  \cellcolor[HTML]{DDEBF7}34.92
& \cellcolor[HTML]{DDEBF7}21.93 
& \cellcolor[HTML]{DDEBF7}0.00
&66.28&	76.78 &	71.14
&\cellcolor[HTML]{DDEBF7}25.39
&	\cellcolor[HTML]{DDEBF7}77.45
&	\cellcolor[HTML]{DDEBF7}38.24
\\
\cmidrule(l){2-16}
&\sc Bart\textsubscript{rand} \small{(w)} 
&  55.47 &	57.25&	56.35&90.07
&  52.95&	46.60&	49.57&95.04
&\cellcolor[HTML]{DDEBF7}27.47
&\cellcolor[HTML]{DDEBF7}23.13
&	\cellcolor[HTML]{DDEBF7}25.12
&\cellcolor[HTML]{DDEBF7}18.98
&	\cellcolor[HTML]{DDEBF7}43.75
&\cellcolor[HTML]{DDEBF7}26.48
\\
&\sc Bart\textsubscript{plm} \small{(w)} 

& 57.92 & 74.19 & 64.91 & 87.99
& 57.00 & 65.91 & 61.13 & 94.18
&\cellcolor[HTML]{DDEBF7}35.81 
&\cellcolor[HTML]{DDEBF7}43.00 
&\cellcolor[HTML]{DDEBF7}39.08
&\cellcolor[HTML]{DDEBF7}24.30 
&\cellcolor[HTML]{DDEBF7}78.01 
&\cellcolor[HTML]{DDEBF7}37.06
\\
\cmidrule(l){2-16}
&\sc Bart\textsubscript{rand} \small{(r+w)}   

&52.79 &64.15 &57.92 & 87.45
&51.89& 54.28 & 53.06  &  93.71
&66.87 &72.24 &69.45
& \cellcolor[HTML]{DDEBF7}29.02
& \cellcolor[HTML]{DDEBF7}82.35
&\cellcolor[HTML]{DDEBF7}42.91
\\
&\sc Bart\textsubscript{plm} (\small{(r+w)} 
&54.63&78.43&64.40 & 76.43
&55.22&71.25&62.22&82.59
&66.42 &78.29 & 71.87
& \cellcolor[HTML]{DDEBF7}29.25 
& \cellcolor[HTML]{DDEBF7}86.38
& \cellcolor[HTML]{DDEBF7}43.70
\\
\bottomrule
\end{tabular}
}
\caption{Experiment results for beam search with and without constraint Trie.
P and R refer to precision and recall, respectively, and Acc-N shows the accuracy of the negative examples.
\textsc{Bart\textsubscript{rand}} corresponds to models with BART configuration but randomly initialised weights.
\textsc{Bart\textsubscript{plm}} are models with pretrained weights from \citet{lewis-etal-2020-bart}.
\textsc{(r)}, \textsc{(w)}, \textsc{(r+w)} refer to models trained on REBEL, \textsc{WebIE}, and both datasets, respectively. 
For \webie\ we show the overall performance and the accuracy on negative samples.
Results in blue shades are zero-shot performance.}
\label{tab:main-full}
\end{table}

\subsection{Main Results}
\label{sec:main-results}
\autoref{tab:main-full} shows our benchmarking results on \textsc{WebIE}.
We report results with the constraint Trie in decoding since it overall achieves better results.
Contrary to the findings from \citet{josifoski-etal-2022-genie}, we find that BART models with pre-trained weights are better than randomly initialised weights.
Constraint Trie decoding benefits REBEL, WikiNRE, and the recall performance of \textsc{WebIE}, but may compromise the precision since the models are also trained to handle negative examples.

Models trained on both REBEL and \textsc{WebIE} \textsc{(r+w)} obtain overall better F1 scores on the two datasets compared to models trained on each dataset separately.
Similar performance can also be observed in the zero-shot performance on WikiNRE.
Models trained solely on the REBEL dataset (Wikipedia-domain) show poor generalisability on \webie\footnote{For positive examples it only achieves 20 F1 points.} and always generate false positives thus resulting in 0\% accuracy for negative instances in \webie.
This indicates that Wikipedia-domain data is not adequate for training robust models for the web, and the absence of negative examples in these datasets leads to a prominent issue of hallucination when applied to the web.

\textsc{Bart\textsubscript{plm} (r+w)} also achieves a new state-of-the-art F1 score of 71.87 on REBEL, surpassing the performance of 68.93 from GenIE \citep{josifoski-etal-2022-genie} and 70.74 from KnowGL \citep{knowgl-aaai_2023_demo}, the latter of which trains with additional information including entity type.
The results demonstrate the benefit of \textsc{WebIE}, which contributes to the generalisability of the models.

\subsection{Cross-lingual Transfer with mBART}

\begin{table}[!t]
\centering
\scalebox{0.8}{
\addtolength{\tabcolsep}{-1.5pt}
\begin{tabular}{l|ccccc|ccccc}
\toprule
{\multirow{2}{*}{\sc Language}}
 &\multicolumn{5}{c|} {\sc Unconstrained}  &\multicolumn{5}{c} {\sc Constraint Trie}
\\
&\textit{P} &\textit{R} & \textit{F1} & \textit{Empty-Pos\%} 
& \textit{Acc-N} 
&\textit{P} &\textit{R} & \textit{F1} & \textit{Empty-Pos\%}  & \textit{Acc-N} 
\\
\cmidrule(r){1-1} 
\cmidrule(lr){2-4} \cmidrule(lr){5-5} \cmidrule(lr){6-6} \cmidrule(lr){7-9}  
\cmidrule(lr){10-10} 
\cmidrule(l){11-11} 
\sc English 
&57.72 &61.26 &59.43 &~~2.48 &95.69
&60.29 &64.29 & 62.22 &~~2.63 &96.11
\\
 \sc French 
&43.27 &36.13 &39.38 &11.89 &96.19
&46.52 & 40.26 & 43.16 &12.63 &96.64
\\
 \sc Spanish  
&41.93 & 34.63 & 37.93 &12.34 & 96.74
&45.13 &38.89 & 41.78 & 12.80 &96.97
\\
\sc Portuguese 
&41.17 &32.37 &36.24 &14.07 &96.91
&44.15 & 36.61 & 40.02 & 14.82 &97.22
\\
 \sc Hindi 
 &~~4.28 & ~~1.62 & ~~2.35 &67.38& 98.64
&~~4.23 & ~~1.67 & ~~2.40 & 67.55 & 98.64
\\
\bottomrule
\end{tabular}
}
\caption{Performance on m\textsc{WebIE} with mBART. Results for non-English are zero-shot. 
Empty-Pos(itive)\% shows \textit{false} negatives percentage, revealing zero-shot performance has a high rate of empty results for positive examples.
}
\label{tab:positive-negative-mwebie}
\end{table}

We train mBART on the training set of \webie\ and evaluate the zero-shot cross-lingual transfer on m\webie. 
Similar to prior experiments, results in \autoref{tab:positive-negative-mwebie} show that constraint Trie decoding obtains higher performance than standard decoding.\footnote{We report results using EN as the source language token for mBART, as it produces better performance compared to the actual source language token. See more details in \autoref{sec:src_lang}.}

For English, mBART achieves higher overall performance than \textsc{Bart\textsubscript{plm}} (see \autoref{tab:main-full}). 
The zero-shot results reveal that Hindi has a significant decline in performance compared to the other three non-English languages, French, Spanish, and Portuguese. 
As these three languages utilise the Latin script as in English, which may result in an overlap of entity surface forms. 
In contrast, the transfer is more difficult for Hindi as it employs a different writing system. 
Manual analysis indicates that mBART tends to produce a high rate of false negatives in Hindi examples, where the correct extraction mostly occurs when the entities in the sentences share similar surface forms with the English counterparts.

\begin{table}[!t]
\centering
\scalebox{0.8}{
\addtolength{\tabcolsep}{-2.4pt}
\begin{tabular}{l|cccccc|cccccc}
\toprule
{\multirow{3}{*}{\sc Model}} & \multicolumn{6}{c|} {REBEL} & \multicolumn{6}{c} {WebIE (Anno.)} \\
& \multicolumn{3}{c} {\sc Unconstrained} & \multicolumn{3}{c|} {\sc Constraint Trie} & \multicolumn{3}{c} {\sc Unconstrained} & \multicolumn{3}{c} {\sc Constraint Trie} \\
 & \textit{P} & \textit{R} & \textit{F1 }& \textit{P} &\textit{R} & \textit{F1} & \textit{P} &\textit{R} & \textit{F1 }& \textit{P} & \textit{R} &\textit{ F1 }
\\
\cmidrule(r){1-1} 
\cmidrule(lr){2-4} \cmidrule(lr){5-7} \cmidrule(lr){8-10} \cmidrule(l){11-13}

 \sc Bart\textsubscript{rand}
 & 64.34	&67.90	& 66.07 
 & 66.89	&70.37&	68.58
 & 51.64 &	44.46 &	47.78 	
&  52.95 &	46.60 &	49.57\\
\sc Entity-prompts 
& 63.30	& 63.04	& 63.17	
& 67.91 & 	67.54 &	67.72 
& 49.64 & 51.62 & \textbf{50.61}
& 51.90 & 54.28 & \textbf{53.06}
\\
\sc Artificial-prompt 
&64.23 &	68.23&	66.17 
& 66.41&	70.72&	68.50 
& 52.33 & 46.21 & 49.08
& 53.86 & 48.18 & 50.86
\\
 \sc 2lm-heads & 65.16 &	68.70  &	\textbf{66.88}
 & 67.05&	70.88&	\textbf{68.91}
& 49.13 & 47.67 & 48.39 
& 51.07 & 49.59 & 50.32
 \\
\bottomrule
\end{tabular}
}
\caption{Comparison of various training with entity linking as an auxiliary task, and beam search with and without constraint Trie decoding. 
\textsc{WebIE} results are on the annotated test set.
All models use BART configuration with randomly initialised weights. 
We show in bold the best result among the training objectives.
}
\label{tab:el-models} 
\end{table}

\subsection{Results with Additional EL Training}
\label{sec:aux}
\autoref{tab:el-models} shows the results of training with Entity-Linking as an auxiliary task.
For REBEL, the best results are achieved with the \textsc{2LM-Heads} approach, where the $\alpha$ parameter is set to 0.75.
For \textsc{WebIE} with negative examples, all EL training models achieve better F1 performance than \textsc{Bart\textsubscript{rand}}, with \textsc{Entity-Prompt} particularly resulting in better recall.
This shows the benefit of joint training with EL to improve the faithfulness of web domain data.
\textsc{Artificial-Prompt} achieves the best precision in \textsc{WebIE} but does not show significant differences in performance compared to \textsc{Bart\textsubscript{rand}}. 
We also note that the performance on negative examples does not show significant variations among the different training approaches.
Nevertheless, all three approaches provide better interpretability, i.e., providing the information of the mention spans in the text that contributes to the IE prediction.

\textsc{Entity-Prompt} and \textsc{Artificial-Prompt} do not require additional architectural adaptation over the standard model.
\textsc{Entity-Prompt} also does not introduce training overhead, whereas the other two models may require twice the training time.
\textsc{2LM-Heads} offers the flexibility of adapting the weighted combination of the main task and the auxiliary task by adjusting $\alpha$ in the joint loss formula, which allows more emphasis on the main target. 
Regarding extending to more tasks, the \textsc{Artificial-Prompt} approach can be easily extended by adding additional artificial tokens as task identifiers, whereas \textsc{2LM-Heads} approach would need to add more LM-heads to include more tasks.

\section{Conclusion and Future Work}
We present (m)\textsc{WebIE}, the first large-scale, entity-linked closed IE dataset on the web.
A subset of the dataset is further annotated by humans and translated into four other languages, French, Spanish, Portuguese, and Hindi, via crowdsourcing.
The main contribution lies in the development of faithful and robust extraction pipelines, coupled with effective training methods, for structured knowledge on the web.

\comment{[Chapter 5 - Correction point 4/4]~}\add{
The high approval rate of the triples obtained from the automatic IE pipeline, as confirmed by Human annotation, underscores the efficacy of extracting fact triples through Entity linking, Distance Supervision, and NLI entailment filtering. However, given the recent surge in powerful Language Model (LLM) capabilities, it is intriguing to compare the quality of IE datasets obtained using LLM prompting or via few-shot in-context learning \citep{ma-etal-2023-large}. While the automatic pipeline excels in efficiency, LLMs offer greater flexibility since they don't necessitate restriction to a single knowledge base as required by distance supervision. Nevertheless, both distance supervision and LLM-generated triples are prone to false positives, hence the post-processing such as the NLI filtering remains essential in both scenarios.}

We benchmark \webie\ with generative models and compare the models trained on \webie\ and REBEL (Wikipedia domain).
Our results show that models trained on \webie\ have competitive zero-shot performance when applied to REBEL and WikiNRE, whereas models trained only on REBEL have 0\% accuracy on the negative examples in \webie. This highlights the importance of including negative examples for training more robust models and reducing hallucination in generative IE on the web. 
Models trained on both REBEL and \webie\ achieve the best performance on both datasets, as well as zero-shot results on WikiNRE, positioning \textsc{WebIE} as a complementary dataset to existing Wikipedia-centric datasets.

Our exploration of approaches integrating Entity Linking as an auxiliary task reveals that the addition of a task-specific LM head achieves the overall best performance for REBEL. Notably, the \textsc{Entity-Prompt} approach demonstrates significant improvement on \webie, particularly enhancing recall.

While our primary benchmarking involves transformer-based encoder-decoder models on \webie, future work could also explore pipeline frameworks and larger language models for few-shot performance. 
This chapter underscores the importance of developing a faithful and robust pipeline for extracting structured knowledge on the web, which can then be incorporated into many other knowledge-intensive applications.

\section{Limitations}
We identify several limitations in this work:
\begin{itemize}

    \item \textbf{False Negatives}: Our current automatic triple extraction pipeline is built using the DS approach followed by filtering using an NLI model.
However, Wikidata is not complete \citep{tan-etal-2022-revisiting}.
While some triples may not be completely available in \webie, we expect models trained on this dataset can still discover new triples that do not exist in Wikidata.

    \item  \textbf{Limited Relations in Annotation}: The human annotation is only conducted on the most frequent 200 relations.
    \item  \textbf{Limited Languages in m\textsc{WebIE}}: As discussed in \autoref{sec:mwebie} and \autoref{sec:mturk}, the languages in m\textsc{WebIE} are limited to official languages from geographical regions where there is a reasonable amount of MTurk workers to accept the job. 

An alternative solution would be to use professional translators, especially for low-resource languages.
    \item  \textbf{Fixed Dataset}: 

Facts might change in the world (and Wikidata). 
This can lead to a degraded real-world performance if a system relies exclusively on WebIE for evaluation when the dataset is not updated accordingly.

\end{itemize}
\chapter{Grounded Answer and Explanation in Knowledge-Intensive VQA} 

\label{VQA} 

In the upcoming two chapters, we shift our focus to a broader format of knowledge beyond structured knowledge. This chapter, in particular, centres around knowledge-intensive Visual Question Answering. We explore better utilisation of the parametric knowledge embedded in pre-trained multimodal language models through self-explanation.

The main content is an extended version of the paper ``Towards a Unified Model for Generating Answers and Explanations in Visual Question Answering'' \citep{whitehouse-etal-2023-towards} 
published in \textit{Findings of the Association for Computational Linguistics: EACL 2023}.

\section{Background and Introduction}

The focus of this chapter is on knowledge-intensive Visual Question Answering (VQA) tasks.
Contemporary models for Visual Question Answering (VQA) and Visual Commonsense Reasoning are typically trained discriminatively to select the best answers from multiple-choice questions or to classify single-word answers from a predetermined vocabulary \citep{anderson2018bottom}. However, such settings often have limitations, such as encouraging models to find superficial correlations \citep{ye2021case} or penalising model performance, even when the answers are plausible. For example, synonyms, multi-word expressions, and morphological variations are often not considered correct answers.

Moreover, most current explanation generation models are trained independently of the QA model, and explanations are usually generated after the QA model has provided an answer. Consequently, these explanation models lack access to the process that generated the answer, limiting the grounding of the explanation to the answer text.

\begin{figure}[!t]
\centering
    \includegraphics[width=
    \linewidth]{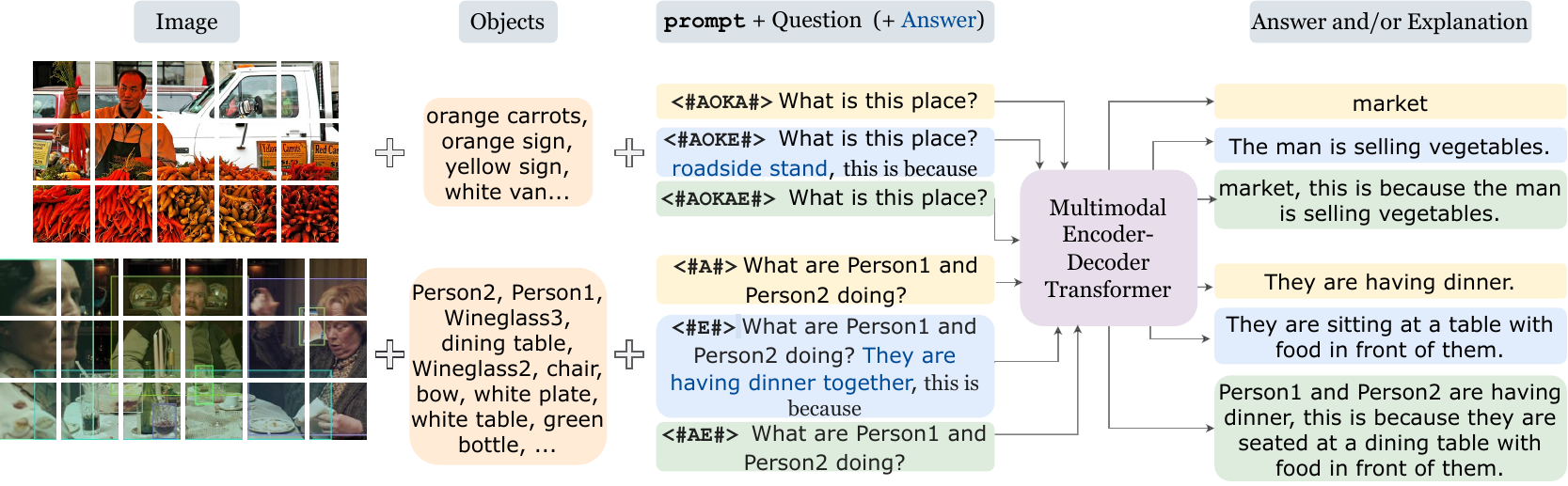}
    \caption{
    Illustration of UMAE:
    we train a multimodal encoder-decoder model on the mix of VQA tasks for jointly optimising answer and explanation, where we distinguish the training instances and target output with artificial prompt tokens (e.g., \texttt{<\#AOKA\#>}).
    The top and bottom examples are from A-OKVQA and VCR, respectively.
    }
\label{fig:illustration}
\end{figure}

We posit that a unified model that simultaneously performs answer prediction and explanation generation is a more effective and consistent approach for VQA. Generative models, such as GPT-3 \citep{brown2020language}, T5 \citep{JMLR:v21:20-074}, or OFA \citep{Wang2022UnifyingAT}, have demonstrated success in rapidly adapting to downstream tasks and generating high-quality open-ended text, making them suitable candidates for this unified approach.

To address this, we propose a multitask learning approach for transformer-based multimodal encoder-decoder models, creating a Unified Model for Answer and Explanation generation (UMAE). In addition to the prevailing trend of separate answer prediction and explanation generation based on the answers, our approach adds the capability of jointly generating answers and explanations. Inspired by the success of artificial prompt tokens in Neural Machine Translation (NMT) \citep{johnson-etal-2017-googles}, we extend and demonstrate the efficacy of the artificial prompt-based method for VQA in a multitask setup. Specifically, we augment training instances with artificial prompt tokens, enabling the model to distinguish different tasks while learning shared semantic features.

\comment{[Chapter 6 - Correction point 1/3]~}\add{
After the model is trained, we propose to determine the best option among the multiple choices by directly ranking the options based on the model's perplexity against each one. We find this approach to be more optimal compared to mapping the generated answer to the closest word embedding from the options \citep{Schwenk2022AOKVQAAB} or using sentence embedding (concatenating the question and answer) with BERTScore \citep{bert-score}. Our method also addresses the surface form penalisation mentioned earlier (synonyms, multi-word expressions, morphological variations, etc.) that involves exact match metrics \citep{Goyal_2017_CVPR}.}

Experiments on a combination of three knowledge-intensive VQA datasets, OK-VQA \citep{marino2019ok}, A-OKVQA \citep{Schwenk2022AOKVQAAB}, and VCR \citep{Zellers_2019_CVPR}, show that UMAE models achieve a new state-of-the-art answer accuracy on A-OKVQA, a new state-of-the-art explanation score on VCR, and competitive out-of-domain performance on VQA-X \citep{Park_2018_CVPR}. UMAE supports the generation of the answer to a question, the explanation for a given question and answer, and both together jointly, making the model efficient and flexible. An illustration of the training setup is shown in \autoref{fig:illustration}.
\comment{[Chapter 6 - Correction point 2/3]~}\add{To the best of our knowledge, our proposal is the first to unit grounded answer and explanation generation for VQA. We specifically focus on knowledge-intensive VQA tasks that are designed to require accessing a knowledge base for answers. By joint training for answer and explanation generation, we hypothesise that the model can more effectively leverage parametric knowledge, potentially enhancing performance even without explicit links to external knowledge sources.}

In summary, our main contributions are as follows:
\begin{itemize}
  \item  The UMAE framework where answers and explanations can be generated by a single unified model (\autoref{sec:umae}).
 \item  A simple and efficient training approach that uses multitask learning with artificial prompts and demonstrates its ability to generalise across domains (\autoref{sec:setup}).
 \item  A method to map generated answers to Multiple-Choice options via evaluating the perplexity of the generation (\autoref{sec:ppl}).
 \item New state-of-the-art results by UMAE particularly for explanation generation, as well as promising out-of-domain performance (\autoref{sec:vqa-results}).
\end{itemize}

\section{Related Work} \label{sec:related_work}
We introduce the related work to this chapter in the following three aspects.

\subsection{Multimodal Transformer-based Models} 
As detailed in Chapter \ref{Background}, multimodal transformer-based encoder-decoder models achieve state-of-the-art performance on various vision-language tasks \citep{chen2020uniter, li2020oscar, cho2021unifying,  zhang2021vinvl, wang2022simvlm}. 
They showcase the possibility of capturing richer multimodal semantic coherence than discriminatively trained models and are further capable of generating self-explanations. 
Pre-trained on multitask settings with natural language instructions, e.g., \textit{``what does the region describe?''}, models like OFA \citep{Wang2022UnifyingAT} are claimed to have the capability to transfer to unseen tasks and domains via similar instructions. However, contrary to these claims, we observe that pre-trained OFA is incapable of generating valid explanations through simple natural language instructions (\autoref{sec:vqa-results}).

\subsection{Artificial Prompt Tokens} 

Artificial prompt tokens, introduced by \citet{johnson-etal-2017-googles}, have primarily found application in Neural Machine Translation (NMT). In their work on Google's multilingual NMT, tokens like \texttt{2es} are added at the beginning of sentences to indicate the target language for translation is Spanish. By using different prompt tokens, they jointly train a multilingual NMT model on various language pairs, allowing the model to learn shared semantics among different instances. This approach proves beneficial for low-resource languages and is effective for zero-shot performance on previously unseen language pairs \citep{johnson-etal-2017-googles}.

Building on the success observed in NMT, our work exploits a similar approach with artificial prompts for answer and explanation generation in VQA with a united model. 
This enables the model to learn shared features among tasks and datasets in various domains.

\subsection{Explanation Generation for VQA} 
There has been a growing interest in incorporating explanations into Visual Question Answering tasks, including Visual Commonsense Reasoning.
Various datasets have been developed where explanations are provided,  e.g., VCR \citep{Zellers_2019_CVPR}, which provides explanations as candidate options in multiple-choice questions, and A-OKVQA \citep{Schwenk2022AOKVQAAB}, which includes multiple explanations for the answer to each question.

Existing VQA datasets have also been extended or corrected with provided textual explanations, for example, VQA-X \citep{Park_2018_CVPR}, CLEVR-X \citep{salewski2022clevr}, and e-SNLI-VE \citep{Kayser_2021_ICCV} are developed from VQAv2 \citep{Goyal_2017_CVPR}, CLEVR \citep{Johnson_2017_CVPR}, and SNLI-VE \citep{xie2019visual}, respectively.

For modelling explanation generation, most recent approaches use separate models to predict answers and generate explanations \citep{dua2021beyond}.
\citet{wu-mooney-2019-faithful} introduce the concept of Faithful Multimodal Explanations (FME) for VQA, wherein textual explanations are linked to relevant image regions attended to by the underlying VQA system. \citet{dua2021beyond} take a two-step approach, training separate modules for answer generation and explanation generation, with the latter based on previously computed answers. \citet{Kayser_2021_ICCV} develop a model called e-UG, which combines UNITER \citep{chen2020uniter} for processing multimodal input and GPT-2 \citep{radford2019language} for generation. 
In contrast, we propose using a single united model for more grounded answer and explanation generation.

\section{Methodology}
\label{sec:methodology}
We introduce the methodology in the following two aspects.

\subsection{Multitask Learning with Artificial Prompt Tokens}
\label{sec:umae}

We formulate three generation settings: \texttt{Q$\rightarrow$A}: answer prediction; \texttt{QA$\rightarrow$E}: explanation generation conditioned on the answer; and \texttt{Q$\rightarrow$AE}: 
\textit{joint} answer and explanation generation for a given question.
We hypothesise that by training the model to generate both the answer and its explanation \textit{simultaneously}, the result answer and explanation will be more grounded and consistent.

We use a pre-trained multimodal encoder-decoder transformer as our base model (here we build on the openly released version of OFA as a strong baseline), and fine-tune the model on a mix of VQA datasets from different domains. 

Different from OFA, for each image in the VQA datasets, we first extract objects and attributes using a bottom-up top-down attention-based model, which is crucial for open-domain VQA tasks \citep{anderson2018bottom}. 
We then add artificial prompt tokens at the beginning of the textual input to signal the generation task (answer, explanation, or both) and the dataset.\footnote{Artificial prompt tokens are added as special tokens to the tokeniser to avoid bias in the pre-trained embeddings. However, we note that these tokens may be biased regarding their association with specific tasks after training, which is an intended effect.}
For \texttt{Q$\rightarrow$AE}, we concatenate answers and explanations with a separator in between.
Finally, we mix all training instances, each consisting of an image (processed in patches), objects and attributes, and textual input with artificial prompts.

\subsection{Perplexity as Multiple Choice Metric}
\label{sec:ppl}
To map the generated output to Multiple-Choice options, in previous work the predictions are loosely matched with options or gold answers using embedding-based methods, such as GloVe embedding similarity \citep{Schwenk2022AOKVQAAB}. 
In contrast to these approaches, we propose to evaluate each option as a \textit{text generation} task, by feeding the model the information that was used to generate the answer as a prompt and calculating the likelihood of each option being generated.
Formally, given an option $Y = (y_1, y_2, ..., y_t)$ with $t$ tokens, we calculate the probability of each token $y_i$ being generated by feeding the image, objects, and question, as well as the first $i-1$  tokens from $Y$ to the model $p_\theta$.
The perplexity is then calculated with: $\mathrm{PPL}(Y) = \mathrm{exp}\left\{ -\frac{1}{t}\sum_{i}^{t}\mathrm{log}\,p_{\theta}\left(y_{i}|y_{< i}  \right) \right\}$, which reflects the probability of option $Y$ being generated by the model.
Finally, the option with the lowest perplexity is chosen as the answer.

We also compare the performance of our approach, using perplexity as the metric, with GloVe embedding similarity for A-OKVQA (see \autoref{tab:ans_acc}).

\section{Experimental Setup}
\label{sec:setup}

We primarily evaluate our proposed UMAE approach using pre-trained OFA\footnote{\url{https://github.com/OFA-Sys/OFA}} as the base model on three knowledge-intensive VQA datasets: 
OK-VQA, A-OKVQA and VCR.
The datasets are introduced below.

\subsection{Datasets}
\label{sec:datasets}

\paragraph*{OK-VQA}\hspace{-.3cm} \citep{marino2019ok} is a knowledge-based VQA dataset that requires outside knowledge beyond the images to answer the questions.
It has train and test splits of size 9,009 and 5,046.
Each question is provided answers by five annotators.
To use the VQA \citep{antol2015vqa} metric, each annotated answer is then repeated twice to form a gold answer set with 10 answers.
Since no explanation is provided, we only train \texttt{Q$\rightarrow$A} task on OK-VQA.

\paragraph*{A-OKVQA}\hspace{-.3cm} 
\citep{Schwenk2022AOKVQAAB} is currently a large-scale knowledge-based VQA dataset split into 17.1K, 1.1K, and 6.7K for train, validation, and test, respectively.
The questions cover four knowledge types: visual, commonsense, knowledge bases, and physical.
For each question, it provides both multiple-choice answers and 10 free-form answers (annotated by 10 different people), as well as three explanations. 
Images in both OK-VQA and A-OKVQA are from MSCOCO \citep{lin2014microsoft}, and answers in both datasets are in single words or short phrases.

\paragraph*{VCR}\hspace{-.3cm}
\label{sec:vcr}
\citep{Zellers_2019_CVPR} is a large multiple-choice dataset for Visual Commonsense Reasoning. 
The train, validation, and test splits have 191.6k, 21.3k, and 26.5k instances, respectively.
Each question has four answer options in sentences, and the correct answer is further provided with four explanation options.
Images in VCR are from movie clips \citep{rohrbach2017movie}.
Bounding boxes of entities are provided associated with mentions such as \texttt{Person1} in questions, answers and explanations.
We follow \citet{NEURIPS2021_c6d4eb15} and draw coloured highlights around the referenced entity on the images, where entity names and the coloured highlights are consistent in the entire dataset, expecting the model to learn the association between the coloured bounding box and the entity.

\paragraph*{VQA-X}\hspace{-.3cm}
\citep{Park_2018_CVPR} contains a subset from the VQAv2 \citep{Goyal_2017_CVPR} dataset and further provides three explanations for each question. 
The image-question pairs are split into train, validation, and test with 29.5k, 1.5k, and 2k instances, respectively. 
We only use the original test set to evaluate the zero-shot performance of the trained models.

\subsection{Training Setup and Hyper-parameters}

We begin with the pre-trained weights from the original OFA-large,\footnote{\url{https://github.com/OFA-Sys/OFA}} which is trained on vision-only tasks including Image Classification,
language-only tasks including Sentence Classification and text Summarisation, as well as various vision-language tasks including Image Captioning, Visual Question Answering and Visual Entailment \citep{Wang2022UnifyingAT}. 

We split the original train set into train and validation sets (95/5 split) for all three datasets. Since the test set is not publicly available for A-OKVQA and VCR, we use the original validation set for experimental analyses. 
We prepare training instances as introduced in \autoref{sec:umae}. Specifically, we add \texttt{<\#OKA\#>} for OK-VQA (only answers are available), \texttt{<\#A\#>}, \texttt{<\#E\#>}, \texttt{<\#AE\#>} for VCR, and \texttt{<\#AOKA\#>}, \texttt{<\#AOKE\#>}, \texttt{<\#AOKAE\#>} for A-OKVQA.
Additionally, for VCR, we draw coloured highlighted boxes around the referenced entity on the images as described.
To account for the imbalance in size among the datasets, we up-sample 
instances in OK-VQA and A-OKVQA, and shuffle all instances to train a model denoted as \textsc{umae\textsubscript{all}}.

For ablation studies, we fine-tune OFA for separate answer prediction (\textsc{ofa\textsubscript{q->a}}) and explanation generation conditioned on answers (\textsc{ofa\textsubscript{qa->e}}). 
To better understand the impact of mixing datasets from different domains, we also train models \textsc{umae\textsubscript{a-okvqa}} and {\textsc{umae\textsubscript{vcr}}},  focusing on all three answer and explanation generation tasks but only using data from a single dataset: either with A-OKVQA or with VCR.
Adam is used as the optimiser and cross-entropy is the loss function.

We set the learning rate to $10e^{-5}$, the warm-up ratio to 0.4, and the patch image size to 480. 
We shuffle all the training examples and use batch size 16.
Due to the large size of VCR, we train for 30 epochs on models involving VCR (\textsc{ofa\textsubscript{q->a}} for VCR, \textsc{umae\textsubscript{vcr}} and \textsc{umae\textsubscript{all}}), and up to 100 epochs for other models.
We report the empirical performance with checkpoints that perform best on the validation set (the 5\% split from the original train set).
For A-OKVQA, we additionally report the answer accuracy on the original test set.

\subsection{NLG Evaluation Metrics}

We use beam search for generating answers and additionally experiment with different decoding methods including top-k sampling, Nucleus sampling \citep{Holtzman2020The}, and Typical sampling \citep{meister-etal-2023-locally}, for generating explanations.
We evaluate answer accuracy as well as explanation quality with automatic NLG metrics and e-ViL scores, following \citet{Kayser_2021_ICCV} for better comparison.
Specifically, e-ViL scores consist of \texttt{S\textsubscript{T}} (task, i.e., answer accuracy),
\texttt{S\textsubscript{E}} (explanation score), and overall \texttt{S\textsubscript{O}} (product of \texttt{S\textsubscript{T}} and \texttt{S\textsubscript{E}}), where \texttt{S\textsubscript{E}} is defined by \citet{Kayser_2021_ICCV} as the harmonic mean of NGRAMScore (the harmonic mean of n-gram scores ROUGE-L \citep{lin-och-2004-automatic}, METEOR \citep{banerjee-lavie-2005-meteor}, CIDEr \citep{ vedantam2015cider}, and  SPICE \citep{anderson2016spice}) and additionally the BERTScore \citep{bert-score}, a learned similarity metric over contextual representations of sentences. 

\comment{[Chapter 6 - Correction point 1/3]~}\add{We elaborate and discuss different aspects of these NLG metrics.
Each of the n-gram scores targets distinct aspects of the output.
ROUGE-L focuses on assessing the overlap of contiguous sequences of words between the generated text and the reference text by measuring the precision, recall, and F1 score of the longest common subsequences between the generated and reference texts. METEOR emphasises semantic similarity and incorporates stemming, synonymy, and word order information to compute a score that better reflects the semantic equivalence between the generated and reference texts. CIDEr assesses the consensus between human judgments and machine-generated texts by calculating a score based on the similarity of n-grams between the generated text and multiple reference texts provided by human annotators. SPICE focuses on the precision and recall of generated linguistic structures and evaluates the structural similarity between the generated and reference texts by analysing syntactic and semantic components. On the other hand, BERTScore is a trained metric that leverages contextual embeddings obtained from pre-trained transformer models like BERT to measure the similarity between sentences at a more granular level.
}

\add{To summarise, higher ROUGE-L scores suggest higher degrees of lexical overlap between the generated and reference texts; higher METEOR scores indicate greater semantic similarity between the generated and reference texts; higher CIDEr scores imply better alignment with human judgments and consensus among multiple reference texts; higher SPICE scores reflect greater precision and recall of generated linguistic structures; and higher BERTScores signify increased similarity between sentences at a contextual level.
Although these metrics target different aspects of the generated text, experimental results reveal a consistent trend across various metrics: if a generation scores higher in one metric, it often scores higher in others as well (refer to \autoref{sec:explanation_results}). 
}

\add{However, it is worth noting that none of these metrics directly measure the human interpretability of the generated explanations. N-gram metrics overlook semantic meanings, coherence, and relevance of the generated text, while BERTScore is also biased with the lengths of the generation greatly differing. More discussions on the limitation of the NLG metric are included in \autoref{sec:explanation_results} and \autoref{sec:limitations}.}

\section{Main Results}
\label{sec:vqa-results}
 In this section, we show the results for generated answers and explanations, compare models and analyse errors.

\begin{table}[t!]
\centering
\scalebox{0.85}{
\addtolength{\tabcolsep}{-3pt}
\begin{tabular}{l|cccccccc}
\toprule
{\multirow{3}{*}{\sc Model}} & {\sc Ok-vqa} & \multicolumn{5}{c}{\sc A-okvqa}  & \multicolumn{2}{c}{\sc Vcr}\\
& \textit{direct answer} & \multicolumn{3}{c}{\textit{multiple choice}} & \multicolumn{2}{c}{\textit{direct answer}}  & \textit{multiple choice} & \em BERTScore\\

& \sc test &\textsc{val} \small{(\textit{ppl})} &\textsc{val} \small{(\textit{GloVe})} &\sc test  & \sc val   & \sc test    & \textsc{val} \small{(\textit{ppl})} & \sc val \\
\cmidrule(r){1-1} \cmidrule(lr){2-2} \cmidrule(lr){3-5} \cmidrule(lr){6-7} \cmidrule(l){8-9}
\sc ofa* & 40.40 & 24.54 &  56.19 & 47.40 & 48.09 & 39.77 & 33.55 & 64.55 \\
\sc ofa\textsubscript{q->a}  & 49.93 & 74.32 & 65.30   & 61.71 &  63.00  & 53.91  & 54.89 & 83.85 \\

\sc umae\textsubscript{all}  & \textbf{51.77} & \textbf{74.59} &  \textbf{65.67}   & \textbf{63.26}  & \textbf{63.29}   & \textbf{56.14}  & \textbf{56.66} &\textbf{85.97} \\
\midrule
\sc Prior-best  & 54.41 &  -- &60.30 & 53.70  & 48.60  & 40.70 & (77.10)\rlap{\textsuperscript{\dag}} & --  \\
\bottomrule
\end{tabular}
}
\caption{Performance of models for answer generation.
Better results are in bold.
\textsc{ofa}* refers to the pre-trained OFA.
Prior-best results
for the three datasets 
are from \citet{gui-etal-2022-kat}, \citet{Schwenk2022AOKVQAAB}, \citet{Wang_2023_ICCV}, respectively.
\dag{} is from a discriminative model and thus not comparable \citep[see][]{ye2021case}.
}
 \label{tab:ans_acc}
\end{table}

\subsection{Answer Accuracy}
\autoref{tab:ans_acc} presents our observations for answer accuracy on \texttt{Q->A} task over the three datasets. 
We also evaluate VCR answers using BERTScore as the answers for VCR are usually sentences.
We observe that \textsc{umae\textsubscript{all}} outperforms \textsc{ofa\textsubscript{q->a}} on all datasets, improves the prior state-of-the-art on A-OKVQA by 10$\sim$15\%, and achieves competitive results on OK-VQA.
For models that are fine-tuned on A-OKVQA, we also see a salient improvement (+9\%) with the proposed mapping of options by perplexity in Multiple-Choice, instead of GloVe embeddings similarity.\footnote{Preliminary experiments with NLG metrics (BERTScore and BLEU) for selecting the options given generation were sub-optimal.}

\begin{table}[ht]
\centering
\vspace{1.5ex}
\scalebox{0.95}{
\begin{tabular}{ccc|c}
\toprule
   \sc Question & \sc Objects & \sc Images & \sc Accuracy \\
   \midrule
\sc \cmark     & \cmark   & original    & 50.39        \\
\sc \cmark   & \xmark  & \xmark      & 39.16         \\
\sc \cmark    & \xmark  & random    & 33.48        \\
\sc \cmark    & \cmark  & \xmark    & 33.28        \\
\bottomrule

\end{tabular}
}
\caption{Ablation on the modality dependency for answer accuracy of A-OKVQA.
}
 \label{tab:modality}
\end{table}

 We conducted several ablation studies to investigate the dependency of object features and images on the performance of our model \textsc{UMAE\textsubscript{all}} for answer accuracy of A-OKVQA, where we removed images, replaced them with random images, and removed extracted attributes and features.
Results in \autoref{tab:modality} show that the visual encoder is crucial for performance and that visual objects alone are not sufficient for answer prediction.
Using a random image would introduce noise and therefore performs worse than not including the image at all.
 We did not test removing the question because we believe the model needs the questions to be able to provide answers.

\begin{table}[!t]
\centering
\scalebox{0.85}{
\addtolength{\tabcolsep}{-3.5pt}
\begin{tabular}{ll|rrrrrrrrc}
\toprule
{\multirow{2}{*}{\sc Dataset}}
 & {\multirow{2}{*}{\sc Model}} & \multicolumn{3}{c} {e{\sc-V}i{\sc l Scores}} & \multicolumn{5}{c} {\sc {N-gram Scores}} & \sc Learnt Sc.\\

 & & \sc \textit{S}\textsubscript{O}    & \sc \textit{S}\textsubscript{T}     & \sc \textit{S}\textsubscript{E}   & \multicolumn{1}{c}{\sc bleu\small{4}}  & \sc r-l   &\sc met.  & {\sc cide}r & \sc spice & \sc bertscore    \\
\cmidrule(r){1-2} 
\cmidrule(lr){3-5} \cmidrule(lr){6-10} \cmidrule(l){11-11}

{\multirow{4}{*}{\sc {A-okvqa}}}
&\sc ofa* &   4.44
 & 56.19 & 
7.90			& 	0.30&	4.45& 	3.26	& 4.82	&4.62& 	68.64
\\
&\sc ofa\textsubscript{q->a}+ofa\textsubscript{qa->e} &   
35.82 & 74.32 &
48.29&	22.18&	48.51	&23.56	&86.76	&22.46	&	85.96
\\

& \sc umae\textsubscript{a-okvqa} & 37.10 & 73.97 & 50.15  & \textbf{27.61} & 52.23 & 24.06 & \textbf{104.39} & 22.88 & 87.86    \\

&\sc umae\textsubscript{all} & \textbf{37.91} & \textbf{74.59} &\textbf{50.82} & 27.35 & \textbf{52.56} & \textbf{24.83} & 101.09 & \textbf{23.33} & \textbf{88.21}    \\
\midrule
{\multirow{3}{*}{\sc {Vcr}}} 
& e{\sc-ug} &19.30 & \textbf{69.80} & 27.60  &4.30 & 22.50 & 11.80 & 32.70 & 12.60 & 79.00   \\

& \sc umae\textsubscript{vcr} & 22.57 & \textbf{56.68} & 39.82 & 12.25 & 28.87 & 16.67 & \textbf{48.14} & 27.36  & 81.77\\
& \sc umae\textsubscript{all} & \textbf{22.82} & 56.66 & \textbf{40.27} & \textbf{13.44} & \textbf{29.53} & \textbf{17.54} & 47.33 & \textbf{26.45}   & \textbf{81.91} \\

\midrule
{\multirow{2}{*}{\sc {Vqa-x}}}
&  e{\sc-ug} & 36.50 & 80.50 & 45.40 &  23.20 & 45.70 & 22.10 & 74.10 & 20.10 & 87.00\\
&\cellcolor[HTML]{DDEBF7}\sc umae\textsubscript{all} & \cellcolor[HTML]{DDEBF7}{31.58}  &\cellcolor[HTML]{DDEBF7}{77.65}   &\cellcolor[HTML]{DDEBF7}{40.67} & \cellcolor[HTML]{DDEBF7}{14.63} & \cellcolor[HTML]{DDEBF7}{35.12} & \cellcolor[HTML]{DDEBF7}{20.29} &\cellcolor[HTML]{DDEBF7}{50.35} &\cellcolor[HTML]{DDEBF7}{19.13}  & \cellcolor[HTML]{DDEBF7}{85.40}    \\

\bottomrule
\end{tabular}
}
\caption{Explanation Scores. 
\textsc{r-l}, \textsc{met.} stand for \textsc{rouge-l}, \textsc{meteor}, respectively.
\textsc{ofa}* is the pre-trained OFA, showing the transferability of OFA for generating explanations with natural language instructions.
Results with e-UG are from \citet{Kayser_2021_ICCV}.
We show the best results of A-OKVQA and VCR in bold.
The last row in blue shade shows \textit{out-of-domain} performance.
}
\label{tab:explanation}
\end{table}

\subsection{Explanation Evaluation}
\label{sec:explanation_results}
\begin{figure}[!t]
\centering
    \includegraphics[width=\linewidth]{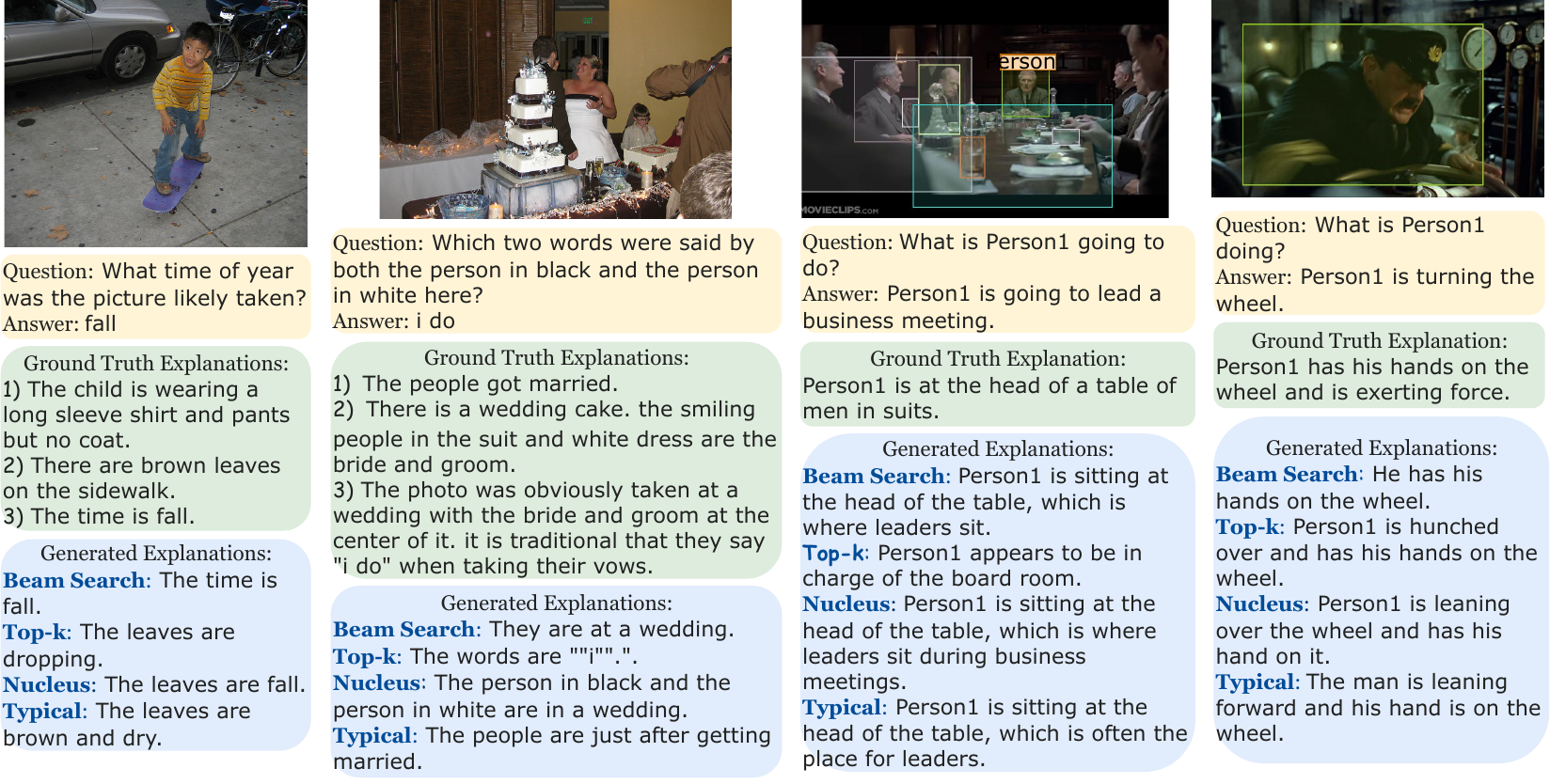}
    \caption{Examples of generated explanations from \textsc{mix} model with different decoding strategies.
    Two examples on the left are from A-OKVQA and the other two on the right are from VCR.}
\label{fig:explanations}
\end{figure}

\autoref{tab:explanation} shows e-ViL sores and individual NLG evaluation results for explanations using automatic NLG metrics.\footnote{Nucleus sampling shows the best results and is reported. 
Detailed scores with different decoding methods are shown in \autoref{sec:explanation_scores}.}
Following the same setup as in \citet{Kayser_2021_ICCV}, an explanation is evaluated only if the answer predicted by the system is correct.\footnote{A limitation of evaluating all explanations is that explanations of wrong answers may get high scores with n-gram metrics, even though they are justifying wrong answers and should be penalised.}
We observe that pre-trained OFA with natural language prompts, e.g., \textit{``what is the explanation for the answer?''} or \textit{``this is because''} performs poorly, as most generated explanations are words (\textit{``yes/no''}) or short-phrases.\footnote{BERTScore in not representative of the validity of outputs from \textsc{OFA}*. We refer the reader to an exposition of the problems associated with NLG metrics in \citet{caglayan-etal-2020-curious}.}
We compare UMAE models (on all and individual datasets) with prior best results from e-UG (see \autoref{sec:related_work}), and standard separated trained baselines (\textsc{ofa\textsubscript{q->a}}+\textsc{ofa\textsubscript{qa->e}}).
\textsc{umae\textsubscript{all}} achieves better results across all datasets, showing the advantage of mixing tasks and datasets in different domains.
For out-of-domain evaluation on VQA-X, \textsc{umae\textsubscript{all}} also shows mostly competitive results.
Examples of explanation generation are shown in \autoref{fig:explanations} as well as in  \autoref{sec:explanations} in \autoref{sec:VQA-appendix}.

Since e-ViL only evaluates an explanation if a model generates the correct answer, the subset of explanations evaluated varies by model.
To \textit{fairly} compare explanations on the same subset, we propose only using the subset of samples where all models provide correct answers for explanation prediction. 
\autoref{tab:exp} shows the results on A-OKVQA with such a subset of 770 candidates, where \textsc{umae\textsubscript{all}} shows an even higher explanation score. This highlights that \textsc{umae\textsubscript{all}} generates explanations that overlap significantly better with gold explanations.

\begin{table}[ht]
\centering
\scalebox{0.95}{
\addtolength{\tabcolsep}{0pt}
\begin{tabular}{l|cccccccc}
\toprule
\sc Model & \sc \textit{S}\textsubscript{E}   & {\sc bleu\small{4}}  & \sc r-l   &\sc met.  & {\sc cide}r & \sc spice & {\sc bertscore} 
\\
\midrule
\textsc{ofa\textsubscript{q->a}+ofa\textsubscript{qa->e}}& 42.4 & 20.0	&44.2	&19.3&	66.7	&19.1	&	85.1\\

\sc {umae\textsubscript{a-okvqa}}& 45.8 &	 23.6&		47.9&		21.7	&	78.0	&	20.5	&	86.9\\
\sc {umae\textsubscript{all}} & \textbf{46.8}& \textbf{24.9}	& \textbf{49.5} &	\textbf{22.3}  &	\textbf{84.1}   &	\textbf{20.8}	&	\textbf{87.3} \\

\bottomrule
\end{tabular}
}
\caption{
Explanation scores on the same subset of A-OKVQA.
}
\label{tab:exp}
\end{table}

\subsection{Joint Answer and Explanation Generation}

We further present the results of the proposed \texttt{Q$\rightarrow$AE} task where answers and explanations are jointly generated. 
We parse the generated sequence to the answer and the explanation and use the same sets of metrics as the separate generation for evaluation. 
Results for answers in \autoref{tab:AR_ans}  and explanations in \autoref{tab:AR}.
For answers, since the perplexity metric does not directly compare the generation, we show the Multiple-Choice accuracy using the Glove metric for A-OKVQA and \texttt{BERTScore} for VCR answer sentences.

\begin{table}[ht]
\centering
\vspace{1ex}
\scalebox{0.95}{
\addtolength{\tabcolsep}{0pt}
\begin{tabular}{l|cccccc}
\toprule
\multirow{2}{*}{\sc Task} & {\sc A-okvqa} & {\sc Vcr} &  {\sc Vqa-x}    \\

& \sc multiple-choice  & \sc bertscore & \sc direct answer\\
\cmidrule(r){1-1} \cmidrule(lr){2-2} \cmidrule(lr){3-3} \cmidrule(l){4-4} 
\sc{q}{\small{->}}{\sc a} & 65.67 & 81.91  & \cellcolor[HTML]{DDEBF7}77.65\\
\sc{q}{\small{->}}{\sc ae} & 65.67 & 82.30 & \cellcolor[HTML]{DDEBF7}69.60\\

\bottomrule
\end{tabular}
}
\caption{Evaluation of answers generated given questions (\texttt{Q->A}) and jointly generated with explanations (\texttt{Q->AE}).
The last column with a blue shadow indicates out-of-domain performance.
For simplicity, we use the Glove metric for selecting answers for Multiple-Choice questions in A-OKVQA.}
\label{tab:AR_ans}
\end{table}

\begin{table}[ht]
\centering
\scalebox{0.95}{
\addtolength{\tabcolsep}{0pt}
\begin{tabular}{l|cccccc}
\toprule
\multirow{2}{*}{\sc Dataset} & \multicolumn{2}{c}{\sc \textit{S}\textsubscript{E}}   & \multicolumn{2}{c}{\sc Ngramscore}   &  \multicolumn{2}{c}{\sc Bertscore}    \\

& \sc{qa}{\small{->}}{\sc e} &  \sc{q}{\small{->}}{\sc ae} & \sc{qa}{\small{->}}{\sc e} &  \sc{q}{\small{->}}{\sc ae} & \sc{qa}{\small{->}}{\sc e} &  \sc{q}{\small{->}}{\sc ae} \\
\cmidrule(r){1-1} \cmidrule(lr){2-3} \cmidrule(lr){4-5} \cmidrule(l){6-7}
\sc {A-okvqa}& 50.82 & 47.01 & 35.69 & 32.15 & 88.21 & 87.39  \\

\sc {Vcr}& 40.27 & 37.02 &  26.70 &24.02 & 81.91 & 80.68\\
\sc {Vqa-x}  & \cellcolor[HTML]{DDEBF7}40.67 &\cellcolor[HTML]{DDEBF7}39.67 & \cellcolor[HTML]{DDEBF7}26.69 & \cellcolor[HTML]{DDEBF7}25.85 & \cellcolor[HTML]{DDEBF7}85.40 &\cellcolor[HTML]{DDEBF7}85.21 \\

\bottomrule
\end{tabular}
}
\caption{Scores of explanations generated given answers (\texttt{QA->E}) and jointly generated with answers (\texttt{Q->AE}).
The last row with a blue shadow indicates out-of-domain performance.}
\label{tab:AR}
\end{table}

In summary, our experiments demonstrate that the UMAE model leads to improved answer and explanation generation and allows for the flexibility to generate different types of outputs, including answers, explanations, or both. We observe that UMAE exhibits promising results in jointly generating both the answer and explanation.

\section{Analysis and Discussion}
\label{sec:analysis}

In this section, we analyse the errors in answer and explanation generation, showcase the issues existing in the datasets, and discuss the limitations in the explanation evaluation metrics. 

\subsection{Error Analysis}  \label{sec:error_analysis}
\begin{figure}[!t]
\centering

    \includegraphics[width=0.85\linewidth]{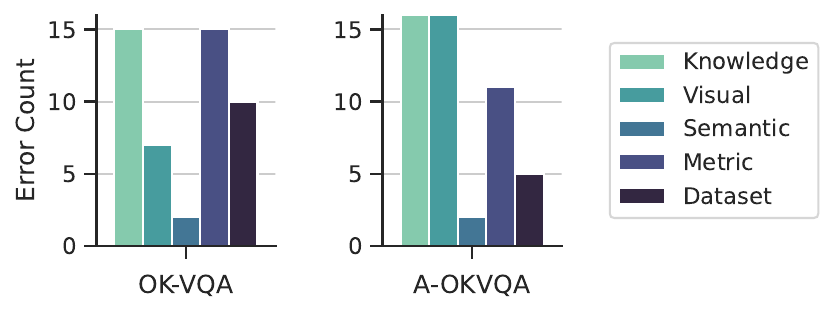}
    \caption{Distribution of error types in direct answers from 50 error samples in OK-VQA and A-OKVQA.}
\label{fig:errors}
\end{figure}

To better understand the generated answers and errors, we randomly sample 50 errors in OK-VQA and A-OKVQA.
Our analysis reveals the following main error types, where the first three are related to model performance:
\begin{itemize}

    \item  \textit{Knowledge}:  
the implicit knowledge learned by the model is insufficient for answering some of the knowledge-intensive questions, such as questions asking \textit{when} a certain sport was invented.
     \item \textit{Visual}: 
the model fails to identify the visual attributes correctly, such as questions about \textit{recognising object shape or material}.
\item  \textit{Semantic disassociation}:
the model misinterprets questions or fails to match the intended semantic meaning. For example, it may answer what \textit{an object is} instead of a more complex question such as \textit{what is commonly packed in it} (e.g., answering ``suitcase'' instead of ``clothes'').
    \item  \textit{Metric}: 
the evaluation metric may penalise some of the plausible answers, especially when searching for exact match answers (mostly due to the difference of singular/plural or phrases with/without space in between).
    \item  \textit{Dataset}:  
errors due to issues in the datasets themselves.

\end{itemize}
We discuss prominent issues in dataset quality briefly in \autoref{sec:VQA-appendix} and further present the distribution of error types in \autoref{fig:errors}.

\begin{figure}[!t]
\centering
\vspace{0.5cm}
    \includegraphics[width=0.7\linewidth]{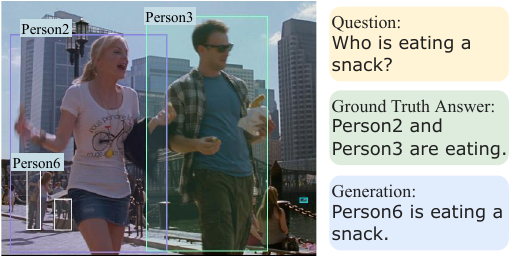}

    \caption{Example of misassociation between coloured highlights and entities.
    The model fails to associate \textit{purple} box with \texttt{Person2} and \textit{green} box with \texttt{Person3}.}
\label{fig:mismatch}
\end{figure}

We also observe misassociation between bounding boxes and entities in VCR.
Many questions in VCR provide extra context assisting the association, for example, questions like \textit{``Why are the eyes of Person2 closed?''} helps the model to identify \texttt{Person2} as ``the person with closed eyes'' in the image, but questions that specifically need the correct association without context can become problematic.
\autoref{fig:mismatch} shows an example revealing that the model struggles with the association.
This may be related to the limitation of patch image processing used in OFA.
The assumption that the model would learn the association via the coloured highlights in the images is not always valid.

\subsection{Dataset Quality} \label{sec:dataset_issue}
We observe the following issues in the existing datasets:
(i) wrong answers, 
(ii) subjective or unanswerable questions,
(iii) typos or unclear expressions,
(iv) not requiring images or knowledge to answer the question as designed.

Unlike OK-VQA and A-OKVQA where the ground truth answers are provided by multiple people, in VCR instead, answers and explanations for a question are obtained from the same person who authored the question.
This makes the answers or explanations contain a severe amount of subjectivity.
We find that a significant number of the examples expect an understanding of the \textit{movie plot} from which the image is extracted, rather than requiring \textit{commonsense} reasoning. 
While human annotators have an implicit understanding of the movies, the dataset itself does not contain relevant contextual information. 
We provide specific examples of the issues in \autoref{sec:datasets_issues} in \autoref{sec:VQA-appendix}.

\begin{table}[ht]
\centering
\scalebox{0.95}{
\begin{tabular}{l|ccc}
\toprule
     & \sc Ok-vqa     & \multicolumn{2}{c}{\sc A-okvqa}           \\
      & \sc direct answer & \sc multiple-choice  & \sc direct answer \\ 
       \cmidrule(r){1-1} \cmidrule(lr){2-2} \cmidrule(l){3-4}
\sc Best     & 80.94  & 80.74    & 66.20        \\
\sc Average  & 54.98  & 71.53      & 57.29         \\
\sc Worst   & 16.37 & 59.35    & 41.46        \\
\bottomrule
\end{tabular}
}
\caption{Human performance on OK-VQA and A-OKVQA measured from the ground truth answers.
For simplicity, we use the Glove metric for selecting answers for Multiple-Choice questions in A-OKVQA.
}
 \label{tab:human}
\end{table}

We further measure the best, average and worst human performance on OK-VQA and A-OKVQA by selecting the most common answer, a random answer, and the least common answer, respectively, from the 10 ground truth answers for each question.
We calculate the performance using the VQA metric for direct answers, and the GloVe metric for multiple choices for simplicity.
Note that we also remove the answer selected from the ground truth answers when measuring human performance.
From the results in \autoref{tab:human} we can see that the average performance on both datasets is relatively poor, which indicates the noise in the datasets.
The quality of the datasets needs to be more carefully inspected so that the model performance evaluated on these datasets can be more meaningful.

\subsection{Explanation Evaluation}
Current NLG metrics predominantly evaluate the n-gram scores between generation and reference.
Although human evaluation may be the ultimate criterion \citep{Kayser_2021_ICCV}, this does not scale.
Humans' judgments on generated explanations are contextual, especially the images that are predominantly taken into consideration. 
However, none of the current widely used metrics considers visual information. 
Multimodal evaluation metrics \citep{madhyastha-etal-2019-vifidel,jiang-etal-2019-tiger,hessel-etal-2021-clipscore} are potentially better approaches to obtain visually grounded measures for evaluation.

\section{Conclusion and Future Work}
\label{sec:limitations}
In summary, our proposed Unified Model for Answer and Explanation generation (UMAE) leverages a multitask learning approach within a multimodal encoder-decoder framework, incorporating artificial prompt tokens to distinguish distinct tasks while learning shared semantics.

Evaluation of our approach on various VQA tasks shows that UMAE outperforms prior best models and separately trained baselines in both answer and explanation scores, where we also demonstrate the benefit of using perplexity as the metric for mapping generated answers to Multiple-Choice options. 

Additionally, UMAE offers flexibility in output and can generate explanations for datasets without explanations for training, e.g., OK-VQA, while also improving answer quality. 
In-depth case studies and error analyses undertaken in this chapter reveal valuable insights for future enhancements, underscoring the importance of continuous improvement in dataset quality. 

This chapter highlights the benefit of grounded answer and explanation generation towards both, showcasing better utilisation of the parametric knowledge stored in the parameters of the multimodal language models.

\section{Limitations}

We address the limitations of our work in the following two aspects:

\begin{itemize}
\item \textbf{Model Specificity}: Firstly, the experiments conducted with our proposed framework and fine-tuning approach are primarily on the OFA model. While we posit that our approach applies to any multimodal generative model, broader insights could be gained by experimenting with a more diverse set of models.

\item \textbf{Evaluation Methodology}: Secondly, in terms of evaluating our proposed joint framework, assessing the generated explanation quality, especially differentiating between explanations generated jointly with answers and those conditioned on the answers, necessitates human judgment. Relying on human evaluation becomes essential for a nuanced assessment compared to the use of automatic NLG metrics.

\end{itemize}

\chapter{Knowledge Distillation via LLM-powered Data Augmentation}

\label{LLM} 

In the final chapter, we explore the utilisation of knowledge from the latest trend of powerful large language models, in complex and challenging multilingual commonsense reasoning tasks.

The main content of this chapter is based on the paper ``LLM-powered Data Augmentation for Enhanced Cross-lingual Performance'' \citep{whitehouse-etal-2023-llm} 
published in \textit{
Proceedings of the 2023 Conference on Empirical Methods in Natural Language Processing: EMNLP 2023}.

\section{Background and Introduction}

The success of NLP models greatly depends on the availability and quality of training data. This poses a significant challenge for multilingual NLP, as data for languages other than English is typically limited \citep{ponti-etal-2019-modeling, joshi-etal-2020-state, whitehouse-etal-2022-entitycs}. An approach to address the data scarcity challenge is through zero-shot cross-lingual transfer or multitask training, in which a model is trained across data of diverse tasks and languages, exhibiting the capability to handle unseen tasks, particularly in larger models \citep{artetxe-schwenk-2019-massively, nooralahzadeh-etal-2020-zero, huang-etal-2021-improving-zero}. However, when aiming for task-specific objectives, a smaller, fine-tuned model dedicated to that particular task often outperforms larger general-purpose, zero-shot models. In addition, a smaller task-specific model is more practical and cost-effective for training and deployment. Nevertheless, developing a powerful task-specific model becomes challenging in the absence of training data \citep{lauscher-etal-2020-zero}. 

Conversely, recent powerful Large Language Models (LLMs) excel at handling general instructions and have shown promise in data generation tasks \citep{wang-etal-2023-self-instruct}. 
In this work, we leverage LLMs to generate synthetic data for various multilingual commonsense reasoning tasks, XCOPA \citep{ponti-etal-2020-xcopa}, XWinograd \citep{tikhonov-ryabinin-2021-heads}, and XStoryCloze \citep{lin-etal-2022-shot}, where the training data is limited even for English (see \autoref{tab:data}). 
To augment the training data, we provide LLMs with instructions and examples from the original training data, prompting them to generate new and diverse examples.
We explore the generation of synthetic data in English using different LLMs, including open-source models like  Dolly-v2\footnote{\url{https://github.com/databrickslabs/dolly}} and StableVicuna\footnote{\url{https://github.com/Stability-AI/StableLM}}, as well as ChatGPT and GPT-4.
Although the weights and capabilities of the latter two models remain undisclosed, we explore them as they extend the capability of generating texts in languages beyond English.

\begin{table}[!t]
\centering
\sisetup{table-format=4.0}
\addtolength{\tabcolsep}{-1.5pt}
\scalebox{0.9}{
\begin{tabular}{l|cccccc}
\toprule
\multirow{2}{*}{\sc\textbf{Dataset}}

& \multicolumn{2}{c}{\textbf {Train}}
& \multicolumn{2}{c}{\textbf {Validation}}
& \multicolumn{2}{c}{\textbf {Test}}
  \\
  \cmidrule(lr){2-3} \cmidrule(lr){4-5} \cmidrule(l){6-7}

  & {English} & {Non-English}
& {English} & {Non-English}
& {English} & {Non-English}
\\
  \midrule
 {XCOPA} & 400 & 0 & 100 & 100 & 500 & 500
\\
 {XWinograd}  & 1858 & 0 & 233 & 0 & 233 & 424
\\
 {XStoryCloze} & 300 & 300 & 60 & 60 &1511  & 1511 \\

\bottomrule
\end{tabular}}
\caption{Number of examples available in XCOPA, XWinograd, and XStoryCloze per language. Since a validation split is not specified in XStoryCloze, we take 60 random examples from the train split for validation.
}
\label{tab:data}
\end{table} 
We develop task-specific models by fine-tuning multilingual pre-trained language models, namely mBERT \citep{devlin-etal-2019-bert} and XLM-R \citep{conneau-etal-2020-unsupervised}, using the generated data.
We then compare their performance against models trained on a limited set of human-created data in the target language whenever available, and otherwise through zero-shot transfer learning from manually created English training data.
Our experiments demonstrate that training the models with {\em relatively large} synthetically generated datasets yields better performance than training with {\em limited} manually-created datasets. This finding empirically confirms the utility of synthetic data generated by LLMs for improving downstream task-specific models.

We expand the multilingual data synthesis using ChatGPT and GPT-4 on XCOPA and find that generating multilingual datasets generally surpasses the effectiveness of the zero-shot cross-lingual transfer.
We further assess the quality of the generated dataset in different languages by asking native speakers to evaluate the naturalness and logical soundness of the generated dataset compared to the human-written examples. 
The annotation results reveal that while ChatGPT and GPT-4 successfully generate natural text in most languages, they struggle with generating understandable text in certain languages such as Tamil.
Moreover, a noticeable gap is observed in terms of commonsense coherence when comparing ChatGPT-generated data to human-constructed data. On the other hand, GPT-4 significantly narrows this difference.

To summarise, our work has the following key contributions:
\begin{itemize}
  \item  Leveraging and prompting four LLMs to augment three low-resource, multilingual commonsense reasoning datasets.
  \item Fine-tuning smaller models, mBERT and XLMR, using the synthesised data and showcasing the practical value of the LLM-generated data.
  \item Performing an extensive analysis of the effects of various target languages in data generation and scaling, as well as a human evaluation of the naturalness and logical coherence of the data generated in various languages.
  \item Releasing the synthesised datasets for public use and reproducibility.
\end{itemize}

\section{Related Work}

We review the related work of the chapter in the following two aspects.
\subsection{Multilingual and Low-Resource NLP}
Recently, there has been increased attention on expanding NLP beyond English, including the development of multilingual models~\citep{devlin-etal-2019-bert, conneau-etal-2020-unsupervised, xue-etal-2021-mt5, scao2022bloom} as well as the creation of benchmarks to address multilingual challenges~\citep{conneau2018xnli, artetxe-etal-2020-cross, adelani-etal-2021-masakhaner, winata-etal-2023-nusax}. Among the prevailing challenges faced across various languages, a common theme is the scarcity of available data. 

Consequently, when data is lacking, one approach is to employ zero-shot cross-lingual transfer. Studies conducted by \citet{winata-etal-2023-nusax} have demonstrated the effectiveness of zero-shot cross-lingual transfer for related languages. Additionally, \citet{muennighoff2022crosslingual} show that models fine-tuned only with English instruction data are capable of understanding multilingual instructions.
In this work, we are tackling a similar scenario where the availability of data is limited.

\subsection{Multilingual Data Augmentation}
\citet{lauscher-etal-2020-zero} show that few-shot can drastically increase the cross-lingual performance of small models, proving that multilingual data augmentation is an effective strategy. 
A series of works try to predict the cross-lingual accuracy of models through measurements and modelling \citep{xia-etal-2020-predicting}, and study strategies for multilingual data augmentation, such as choosing the transfer languages \citep{lin-etal-2019-choosing}, and predicting multilingual few-shot accuracy leading for optimal data augmentation approaches \citep{tool_2022}.

Many works focus on synthetic data augmentation for code-mixing, including utilising linguistic theories \citep{lee19d_interspeech, pratapa-etal-2018-language}, machine translation models \citep{tarunesh-etal-2021-machine}, parallel corpus and Wikipedia \citep{winata-etal-2019-code, whitehouse-etal-2022-entitycs}, and employing ChatGPT \citep{dai2023chataug}. 
\comment{[Chapter 7 - Correction point 2/5]~}\add{To the best of our knowledge, this is the first work that utilises LLMs for multilingual data augmentation, comparing data generation in English and then translating and generating in target languages, with a special focus on the challenging multilingual commonsense reasoning tasks.}

\section{Dataset Augmentation}

This section explains the datasets used in the experiments and the detailed instruction setup.

Our experiments use XCOPA, XWinograd, and XStoryCloze, which are selected due to (i) the limited availability of training data and (ii) commonsense reasoning datasets present greater challenges for data synthesis.
\autoref{tab:data} summarises the statistics of the three datasets.

\paragraph*{XCOPA}\hspace{-.3cm} is a cross-lingual Choice of Plausible Alternatives dataset that translates and re-annotates the validation and test sets of English (EN) COPA \citep{roemmele2011choice} into 11 target languages (ET: Estonian, HT: Haitian Creole, ID: Indonesian, IT: Italian, QU: Quechua, SW: Swahili, TA: Tamil, TH: Thai, TR: Turkish, VI: Vietnamese, and ZH: Chinese).\footnote{\url{https://huggingface.co/datasets/xcopa}}
Each instance consists of a premise, a question
(\textit{cause}/\textit{result}), and two alternatives. The task is to predict the more plausible alternative.

\paragraph*{XWinograd}\hspace{-.3cm} is expanded from the original English Winograd Schema Challenge (WSC) \citep{levesque2012winograd} to five other languages (FR: French, JA: Japanese, PT: Portuguese, RU: Russian, and ZH),\footnote{\url{https://huggingface.co/datasets/Muennighoff/xwinograd}}
which consists of pronoun resolution problems aiming to evaluate the commonsense reasoning ability of a machine.
Given a statement with two noun phrases and a pronoun, the challenge of WSC is to determine the referent of the pronoun, which can only be inferred from the context.

\paragraph*{XStoryCloze}\hspace{-.3cm} is collected by
\citet{lin-etal-2022-shot},
where the validation split of the original English StoryCloze dataset \citep{mostafazadeh-etal-2016-corpus} is translated into 10 other typologically diverse languages (RU, ZH, ES: Spanish, AR: Arabic, HI: Hindi, ID, TE: Telugu, SW, EU: Basque, and MY: Burmese).
Each example consists of a four-sentence commonsense story, a correct ending, as well as a wrong ending.

\subsection{LLMs for Data Generation}
Our preliminary experiments reveal that language models that are specifically fine-tuned on downstream NLP tasks, such as BLOOMZ~\citep{scao2022bloom} and Flan-T5~\citep{chung2022scaling}, struggle to follow the complex instructions. 
Conversely, more recent LLMs such as Dolly-v2, StableVicuna,
ChatGPT, and GPT-4,
which are designed to handle more intricate and general-purpose instructions, have demonstrated success in following our instructions for data generation.
ChatGPT and GPT-4 also stand out with the capability of generating examples in non-English languages. 

 We explore synthetic data generation with the four aforementioned LLMs, balancing between open-access models and closed models (see \autoref{sec:llm-main-results}).
Specifically, we use \texttt{dolly-v2-12b}, which is derived from EleutherAI’s Pythia-12b \citep{biderman2023pythia} and fine-tuned on approximately 15K instructions generated by Databricks employees; and \texttt{StableVicuna-13B}, an RLHF (reinforcement learning from human feedback) fine-tuned Vicuna model on various conversational and instructional datasets - Vicuna is an open-source LLaMA model \citep{touvron2023llama} fine-tuned on user-shared conversations collected from ShareGPT.\footnote{\url{https://github.com/lm-sys/FastChat}}

\subsection{Instructions and Responses}
\label{sec:wino_}
\begin{table}[ht!]
\centering
\scalebox{0.85}{
\addtolength{\tabcolsep}{0pt}
\begin{tabular}{p{4.6cm}|p{4.5cm}|p{5.7cm}}
\toprule
\sc{XCOPA}
& \sc{XWinograd}
& \sc{XStoryCloze}                \\
\midrule

\begin{minipage}[t]{\linewidth}\includegraphics[width=0.4cm]{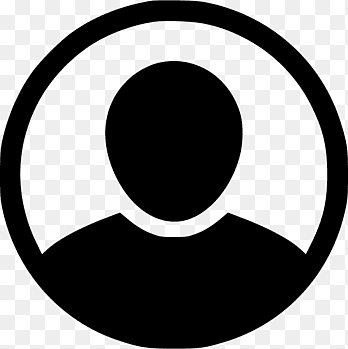} We are collecting more examples for the COPA dataset which will be used to test a system's ability of Commonsense Causal Judgments. The format of the data:\\
A premise: a statement of something that happened, and two choices that could plausibly \{\textit{occur as the result} / \textit{be the cause}\} of the premise. The correct choice is the alternative that is more plausible than the wrong choice. \\
Here are $n$ examples in \texttt{\{language\}}: \\
Example 1: \textcolor{MidnightBlue}{Premise}: The man wanted to save money. What happened as a result?  \textcolor{MidnightBlue}{Correct choice}: He cut back on making frivolous purchases. \textcolor{MidnightBlue}{Wrong choice}: He withdrew money from his savings account.  …
Example $n$: …\\
Based on the examples above, generate $m$ new examples in \texttt{\{language\}}.\end{minipage}
&
\begin{minipage}[t]{\linewidth}\includegraphics[width=0.4cm]{Figures/LLM/user.png} We are collecting more examples for the Winograd Schema Challenge. Each example has a short sentence that contains two noun phrases and one pronoun replaced by ``\_'', and the challenge is to determine the referent of the pronoun, which can only be inferred from the context.\\
Here are $n$ examples of the data:  \\
Example 1:
\textcolor{MidnightBlue}{Sentence}: Harley hides from Dyna because \_ is scary. Who/What is scary? \textcolor{MidnightBlue}{Correct answer}: Dyna.
\textcolor{MidnightBlue}{Wrong answer}: Harley. … Example $n$: … \\
Based on the examples above, generate $m$ new examples. Both noun phrases in each example can be males, females, inanimate objects, or groups of people or objects. There should only be one ``\_'' in the sentence. The correct and wrong answer should be one of the noun phrases mentioned in the sentence.\end{minipage}
&
\begin{minipage}[t]{\linewidth}\includegraphics[width=0.4cm]{Figures/LLM/user.png} We are collecting more examples for the Story Cloze dataset. Each example consists of a 4-sentence story, one correct ending sentence which is a plausible continuation of the story, and one wrong ending sentence which is logically inconsistent with the context. \\
Here are $n$ examples of the data:\\
Example 1:
\textcolor{MidnightBlue}{Sent-1}: Tina is very tired every single morning. \textcolor{MidnightBlue}{Sent-2}: She does not get enough sleep because of her two jobs. \textcolor{MidnightBlue}{Sent-3}: Tina decides to quit one of the jobs. \textcolor{MidnightBlue}{Sent-4}: She now gets enough sleep to function everyday. \textcolor{MidnightBlue}{Correct ending}: Tina is well rested. \textcolor{MidnightBlue}{Wrong ending}: Tina is more tired than ever before. … Example $n$: …\\
Based on the examples above, provide $m$ new similar examples. Requirements: 1) the story should read like a coherent story, with a specific beginning and ending, where something happens in between 2) both ending sentences should be entirely reasonable, realistic and sensible when read in isolation, and 3) both ending sentences should follow up the story by sharing at least one of the characters of the story.\end{minipage}
\\
\midrule
\begin{minipage}[t]{\linewidth}\includegraphics[width=0.4cm]{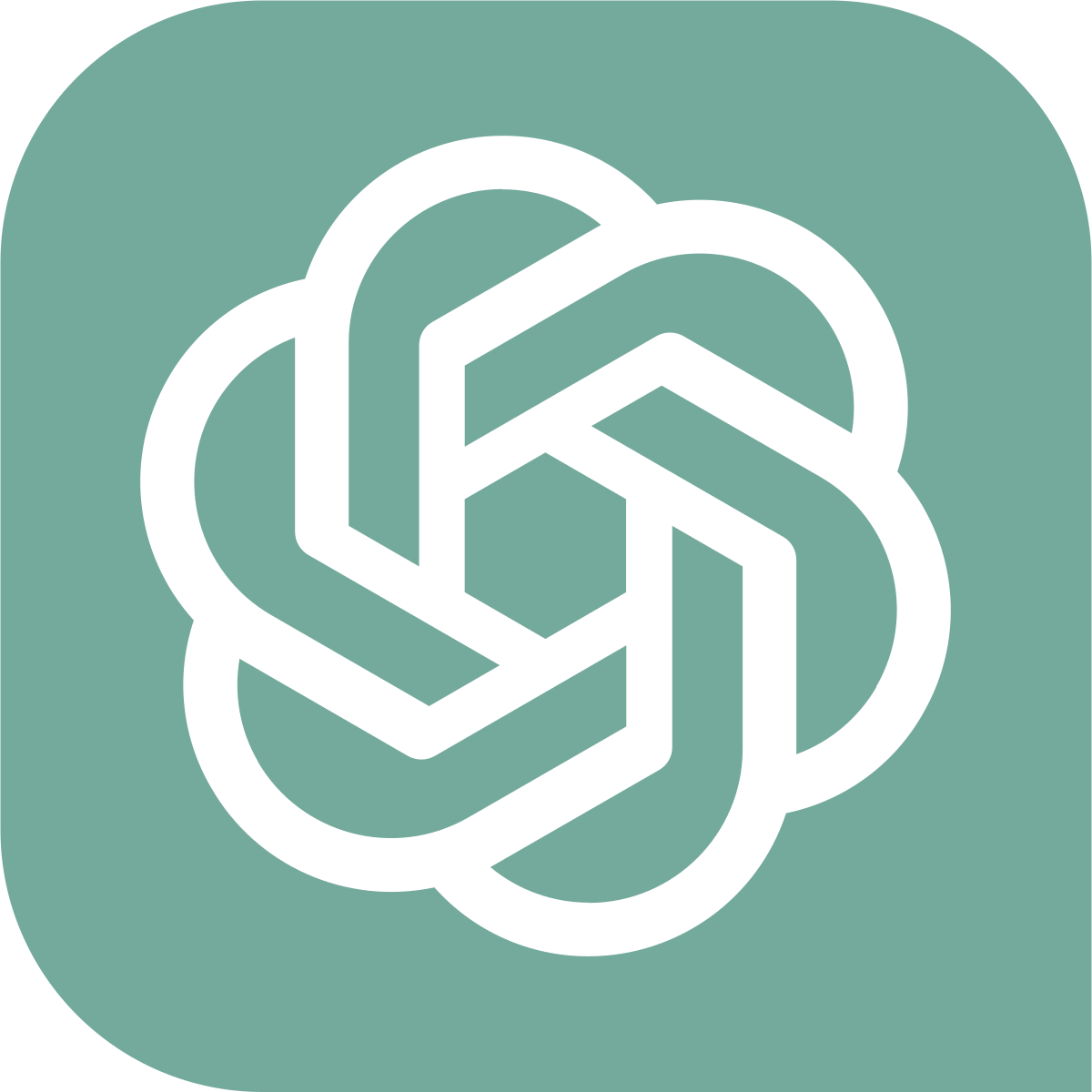}
\textcolor{MidnightBlue}{Premise}: The politician made a controversial statement. What happened as a result?
\textcolor{MidnightBlue}{Correct choice}: The politician faced criticism from the media.
\textcolor{MidnightBlue}{Wrong choice}: The politician's approval ratings increased.\\
\includegraphics[width=0.4cm]{Figures/LLM/chatgpt.png}
\textcolor{MidnightBlue}{Premise}: \begin{CJK}{UTF8}{gbsn}我裤子口袋里的钥匙不见了。\end{CJK} What was the cause? \textcolor{MidnightBlue}{Correct choice}: \begin{CJK}{UTF8}{gbsn}这个口袋上有一个洞。\end{CJK} \textcolor{MidnightBlue}{Wrong choice}: \begin{CJK}{UTF8}{gbsn}裤子是新的。\end{CJK}\end{minipage}
&
\begin{minipage}[t]{\linewidth}\includegraphics[width=0.4cm]{Figures/LLM/chatgpt.png} \textcolor{MidnightBlue}{Sentence}: Sam gave Andrew the book because \_ had already read it. Who/What had already read the book?
\textcolor{MidnightBlue}{Correct answer}: Sam.
\textcolor{MidnightBlue}{Wrong answer}: Andrew. \\
\includegraphics[width=0.4cm]{Figures/LLM/chatgpt.png} \textcolor{MidnightBlue}{Sentence}: The dog chased the cat , but \_ was too fast. Who/What was too fast? \textcolor{MidnightBlue}{Correct answer}: the cat. \textcolor{MidnightBlue}{Wrong answer}: The dog.\end{minipage}
&
\begin{minipage}[t]{\linewidth}\includegraphics[width=0.4cm]{Figures/LLM/chatgpt.png} \textcolor{MidnightBlue}{Sent-1}: Jordan was a high school student who wanted to become a doctor. \textcolor{MidnightBlue}{Sent-2}: He spent all his free time studying biology and chemistry. \textcolor{MidnightBlue}{Sent-3}: One day, his school hosted a science fair competition. \textcolor{MidnightBlue}{Sent-4}: Jordan's project won first place. \textcolor{MidnightBlue}{Correct ending}: Jordan went on to study medicine in college. \textcolor{MidnightBlue}{Wrong ending}: Jordan gave up his dream of becoming a doctor.\end{minipage}  \\
\bottomrule
\end{tabular}
}
\caption{Examples of instructions and LLM-responses for XCOPA, XWinograd, and XStoryCloze.
We use ChatGPT for demonstration.
}
\label{tab:instructions}
\end{table}

We utilise LLMs to generate synthetic examples for all datasets by prompting them.
We construct instructions using the descriptions from the dataset papers as a reference and provide LLMs with some examples, randomly sampled from the  \textit{train (+validation)} split of the original dataset, then ask LLMs to generate similar data points.
We experiment with various instructions and evaluate the synthesised data on a smaller scale, update the instructions based on the errors, and then choose the best instruction to generate the final datasets. 

The final instructions and responses are in \autoref{tab:instructions}. 
Our data generation process comprises the following key steps:
(i) We establish the desired total number of examples to generate. This quantity can be determined by various factors such as budget constraints, a fixed ratio concerning the original dataset, etc.
(ii) We proceed to generate examples through the following iterative process:
(a) To ensure diversity, we randomly sample a set of $n$ examples from the training datasets.
(b) We append these sampled examples to the instructions and prompt the model to generate an additional set of $m$ new examples.
(c) Afterwards, we perform post-processing and only add valid and unique examples to the generated set.
Typically, the values of $n$ and $m$ are set to 5 to 10.

\begin{table}[!t]
\centering
\scalebox{0.95}{
\addtolength{\tabcolsep}{0pt}
\begin{tabular}{l|ccc}
    \toprule
         \textbf{Model} &  \textbf{XCOPA} &  \textbf{XWinograd} & \textbf{XStoryCloze} \\
    \midrule
       \sc  Dolly-v2 & 41.6\% & 22.4\% & 41.2\% \\
        \sc StableVicuna & 36.1\%& 33.8\% & 36.1\% \\
       \sc  ChatGPT & 86.4\% & 43.8\% & 77.6\%\\
       \sc  GPT-4 & 89.7\% & 85.0\% & 89.3\%\\
    \bottomrule
\end{tabular}}
\caption{Generation Success Rate in English (valid examples obtained / total examples requested) with different LLMs on the three datasets. }
\label{tab:success}
\end{table}

We focus on a fixed-budget scenario and first generate a total of 3-4K data points for each dataset with LLMs. 
LLMs tend to generate fewer samples than requested or inconsistent output in invalid formats.
We report the success rate for different LLMs on the three datasets in \autoref{tab:success}, which indicates that GPT-4 has the most robustness.

Among the datasets, LLMs have the lowest generation success rate for XWinograd, which is more challenging.
XWinograd requires both answers to be from the generated sentence, with only one pronoun being replaced. 
One failed example in generated XWinograd: \textcolor{MidnightBlue}{Sentence}: ''The computer crashed and \_ lost all of their files''. \textcolor{MidnightBlue}{Correct answer}: the user. \textcolor{MidnightBlue}{Wrong answer}: the computer.
In addition, we observed pronoun inconsistency in the generated XWinograd data. Despite the requirement for interchangeable pronouns in the options, models frequently fail to comply.
For example, ``The dog bit the mailman because \_ entered the yard.'' is generated by ChatGPT with the options ``The dog'' or ``the mailman'', however, ``\_'' in the sentence cannot be replaced by the same pronoun for the given two options, hence it may make the task easier and the example is considered sub-optimal.

\comment{[Chapter 7 - Correction point 3/5]~}
\add{Despite multiple instructions emphasizing the necessity of maintaining consistency between the two phrases to be replaced, all Language Model Models (LLMs) experimented with, failed to consistently follow this requirement. The best-performing instruction, as illustrated in \autoref{tab:instructions}, still fell short of achieving perfect consistency. We retain these instances within the dataset and further include a human evaluation in \autoref{sec:human}. Specifically, we found that among the LLMs studied, GPT-4 demonstrated the highest level of consistency, followed by Chat-GPT, with 76.6\% and 48.9\% of the annotated examples exhibiting adherence to the rule, respectively, as confirmed by annotations from native speakers.}

\subsection{Topic Diversity}
As the StoryCloze dataset contains more sentences and has richer content, we analyse the diversity of the generation as well as topic coverage and the most frequent events, following 
Specifically, an event is counted as any hyponym of ``event'' or ``process'' in WordNet. 

\begin{figure}

\centering
    \includegraphics[width=0.95\linewidth]{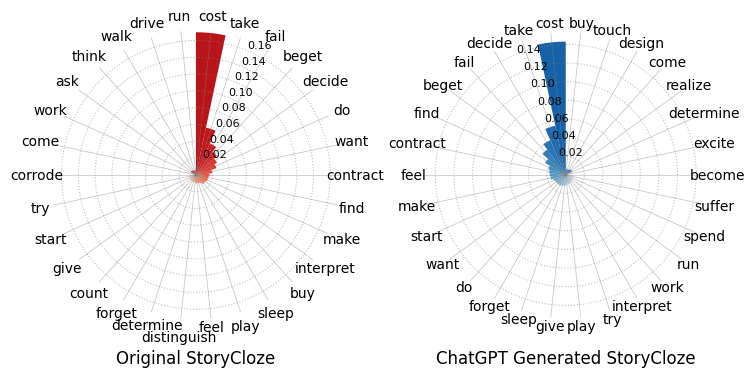}
\caption{Comparison between the 30 most frequent events in the original and the ChatGPT-generated English StoryCloze dataset.}
\label{fig:story}
\end{figure}

This helps us to determine whether LLM-generated data can capture the corpus distribution by randomly sampling $n$ examples from the dataset in the instructions.
ChatGPT-generated data is used for demonstration.

In \autoref{fig:story}, we present the results of comparing the generated data points with the original 300 train set used as few-shot examples in the generation instructions.
We can see that 23 of the 30 most frequent events in the original dataset can also be found in the 30 most frequent events of the ChatGPT-generated data.

\section{Experimental Setup}
\label{sec:exp}

We first generate synthetic English examples for XCOPA, XWinograd, and XStoryCloze, with Dolly-v2,  StableVicuna, ChatGPT, and GPT-4.
The size of the final filtered synthesised data for the three datasets is 3.7K, 2K, and 1.7K, respectively. 
We then fine-tune mBERT, XLMR-base, and XLMR-large with the synthesised data and compare the zero-shot cross-lingual transfer performance across different languages, where we use the original validation set in target languages.

For XCOPA, we additionally experiment with generating data points directly in non-English languages, by providing examples in the target language and specifying the language desired for the generated data (see \autoref{tab:instructions}). 
However, since no examples for \textit{cause} are included in TH and TR train/validation data (they do appear in the test split), we do not generate XCOPA for the two languages. We use ChatGPT and GPT-4 for multilingual synthetic data generation, as both Dolly-v2 and StableVicuna exhibit limitations in effectively generating multilingual text.
The size of the multilingual synthesised data is \mytilde3.6K in each language.

We fine-tune models on all datasets as multiple-choice tasks\footnote{In our preliminary experiments, we find that formulating XWinograd as a binary text classification results poorly, in line with the observation from \citet{liu-etal-2020-precise} that the task formulation is essential to the performance of Winograd.} by searching the best learning rate from \{$5e^{-6}$, $10e^{-6}$\}, and batch size from \{8, 16, 32\}.
All the fine-tuning experiments are conducted on a single 40G A100. 
For generating data with Dolly-v2 and StableVicuna, we use 2$\times$40G A100. 
Additionally, for XLMR-large trained on the original XWinograd dataset, we found that a lower learning rate was necessary. Therefore, we performed an additional tuning process with a learning rate of $10e^{-7}$.

\section{Results and Discussion}
We now present the main results of fine-tuned models on the three datasets and compare performance with generated data in different LLMs, languages, and scales.
\begin{table*}[t]
\centering
\scalebox{0.8}{
\addtolength{\tabcolsep}{-3.5pt}
\begin{tabular}{ll|ccccccccc}
\toprule
{\multirow{2}{*}{\textbf{\parbox{0.12\textwidth}{Finetuned Model}}}} &
{\multirow{2}{*}{\textbf{ \parbox{0.11\textwidth}{LLM for \\Generation}}}} & \multicolumn{3}{c}{\sc \textbf{Xcopa}}                       
  & \multicolumn{3}{c}{\sc \textbf{XWinograd} }                
  & \multicolumn{3}{c}{\sc  \textbf{XStoryCloze}}                   \\
 
\cmidrule(lr){3-5}  \cmidrule(lr){6-8}  \cmidrule(l){9-11}
& & 
 {\textit{ORI}}\textcolor{MidnightBlue}{$_{ 400}$} 
 & {\textit{GEN}}\textcolor{MidnightBlue}{$_{ 3.7k}$} 
 & {\textit{O+G}}\textcolor{MidnightBlue}{$_{ 4.1k}$} 
 &{\textit{ORI}}\textcolor{MidnightBlue}{$_{ 1.8k}$}
 & {\textit{GEN}}\textcolor{MidnightBlue}{$_{ 2k}$}
 & {\textit{O+G}}\textcolor{MidnightBlue}{$_{ 3.8k}$} 
  &{\textit{ORI}}\textcolor{MidnightBlue}{$_{ 300}$} 
 &{\textit{GEN}}\textcolor{MidnightBlue}{$_{ 1.7k}$}
 &{\textit{O+G}}\textcolor{MidnightBlue}{$_{ 2k}$}
 \\
 \midrule
\multirow{4}{*}{mBERT}      
& Dolly-v2     
& 47.9 
& 53.3\textcolor{DarkGreen}{\textsubscript{~$\uparrow$5.4}} 
& 54.0\textcolor{DarkGreen}{\textsubscript{~$\uparrow$6.1}}
& 52.9
&  \textbf{59.6}\textcolor{DarkGreen}{\textsubscript{~$\uparrow$6.7}}
&  \textbf{59.3}\textcolor{DarkGreen}{\textsubscript{~$\uparrow$6.4}}
& 65.0 
& \textbf{68.7}\textcolor{DarkGreen}{\textsubscript{~$\uparrow$3.7}}  
& 68.1\textcolor{DarkGreen}{\textsubscript{~$\uparrow$3.1}} 
\\

& StableVicuna 
 & 47.9 
 & 52.9\textcolor{DarkGreen}{\textsubscript{~$\uparrow$5.0}}
 & 54.7\textcolor{DarkGreen}{\textsubscript{~$\uparrow$6.8}}
 & 52.9 
 & 53.7\textcolor{DarkGreen}{\textsubscript{~$\uparrow$0.8}}
 & 58.5\textcolor{DarkGreen}{\textsubscript{~$\uparrow$5.6}}  
 & 65.0 
 & 64.6\textcolor{Maroon}{\textsubscript{~$\downarrow$0.4}}
 & 67.3\textcolor{DarkGreen}{\textsubscript{~$\uparrow$2.3}} 
   \\
 & ChatGPT      
 & 47.9 
 & 55.0\textcolor{DarkGreen}{\textsubscript{~$\uparrow$7.1}} 
 & 54.1\textcolor{DarkGreen}{\textsubscript{~$\uparrow$6.2}} 
 & 52.9 
 & 56.0\textcolor{DarkGreen}{\textsubscript{~$\uparrow$3.1}} 
 & 58.3\textcolor{DarkGreen}{\textsubscript{~$\uparrow$5.4}} 
 & 65.0 
 & 64.3\textcolor{Maroon}{\textsubscript{~$\downarrow$0.7}} 
 & 68.3\textcolor{DarkGreen}{\textsubscript{~$\uparrow$3.3}} 
 \\
& GPT-4    
& 47.9 
& \textbf{56.4}\textcolor{DarkGreen}{\textsubscript{~$\uparrow$8.5}}  
& \textbf{57.2}\textcolor{DarkGreen}{\textsubscript{~$\uparrow$9.3}}  
& 52.9 
& 54.9\textcolor{DarkGreen}{\textsubscript{~$\uparrow$2.0}}  
& 57.5\textcolor{DarkGreen}{\textsubscript{~$\uparrow$4.6}} 
& 65.0 
& 68.0\textcolor{DarkGreen}{\textsubscript{~$\uparrow$3.0}}  
& \textbf{69.8}\textcolor{DarkGreen}{\textsubscript{~$\uparrow$4.8}}    
\\
\midrule
{\multirow{4}{*}{\parbox{0.1\textwidth}{XLMR-\\Base}}}
& Dolly-v2      
&54.8 
& 58.1\textcolor{DarkGreen}{\textsubscript{~$\uparrow$3.3}}  
& 58.1\textcolor{DarkGreen}{\textsubscript{~$\uparrow$3.3}}  
& 53.5 
& 56.5\textcolor{DarkGreen}{\textsubscript{~$\uparrow$3.0}}   
&  66.3\textcolor{DarkGreen}{\textsubscript{~$\uparrow$12.8}}
& 73.0 
& 75.8\textcolor{DarkGreen}{\textsubscript{~$\uparrow$2.8}}  
& 76.5\textcolor{DarkGreen}{\textsubscript{~$\uparrow$3.5}}   
\\
& StableVicuna 
& 54.8 
& 57.6\textcolor{DarkGreen}{\textsubscript{~$\uparrow$2.8}} 
& 59.3\textcolor{DarkGreen}{\textsubscript{~$\uparrow$4.5}} 
& 53.5 
& 59.0\textcolor{DarkGreen}{\textsubscript{~$\uparrow$5.5}} 
& 66.0\textcolor{DarkGreen}{\textsubscript{~$\uparrow$12.5}} 
& 73.0 
& 69.6\textcolor{Maroon}{\textsubscript{~$\downarrow$3.4}} 
& 74.2\textcolor{DarkGreen}{\textsubscript{~$\uparrow$1.2}}
\\
 & ChatGPT     
& 54.8 
& 58.2\textcolor{DarkGreen}{\textsubscript{~$\uparrow$3.4}} 
& 59.4\textcolor{DarkGreen}{\textsubscript{~$\uparrow$4.6}} 
& 53.5 
& 62.7\textcolor{DarkGreen}{\textsubscript{~$\uparrow$9.2}} 
& 65.9\textcolor{DarkGreen}{\textsubscript{~$\uparrow$12.4}} 
& 73.0 
& 67.4\textcolor{Maroon}{\textsubscript{~$\downarrow$5.6}} 
& 74.5\textcolor{DarkGreen}{\textsubscript{~$\uparrow$1.5}}  
\\
& GPT-4        
& 54.8 
& \textbf{62.7}\textcolor{DarkGreen}{\textsubscript{~$\uparrow$7.9}} 
& \textbf{63.0}\textcolor{DarkGreen}{\textsubscript{~$\uparrow$8.2}}  
& 53.5 
& \textbf{63.3}\textcolor{DarkGreen}{\textsubscript{~$\uparrow$9.8}}
& \textbf{66.9}\textcolor{DarkGreen}{\textsubscript{~$\uparrow$13.4}} 
& 73.0
& \textbf{74.6}\textcolor{DarkGreen}{\textsubscript{~$\uparrow$1.6}} 
& \textbf{79.3}\textcolor{DarkGreen}{\textsubscript{~$\uparrow$6.3}}       
\\
 \midrule
{\multirow{4}{*}{\parbox{0.1\textwidth}{XLMR-\\Large}}}

& Dolly-v2  
&   63.0 
& 58.6\textcolor{Maroon}{\textsubscript{~$\downarrow$4.4}} 
& 65.0\textcolor{DarkGreen}{\textsubscript{~$\uparrow$2.0}} 
& 80.1 
&  \textbf{76.9}\textcolor{Maroon}{\textsubscript{~$\downarrow$3.2}} 
&  83.1\textcolor{DarkGreen}{\textsubscript{~$\uparrow$3.0}} 
& 85.0
& 84.8\textcolor{Maroon}{\textsubscript{~$\downarrow$0.2}} 
& 86.4\textcolor{DarkGreen}{\textsubscript{~$\uparrow$1.4}}   
\\
& StableVicuna 
&   63.0 
&64.4\textcolor{DarkGreen}{\textsubscript{~$\uparrow$1.4}}  
& 68.7\textcolor{DarkGreen}{\textsubscript{~$\uparrow$5.7}}  
& 80.1 
& 68.2\textcolor{Maroon}{\textsubscript{~$\downarrow$11.9}}  
& 82.0\textcolor{DarkGreen}{\textsubscript{~$\uparrow$1.9}}  
& 85.0 
& 74.6\textcolor{Maroon}{\textsubscript{~$\downarrow$10.4}}  
& 84.8\textcolor{Maroon}{\textsubscript{~$\downarrow$0.2}}   
\\
 & ChatGPT      
& 63.0 
& 64.6\textcolor{DarkGreen}{\textsubscript{~$\uparrow$1.6}} 
& 68.1\textcolor{DarkGreen}{\textsubscript{~$\uparrow$5.1}} 
& 80.1 
& 73.2\textcolor{Maroon}{\textsubscript{~$\downarrow$6.9}} 
& 83.2\textcolor{DarkGreen}{\textsubscript{~$\uparrow$3.1}}
& 85.0 
& 77.3\textcolor{Maroon}{\textsubscript{~$\downarrow$7.7}}
& 85.8\textcolor{DarkGreen}{\textsubscript{~$\uparrow$0.8}}
\\
& GPT-4       
& 63.0 
& \textbf{72.1}\textcolor{DarkGreen}{\textsubscript{~$\uparrow$9.1}}
& \textbf{72.2}\textcolor{DarkGreen}{\textsubscript{~$\uparrow$9.2}} 
& 80.1 
& 76.4\textcolor{Maroon}{\textsubscript{~$\downarrow$3.7}}
& \textbf{83.5}\textcolor{DarkGreen}{\textsubscript{~$\uparrow$3.4}}
& 85.0 
& \textbf{86.0}\textcolor{DarkGreen}{\textsubscript{~$\uparrow$1.0}} 
& \textbf{88.4}\textcolor{DarkGreen}{\textsubscript{~$\uparrow$3.4}} 
\\
\bottomrule
\end{tabular}
}
\caption{Comparison of Average Accuracy across all languages for mBERT, XLMR-Base, and XLMR-Large on XCOPA, XStoryCloze, and XWinograd. Training datasets include {\textit{ORI}} (original EN data), {\textit{GEN}} (LLM-generated EN data), and {\textit{O+G}} (both), with the number of examples used for training indicated by the subscripts. The best results obtained with the same amount of training data are highlighted in bold. Green and red subscripts denote improvement and decline in performance compared to the baseline (\textit{ORI}). See per language results in \autoref{sec:additional}.
}
 \label{tab:main}
\end{table*}

\subsection{General Result}

\label{sec:llm-main-results}
\autoref{tab:main} presents the average accuracy of fine-tuned mBERT, XLMR-Base, and XLMR-Large models across all languages on the three datasets.
The models are trained using original data (\textit{ORI}), different LLM-generated data (\textit{GEN}), as well as a combination of both sources (\textit{O+G}) in English. 

Across different datasets, LLMs, and fine-tuned models, consistent improvements are observed when using both original and LLM-generated data. Among the models, Dolly-v2 performs the best on XWinograd when fine-tuned on mBERT, while GPT-4 achieves the highest accuracy in other settings.
The most significant improvement is shown in XWinograd with XLMR-Base, where the addition of an extra 2k data points leads to an average accuracy enhancement of 12.8 compared to the baseline, across all four LLMs.

\comment{[Chapter 7 - Correction point 4/5]~}
\add{We observe the following from the results: 
(i) In instances where the baseline performance is below 60, exemplified by XCOPA and XWinograd tasks using small models like mBERT and XLMR-Base, leveraging LLM-generated data either solely or in conjunction with original data consistently enhances performance across the board;  
(ii)  As the baseline performance increases, notably exceeding 80 accuracy as seen in the case of XStoryCloze, relying solely on LLM-generated data might adversely impact performance. However, even in such cases, utilising data generated solely by GPT-4 still yields improvements across all fine-tuned models;
(iii) Combining original data with LLM-generated data consistently demonstrates enhancements over using original data alone, irrespective of the baseline performance or the specific datasets and fine-tuned models employed. This trend holds true across various scenarios, with the exception of XStoryCloze with XLMR-Large, where there's only a marginal 0.2 score difference.}

\add{Overall, we can see that LLM-based data augmentation particularly benefits scenarios where baseline performance is moderate to low, while stronger LLMs show more promises in generating high-quality data, as exemplified by GPT-4.}

For smaller models with 
When using only LLM-generated data, smaller models like mBERT and XLMR-Base generally outperform the baseline. However, with XLMR-Large, which achieves stronger baselines, e.g., >80 in XWinograd and XStoryCloze, the accuracy remains similar or even worse compared to using the original data.
GPT-4-generated data demonstrates the best robustness but still experiences a decline in performance in XWinograd when the generated data size is similar to the original data. This highlights the challenges of generating data at a human-level quality.

\subsection{Multilingual Data Generation}

We investigate whether the synthetically generated multilingual dataset outperforms training solely in English. 
We choose the XCOPA dataset and explore two settings: synthetic multilingual data by asking LLMs to generate responses in the target languages directly and translating the English-generated data to target languages with Google Translate API. 
We exclude Dolly-v2 and StableVicuna due to their limited effectiveness in generating non-English text. 
Although GPT-4 exhibits the most promising performance, it is significantly costlier compared to ChatGPT. Therefore, we also consider ChatGPT as a contrasting experiment under resource-constrained conditions.

\begin{table*}[t]
\centering
\scalebox{0.8}{
\addtolength{\tabcolsep}{-1.5pt}
\begin{tabular}{lll|cccccccccc}
\toprule
\textbf{Finetuned} &
  \textbf{LLM} &
  \textbf{Training data} &
  \multicolumn{1}{l}{\textbf{AVG}} &
  \multicolumn{1}{l}{\textbf{EN}} &
  \multicolumn{1}{l}{\textbf{ET}} &
  \multicolumn{1}{l}{\textbf{HT}} &
  \multicolumn{1}{l}{\textbf{ID}} &
  \multicolumn{1}{l}{\textbf{IT}} &
  \multicolumn{1}{l}{\textbf{SW}} &
  \multicolumn{1}{l}{\textbf{TA}} &
  \multicolumn{1}{l}{\textbf{VI}} &
  \multicolumn{1}{l}{\textbf{ZH}} \\
  \midrule
 &
    Baseline &
  \small{$ORI$} &
  \cellcolor[HTML]{FFFFFF}47.2 &
  \cellcolor[HTML]{FFFFFF}53.8 &
  \cellcolor[HTML]{FFFFFF}44.2 &
  \cellcolor[HTML]{FFFFFF}48.6 &
  \cellcolor[HTML]{FFFFFF}47.2 &
  \cellcolor[HTML]{FFFFFF}46.2 &
  \cellcolor[HTML]{FFFFFF}45.4 &
  \cellcolor[HTML]{FFFFFF}48.4 &
  \cellcolor[HTML]{FFFFFF}43.6 &
  \cellcolor[HTML]{FFFFFF}47.4 \\
  \cmidrule{2-13}
 &
   &
  \small{$GEN_{EN}+ORI$} &
  \cellcolor[HTML]{A2DABF}54.6 &
  \cellcolor[HTML]{AEDFC7}59.6 &
  \cellcolor[HTML]{83CDA9}56.4 &
  \cellcolor[HTML]{B9E3CF}53.6 &
  \cellcolor[HTML]{A8DCC3}53.8 &
  \cellcolor[HTML]{B7E2CD}51.4 &
  \cellcolor[HTML]{AADDC5}51.6 &
  \cellcolor[HTML]{E3F4EC}50.4 &
  \cellcolor[HTML]{88CFAD}55.0 &
  \cellcolor[HTML]{85CEAB}59.2 \\
 &
   &
    \small{$GEN_{XX}+ORI$} &
  \cellcolor[HTML]{94D4B5}56.8 &
  \cellcolor[HTML]{AEDFC7}59.6 &
  \cellcolor[HTML]{73C79E}58.8 &
  \cellcolor[HTML]{ABDDC5}54.6 &
  \cellcolor[HTML]{98D6B8}56.2 &
  \cellcolor[HTML]{70C69C}61.2 &
  \cellcolor[HTML]{96D5B7}54.6 &
  \cellcolor[HTML]{B7E2CD}53.6 &
  \cellcolor[HTML]{9CD7BA}52.0 &
  \cellcolor[HTML]{7FCBA6}60.2 \\
 &
  \multirow{-3}{*}{ChatGPT} &
\small{$GEN_{EN}^{Trans}+ORI$} &
  \cellcolor[HTML]{87CFAC}58.7 &
  \cellcolor[HTML]{AEDFC7}59.6 &
  \cellcolor[HTML]{6CC499}59.8 &
  \cellcolor[HTML]{94D4B5}58.2 &
  \cellcolor[HTML]{6CC499}62.8 &
  \cellcolor[HTML]{72C69D}61.0 &
  \cellcolor[HTML]{A4DAC0}52.6 &
  \cellcolor[HTML]{9CD7BA}56.8 &
  \cellcolor[HTML]{73C79E}58.2 &
  \cellcolor[HTML]{84CEAA}59.4 \\
  \cmidrule{2-13}
 &
   &
  \small{$GEN_{EN}+ORI$} &
  \cellcolor[HTML]{84CDA9}59.3 &
  \cellcolor[HTML]{57BB8A}72.6 &
  \cellcolor[HTML]{73C79E}58.8 &
  \cellcolor[HTML]{C2E7D5}53.0 &
  \cellcolor[HTML]{72C69D}62.0 &
  \cellcolor[HTML]{72C69D}61.0 &
  \cellcolor[HTML]{BFE5D3}50.0 &
  \cellcolor[HTML]{B1E0C9}54.0 &
  \cellcolor[HTML]{77C8A1}57.6 &
  \cellcolor[HTML]{62C092}64.6 \\
 &
   &
    \small{$GEN_{XX}+ORI$} &
\cellcolor[HTML]{6CC499}61.8 &
  \cellcolor[HTML]{57BB8A}72.6 &
  \cellcolor[HTML]{63C093}61.2 &
  \cellcolor[HTML]{94D4B5}58.2 &
  \cellcolor[HTML]{70C69C}62.2 &
  \cellcolor[HTML]{57BB8A}66.4 &
  \cellcolor[HTML]{84CEAA}57.4 &
  \cellcolor[HTML]{C5E8D7}53.4 &
  \cellcolor[HTML]{57BB8A}63.0 &
  \cellcolor[HTML]{72C69D}61.8 \\
\multirow{-7}{*}{mBERT} &
  \multirow{-3}{*}{GPT-4} &
\small{$GEN_{EN}^{Trans}+ORI$} &
  \cellcolor[HTML]{6EC49A}62.6 &
  \cellcolor[HTML]{57BB8A}72.6 &
  \cellcolor[HTML]{74C79F}58.6 &
  \cellcolor[HTML]{A8DCC3}55.2 &
  \cellcolor[HTML]{5ABD8C}65.6 &
  \cellcolor[HTML]{57BB8A}65.4 &
  \cellcolor[HTML]{9CD7BA}53.8 &
  \cellcolor[HTML]{76C8A0}62.6 &
  \cellcolor[HTML]{57BB8A}64.6 &
  \cellcolor[HTML]{5DBE8E}65.4 \\
  \cmidrule{1-13}
 & Baseline &
  \small{$ORI$} &
  \cellcolor[HTML]{FFFFFF}55.6 &
  \cellcolor[HTML]{FFFFFF}57.6 &
  \cellcolor[HTML]{FFFFFF}54.6 &
  \cellcolor[HTML]{FFFFFF}50.6 &
  \cellcolor[HTML]{FFFFFF}59.6 &
  \cellcolor[HTML]{FFFFFF}54.8 &
  \cellcolor[HTML]{FFFFFF}55.0 &
  \cellcolor[HTML]{FFFFFF}53.4 &
  \cellcolor[HTML]{FFFFFF}54.8 &
  \cellcolor[HTML]{FFFFFF}59.6 \\
  \cmidrule{2-13}
 &
   &
  \small{$GEN_{EN}+ORI$} &
  \cellcolor[HTML]{C4E7D6}59.8 &
  \cellcolor[HTML]{AADDC5}63.8 &
  \cellcolor[HTML]{A5DBC1}61.6 &
  \cellcolor[HTML]{F1FAF6}51.6 &
  \cellcolor[HTML]{D5EEE2}62.6 &
  \cellcolor[HTML]{B9E3CF}59.8 &
  \cellcolor[HTML]{F4CDCD}51.6 &
  \cellcolor[HTML]{A5DBC1}60.4 &
  \cellcolor[HTML]{91D3B3}64.8 &
  \cellcolor[HTML]{DEF2E8}62.0 \\
 &
   &
    \small{$GEN_{XX}+ORI$} &
  \cellcolor[HTML]{C3E7D6}59.9 &
  \cellcolor[HTML]{AADDC5}63.8 &
  \cellcolor[HTML]{ABDDC5}60.6 &
  \cellcolor[HTML]{C2E7D5}55.0 &
  \cellcolor[HTML]{BAE3CF}64.6 &
  \cellcolor[HTML]{BCE4D1}59.6 &
  \cellcolor[HTML]{FFFFFF}54.6 &
  \cellcolor[HTML]{D5EEE2}56.4 &
  \cellcolor[HTML]{BCE4D1}59.6 &
  \cellcolor[HTML]{B7E2CD}64.8 \\
 &
  \multirow{-3}{*}{ChatGPT} &
\small{$GEN_{EN}^{Trans}+ORI$} &
  \cellcolor[HTML]{B1E0CA}61.1 &
  \cellcolor[HTML]{AADDC5}63.8 &
  \cellcolor[HTML]{B4E1CB}60.0 &
  \cellcolor[HTML]{A2DABF}58.0 &
  \cellcolor[HTML]{B4E1CB}65.0 &
  \cellcolor[HTML]{ABDDC5}60.8 &
  \cellcolor[HTML]{FCF1F1}53.8 &
  \cellcolor[HTML]{A6DBC2}60.2 &
  \cellcolor[HTML]{A0D9BD}62.6 &
  \cellcolor[HTML]{A9DCC4}66.0 \\
  \cmidrule{2-13}
 &
   &
  \small{$GEN_{EN}+ORI$} &
  \cellcolor[HTML]{9ED8BC}63.6 &
  \cellcolor[HTML]{84CEAA}69.6 &
  \cellcolor[HTML]{96D5B7}63.8 &
  \cellcolor[HTML]{F7FCFA}51.2 &
  \cellcolor[HTML]{A1D9BE}67.2 &
  \cellcolor[HTML]{A1D9BE}62.4 &
  \cellcolor[HTML]{D0ECDF}58.4 &
  \cellcolor[HTML]{8FD2B1}63.8 &
  \cellcolor[HTML]{84CEAA}66.8 &
  \cellcolor[HTML]{93D3B4}69.4 \\
 &
   &
    \small{$GEN_{XX}+ORI$} &
  \cellcolor[HTML]{99D6B9}64.0 &
  \cellcolor[HTML]{84CEAA}69.6 &
  \cellcolor[HTML]{A1D9BE}62.2 &
  \cellcolor[HTML]{B1E0C9}56.2 &
  \cellcolor[HTML]{98D6B8}68.6 &
  \cellcolor[HTML]{98D6B8}63.8 &
  \cellcolor[HTML]{D8F0E4}57.8 &
  \cellcolor[HTML]{A2DABF}61.2 &
  \cellcolor[HTML]{84CEAA}66.8 &
  \cellcolor[HTML]{8FD2B1}70.0 \\
\multirow{-7}{*}{XLMR-Base} 
&
  \multirow{-3}{*}{GPT-4} &
\small{$GEN_{EN}^{Trans}+ORI$} &
  \cellcolor[HTML]{9CD7BB}63.9 &
  \cellcolor[HTML]{84CEAA}69.6 &
  \cellcolor[HTML]{A5DBC1}61.6 &
  \cellcolor[HTML]{ABDDC5}56.6 &
  \cellcolor[HTML]{99D6B9}68.4 &
  \cellcolor[HTML]{8FD2B1}65.2 &
  \cellcolor[HTML]{D3EDE1}58.2 &
  \cellcolor[HTML]{A6DBC2}60.2 &
  \cellcolor[HTML]{89D0AE}66.0 &
  \cellcolor[HTML]{91D3B3}69.6 \\
  \cmidrule{1-13}
 & Baseline &
  \small{$ORI$} &
  \cellcolor[HTML]{FFFFFF}64.4 &
  \cellcolor[HTML]{FFFFFF}71.4 &
  \cellcolor[HTML]{FFFFFF}62.8 &
  \cellcolor[HTML]{FFFFFF}51.4 &
  \cellcolor[HTML]{FFFFFF}69.0 &
  \cellcolor[HTML]{FFFFFF}65.8 &
  \cellcolor[HTML]{FFFFFF}60.6 &
  \cellcolor[HTML]{FFFFFF}62.0 &
  \cellcolor[HTML]{FFFFFF}69.4 &
  \cellcolor[HTML]{FFFFFF}66.8 \\
  \cmidrule{2-13}
 &
   &
  \small{$GEN_{EN}+ORI$} &
  \cellcolor[HTML]{B7E2CE}69.5 &
  \cellcolor[HTML]{B9E3CF}76.4 &
  \cellcolor[HTML]{A5DBC1}69.8 &
  \cellcolor[HTML]{F5D0D0}48.2 &
  \cellcolor[HTML]{A5DBC1}76.0 &
  \cellcolor[HTML]{A5DBC1}72.8 &
  \cellcolor[HTML]{D8F0E4}63.4 &
  \cellcolor[HTML]{AEDFC7}67.8 &
  \cellcolor[HTML]{C7E9D9}73.4 &
  \cellcolor[HTML]{8BD0AE}77.8 \\
 &
   &
    \small{$GEN_{XX}+ORI$} &
  \cellcolor[HTML]{F3FAF7}65.2 &
  \cellcolor[HTML]{B9E3CF}76.4 &
  \cellcolor[HTML]{FFFFFF}62.4 &
  \cellcolor[HTML]{CAEADB}55.2 &
  \cellcolor[HTML]{ABDDC5}75.0 &
  \cellcolor[HTML]{F4CACA}62.2 &
  \cellcolor[HTML]{F8DEDE}58.2 &
  \cellcolor[HTML]{EA9999}55.4 &
  \cellcolor[HTML]{F5D0D0}66.2 &
  \cellcolor[HTML]{95D4B6}76.2 \\
 &
  \multirow{-3}{*}{ChatGPT} &
\small{$GEN_{EN}^{Trans}+ORI$} &
  \cellcolor[HTML]{DAF1E6}67.0 &
  \cellcolor[HTML]{B9E3CF}76.4 &
  \cellcolor[HTML]{F6D7D7}60.0 &
  \cellcolor[HTML]{9DD8BB}59.6 &
  \cellcolor[HTML]{F6D7D7}66.2 &
  \cellcolor[HTML]{F4FBF8}66.6 &
  \cellcolor[HTML]{FAEBEB}59.0 &
  \cellcolor[HTML]{D8F0E4}64.8 &
  \cellcolor[HTML]{B4E1CB}74.8 &
  \cellcolor[HTML]{99D6B9}75.6 \\
  \cmidrule{2-13}
 &
   &
  \small{$GEN_{EN}+ORI$} &
  \cellcolor[HTML]{95D5B6}73.7 &
  \cellcolor[HTML]{7CCAA4}84.6 &
  \cellcolor[HTML]{A1D9BE}70.4 &
  \cellcolor[HTML]{FBEEEE}50.0 &
  \cellcolor[HTML]{85CEAB}80.8 &
  \cellcolor[HTML]{74C79F}80.2 &
  \cellcolor[HTML]{B7E2CD}65.8 &
  \cellcolor[HTML]{8CD1AF}72.8 &
  \cellcolor[HTML]{98D6B8}78.4 &
  \cellcolor[HTML]{7AC9A2}80.4 \\
 &
   &
    \small{$GEN_{XX}+ORI$} &
  \cellcolor[HTML]{8FD2B1}74.6 &
  \cellcolor[HTML]{7CCAA4}84.6 &
  \cellcolor[HTML]{76C8A0}77.0 &
  \cellcolor[HTML]{BFE5D3}56.0 &
  \cellcolor[HTML]{7CCAA4}82.2 &
  \cellcolor[HTML]{89D0AE}77.0 &
  \cellcolor[HTML]{C2E7D5}65.0 &
  \cellcolor[HTML]{8BD0AE}73.8&
  \cellcolor[HTML]{A6DBC2}76.2 &
  \cellcolor[HTML]{7CCAA4}80.0 \\
\multirow{-7}{*}{XLMR-Large} &
  \multirow{-3}{*}{GPT-4} &
\small{$GEN_{EN}^{Trans}+ORI$} &
  \cellcolor[HTML]{93D4B4}74.1 &
  \cellcolor[HTML]{7CCAA4}84.6 &
  \cellcolor[HTML]{88CFAD}74.2 &
  \cellcolor[HTML]{AEDFC7}57.2 &
  \cellcolor[HTML]{7ECBA5}82.0 &
  \cellcolor[HTML]{87CFAC}77.4 &
  \cellcolor[HTML]{E9F6F0}62.2 &
  \cellcolor[HTML]{7ECBA5}75.0 &
  \cellcolor[HTML]{B9E3CF}74.4 &
  \cellcolor[HTML]{7FCBA6}79.6\\
  \bottomrule
\end{tabular}
}
\caption{Accuracy on XCOPA.
{\footnotesize{$ORI$}} corresponds to the original data, 
{\footnotesize{$GEN_{EN}$}} and {\footnotesize{$GEN_{XX}$}} represents data generated in English and target languages.
{\footnotesize{$Trans$}} denotes translations of the English-generated data.
We show languages that are available in all settings. 
Improvement and decline in performance are represented with green and red shadows.
}
 \label{tab:xcopa}
\end{table*}

\autoref{tab:xcopa} shows the results for the languages that are available for all settings, excluding TR and TH (unavailable for LLM-generation, refer to \autoref{sec:exp}), and QU (not supported by the Google Translate API).
We can see the impact of the generated data varies across different fine-tuned models and languages, aligning with the findings of \citet{kumar-etal-2022-diversity}.
Training on GPT-4 synthesised data displays consistent improvement across all scenarios and languages, except the zero-shot cross-lingual result on HT with XLMR-Large.

More fluctuating results can be observed with ChatGPT-generated data. 
A comparison between {\small{$GEN_{EN} + ORI$}} and {\small{$GEN_{XX} + ORI$}} indicates that utilising data generated in target languages generally leads to improved performance with GPT-4 generated data, as well as in base models with ChatGPT-generated data.
However, for XLMR-Large, employing ChatGPT-generated data in target languages mostly yields negative outcomes.
In languages such as TA and VI, training on generated data in the target languages results in more performance degradation compared to zero-shot cross-lingual transfer. This suggests that ChatGPT performs worse in those languages than XLMR-Large \citep{ahuja-etal-2023-mega}.

Translating the English dataset generally shows overall better results than training on the data generated directly in the target languages, except for XLMR-Large with GPT-4.
For SW, both XLMR-Base and XLMR-Large models fined-tuned with ChatGPT-generated data exhibit performance decline in most cases, even when the English-generated data benefits all other languages. This observation suggests that XLMR struggles with SW.
In \autoref{sec:human}, we select TA, SW, and the two best languages, ID and ZH, along with EN, for human evaluation.

\comment{[Chapter 7 - Correction point 5/5]~}
\add{Furthermore, we explore the effects of incorporating Target Languages in Validation (TLV). This approach involves training on English examples but evaluating and testing on the target language. The detailed results are presented in Table \ref{tab:xcopa_full} in Section \ref{sec:additional}. Notably, for smaller models trained on limited data (e.g., 4.1K examples), integrating target languages during validation led to significant performance boosts of 3.0 and 0.9 for mBERT and XLMR-Base, respectively.
However, when considering larger models like XLMR-large and smaller models trained with more extensive datasets (e.g., 29K examples), the impact of including target languages during validation was less pronounced. In these cases, we observe only minor variations in performance. These findings align with those of \citet{ponti-etal-2020-xcopa}, suggesting that the effectiveness of TLV may vary depending on factors such as model size and training data availability.}

\subsection{Dataset Scaling Up}

\begin{table}[!t]
\centering
\scalebox{0.95}{
\addtolength{\tabcolsep}{0pt}
\begin{tabular}{l|cc|cc}
\toprule
\multirow{2}{*}{\textbf{Model}}

& \multicolumn{2}{c|} {\small{$GEN_{EN} + ORI_{EN}$} }
& \multicolumn{2}{c}{\small{$GEN_{EN}^{Trans} + ORI_{EN}$} }
  \\
  \cmidrule(lr){2-3} \cmidrule(l){4-5}

  & \textsc{3.7k} & \textsc{28.6k}
& \textsc{3.7k}  & \textsc{28.6k}
\\
\midrule

 {mBERT} &
 54.3 & 56.0 & 58.0 & \textbf{60.1}
\\
{XLMR-Base}
& 60.1 & \textbf{61.8}
& 61.2 & 61.7

\\ {XLMR-Large}
 & 69.7 &\textbf{72.4} & 67.2 & 71.4 \\

\bottomrule
\end{tabular}}
\caption{Accuracy on XCOPA when scaling up the generated data to over 28K with ChatGPT. We report average results on all XCOPA languages excl. QU, since it is not available with the Google Translate API.}
\label{tab:scale}
\end{table}

We now investigate the impact of training on a larger scale of generated data on model performance. 
We focus on the XCOPA dataset and expand the generated data with ChatGPT (more budget-efficient) to 28.6k examples in English. We also compare the results of zero-shot cross-lingual transfer with translating the English-generated data to target languages.

The results in \autoref{tab:scale} demonstrate the positive impact of scaling up the generated data on model performance. Particularly, XLMR-Large exhibits the most significant improvement.

\subsection{Fixed Ratio Data Augmentation}

\begin{table}[!t]
\centering
\scalebox{0.95}{
\addtolength{\tabcolsep}{0pt}
\begin{tabular}{ll|ccc}
    \toprule
\textbf{Model}                       & \textbf{Ratio} & \textbf{XCOPA} & \textbf{XWingrad} & \textbf{XStoryCloze} \\
\midrule
\multirow{4}{*}{mBERT}      
& 1$\times$    &  64.0    &  50.2  &  74.6         \\
 & 2$\times$   & 64.8     &  51.9 &    76.8      \\
& 5$\times$  &   68.0     &  57.1 &     80.6      \\
& 10$\times$  &   69.8    & 65.7 &  80.3     \\
 \midrule
\multirow{4}{*}{XLMR-Base}  
& 1$\times$    &  58.0  &    45.9   &  70.7     \\
& 2$\times$   &  59.0   &   53.7   &   79.7    \\
& 5$\times$  & 63.0  & 67.8 &    81.9      \\
& 10$\times$   &  65.8	   &  71.2   &  84.1    \\
 \midrule
\multirow{4}{*}{XLMR-Large} 
& 1$\times$  &  56.0   &     78.1   &  81.1     \\
& 2$\times$  &  61.2    &   79.8    &    90.9    \\
& 5$\times$  &  81.4     &    82.0      &   89.9   \\
& 10$\times$  &   85.2    &    82.8      &    91.9   \\
\bottomrule
\end{tabular}
}
\caption{Performance on English test examples training on GPT-4-generated English data and the original data. Original data points selected from the three datasets are set to 200. 1$\times$ corresponds to using only the original data, 2$\times$ means using 200 original data and 200 generated data. }
\label{tab:ratio}
\end{table}
\label{sec:ratio}
We experiment with generating data with a fixed ratio of the original datasets. 
Specifically, we compare training with the original English data (200 randomly selected examples) and augment it with different quantities of English examples generated by GPT-4, where we include original training instances in all cases.

The results in \autoref{tab:ratio} showcase the performance on English test examples when fine-tuning mBERT and XLMR models with training data sizes that are 1$\times$, 2$\times$, 5$\times$, and 10$\times$ the size of the original dataset.
We can see that performance consistently improves as we increase the amount of generated data except XStoryCloze, which has the highest baselines, echoing the previous findings. The relative performance gain is generally more pronounced when increasing the data from 2$\times$ to 5$\times$ for the other two datasets.

\section{Human Evaluation}

To better evaluate the quality of the generated datasets and compare them with the human-created data, we ask native speakers to annotate the multilingual data generated by ChatGPT and GPT-4.

For each dataset, we first select 50 generated examples in English, and then request two annotators to evaluate the examples in two categories: 
(i) \textbf{Text Naturalness}. The annotators are asked to choose one of the following options for each example:  ``the text sounds natural'', ``the text sounds awkward but understandable'', or ``the text is not understandable'', and 
(ii) \textbf{Logic Soundness}. This category focuses on the commonsense aspect of the examples. The annotators are required to select the most appropriate description from: ``the correct option is (clearly) more plausible'', ``both options are equally plausible'', ``both options are implausible'', or ``the wrong option is more plausible''. We only ask the annotators to evaluate the logic if the text is at least understandable.

For XWinograd, we introduce an additional evaluation criterion. Annotators are asked to determine whether the two noun phrases in the examples can be replaced by the same pronoun (refer to \autoref{sec:wino_}).
For XCOPA, we extend the annotations to non-English languages, where we choose the two languages that demonstrate the most notable improvement, namely ZH and ID, as well as the two languages that exhibit the least improvement or regression in performance with ChatGPT-generated data, namely TA and SW (see \autoref{tab:xcopa}).
In addition to the original examples and the generated examples in the target languages, we include 50 examples that are translated from the same English-generated examples (that were selected for annotation).

To ensure impartiality, all the examples are shuffled, and the annotators are not provided with information regarding the source of the examples (human-created, LLM-generated, or translated).

\begin{figure}[t!]
\centering
    \includegraphics[width=\linewidth]{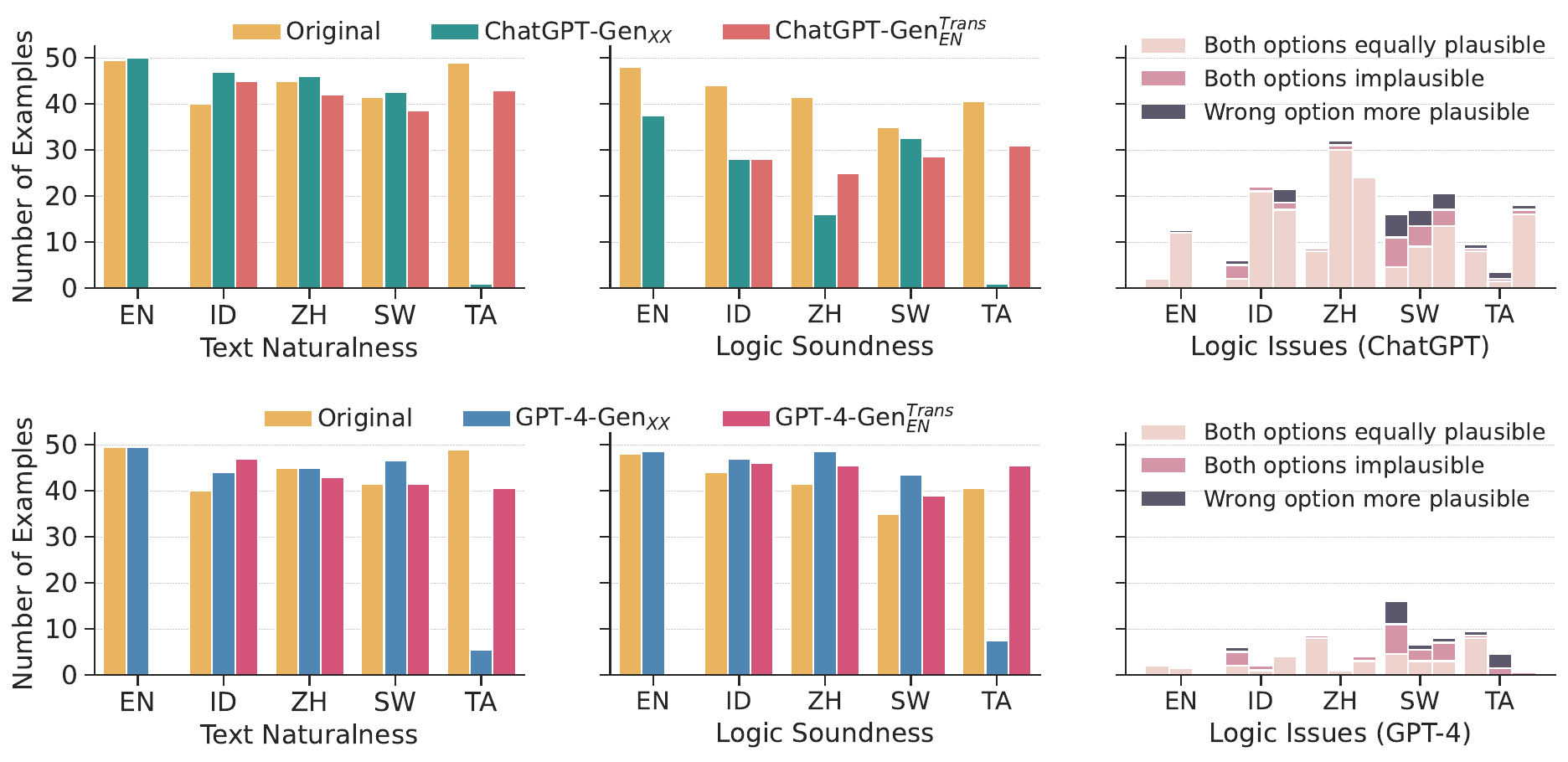}
\caption{Human evaluation of 50 random examples from the original XCOPA, ChatGPT (top) and GPT-4 (bottom) generated data in target languages, and translation of English generated data. Examples are annotated by two native speakers in each language. 
    The subplots in the last column show the logic issues of the XCOPA data, where the three bars for each language represent \textit{Original}, {\footnotesize{$Gen_{XX}$}}, and {\footnotesize{$Gen_{EN}^{Trans}$}} (from left to right).
}
\label{fig:human}
\end{figure}

\subsection{Text Naturalness}
\label{sec:human}
\autoref{fig:human} presents the annotation results for XCOPA, averaged from two annotators for each language.
For Text Naturalness,
we can see that in EN, ID, ZH, and SW, both ChatGPT and GPT-4 achieved higher naturalness than the original dataset.
This is particularly prominent in ID, revealing the fluency issue in the original ID data in XCOPA, which is also confirmed by a native speaker.

\subsection*{Issues with Tamil}

In contrast, the performance of the TA dataset is surprisingly low, with a majority of examples classified as "not understandable." Upon consulting language experts, we have identified several main issues in Tamil, including 
(i) the insertion of redundant words with the same meaning, such as ``I will retry to try it again'' 
(ii) verb agreement errors, and 
(iii) the presence of uncommon and out-of-context words.

It is worth noting that generating Tamil using GPT-4 is both slow and costly. We suspect that the tokeniser for Tamil, as well as similar languages like Telugu and Kannada, are poorly trained, resulting in unusable generation in those languages.
While the low quality of the generated data could explain the significant decline in the performance of the XLMR-Large model when trained on ChatGPT-generated data in Tamil, intriguingly, models trained on Tamil data generated by GPT-4 show improvement over the baselines.

To further investigate this issue, we conduct an experiment where we fine-tune the models using only five examples from the TA examples generated by GPT-4 that are identified as natural and sound by the annotators. 
The improvement on mBERT under this setting is 50\% of the total improvement seen with the entire 3.6K TA examples. For XLMR-base and XLMR-large, 15\% and 3\% of the total improvement can be observed, respectively.
Considering that the estimated number of correct samples in the 3.6k dataset is around 360, it is plausible that training solely on those examples could raise the accuracy level, or even surpass, what we observe for the entire dataset.\footnote{We could not conduct this experiment as the entire dataset was not manually labelled.}
An intriguing question that remains to be investigated in future research is why the remaining 3.2k incorrect or unnatural examples do not negatively impact the model's performance.

The translated text is typically less natural than the original and generated data (apart from ID due to issues in the original data). This result affirms that LLMs generally excel in generating fluent text for the languages it supports.

\subsection{Logic Soundness}
In terms of logic soundness, ChatGPT falls short compared to the original dataset.
We further illustrate the categorised issues in the last column of the plots in \autoref{fig:human}.
We can see that for ChatGPT, the majority of the examples are labelled as ``both options are equally plausible'', only SW has more problematic examples with ``the wrong option is more plausible''. 
We suspect that this issue arises from the instruction provided (taken from the description of the original COPA dataset), which states that ``both options could be plausible, but one is more plausible.''
In some cases, ChatGPT generates two choices that are excessively similar in terms of plausibility. 
On the other hand, GPT-4 tends to generate options with more clear-cut differences in plausibility, mirroring the original data.
We note that despite the description and instruction that both alternatives could happen, both the original dataset and the data synthesised by GPT-4 tend to present one plausible and one \textit{implausible} option.

For English XWinograd and XStoryCloze, the majority of the examples in both original and generated examples are evaluated as natural and logically sound. 
For XWinograd, although more than 47 examples are evaluated to exhibit high text quality and follow commonsense logic, \add{as mentioned in \autoref{sec:wino_}}, only 23 ChatGPT-generated examples fulfil the requirement that both noun phrases should be interchangeable with the same pronoun.
GPT-4 examples demonstrate better consistency, with 36 following this rule, whereas all original examples are found satisfactory.

\section{Conclusion and Future Work}
This chapter explores the effectiveness of utilising LLMs for data augmentation in cross-lingual datasets with limited training data. 
We specifically focus on commonsense reasoning tasks that are challenging for data synthesis.
Our experiments including four LLMs for data generation on three datasets, showcase enhanced cross-lingual zero-shot transfer on smaller fine-tuned task-specific language models. However, the impact varies across different datasets and languages. Notably, larger models such as XLMR-Large, which have higher baselines, demonstrate more difficulty in achieving performance improvements with LLM-generated data. Among the four LLMs, GPT-4-generated data exhibits mostly consistent superior performance.

Expanding data generation directly in target languages also shows general improvements compared to cross-lingual zero-shot with the English-generated data.
Human evaluation of the synthesised multilingual dataset shows that the ChatGPT and GPT-4 generated data demonstrate high naturalness in most languages, even surpassing the original data. However, in certain languages like TA, both models fail to generate natural text. Additionally, when assessing the logical soundness of the dataset, examples synthesised by ChatGPT reveal notable inconsistencies regarding more plausible options compared to the original human-created data. In contrast, GPT-4 exhibits a performance on par with human-written data.

In conclusion, leveraging LLMs for data augmentation shows promise. However, the choice of LLM used for data generation significantly influences the quality of the resulting data, as well as its applicability to the language under consideration. In circumstances where a more advanced model such as GPT-4 cannot be accessed, other models can be utilised, though this might result in performance difficulties in certain non-English languages, a challenge that also exists for GPT-4, and concerns regarding logical coherence. A compelling direction for future research could involve exploring the efficacy of more recent instruction-tuned or aligned open-source LLMs, such as LLaMA 2 \citep{touvron2023llama2} or TÜLU \citep{wu2023fine}, in enhancing data augmentation.

\section{Limitations}

The identified limitations in this chapter are as follows:

\begin{itemize}
\item \textbf{Language Resource Challenges}: While LLMs, particularly GPT-4, showcase promising results in the realm of multilingual commonsense data augmentation, challenges may arise when applied to extremely low-resource languages. The effectiveness of these models could be constrained by the availability of language-specific resources.

\item \textbf{Dependency on Few-Shot Examples}: To achieve optimal performance, the incorporation of few-shot examples in the target language remains necessary for generating new examples. However, obtaining such examples may pose challenges, particularly for languages with limited available resources.

\item \textbf{Closed Model Accessibility}: The usage of closed models like GPT-4 is restricted by licensing limitations, and the reproducibility of results obtained from these models may be compromised. Despite these constraints, the conducted experiments in this chapter illustrate the potential advantages of leveraging LLMs for multilingual dataset augmentation.

\item \comment{[Chapter 7 - Correction point 1/5]~}\add{\textbf{Black-box LLMs}: In addition to the limitation of the closed models, currently, only a handful of models are \textit{truly} open with training code, model checkpoints, and training data (i.e., BLOOM \citep{scao2022bloom}, the Pythia suite \citep{pmlr-v202-biderman23a}, and  OLMO \citep{groeneveld2024olmo}). All LLMs studied in this chapter are not \textit{truly} open or with detailed pre-training data released. This lack of transparency raises concerns regarding the potential biases and shortcomings embedded within these "black-box" LLMs. Without full disclosure of their training data sources and methodologies, there exists the possibility that such models were pre-trained on datasets that overlap with the test sets of public benchmarks, including those used in our experiments. This scenario could introduce a confounding factor, rendering performance evaluations of LLMs on downstream tasks and data augmentation unreliable and potentially skewed.}

\end{itemize}

\chapter{Conclusion}

\label{Conclusion} 

This thesis presents five focused studies that explore knowledge-grounded natural language understanding and generation, covering knowledge-enhanced fake news detection, multilingual knowledge-enhanced cross-lingual transfer, faithful and robust knowledge extraction from the web, grounded answer and explanation generation for knowledge-intensive VQA, and the employment of LLMs for data augmentation in tasks requiring multilingual commonsense knowledge.

Throughout this thesis, our focus centres on studying the utilisation and extraction of knowledge, whether in the form of structured knowledge, multilingual knowledge, parametric knowledge, or knowledge represented as augmented data distilled from powerful LLMs.

Addressing the research questions outlined in Chapter \ref{Introduction}, we hereby present the conclusions derived from this thesis.

\subsection*{Application of Structured Knowledge}
This thesis investigates the broader applicability of structured knowledge beyond entity-centric tasks. Through our experiments involving four distinct techniques to incorporate knowledge related to entities across two diverse fake news detection datasets, we find that leveraging structured knowledge improves applications such as fake news detection, if the applied knowledge is relevant and current.

\subsection*{Multilingual Entity Knowledge}
This thesis broadens the scope of structured knowledge to a multilingual setup. Specifically, we find that utilising multilingual entity knowledge via (i) the creation of an entity-centric code-switched corpus using data from Wikipedia and Wikidata, (ii) the intermediate training of a pre-trained multilingual language model, incorporating masked language modelling and entity prediction objectives, and (iii) fine-tuning the intermediate-trained model on downstream tasks, can consistently and significantly enhance the cross-lingual transferability.

\subsection*{Knowledge Extraction from Web Text}
This thesis explores approaches that facilitate the effective and accurate extraction of information of knowledge from the vast and often noisy web text. 
Upon collecting a novel m\textsc{WebIE} dataset and benchmarking against m\textsc{WebIE}, we conclude that (i) the inclusion of negative examples within the dataset, and (ii) the integration of entity-centric auxiliary tasks, are beneficial to the successful extraction of structured knowledge from the inherently noisy web text.

\subsection*{Grounded Answer and Explanation Learning}
This thesis highlights the benefit of multi-task grounded answer and explanation generation for knowledge-intensive VQA.  
Notably, we find that even without explicitly introducing external knowledge, which is commonly deemed a prerequisite in the datasets experimented, the adoption of this multi-task learning approach contributes to a notable enhancement in both answer accuracy and explanation quality, showcasing the advantage of better utilisation of the parametric knowledge stored within the model's parameters.

\subsection*{LLM-Powered Data Augmentation}
This thesis leverages the inherent knowledge of recent powerful LLMs to enhance the performance of smaller, task-specific models.
We find that data augmentation by prompting LLMs significantly improves complex and scarce multilingual commonsense reasoning tasks for smaller models, underscoring promising directions in data augmentation with LLMs and broader knowledge distillation methods.

\section{Future Work}
Having summarised key insights aimed at optimising the utilisation of knowledge in NLP applications throughout this thesis, we conclude with a vision for promising future research directions as follows.

\subsection*{Dynamic Knowledge Integration with Emphasised Grounding}

As the capabilities of LLMs continue to advance, the incorporation of knowledge emerges as a crucial supplement to maintain the generality of LLMs while ensuring they remain up-to-date and finely adapted to domain-specific tasks.

Current approaches, including the studies in this thesis, often involve single or pre-defined knowledge sources based on the targeted downstream tasks. For instance, in Chapters \ref{FakeNews}-\ref{WebIE}, we employed \textit{static} knowledge bases for both pre-training and downstream tasks. Although these strategies demonstrate effectiveness in benchmark tasks, they risk obsolescence without re-training. Hence, there is a growing need for more adaptive and dynamic knowledge integration methods in real-world applications.

A promising approach introduced in Chapter \ref{Background} is retrieval-augmented language models for text generation, demonstrating the potential of leveraging dynamic unstructured knowledge \citep{NEURIPS2020_6b493230, pmlr-v119-guu20a, izacard2022few}. While extensive research focuses on effective \textit{retrieval} of documents \citep{retro2022, ma-etal-2023-query, soares-etal-2023-nail} and \textit{ranking} of these documents \citep{zerveas-etal-2023-enhancing}, limited attention is given to how these models are \textit{grounded} with the provided context. Questions arise, such as whether the model can learn to disregard irrelevant retrieved context \citep{yoran2024making}, and how the generation is impacted by the context retrieved from end-to-end systems versus post hoc retrieval systems. We posit that a more thorough investigation into the grounding of retrieval-augmented systems is needed.

\subsection*{Knowledge Integration with Mixture-of-Experts}

In addition to dynamic knowledge integration, aligned with the concept of \textit{general-purpose} LLMs, there is a compelling need for a \textit{general-purpose} knowledge augmentation approach, one that does not necessitate the explicit specification of domain knowledge.

Revisiting the investigation on fake news detection in Chapter \ref{FakeNews}, we noted the importance of knowledge base relevance to the target task. Specifically, the integration significantly enhances the \texttt{LIAR} datasets but only marginally improves \texttt{COVID-19}. Consequently, it may become desirable to alleviate the need for predetermined knowledge bases for model training.

A promising direction for such exploration involves extending the adapter approach \citep{wang-etal-2021-k} as introduced in Chapter \ref{Background}. This entails leveraging different knowledge sources to train distinct adapters, thereby enhancing the adaptability of LLMs to various domains. Beyond adapters, broader parameter-efficient fine-tuning methods, such as Low-Rank Adaptation (LoRA) \citep{hu2022lora}, are also promising. By retaining the pre-trained weights of powerful LLMs, it becomes possible to train different versions of knowledge adapters or low-rank matrices. Subsequently, a carefully designed routing mechanism becomes crucial to determine the appropriate knowledge component for a given context.

While related work has demonstrated the advantages of the mixture of experts \citep{pmlr-v162-du22c, shen2024mixtureofexperts}, the mixture of adapters \citep{wang-etal-2022-adamix}, and the mixture of few-shot LoRA learning \citep{huang2023lorahub} for cross-task generalisation, we believe there exists unexplored potential for knowledge augmentation combined with the mixture of experts \citep{diao-etal-2023-mixture}.

\subsection*{Factuality in Text Generation}

Recent advanced LLMs excel at text fluency, typically empowered by large-scale pre-training and methods that leverage rankings over responses, such as reinforcement learning from human feedback (RLHF) \citep{ziegler2019fine, NEURIPS2022_b1efde53}. However, language models are susceptible to producing convincing yet factually inaccurate claims, commonly referred to as ``hallucinations'' \citep{tian2023fine}. This underscores the critical need for advancements in factuality and faithfulness in text generation \citep{augenstein2023factuality}. An expanded exploration of this thesis on knowledge grounding, in this context, plays a crucial role in determining what is \textit{factual} or \textit{faithful}, and in controlling the generation \citep{xu-etal-2020-megatron, rashkin-etal-2021-increasing, brahman-etal-2022-grounded}.

Researchers have made significant efforts in enhancing actuality in text generation, such as fine-tuning with automatically generated factuality preference rankings \citep{tian2023fine}, employing factual-nucleus sampling \citep{NEURIPS2022_df438caa}, training models to self-evaluate for faithfulness \citep{kadavath2022language}, among others. Despite these advancements, there remains a considerable amount of unexplored directions, such as the development of better evaluation metrics. Current metrics, whether n-gram-based or relying on trained neural models like BERTScore \citep{bert-score}, often fall short in capturing factuality \citep{clark-etal-2023-seahorse, aharoni-etal-2023-multilingual}. Moreover, commonly used entailment-based Natural Language Inference scores for faithfulness evaluation \citep{maynez-etal-2020-faithfulness} offer limited interpretability and may be biased by the underlying pre-trained models.

Therefore, we anticipate that training models for more faithful and grounded generation, as well as developing metrics to evaluate factuality performance, will be highly relevant research directions.

\section{Summary}
In this chapter, we highlight key findings and conclusions of the thesis, along with exciting and promising directions for future work including
dynamic knowledge integration with emphasised knowledge grounding, knowledge integration with mixture of experts, and knowledge-grounded text generation for enhanced factuality.
Quoting Prof Christopher D. Manning,\footnote{\url{https://2023.emnlp.org/program/keynotes}} we echo the sentiment of NLP research ``Nothing but blue skies!'' as we look forward to the boundless possibilities that lie ahead in the ever-evolving language technology.


\appendix 
\chapter{Language-Specific Results for \textsc{EntityCS} Experiments}

\label{sec:entitycs_additional_results}

We report the per-language results of the experiments on \textsc{EntityCS} in the following tables. 
WikiAnn results can be found in \autoref{tab:ner_all_langs_others}.
X-FACTR results in \autoref{tab:xfactr_results}, MultiATIS++ Slot Filling-only training in \autoref{tab:multiatis_main},
and XL-WiC in \autoref{tab:xlwic_results}.

\begin{table}[!ht]
\centering
\vspace{1.5ex}
\scalebox{0.75}{
\addtolength{\tabcolsep}{0pt}
\begin{tabular}{l|cccccccccccccc}
\toprule
\sc\textbf{model} & \sc{ar} & \sc{he} & \sc{vi} & \sc{id} & \sc{jv} & \sc{ms} & \sc{tl} & \sc{eu} & \sc{ml} & \sc{ta} & \sc{te} & \sc{af} & \sc{nl} & \sc{en}   \\ 
\midrule
\sc {xlm-r\textsuperscript{ours}} & 44.6 & 51.9 & 68.3 & 48.6 & 59.6 & 63.3 & 72.5 & 61.2 & 63.2 & 54.3 & 49.3 & 76.3 & 80.7   & 83.4 \\
\midrule
\sc {mlm} & 50.7 & 53.7 & 72.7 & 56.4 & 59.2 & 68.4 & 75.1 & 58.4 & 65.1 & 58.1 & 53.0 & 76.3 & 80.9 & 84.2  \\
\sc {wep} & 49.9 & 52.4 & 69.8 & 57.4 & 60.1 & 66.7 & 74.0 & 60.1 & 60.8 & 56.1 & 48.2 & 76.5 & 80.3 & 83.8 \\
\sc {pep\textsubscript{mrs}}  & 47.1 & 52.6 & 69.8 & 56.0 & 60.1 & 62.4 & 74.8 & 56.1 & 61.6 & 56.1 & 50.9 & 77.9 & 81.4 & 83.8\\
\sc {pep\textsubscript{m}}  & 47.7 & 52.9 & 68.9 & 59.1 & 63.1 & 65.5 & 76.3 & 60.0 & 64.0 & 57.5 & 51.6 & 76.8 & 80.9 & 83.9\\
\sc {wep+mlm}  & 50.3 & 53.2 & 69.8 & 60.8 & 60.7 & 69.8 & 74.5 & 59.2 & 64.8 & 57.2 & 51.7 & 76.4 & 80.9 & 84.1 \\
\sc {pep\textsubscript{mrs}+mlm}  & 46.7 & 53.6 & 69.6 & 64.0 & 60.2 & 69.2 & 74.3 & 57.5 & 65.6 & 55.8 & 52.3 & 77.5 & 81.3 & 84.1\\
\sc {pep\textsubscript{m}+mlm}   & 52.4 & 53.5 & 70.4 & 60.3 & 59.7 & 69.2 & 75.5 & 58.4 & 66.6 & 58.1 & 54.5 & 77.7 & 81.2 & 84.1  \\

 \midrule
& \sc{de} & \sc{el} & \sc{bn} & \sc{hi} & \sc{mr} & \sc{ur} & \sc{fa} & \sc{fr} & \sc{it} & \sc{pt} & \sc{es} & \sc{bg} & \sc{ru} & \sc{ja}  \\ \midrule
\sc {xlm-r\textsuperscript{ours}} & 75.4 & 74.2 & 67.9 & 68.3 & 61.8 & 55.8 & 47.6 & 78.0 & 78.2 & 78.9 & 76.2 & 77.3 & 63.9 & 22.9  \\ 
\midrule
\sc {mlm} & 75.2 & 76.3 & 73.9 & 69.9 & 64.5 & 67.0  &51.6 & 79.0 & 78.6 & 79.5 & 77.6 & 78.6 & 67.2 & 22.7 \\ 
\sc {wep}  & 74.7 & 74.5 & 70.8 & 67.5 & 61.1 & 60.7 & 50.9 & 77.6 & 77.7 & 77.3 & 74.1 & 78.7 & 66.3 & 20.7 \\
\sc {pep\textsubscript{mrs}} & 75.4 & 74.8 & 69.6 & 68.3 & 64.1 & 48.7 & 53.0 & 78.9 & 78.6 & 78.7 & 77.1 & 78.6 & 67.3 & 21.9 \\
\sc {pep\textsubscript{m}}& 75.1 & 76.5 & 73.0 & 69.6 & 65.8 & 63.3 & 55.3 & 78.5 & 78.4 & 78.6 & 74.3 & 78.2 & 67.2 & 21.0  \\
\sc {wep+mlm} & 75.4 & 75.2 & 72.1 & 68.9 & 63.9 & 58.6 &53.4 & 78.1 & 78.3 & 79.0 & 74.9 & 78.3 & 66.6 & 23.0\\
\sc {pep\textsubscript{mrs}+mlm} & 75.5 & 76.2 & 72.2 & 68.4 & 64.8 & 60.1 & 54.3 & 79.2 & 79.1 & 80.0 & 76.4 & 79.1 & 67.6 & 23.6 \\
\sc {pep\textsubscript{m}+mlm} & 75.3 & 76.2 & 74.2 & 70.0 & 67.1 & 64.5 & 50.0 & 79.9 & 78.9 & 79.7 & 78.4 & 79.3 & 68.2 & 22.7 \\

\midrule
 & \sc{ka} & \sc{ko} & \sc{th} & \sc{sw} & \sc{yo} & \sc{my} & \sc{zh} & \sc{kk} & \sc{tr} & \sc{et} & \sc{fi} & \sc{hu} \\ \midrule
\sc {xlm-r\textsuperscript{ours}} &  66.4 & 48.8 & 4.3 & 68.3 & 45.4 & 52.7 & 27.7 & 44.2 & 76.9 & 72.4 & 75.6 & 76.9 \\ 
\midrule
\sc {mlm}  & 66.1 & 50.8 & 2.5 & 65.1 & 42.9 & 55.7 & 29.7 & 50.7 & 77.8 & 71.4 & 76.2 & 78.1  \\ 
\sc {wep}& 64.8 & 52.0 & 2.5 & 65.8 & 50.4 & 52.6 & 26.1 & 52.1 & 75.5 & 71.9 & 75.8 & 76.6 \\
\sc {pep\textsubscript{mrs}} & 63.6 & 51.4 & 3.7 & 66.2 & 45.9 & 54.6 & 26.6 & 49.1 & 78.0 & 72.8 & 77.2 & 77.7 \\
\sc {pep\textsubscript{m}} & 66.7 & 50.0 & 5.0 & 66.8 & 52.3 & 56.9 & 26.7 & 48.4 & 77.6 & 73.2 & 76.6 & 77.3  \\
\sc {wep+mlm}   & 65.8 & 50.4 & 2.1 & 63.9 & 44.9 & 56.6 & 29.1 & 51.0 & 75.2 & 71.8 & 76.9 & 77.2\\
\sc {pep\textsubscript{mrs}+mlm}   & 66.1 & 50.9 & 2.4 & 66.6 & 41.4 & 54.5 & 31.1 & 51.8 & 78.4 & 71.4 & 77.1 & 78.7 \\
\sc {pep\textsubscript{m}+mlm}  & 67.7 & 51.1 & 3.3 & 64.5 & 43.7 & 56.6 & 29.3 & 51.7 & 78.2 & 72.0 & 76.8 & 78.6 \\
\bottomrule
\end{tabular}
}
\caption{F1-score per language on the WikiAnn test set. 
Results are averaged across five seeds.
}
 \label{tab:ner_all_langs_others}
\end{table}

\begin{table*}[t!]
    \centering
    \addtolength{\tabcolsep}{-3pt}
    \scalebox{0.63}{
    \begin{tabular}{p{0.3cm}rrrrrrrrrrrrrrrrrrrrrrrrrr}
    \toprule
     \multicolumn{3}{l}{\sc{model}} & &
    \textsc{avg} & \textsc{en} & \textsc{fr} & \textsc{nl} & \textsc{es} &	\textsc{ru} & \textsc{zh} & \textsc{he} & \textsc{tr} &	\textsc{ko} & \textsc{vi} & \textsc{el} & \textsc{mr} &	\textsc{ja} & \textsc{hu} & \textsc{bn} & \textsc{ceb} & \textsc{war} & \textsc{tl} & \textsc{sw} & \textsc{pa} & \textsc{mg} & \textsc{ilo}
    \\ 
    \midrule
    \multirow{6}{*}{\rotatebox[origin=c]{90}{\textsc{xlm-r}\textsuperscript{\textsc{ours}}}} & \multirow{6}{*}
    & \multirow{3}{*}{\rotatebox[origin=c]{90}{\textsc{ind}}} 
    &	\sc{A}
    & 3.5 &	8.2 &	4.7 &	4.4 &	6.5 &	5.3 &	4.6 &	2.5 &	3.1 &	5.1 &	8.5 &	6.3 &	2.7 &	2.3 &	0.9 &	0.1 &	1.4 &	1.2 &	2.8	 & 3.7 &	0.2 &	1.9 &	0.1  \\
     
    && & \sc{S}
    & 9.4  &	15.2 &	11.3 &	11.0 &	13.4 &	14.4 &	11.9 &	12.3 &	4.0 &	16.7 &	14.2 &	27.3 &	19.5 &	9.2 &	2.2 &	0.0 &	1.7 &	1.3 &	5.1 &	5.6 &	5.8 &	3.7 &	0.4  \\
    
        && & \sc{M}
    & 2.1 &	3.3 &	2.3 &	2.6 &	3.3 &	3.8 &	4.5 &	2.2 &	2.5 &	2.6 &	5.1 &	2.9 &	1.1 &	2.1 &	0.2 &	0.1 &	1.0 &	1.1 &	1.4 &	1.9 &	0.0 &	1.6 &	0.0  \\ \cmidrule{3-27}
    
    && \multirow{3}{*}{\rotatebox[origin=c]{90}{\textsc{conf}}} 
    & 	\sc{A}
    & 3.3  &	4.4 &	2.9	 & 2.7  &	4.3 &	5.5 &	5.3	 & 3.0 &	3.0 &	5.6 &	9.5 &	7.3 &	3.4 &	4.4 &	0.9 &	0.1 &	1.2 &	1.1 &	2.3 &	2.9 &	0.6 &	1.8 &	0.5  \\
    
        && & \sc{S}
    & 7.5&	5.2 &	4.4 &	3.6 &	4.9 &	14.2 &	11.8 &	11.4 &	3.9 &	15.9 &	12.6 &	25.6 &	18.9 &	8.8 &	2.0 &	0.0 &	1.4 &	1.4 &	4.4 &	4.3 &	5.8 &	3.5 &	0.5  \\
    
        && & \sc{M}
    & 2.6 &	3.9 &	2.3 &	2.7	 & 4.2 &	4.1 &	5.2 &	2.7 &	2.4 &	3.4 &	7.0 &	4.3 &	2.07 &	4.2 &	0.3 &	0.1 &	1.0 &	1.1 &	1.3 &	1.9 &	0.4 &	1.5&	0.5 \\ \midrule

    \multirow{6}{*}{\rotatebox[origin=c]{90}{\textsc{mlm}}}  & \multirow{6}{*}{\rotatebox[origin=c]{90}{\sc 39}}
    & \multirow{3}{*}{\rotatebox[origin=c]{90}{\textsc{ind}}} 
       & 	\sc{A}	 
    & 2.3 & 2.1 & 3.7 & 2.9 & 3.9 & 2.9 & 1.9 & 3.4 & 1.2 & 5.0 & 4.6 & 4.2 & 3.6 & 0.3 & 0.7 & 0.0 & 2.1 & 1.0 & 1.4 & 5.2 & 0.0 & 0.0 & 0.1 \\
	    && & \sc{S}	 
	& 6.4 & 5.1 & 8.7 & 6.4 & 9.4 & 6.0 & 8.3 & 8.7 & 3.1 & 16.6 & 9.1 & 19.3 & 17.9 & 2.5 & 1.8 & 0.6 & 2.9 & 1.1 & 4.4 & 8.3 & 0.3 & 0.5 & 0.1 \\
	    && & \sc{M} 
	& 1.3 & 0.9 & 2.0 & 2.0 & 1.9 & 2.0 & 1.8 & 1.9 & 0.6 & 2.3 & 2.4 & 2.8 & 2.1 & 0.2 & 0.4 & 0.0 & 1.8 & 1.0 & 0.5 & 2.2 & 0.0 & 0.0 & 0.1 \\ \cmidrule{3-27}

    && \multirow{3}{*}{\rotatebox[origin=c]{90}{\textsc{conf}}} 
         & 	\sc{A}	 &2.5 & 2.5 & 3.6 & 2.9 & 4.3 & 2.6 & 2.0 & 4.8 & 1.1 & 5.7 & 6.3 & 5.2 & 4.2 & 0.4 & 0.6 & 0.1 & 2.0 & 1.0 & 1.2 & 5.2 & 0.0 & 0.0 & 0.1\\
	&& & \sc{S} & 5.9 & 4.9 & 7.6 & 5.9 & 9.0 & 4.4 & 7.6 & 7.4 & 2.5 & 16.1 & 8.5 & 17.2 & 16.7 & 2.5 & 1.6 & 0.6 & 2.8 & 1.1 & 3.9 & 7.8 & 0.3 & 0.5 & 0.0  \\
	&& & \sc{M} & 	1.7 & 1.8 & 2.3 & 2.2 & 2.6 & 2.3 & 1.9 & 3.4 & 0.5 & 3.4 & 4.6 & 4.2 & 2.9 & 0.3 & 0.4 & 0.0 & 1.7 & 1.0 & 0.4 & 2.4 & 0.0 & 0.0 & 0.1\\ \midrule
         
    \multirow{18}{*}{\rotatebox[origin=c]{90}{\textsc{wep}}}  & \multirow{6}{*}{\rotatebox[origin=c]{90}{\sc en}}
    & \multirow{3}{*}{\rotatebox[origin=c]{90}{\textsc{ind}}} 
    & \textsc{A}
    & 3.3 & 18.2 & 6.1 & 6.0 & 5.8 & 1.1 & 0.4 & 0.4 & 1.1 & 0.5 & 8.0 & 3.5 & 0.4 & 0.6 & 3.7 & 0.0 & 3.5 &   0.6 
    & 5.0 & 4.2 & 0.1 & 1.7 & 1.6  \\ 
    
        && & \textsc{S}
    & 8.5 & 38.3 & 16.4 & 18.7 & 14.9 & 4.4 & 3.4 & 1.4 & 5.6 & 2.7 & 16.8 & 7.3 & 4.1 & 2.5 & 8.5 & 0.0 & 6.9 & 2.6 & 9.7 & 10.3 & 0.0 & 6.9 & 5.4\\ 
    
        && & \textsc{M}
    & 1.6 & 9.4 & 2.7 & 2.9 & 2.9 & 0.6 & 0.3 & 0.4 & 0.3 & 0.3 & 3.5 & 1.2 & 0.1 & 0.5 & 2.6 & 0.0 & 1.5 & 0.3 & 1.5 & 1.9 & 0.1 & 1.1 & 0.5
    \\ \cmidrule{3-27}
    
    && \multirow{3}{*}{\rotatebox[origin=c]{90}{\textsc{conf}}} 
       &  \textsc{A}
    & 3.1 & 16.2 & 6.4 & 5.6 & 5.4 & 1.1 & 0.3 & 0.4 & 1.1 & 0.5 & 7.6 & 3.4 & 0.4 & 0.6 & 3.6 & 0.0 & 3.4 & 0.5 & 4.5 & 4.1 & 0.1 & 1.3 & 1.7
    \\
    
       && & \textsc{S}
    & 7.9 & 35.8 & 15.9 & 17.2 & 13.3 & 4.5 & 2.7 & 1.6 & 5.4 & 2.4 & 15.8 & 7.3 & 4.1 & 2.5 & 8.2 & 0.0 & 6.6 & 2.5 & 8.6 & 7.8 & 0.0 & 6.4 & 5.4
    \\
    
        && & \textsc{M}
    & 1.5 & 7.5 & 3.3 & 2.9 & 2.9 & 0.6 & 0.3 & 0.4 & 0.2 & 0.3 & 3.6 & 1.1 & 0.1 & 0.5 & 2.6 & 0.0 & 1.5 & 0.2 & 1.5 & 2.0 & 0.1 & 1.0 & 0.6
    \\
     \cmidrule{2-27}
    

    & \multirow{6}{*}{\rotatebox[origin=c]{90}{\sc 39}}
    & \multirow{3}{*}{\rotatebox[origin=c]{90}{\textsc{ind}}} 
    & \textsc{A}
    
    & 6.1 & 15.6 & 9.1 & 11.5 & 10.5 & 2.8 & 6.7 & 3.7 & 3.2 & 6.7 & 13.2 & 7.9 & 4.0 & 4.6 & 6.7 & 0.9 & 4.3 & 2.1 & 7.4 & 7.2 & 0.0 & 2.3 & 3.3 \\ 
    
        && & \textsc{S}
    & 19.4 & 36.4 & 24.1 & 30.3 & 25.6 & 14.3 & 18.5 & 34.7 & 12.2 & 31.5 & 23.4 & 36.0 & 29.8 & 17.8 & 18.5 & 6.1 & 8.5 & 5.0 & 16.9 & 21.3 & 0.0 & 5.4 & 9.3 \\ 
    
        && & \textsc{M}
    & 3.0 & 7.2 & 3.9 & 4.9 & 4.6 & 1.5 & 6.3 & 2.6 & 1.0 & 2.9 & 7.3 & 3.9 & 1.5 & 4.1 & 2.5 & 0.0 & 2.2 & 1.4 & 1.8 & 3.3 & 0.0 & 1.4 & 0.6
    \\ \cmidrule{3-27}
    
    && \multirow{3}{*}{\rotatebox[origin=c]{90}{\textsc{conf}}} 
       &  \textsc{A}
    & 4.9 & 12.1 & 8.2 & 9.6 & 8.8 & 2.4 & 3.1 & 3.3 & 2.9 & 5.9 & 9.3 & 7.4 & 3.5 & 1.9 & 5.6 & 0.8 & 4.1 & 1.7 & 6.8 & 5.7 & 0.0 & 1.8 & 3.3
    \\
    
       && & \textsc{S}
    &17.4 & 32.6 & 22.9 & 26.5 & 23.4 & 12.2 & 16.7 & 32.4 & 11.2 & 28.3 & 19.3 & 34.3 & 27.1 & 15.9 & 16.0 & 5.6 & 8.2 & 4.7 & 14.9 & 17.2 & 0.0 & 5.1 & 9.2
    \\
    
        && & \textsc{M}
    & 2.1 & 4.6 & 3.3 & 3.6 & 3.0 & 1.2 & 2.6 & 2.3 & 0.8 & 2.7 & 3.9 & 3.7 & 1.4 & 1.5 & 1.8 & 0.0 & 2.1 & 1.0 & 1.9 & 2.0 & 0.0 & 1.0 & 0.7
    \\
    \cmidrule{2-27}
    
 
    & \multirow{6}{*}{\rotatebox[origin=c]{90}{\sc 93}}
    & \multirow{3}{*}{\rotatebox[origin=c]{90}{\textsc{ind}}} 
    & \textsc{A}
    & 5.8 & 13.9 & 7.6 & 10.1 & 11.2 & 2.8 & 7.2 & 2.9 & 2.9 & 5.8 & 13.6 & 8.1 & 4.4 & 3.2 & 7.2 & 0.6 & 3.1 & 2.4 & 6.8 & 6.6 & 1.0 & 2.5 & 3.2 \\ 
    
        && & \textsc{S}
    & 18.5 & 34.5 & 20.0 & 28.9 & 25.3 & 14.0 & 20.1 & 26.0 & 13.0 & 28.6 & 25.4 & 35.0 & 25.6 & 17.3 & 18.2 & 4.9 & 6.7 & 7.2 & 13.6 & 17.6 & 11.6 & 5.7 & 8.3 \\ 
    
        && & \textsc{M}
    & 2.7 & 6.6 & 3.0 & 4.7 & 5.2 & 1.3 & 6.8 & 2.3 & 0.9 & 2.5 & 7.6 & 4.3 & 2.1 & 2.7 & 3.1 & 0.0 & 1.8 & 0.9 & 1.3 & 1.4 & 0.2 & 1.3 & 0.4
    \\ \cmidrule{3-27}
    
    && \multirow{3}{*}{\rotatebox[origin=c]{90}{\textsc{conf}}} 
       &  \textsc{A}
    & 4.6 & 11.3 & 6.4 & 8.6 & 9.1 & 2.2 & 2.7 & 2.5 & 2.7 & 4.9 & 10.5 & 7.2 & 3.5 & 1.8 & 6.1 & 0.6 & 2.8 & 2.0 & 6.2 & 5.1 & 0.8 & 1.4 & 2.5
    \\
    
       && & \textsc{S}
    & 16.3 & 31.5 & 18.4 & 26.3 & 22.3 & 11.8 & 18.0 & 24.3 & 12.2 & 25.5 & 22.3 & 31.3 & 20.8 & 14.5 & 16.4 & 4.4 & 6.1 & 6.6 & 11.6 & 15.3 & 8.9 & 2.4 & 7.6
    \\
    
        && & \textsc{M}
    & 1.8 & 4.7 & 2.1 & 3.5 & 3.4 & 0.9 & 2.3 & 2.0 & 0.8 & 2.1 & 4.9 & 3.5 & 1.5 & 1.4 & 2.2 & 0.0 & 1.6 & 0.7 & 1.1 & 0.9 & 0.1 & 0.6 & 0.4
    \\
    \midrule

    \multirow{6}{*}{\rotatebox[origin=c]{90}{\textsc{pep\textsubscript{ms}}}}  & \multirow{6}{*}{\rotatebox[origin=c]{90}{\sc 39}}
    & \multirow{3}{*}{\rotatebox[origin=c]{90}{\textsc{ind}}} 
    & \textsc{A}
    & 4.7 & 15.1 & 6.9 & 11.0 & 9.6 & 5.0 & 3.8 & 3.2 & 2.0 & 7.3 & 9.0 & 5.5 & 3.0 & 3.3 & 1.9 & 0.2 & 3.3 & 1.5 & 5.9 & 5.5 & 0.0 & 0.7 & 0.6
    \\
    
        && & \textsc{S}
    & 15.0 & 35.2 & 18.6 & 29.4 & 22.0 & 16.7 & 15.7 & 19.4 & 8.7 & 29.3 & 19.2 & 30.2 & 24.5 & 19.9 & 4.6 & 1.7 & 6.4 & 2.5 & 10.3 & 12.8 & 0.0 & 1.1 & 1.2
    \\
    
        && & \textsc{M}
    & 2.4 & 7.1 & 2.4 & 4.5 & 4.2 & 2.1 & 3.5 & 2.6 & 0.7 & 3.3 & 4.4 & 3.5 & 0.5 & 2.7 & 1.0 & 0.0 & 2.0 & 1.2 & 2.4 & 2.8 & 0.0 & 0.5 & 0.4
    \\ 
    \cmidrule{3-27}
    
    && \multirow{3}{*}{\rotatebox[origin=c]{90}{\textsc{conf}}} 
      &  \textsc{A}
    &6.0 & 15.7 & 8.1 & 12.5 & 11.7 & 5.7 & 6.9 & 5.2 & 2.9 & 9.2 & 14.0 & 6.3 & 5.1 & 6.7 & 3.4 & 0.4 & 3.4 & 1.5 & 6.4 & 5.6 & 0.0 & 0.5 & 0.5
    \\
    
        && & \textsc{S}
    & 13.1 & 31.9 & 17.1 & 27.1 & 19.6 & 12.1 & 13.6 & 17.7 & 7.6 & 26.1 & 16.0 & 27.1 & 21.4 & 16.9 & 3.9 & 1.6 & 5.3 & 2.2 & 9.4 & 9.0 & 0.0 & 0.8 & 1.1
    \\
    
        && & \textsc{M}
    &  4.3 & 10.0 & 4.5 & 7.1 & 8.6 & 4.0 & 6.7 & 4.7 & 2.0 & 6.2 & 11.8 & 4.9 & 3.3 & 6.4 & 2.7 & 0.2 & 2.4 & 1.3 & 3.9 & 3.8 & 0.0 & 0.2 & 0.3
    \\ \midrule
    
    \multirow{18}{*}{\rotatebox[origin=c]{90}{\textsc{pep\textsubscript{ms}+mlm}}}  & \multirow{6}{*}{\rotatebox[origin=c]{90}{\sc en}}
    & \multirow{3}{*}{\rotatebox[origin=c]{90}{\textsc{ind}}} 
    & \textsc{A}
    & 2.6 & 16.8 & 5.0 & 5.2 & 4.9 & 1.5 & 0.2 & 0.6 & 0.2 & 0.6 & 6.3 & 3.0 & 0.4 & 0.6 & 1.0 & 0.0 & 1.2 & 0.4 & 4.5 & 2.3 & 0.6 & 1.8 & 0.5 \\ 
    
        && & \textsc{S}
    & 6.5 & 35.5 & 13.4 & 15.5 & 12.3 & 6.2 & 2.0 & 4.7 & 0.5 & 3.3 & 13.0 & 8.7 & 3.1 & 3.0 & 1.9 & 0.0 & 2.5 & 0.5 & 7.0 & 5.3 & 0.7 & 4.5 & 0.4\\ 
    
        && & \textsc{M}
    & 1.2 & 6.6 & 2.2 & 2.7 & 2.3 & 1.2 & 0.2 & 0.5 & 0.1 & 0.3 & 2.8 & 0.7 & 0.1 & 0.5 & 0.3 & 0.0 & 0.6 & 0.4 & 1.4 & 1.0 & 0.5 & 1.3 & 0.5
    \\ \cmidrule{3-27}
    
    && \multirow{3}{*}{\rotatebox[origin=c]{90}{\textsc{conf}}} 
       &  \textsc{A}
    & 2.7 & 18.0 & 5.3 & 5.1 & 4.6 & 1.5 & 0.6 & 1.0 & 0.2 & 0.9 & 6.3 & 3.0 & 0.6 & 0.9 & 0.9 & 0.0 & 1.5 & 0.3 & 4.6 & 2.1 & 0.6 & 1.6 & 0.4
    \\
    
       && & \textsc{S}
    & 5.7 & 33.1 & 12.0 & 13.3 & 10.2 & 5.8 & 2.0 & 4.4 & 0.5 & 2.7 & 11.1 & 7.3 & 3.1 & 3.0 & 1.7 & 0.0 & 1.7 & 0.3 & 6.7 & 4.6 & 0.6 & 1.0 & 0.3
    \\
    
        && & \textsc{M}
    & 1.6 & 10.4 & 3.0 & 3.2 & 2.6 & 1.2 & 0.5 & 0.9 & 0.1 & 0.8 & 3.8 & 1.2 & 0.3 & 0.8 & 0.3 & 0.0 & 1.1 & 0.3 & 1.8 & 1.1 & 0.5 & 1.2 & 0.5
    \\
     \cmidrule{2-27}
    

    & \multirow{6}{*}{\rotatebox[origin=c]{90}{\sc 39}}
    & \multirow{3}{*}{\rotatebox[origin=c]{90}{\textsc{ind}}} 
    & \textsc{A}
    & 4.9 & 14.9 & 9.7 & 10.5 & 10.5 & 7.3 & 5.5 & 4.4 & 1.3 & 7.0 & 9.4 & 5.8 & 2.0 & 2.6 & 1.6 & 0.0 & 2.8 & 0.9 & 4.5 & 6.5 & 0.0 & 0.3 & 0.4 \\ 
    
        && & \textsc{S}
    & 13.9 & 34.8 & 24.2 & 27.7 & 23.1 & 20.2 & 16.0 & 17.5 & 5.8 & 28.6 & 19.3 & 25.2 & 13.5 & 15.3 & 4.0 & 2.4 & 5.5 & 1.7 & 8.3 & 11.2 & 0.0 & 0.3 & 0.7\\ 
    
        && & \textsc{M}
    &2.4 & 6.3 & 3.4 & 4.6 & 4.2 & 3.8 & 5.2 & 2.7 & 0.5 & 3.0 & 4.6 & 4.1 & 0.6 & 2.1 & 0.7 & 0.0 & 1.7 & 0.8 & 1.9 & 3.1 & 0.0 & 0.3 & 0.2
    \\ \cmidrule{3-27}
    
    && \multirow{3}{*}{\rotatebox[origin=c]{90}{\textsc{conf}}} 
       &  \textsc{A}
    & 5.7 & 16.3 & 10.5 & 11.0 & 11.3 & 7.9 & 7.0 & 6.2 & 1.3 & 9.2 & 14.2 & 6.6 & 2.8 & 4.2 & 1.5 & 0.3 & 3.4 & 1.3 & 4.1 & 6.6 & 0.0 & 0.3 & 0.3
    \\
    
       && & \textsc{S}
    & 12.0 & 30.6 & 21.6 & 24.8 & 20.1 & 16.6 & 15.7 & 16.0 & 3.0 & 27.1 & 15.8 & 20.8 & 12.3 & 12.8 & 3.4 & 0.0 & 4.9 & 2.0 & 7.0 & 9.5 & 0.0 & 0.2 & 0.4
    \\
    
        && & \textsc{M}
    & 3.9 & 9.4 & 5.4 & 6.2 & 6.7 & 5.0 & 6.7 & 4.5 & 0.8 & 6.0 & 11.8 & 5.6 & 1.5 & 3.8 & 0.7 & 0.3 & 2.8 & 1.3 & 2.6 & 4.5 & 0.0 & 0.3 & 0.2
    \\
    \cmidrule{2-27}
    
 
    & \multirow{6}{*}{\rotatebox[origin=c]{90}{\sc 93}}
    & \multirow{3}{*}{\rotatebox[origin=c]{90}{\textsc{ind}}} 
    & \textsc{A}
    & 4.5 & 14.1 & 8.6 & 9.1 & 8.8 & 5.8 & 5.0 & 3.1 & 0.8 & 6.3 & 9.4 & 5.7 & 1.8 & 1.5 & 1.5 & 0.1 & 2.9 & 1.7 & 4.0 & 5.2 & 0.7 & 2.3 & 0.4 \\ 
    
        && & \textsc{S}
    &13.2 & 32.7 & 22.0 & 24.4 & 20.3 & 18.6 & 17.0 & 12.8 & 3.8 & 26.9 & 19.4 & 25.9 & 12.2 & 12.0 & 3.4 & 0.9 & 5.0 & 2.1 & 9.7 & 9.5 & 8.1 & 3.1 & 0.9\\ 
    
        && & \textsc{M}
    & 2.3 & 5.9 & 3.3 & 4.4 & 3.8 & 3.1 & 4.7 & 2.6 & 0.4 & 2.8 & 4.4 & 3.8 & 0.4 & 1.1 & 0.7 & 0.0 & 2.2 & 1.5 & 1.2 & 1.3 & 0.0 & 1.9 & 0.2
    \\ \cmidrule{3-27}
    
    && \multirow{3}{*}{\rotatebox[origin=c]{90}{\textsc{conf}}} 
       &  \textsc{A}
    &5.5 & 15.7 & 9.5 & 9.8 & 9.3 & 6.2 & 6.9 & 5.0 & 0.8 & 8.1 & 13.5 & 7.3 & 2.8 & 3.0 & 1.4 & 0.1 & 2.6 & 1.8 & 4.4 & 6.8 & 2.0 & 2.6 & 0.4
    \\
    
       && & \textsc{S}
    & 11.9 & 31.3 & 19.2 & 22.0 & 16.8 & 15.7 & 16.3 & 12.5 & 3.5 & 25.0 & 16.6 & 23.9 & 11.3 & 8.6 & 2.8 & 0.6 & 4.3 & 2.0 & 9.0 & 9.4 & 7.5 & 2.7 & 0.8
    \\
    
        && & \textsc{M}
    & 3.8 & 8.7 & 5.2 & 6.0 & 5.5 & 4.2 & 6.6 & 4.5 & 0.5 & 5.4 & 10.9 & 5.9 & 1.7 & 2.7 & 0.7 & 0.1 & 2.3 & 1.8 & 2.3 & 3.7 & 1.5 & 2.6 & 0.2
    \\
    
    \bottomrule
    \end{tabular}
    }
    \caption{X-FACTR results.}
    \label{tab:xfactr_results}
\end{table*}

\begin{table*}[ht]
\centering
\scalebox{0.85}{
        \addtolength{\tabcolsep}{-1.7pt}
\begin{tabular}{l|ccccccccc}
\toprule
\sc\textbf{model}  & \textsc{en}  & \textsc{es} & \textsc{de} & \textsc{hi} & \textsc{fr}  & \textsc{pt}   & \textsc{zh}   & \textsc{ja} & \textsc{tr}       \\
\midrule
\textsc{xlm-r\textsuperscript{ours}} 
& 95.6~\textsubscript{0.15} & {81.5}~\textsubscript{0.71} & 79.8~\textsubscript{2.04} & 50.6~\textsubscript{5.35} & 
74.8~\textsubscript{1.90} & {76.5}~\textsubscript{1.14} & 77.2~\textsubscript{2.06} & 56.8~\textsubscript{4.99} & 43.0~\textsubscript{2.72} 
\\
\midrule
\textsc{mlm}   
& 95.6~\textsubscript{0.16} & 78.8~\textsubscript{2.88} & 78.0~\textsubscript{2.56} & 61.5~\textsubscript{7.26} & 74.4~\textsubscript{3.18} & 74.6~\textsubscript{1.39} & 76.4~\textsubscript{1.81} & 70.3~\textsubscript{2.00}  & 39.7~\textsubscript{4.09}    \\
\textsc{wep}  
& 95.7~\textsubscript{0.15} & 79.9~\textsubscript{1.34} & 80.3~\textsubscript{0.58} & 52.7~\textsubscript{4.15} & 75.6~\textsubscript{0.87}& 76.3~\textsubscript{0.63} & 78.1~\textsubscript{1.43}  & 60.7~\textsubscript{7.07}    & 40.4~\textsubscript{4.40}  \\
\textsc{pep\textsubscript{ms}}    
& 95.3~\textsubscript{0.06} & 79.3~\textsubscript{2.60} & {79.7}~\textsubscript{2.28} & 62.9~\textsubscript{2.30} & 75.3~\textsubscript{2.10} & 76.2~\textsubscript{1.60} & {77.8}~\textsubscript{1.30} & {69.0}~\textsubscript{4.90} & {45.3}~\textsubscript{2.50}
 \\
\textsc{pep\textsubscript{ms}+mlm}  & 
{95.6}~\textsubscript{0.10} & 81.3~\textsubscript{1.90} & 81.4~\textsubscript{0.90} & {65.8}~\textsubscript{2.20} & 78.2~\textsubscript{0.30} & {76.1}~\textsubscript{1.00} & 78.8~\textsubscript{0.60} & 68.8~\textsubscript{3.30}   & 42.1~\textsubscript{3.30}  \\
\bottomrule
\end{tabular}}
\caption{F1-score (average across five seeds) on MultiATIS++ SF-only training.}
\label{tab:multiatis_main}
\end{table*}

\begin{table}[t]
    \centering
    \scalebox{0.9}{
        \addtolength{\tabcolsep}{-0pt}
    \begin{tabular}{l|cccccccccccc}
    \toprule
      \sc\textbf{model} &  \textsc{bg} &	\textsc{da}	 & \textsc{et} &	\textsc{fa} &	\textsc{hr}	 & \textsc{ja} \\ \midrule
        
        \textsc{xlm-r}\textsuperscript{ours} 
        & 57.5~\textsubscript{1.03}	
        & 60.6~\textsubscript{2.06}
        & 61.7~\textsubscript{3.23}
        & 62.4~\textsubscript{1.05}
        & 61.7~\textsubscript{2.93}
        & 54.0~\textsubscript{1.56}
 \\ \midrule
        
        \textsc{mlm}
        &  59.3~\textsubscript{0.85}
        &  59.0~\textsubscript{0.72}	
        &  60.6~\textsubscript{1.15}	
        &  63.5~\textsubscript{1.41}	
        &  62.3~\textsubscript{1.87}	
        &  52.6~\textsubscript{1.26}	
        \\
        												
        \textsc{wep}
        & 59.0~\textsubscript{1.84}	
        & 61.3~\textsubscript{1.13}	
        & 62.2~\textsubscript{0.74}	
        & 64.9~\textsubscript{1.06}	
        & 63.7~\textsubscript{2.40}	
        & 54.7~\textsubscript{2.69}	
        \\    
        
        \textsc{pep\textsubscript{ms}}
		& 59.4~\textsubscript{0.93}	
		& 60.7~\textsubscript{1.03}
		& 64.4~\textsubscript{1.72}	
		& 63.5~\textsubscript{1.51}	
		& 64.2~\textsubscript{1.85}	
		& 53.6~\textsubscript{2.60}	

		\\					
												
        \textsc{pep\textsubscript{ms}+mlm}
        & 59.7~\textsubscript{1.04}
        & 60.9~\textsubscript{1.20}	
        & 63.9~\textsubscript{1.02}	
        & 63.1~\textsubscript{1.63}	
        & 63.8~\textsubscript{1.89}	
        & 53.2~\textsubscript{2.18}	

        \\
        \midrule
         \sc\textbf{model}  &	\textsc{ko} &	\textsc{nl}	  & \textsc{zh}	  & \textsc{de} &	\textsc{FR}  &	\textsc{IT} \\ \midrule
        
        \textsc{xlm-r}\textsuperscript{ours} 
       
        & 62.4~\textsubscript{1.99}
        & 61.5~\textsubscript{1.94}
        & 56.4~\textsubscript{3.83}
        & 57.7~\textsubscript{1.58}
        & 56.4~\textsubscript{1.40}
        & 57.1~\textsubscript{1.35} \\ \midrule
        
        \textsc{mlm}	
        &  63.1~\textsubscript{1.38}	
        &  62.3~\textsubscript{0.76}	
        &  52.4~\textsubscript{1.03}	
        &  57.2~\textsubscript{0.54}	
        &  56.6~\textsubscript{0.33}	
        &  58.0~\textsubscript{1.09}
        \\
        												
        \textsc{wep}
        & 64.6~\textsubscript{0.74}	
        & 63.8~\textsubscript{0.77}	
        & 55.2~\textsubscript{3.30}	
        & 59.6~\textsubscript{1.04}	
        & 57.0~\textsubscript{0.98}
        & 59.1~\textsubscript{1.03} 
        \\    
        
        \textsc{pep\textsubscript{ms}}
		& 64.6~\textsubscript{2.88}	
		& 63.9~\textsubscript{0.79}	
		& 52.8~\textsubscript{3.01}	
		& 59.5~\textsubscript{1.79}	
		& 57.4~\textsubscript{0.64}	
		& 58.8~\textsubscript{1.25}
		\\					
												
        \textsc{pep\textsubscript{ms}+mlm}
        & 62.1~\textsubscript{3.10}	
        & 63.0~\textsubscript{1.01}	
        & 53.2~\textsubscript{2.24}	
        & 59.0~\textsubscript{0.90}	
        & 57.3~\textsubscript{0.58}	
        & 58.3~\textsubscript{1.00}
        \\
    \bottomrule
    \end{tabular}
    }
    \caption{XL-WiC test set accuracy (average across five seeds) across languages.}
    \label{tab:xlwic_results}
\end{table}

\chapter{Additional Results and Analysis for \textsc{WebIE}}

\label{appendix-webie}

\section{Additional Results}

\label{sec:src_lang}
We show in \autoref{tab:src-lang} the results for non-English languages for m\webie\ when specifying the source language and using the default (English) for the mBART tokenizer. 
These results are from beam search without constraint Trie.
We can see that specifying the source language mostly harms the performance (except French), especially for Portuguese.
We hypothesise that due to the model being trained solely on English as the source token, mBART may have difficulty handling other languages.

\begin{table}[!ht]
\centering
\vspace{0.4cm}
\scalebox{0.8}{
\addtolength{\tabcolsep}{-2pt}
\begin{tabular}{l|ccccc|ccccc}
\toprule
{\multirow{2}{*}{\sc Language}}
 &\multicolumn{5}{c|} { \em EN as Source Language in mBART Tokenizer}  &\multicolumn{5}{c} {\em XX as Source Language in mBART Tokenizer}
\\

&\textit{P} &\textit{R} & \textit{F1} & \textit{Empty-Pos\%} 
& \textit{Acc-N} 
&\textit{P} &\textit{R} & \textit{F1} & \textit{Empty-Pos\%}  & \textit{Acc-N} 

\\
\cmidrule(r){1-1} 
\cmidrule(lr){2-4} \cmidrule(lr){5-5} \cmidrule(lr){6-6} \cmidrule(lr){7-9}  
\cmidrule(lr){10-10} 
\cmidrule(l){11-11} 

  \sc French 
  &43.27 &36.13 &39.38 &11.89 &96.19
&41.29 & 37.73 & 39.43 & ~~8.56 & 94.87
\\
 \sc Spanish  
&41.93 & 34.63 & 37.93 &12.34 & 96.74
& 40.47 & 36.57 & 38.42 & ~~8.56 & 95.82
\\
\sc Portuguese
&41.17 &32.37 &36.24 &14.07 &96.91
&13.81 &~~1.77 & ~~3.14 & 86.33 & 98.21
\\
 \sc Hindi 
  &~~4.28 & ~~1.62 & ~~2.35 &67.38& 98.64
& ~~3.69 & ~~1.69 & ~~2.31 & 60.62 & 98.43
\\
\bottomrule
\end{tabular}
}
\caption{Comparison of the zero-shot performance on m\textsc{WebIE} with mBART when specifying the source language (XX) and keeping the default setting as the source language (EN).
Results are with standard beam search (without the constraint Trie).
}
\label{tab:src-lang}
\end{table}

\begin{table}[ht!]
\centering
\scalebox{0.85}{
\addtolength{\tabcolsep}{0pt}

\begin{tabular}{p{2cm}|p{3.5cm}|p{8.7cm}}
\toprule
   \textbf{Example Id} & \textbf{Sentence} & \textbf{ReFinED Output}
   \\
\midrule
21464177
&
On Thursday, British campaigning group the Environmental Investigation Agency accused Italy of trying to sabotage efforts to reform the EU ETS.
&
[[``Thursday'', None, ``DATE''], [``British'', Entity(wikidata\_entity\_id=Q145, wikipedia\_entity\_title=United Kingdom), ``GPE''], ["Environmental Investigation Agency", Entity(wikidata\_entity\_id=Q1345905, wikipedia\_entity\_title=Environmental Investigation Agency), "ORG"], ["Italy", Entity(wikidata\_entity\_id=Q38, wikipedia\_entity\_title=Italy), ``ORG''], [``EU'', Entity(wikidata\_entity\_id=Q458, wikipedia\_entity\_title=European Union), ``ORG''], [``ETS'', Entity(wikidata\_entity\_id=Q899383, wikipedia\_entity\_title=ETSI), ``ORG'']] 
\\
\midrule
1274217
&
It culminates in the decade-long debate ending in 1913 to turn the Hetch Hetchy valley in Yosemite National Park into a reservoir for San Francisco. 
&
[[``decade-long22'', None, ``DATE''], [``1913'', Entity(parsed\_string=[timepoint: [``1913'']]), ``DATE''], [``Hetch Hetchy'', Entity(wikidata\_entity\_id=Q1616130, wikipedia\_entity\_title=Hetch Hetchy), ``GPE''], [``Yosemite National Park'', Entity(wikidata\_entity\_id=Q180402, wikipedia\_entity\_title=Yosemite National Park), ``FAC''], [``San Francisco'', Entity(wikidata\_entity\_id=Q62, wikipedia\_entity\_title=San Francisco), ``GPE'']]  \\
\bottomrule
\end{tabular}
}
    \caption{ReFinED outputs on \textsc{WebIE} validation examples.}
    \label{tab:refined-examples}
\end{table}

\begin{figure}[ht]
\centering
    \includegraphics[width=0.95\linewidth]{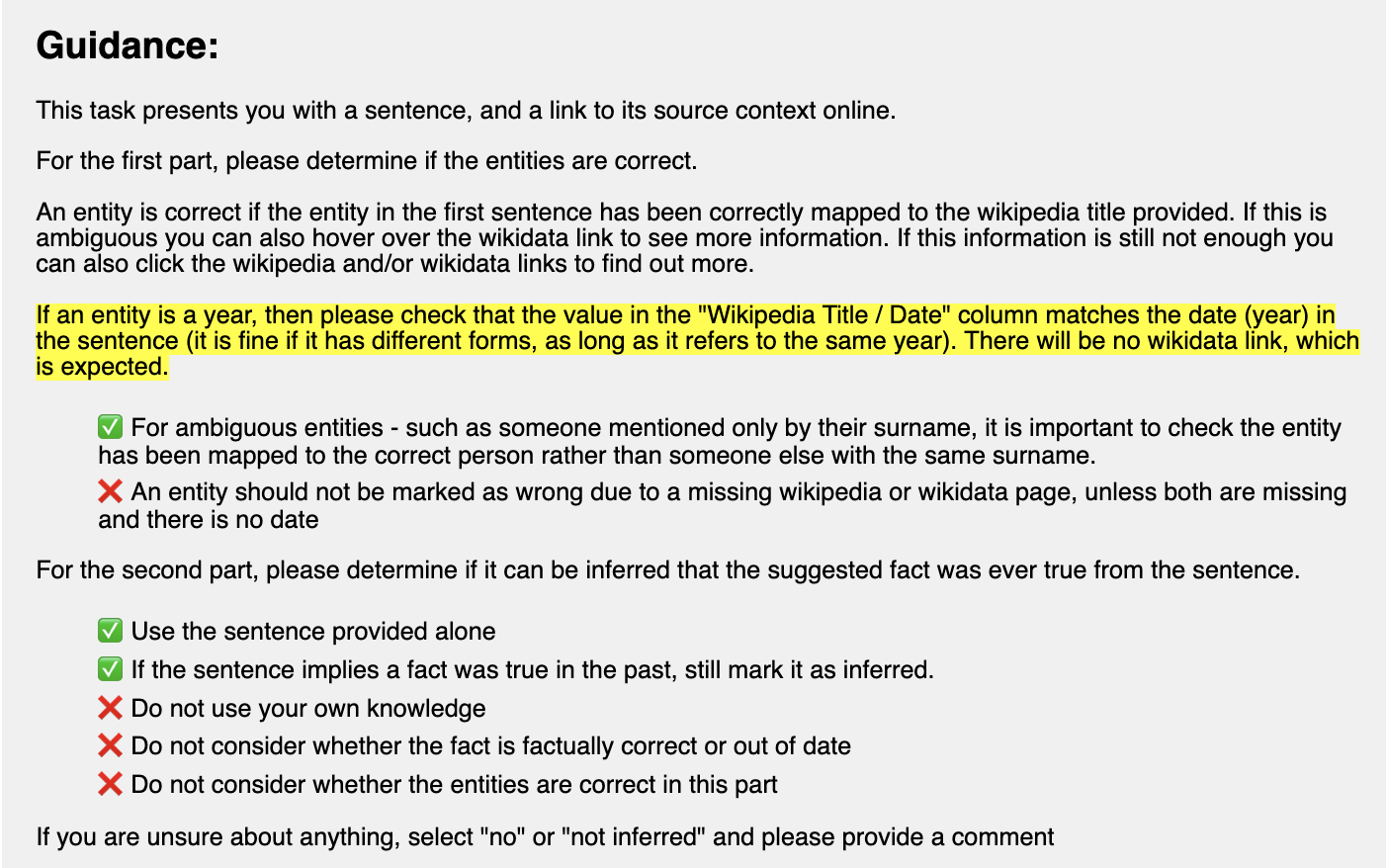}
    \caption{MTurk HIT guidance entity and relation labelling. }
\label{fig:guidance}
\end{figure}

\section{Examples of ReFinED Output}
\label{sec:ReFinED}

We show examples of the sentences processed by ReFinED in \autoref{tab:refined-examples}. For each input sentence, ReFinED identifies the set of entities in that sentence, and outputs mention span, Wikidata id, and Wikipedia title for each entity. For our experiments, we use the \texttt{wikipedia\_model\_with\_numbers} model with \texttt{wikipedia} entity set.

\section{MTurk Annotation Details}
\label{sec:mturk}
In this section, we describe the detailed settings for annotating (m)\webie with MTurk.

\subsection*{\textsc{WebIE}}
\begin{figure}[ht!]
\vspace{1.5ex}
\centering
    \includegraphics[width=0.95\linewidth]{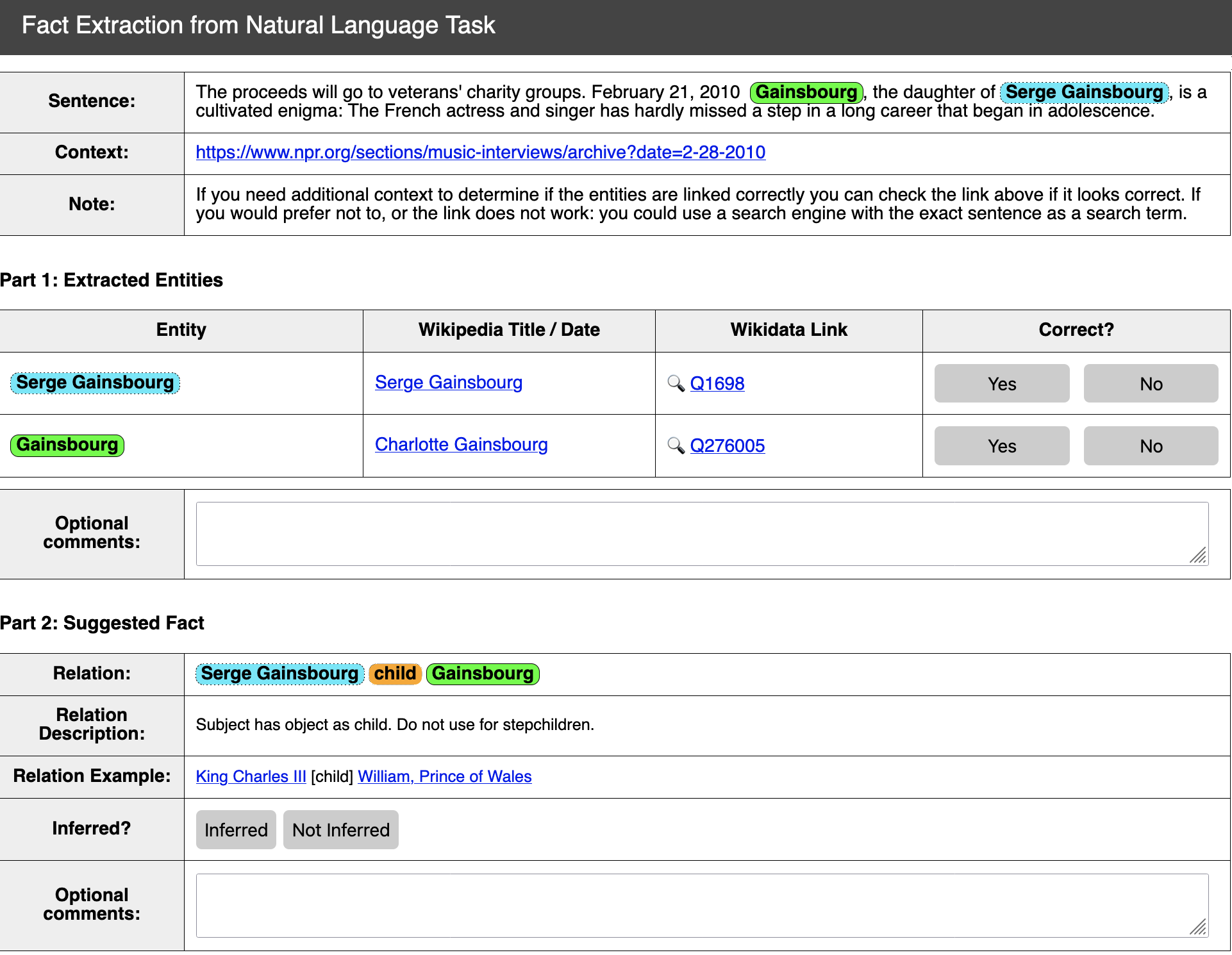}
    \caption{MTurk HIT user interface for entity and relation labelling. }
\label{fig:webie_mturk1}
\end{figure}

The first annotation task (HIT) is to verify the correctness of the triples automatically created from the DS approach and filtered by the NLI model. The guidance and the interface are shown in \autoref{fig:guidance} and \autoref{fig:webie_mturk1}, respectively.

In each HIT, we provide a sentence with its entities highlighted (head entity in blue and tail entity in green) and the URL of the web page which the sentence is extracted from.
For the first EL annotation job, we provide both the links to the Wikipedia and Wikidata pages.
Annotators are asked to choose if the highlighted spans are linked correctly to the KB.
Next, the annotators are asked to verify if a relation (highlighted in orange) can be inferred from the sentence.
We provide the description of the relation and an example use case to facilitate the annotation.

Each triple is annotated by three workers, and we pay \$0.2 per HIT.
We hire MTurk workers with Masters Qualification and set additional requirements including (i) having done 2,000 HITs and (ii) having a job approval rate $\geq$99\%.

\subsection*{m\textsc{WebIE}}
\label{sec:anno-mwebie}

\begin{figure}[ht]
\centering
    \includegraphics[width=0.95\linewidth]{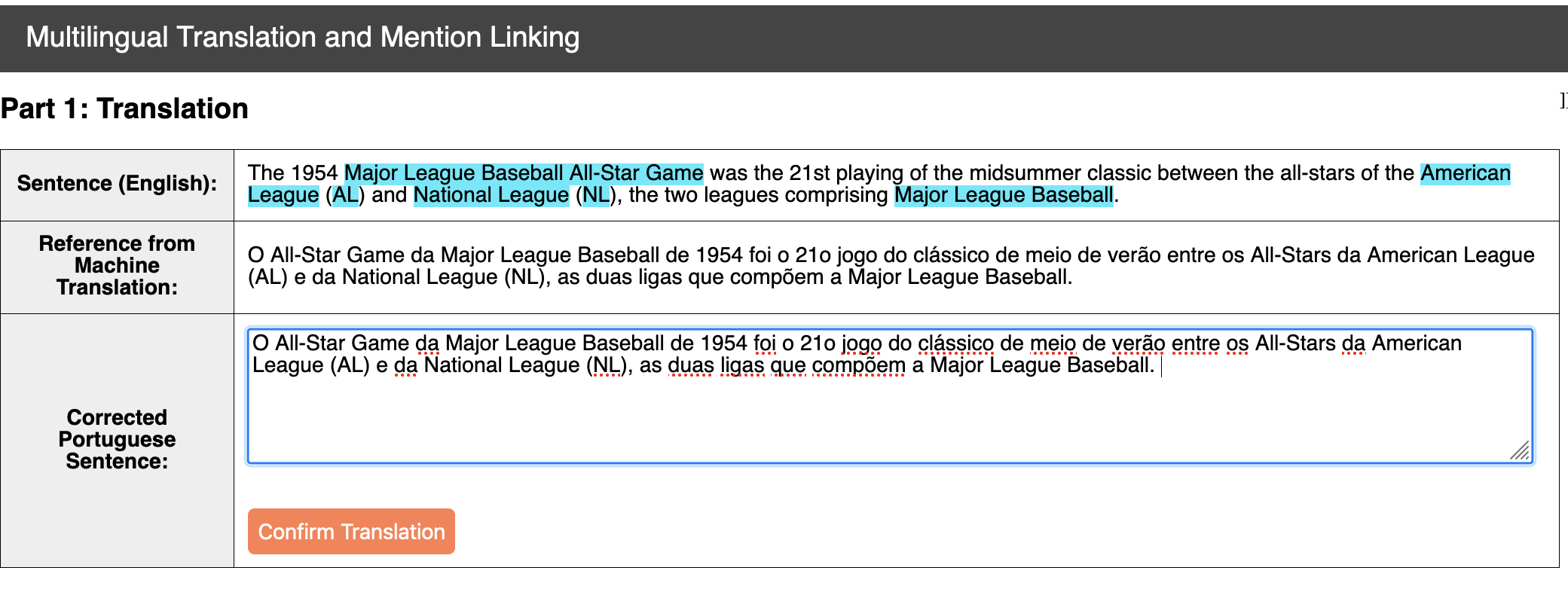}
    \caption{MTurk HIT user interface for correcting the machine-translated text. }
\label{fig:webie_mturk2}
\end{figure}

\begin{figure}[ht!]
\centering
    \includegraphics[width=0.95\linewidth]{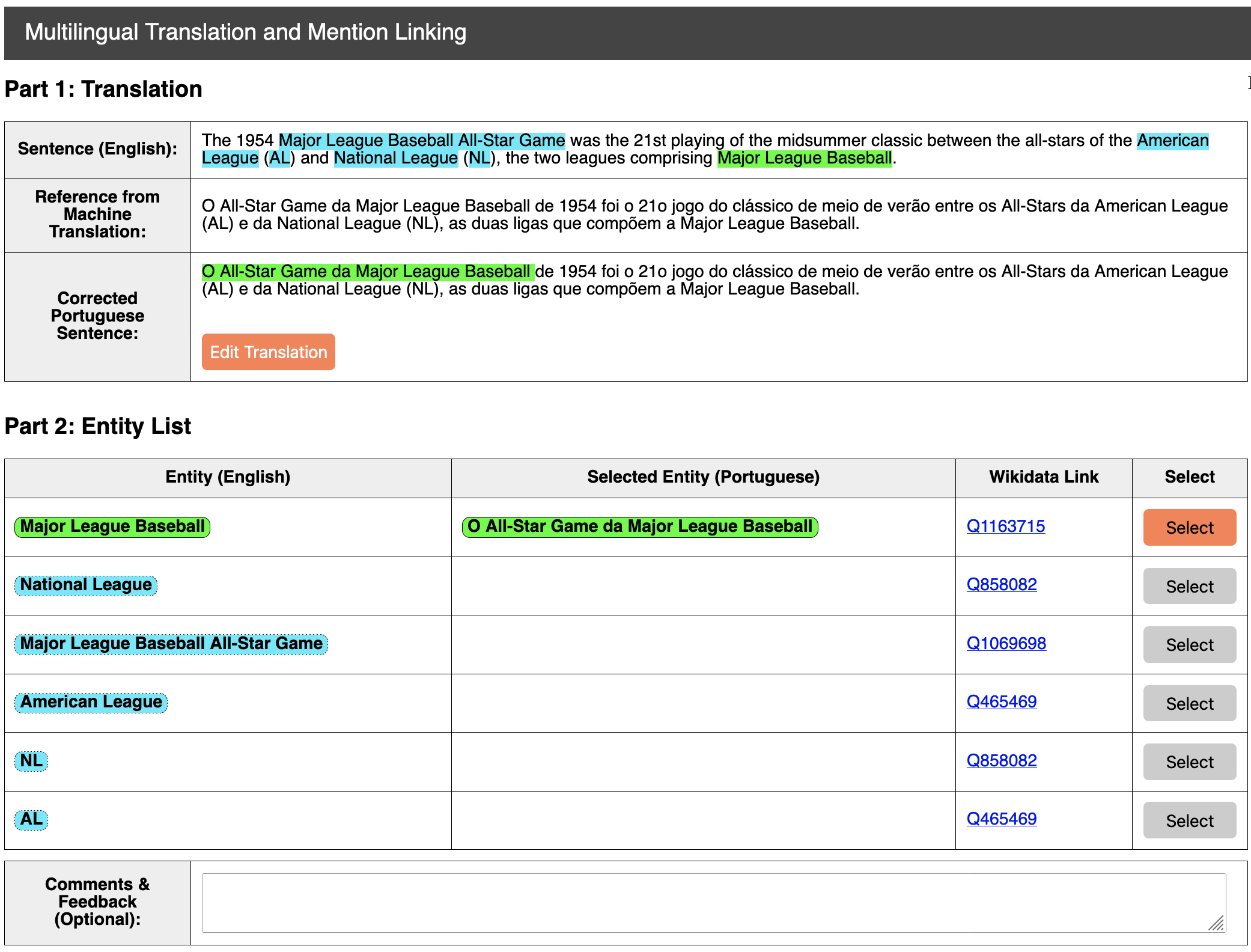}
    \caption{MTurk HIT user interface for entity labelling in the target language. }
\label{fig:webie_mturk3}
\end{figure}

\autoref{fig:webie_mturk2} and \autoref{fig:webie_mturk3} illustrates the interface for correcting machine-translated sentence and identifying corresponding entities in them.
As it is challenging to find qualified crowd workers for the translation task\footnote{Preliminary results where we include the USA for the m\webie\ annotation task indicate that MTurk workers with limited or no knowledge of the target language (or English) still accept the job, despite our specific requirement for proficiency in both English and the target language.}, we set the geographical regions for each language to the countries where the language is one of the official languages.
We find that only India and countries in America have an adequate number of MTurk workers, which highly restricts the options for our target languages.
In the end, the countries we set for the target languages are as follows: 
Portuguese: AO, BR, CV, ST, GW, GQ, MZ; 
Spanish: ES, MX, CO, PE, CL; CA for French, and IN for Hindi\footnote{For the mapping between country codes and countries, please refer to \url{https://docs.aws.amazon.com/AWSMechTurk/latest/AWSMturkAPI/ApiReference_LocaleDataStructureArticle.html}.}.
It was also necessary to remove the Masters Qualification requirement for MTurk workers (except Hindi) to find adequate annotators.
We then conduct pilot annotations, where we deliberately introduce errors in the reference machine translation to verify if the workers under our requirement settings are able to correct them.

We provide the English sentence paired with the original machine-translated sentence for the actual HIT.
The English sentence is highlighted with its entity spans, and we instruct the workers to correct the translation while ensuring that the entities are correctly translated.
After confirming the translation, workers are then asked to highlight the corresponding entities in the target language (in green).
For negative sentences without entity spans, the longest noun phrases were highlighted instead to prevent workers from simply copying the reference translations.
We pay \$0.35 per HIT for positive sentences and \$0.25 for negative sentences (since most sentences in negative examples have only one highlighted entity/noun phrase and it is considered an easier task).

Two MTurk workers were asked for the translation task, and an additional worker was asked to select the better translation, for which \$0.10 per HIT was paid.

\chapter{Additional Results and Analysis for \textsc{UMAE}}

\label{sec:VQA-appendix}

\section{Detailed Explanation Scores} \label{sec:explanation_scores}
For explanation generation on the VQA tasks, we evaluate the performance of beam search with the size of 5, top-k sampling with $k$ from $\{50, 100, 200, ..., 1000\}$, and Nucleus and Typical \citep{meister-etal-2023-locally} sampling, both with $p$ from $\{0.1, 0.2, ..., 0.9\}$.
We show the details of the NLG scores using different decoding strategies for explanations generated from \texttt{QA$\rightarrow$E} in \autoref{tab:sampling}, and \texttt{Q$\rightarrow$AE} in \autoref{tab:ar_detail}.

\begin{table}[th]
\centering
\vspace{1.5ex}
\scalebox{0.8}{
\addtolength{\tabcolsep}{-2.5pt}
\begin{tabular}{ll|cccrrccrcc}
\toprule
{\multirow{2}{*}{\sc Dataset}}
 & {\multirow{2}{*}{\sc Decoding}} & {e{\sc-V}i{\sc l }} & \multicolumn{8}{c} {\sc {N-gram} Scores} & \sc Learnt
\\

&   & \sc \textit{S}\textsubscript{E}     &{\sc b\small{1}} & {\sc b\small{2}} & \multicolumn{1}{c}{\sc b\small{3}} & \multicolumn{1}{c}{\sc b\small{4}}  & \sc r-l   &\sc met.  & {\sc cide}r & \sc spice & \sc bertsc.   \\\cmidrule(r){1-2} 
\cmidrule(lr){3-3} \cmidrule(lr){4-11} \cmidrule(l){12-12}

{\multirow{4}{*}{\sc {A-okvqa}}}  & \sc beam search &  44.71 & 52.01 & 36.69 & 26.72 & 19.88 & 40.39 & 22.06 & 68.48 & 20.94   & 86.05  \\
&  {\sc  top-k} ($k=100$) & 44.34 & 52.56 & 37.06 & 27.06 & 19.72 & 44.45 & 21.58 & 73.44 & 19.38  & 86.27\\
& {\sc  nucleus} ($p=0.4$) &\textbf{50.82} & 58.92 & 44.66 & 35.06 & 27.35 & 52.56 & 24.83 & 101.09 & 23.33 & 88.21 \\
&{\sc typical} ($p=0.6$) & 47.27 & 54.18 & 39.39 & 29.82 & 22.18 & 47.78 & 22.79 & 84.43 & 21.47  & 86.95\\
\midrule
{\multirow{4}{*}{\sc {Vcr}}} & \sc beam search  & 40.23 & 26.41 & 20.15 & 15.95& 12.47 & 29.13 & 16.82 & 49.72  & 27.70 & 81.84\\
&  {\sc top-k} ($k=50$) & 33.19 & 20.98 & 14.89 & 11.18 & 8.33 & 23.65 & 13.72 & 32.73 & 21.99 & 80.31 \\
& {\sc nucleus} ($p=0.1$) & \textbf{40.27} & 31.42 & 22.95 & 17.62 & 13.44 & 29.53 & 17.54 & 47.33 & 26.45   & 81.91 \\
&{\sc typical} ($p=0.4$) &35.12 & 23.42 & 16.88 & 12.83 & 9.64  & 25.36 & 14.70 & 35.85 & 23.32  & 80.70 \\
\midrule
{\multirow{4}{*}{\sc {Vqa-x}}} & \sc beam search & \cellcolor[HTML]{DDEBF7}35.88     & \cellcolor[HTML]{DDEBF7}37.84& \cellcolor[HTML]{DDEBF7}24.91 & \cellcolor[HTML]{DDEBF7} 16.67& \cellcolor[HTML]{DDEBF7}10.97       & \cellcolor[HTML]{DDEBF7}31.32       &\cellcolor[HTML]{DDEBF7}17.90     &\cellcolor[HTML]{DDEBF7}38.23   & \cellcolor[HTML]{DDEBF7}16.23      &\cellcolor[HTML]{DDEBF7}84.39\\
&  {\sc top-k} ($k=50$) & \cellcolor[HTML]{DDEBF7}33.28 &\cellcolor[HTML]{DDEBF7}38.35 & \cellcolor[HTML]{DDEBF7}23.11 &\cellcolor[HTML]{DDEBF7}14.21 &\cellcolor[HTML]{DDEBF7}8.45  &\cellcolor[HTML]{DDEBF7}29.15 &\cellcolor[HTML]{DDEBF7}17.05 & \cellcolor[HTML]{DDEBF7}32.89 & \cellcolor[HTML]{DDEBF7}15.26 &\cellcolor[HTML]{DDEBF7}83.41\\
&{\sc nucleus} ($p=0.1$) & \cellcolor[HTML]{DDEBF7}\textbf{40.67} &\cellcolor[HTML]{DDEBF7}47.56 & \cellcolor[HTML]{DDEBF7}31.44 & \cellcolor[HTML]{DDEBF7}21.47 & \cellcolor[HTML]{DDEBF7}14.63 &\cellcolor[HTML]{DDEBF7}35.12 & \cellcolor[HTML]{DDEBF7}20.29 &\cellcolor[HTML]{DDEBF7}50.35 & \cellcolor[HTML]{DDEBF7}19.13  &\cellcolor[HTML]{DDEBF7}85.40\\
&{\sc typical} ($p=0.5$) & \cellcolor[HTML]{DDEBF7}36.31 &\cellcolor[HTML]{DDEBF7}40.85 &\cellcolor[HTML]{DDEBF7}25.57 &\cellcolor[HTML]{DDEBF7}16.82 &\cellcolor[HTML]{DDEBF7}11.14 &\cellcolor[HTML]{DDEBF7}31.08 &\cellcolor[HTML]{DDEBF7}18.15 &\cellcolor[HTML]{DDEBF7}39.71 &\cellcolor[HTML]{DDEBF7}16.62  &\cellcolor[HTML]{DDEBF7}83.93 \\
\bottomrule
\end{tabular}

}
\caption{Explanation scores with automatic NLG for generated explanations (\texttt{QA$\rightarrow$E}) from \textsc{umae\textsubscript{all}} model with different decoding strategies.
\textsc{b\small{1}}~\textsc{b\small{4}} correspond to \textsc{bleu\small{1}}~\textsc{bleu\small{4}}, \textsc{r-l} means \textsc{rouge-l} and \textsc{met.} means \textsc{meteor}.
The last two rows (with blue shadow) indicate out-of-domain performance.
}
\label{tab:sampling}
\end{table}

\begin{table}[ht]
\centering
\scalebox{0.78}{
\addtolength{\tabcolsep}{-2pt}
\begin{tabular}{ll|cccrrccrcc}
\toprule
{\multirow{2}{*}{\sc Dataset}}
 & {\multirow{2}{*}{\sc Decoding}} & {e{\sc-V}i{\sc l }} & \multicolumn{8}{c} {\sc {N-gram} Scores} & \sc Learnt
\\

&   & \sc \textit{S}\textsubscript{E}      &{\sc b\small{1}} & {\sc b\small{2}} & \multicolumn{1}{c}{\sc b\small{3}} & \multicolumn{1}{c}{\sc b\small{4}}  & \sc r-l   &\sc met. & {\sc cide}r & \sc spice & \sc bertsc.    \\\cmidrule(r){1-2} 
\cmidrule(lr){3-3} \cmidrule(lr){4-11} \cmidrule(l){12-12}

{\multirow{2}{*}{\sc {A-okvqa}}}  & \sc beam search &  \textbf{47.01} & 54.75 & 41.39 & \textbf{32.08} & \textbf{24.25} & \textbf{49.75} & \textbf{22.54} & \textbf{86.28} & \textbf{20.68} & \textbf{87.39}  \\
& {\sc  nucleus} ($p=0.5$) &46.72   & \textbf{55.53} & \textbf{41.63} & 31.91   & 23.67  & 49.16   & 22.48  & 82.37  & 20.67   & 87.18    \\
\midrule
{\multirow{2}{*}{\sc {Vcr}}} & \sc beam search  &\textbf{37.02} & 25.00          & 18.90          & \textbf{14.87} & \textbf{11.54} & \textbf{27.07} & \textbf{15.66} & \textbf{38.77} & \textbf{25.03} & \textbf{80.68} \\
& {\sc nucleus} ($p=0.1$) &  35.10     & \textbf{27.41} & \textbf{19.36} & 14.50   & 10.73  & 26.18    & 15.21 & 34.99    & 21.88  & 80.52  \\

\midrule
{\multirow{2}{*}{\sc {Vqa-x}}} & \sc beam search & \cellcolor[HTML]{DDEBF7}38.13 & \cellcolor[HTML]{DDEBF7}39.91 & \cellcolor[HTML]{DDEBF7}26.30 & \cellcolor[HTML]{DDEBF7}17.99 & \cellcolor[HTML]{DDEBF7}12.46 &\cellcolor[HTML]{DDEBF7}31.69 & \cellcolor[HTML]{DDEBF7}19.11 & \cellcolor[HTML]{DDEBF7}42.10 & \cellcolor[HTML]{DDEBF7}18.15 & \cellcolor[HTML]{DDEBF7}84.95\\
&{\sc nucleus} ($p=0.1$) & \cellcolor[HTML]{DDEBF7}\textbf{39.67} & \cellcolor[HTML]{DDEBF7}\textbf{44.92} & \cellcolor[HTML]{DDEBF7}\textbf{28.88} & \cellcolor[HTML]{DDEBF7}\textbf{19.04} & \cellcolor[HTML]{DDEBF7}\textbf{12.55} & \cellcolor[HTML]{DDEBF7}\textbf{33.08} &\cellcolor[HTML]{DDEBF7}\textbf{20.07} &\cellcolor[HTML]{DDEBF7}\textbf{44.28} & \cellcolor[HTML]{DDEBF7}\textbf{19.19} & \cellcolor[HTML]{DDEBF7}\textbf{85.21}\\

\bottomrule
\end{tabular}
}
\caption{Explanation scores with automatic NLG for generated explanations from \texttt{Q$\rightarrow$AE} with \textsc{umae\textsubscript{all}} model.
\textsc{b\small{1}}~\textsc{b\small{4}} correspond to \textsc{bleu\small{1}}~\textsc{bleu\small{4}}, \textsc{r-l} means \textsc{rouge-l} and \textsc{met.} means \textsc{meteor}.
The last two rows (with blue shadow) indicate out-of-domain performance.
}
\label{tab:ar_detail}
\end{table}

\section{Examples of Generated Explanations} \label{sec:explanations}
\begin{figure}[!t]
\centering
    \includegraphics[width=\linewidth]{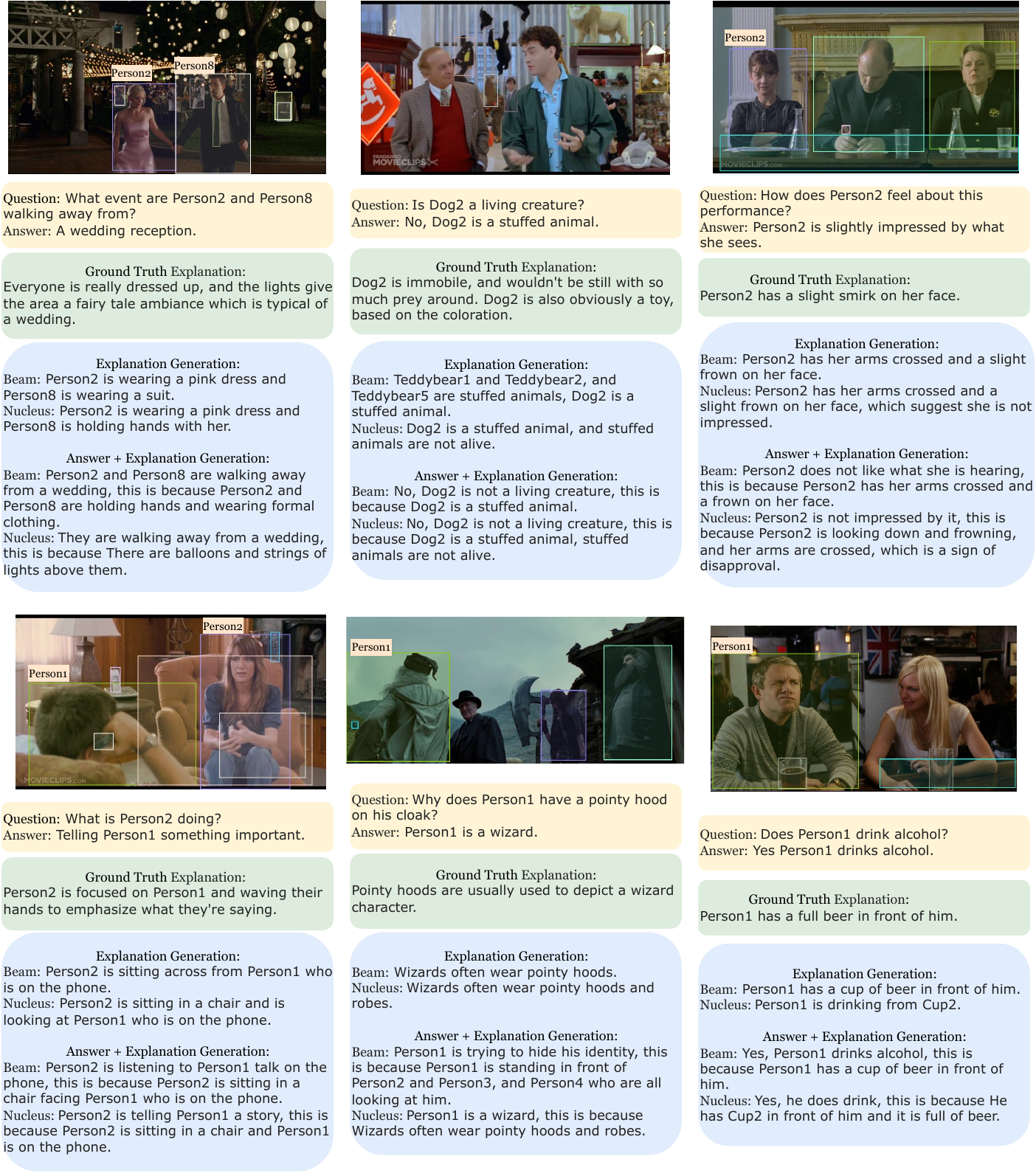}
    \caption{Examples of generated answers and explanations generation for VCR.}
\label{fig:vcr}
\end{figure}

\begin{figure}[!t]
\centering
    \includegraphics[width=\linewidth]{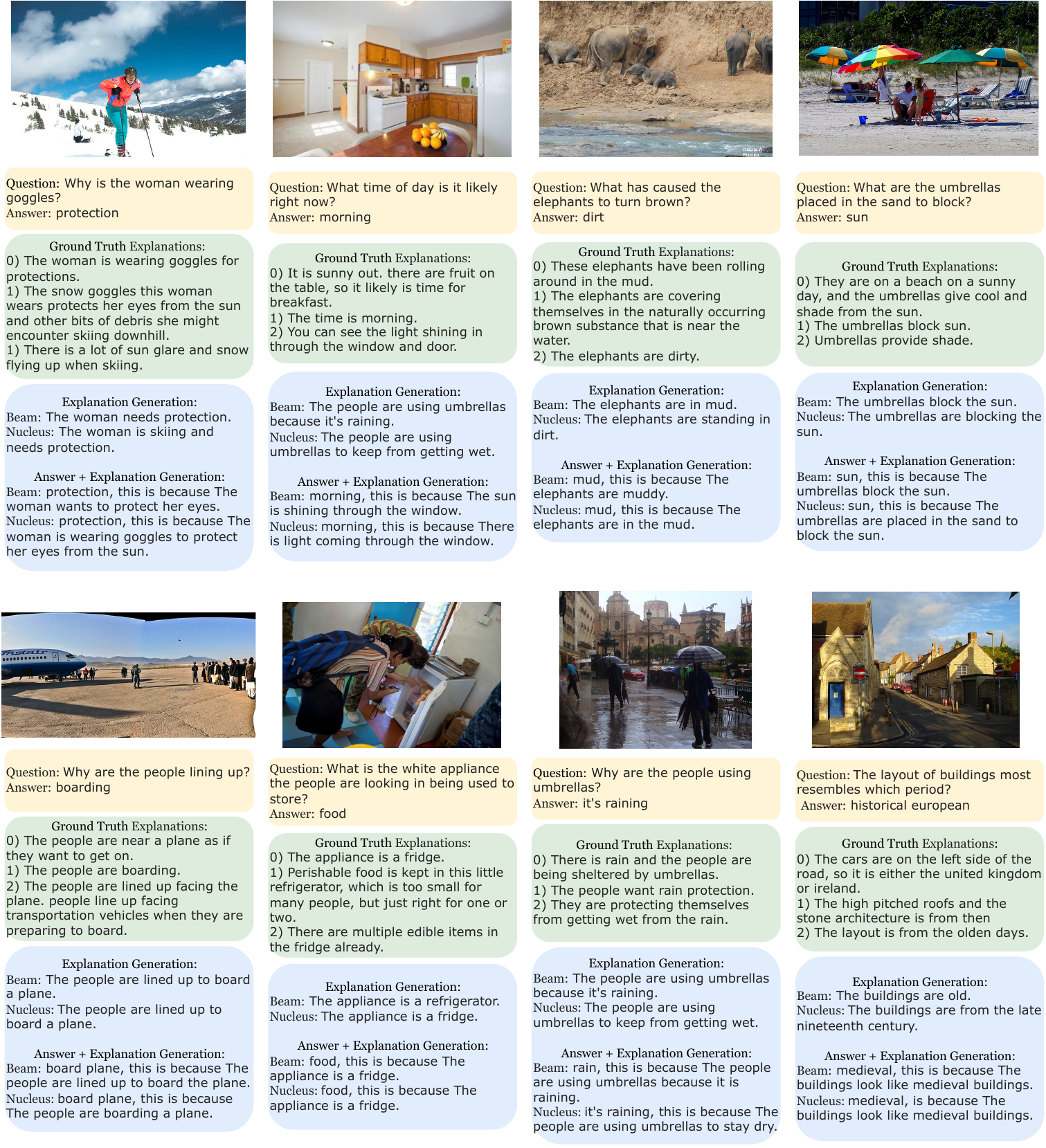}
    \caption{Examples of generated answers and explanations for A-OKVQA.}
\label{fig:aok}
\end{figure}

Examples of the explanations generated with beam search and Nucleus sampling for A-OKVQA are shown in \autoref{fig:aok}, and VCR in \autoref{fig:vcr}.

In general, Nucleus sampling achieves the best performance across the datasets. However, top-k and Typical sampling do not show improvement over beam search.

\section{Examples of Issues in the Datasets} \label{sec:datasets_issues}
We show some of the issues in the datasets described in \autoref{sec:dataset_issue}.

\begin{figure}[!t]
\centering
    \includegraphics[width=\linewidth]{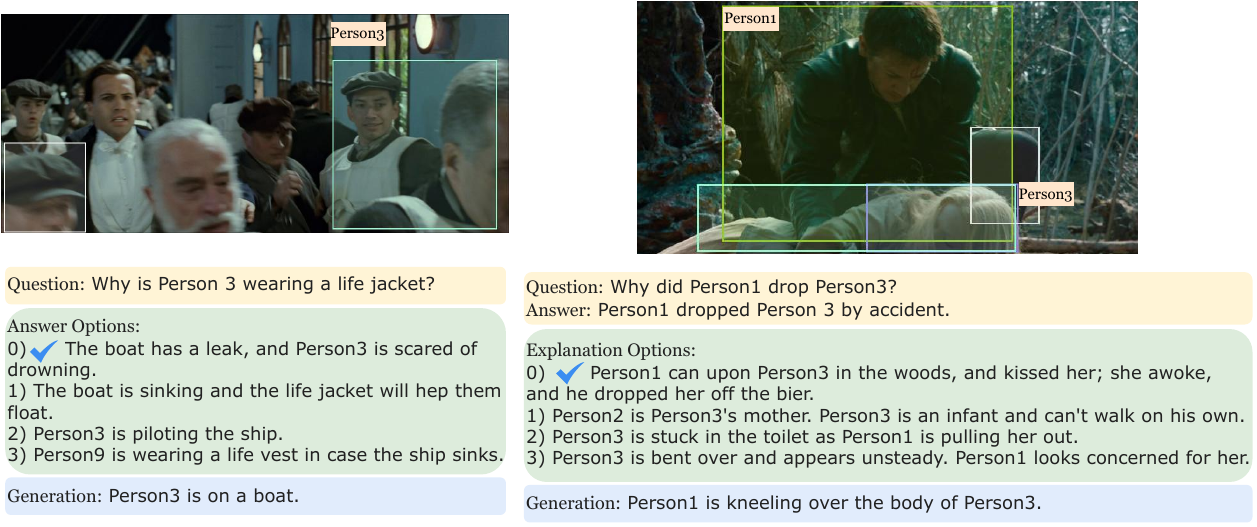}
    \caption{Questions that require knowledge of the movie plots to generate the answers from VCR.}
\label{fig:moive_know}
\end{figure}

\begin{figure}[!t]
\centering
    \includegraphics[width=\linewidth]{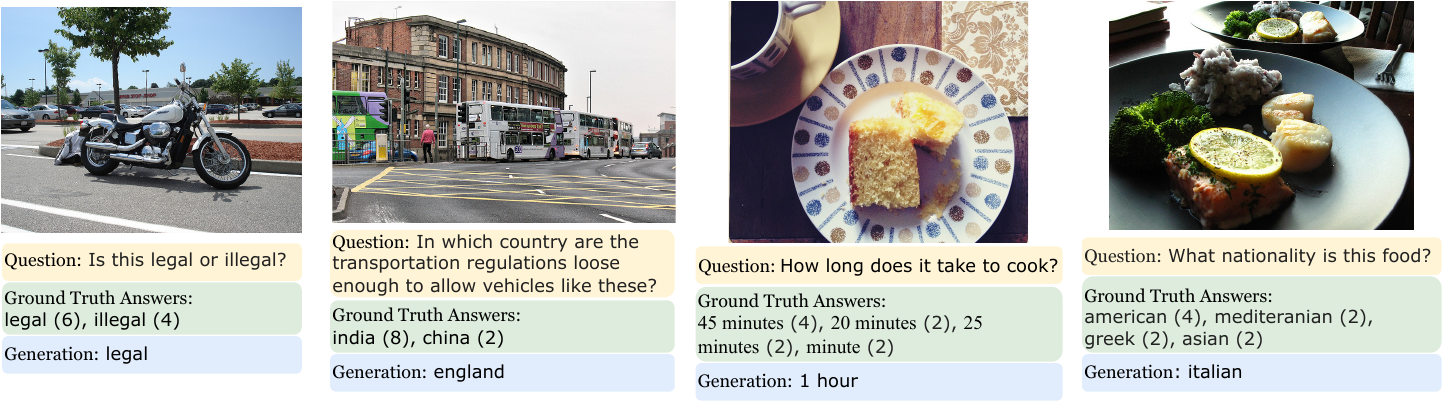}
    \caption{Examples of subjective questions from OK-VQA.}
\label{fig:dataset}
\end{figure}

\begin{figure}[!t]
\centering
    \includegraphics[width=\linewidth]{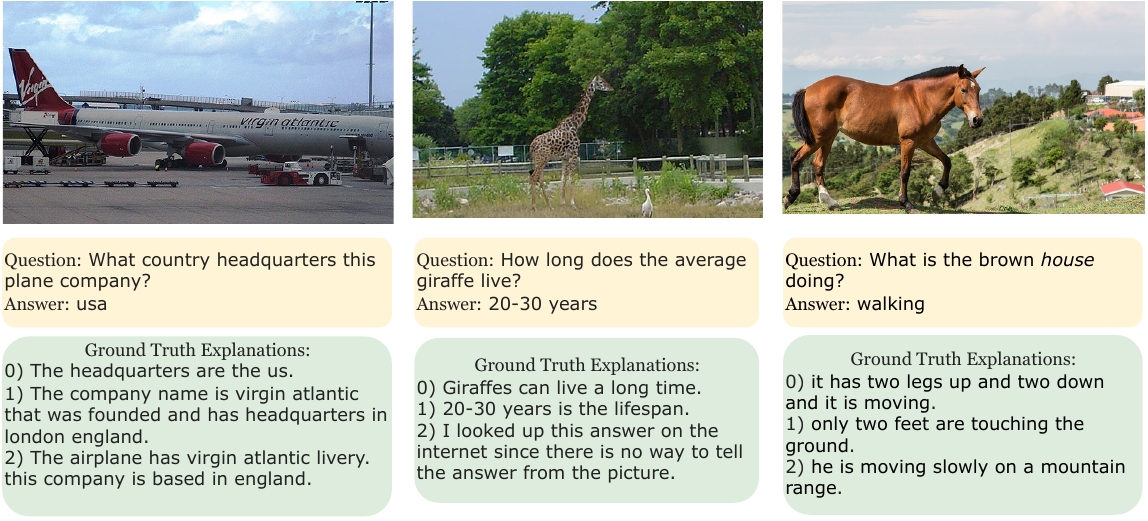}
    \caption{Issues in the datasets that severely impact the model generation: wrong answers (left, from A-OKVQA), questions that do not need visual input to answer (middle, from A-OKVQA), and typo (right, from VQA-X).}
\label{fig:dataset_issues}
\end{figure}

\autoref{fig:moive_know} shows examples from VCR that require an understanding of the movie plot to generate answers.
\autoref{fig:dataset} shows examples from OK-VQA where questions and answers are subjective or ambiguous.
\autoref{fig:dataset_issues} shows examples from A-OKVQA and VQA-X that either contain wrong answers, questions that do not need visual input or typos which severely impact the model generation (``house'' should be ``horse'').

\clearpage

\clearpage

\chapter{Additional Results for LLM-powered Data Augmentation}
\label{sec:additional}

This section includes the following additional results:
\autoref{tab:all-copa}, \autoref{tab:xwino}, and \autoref{tab:xstory} show generated data in English with different LLMs on XCOPA, XWinograd, and XStoryCloze.
\autoref{tab:xcopa_full}  and \autoref{tab:gpt4-copa} show the full result on XCOPA with ChatGPT and GPT-4.

\begin{table*}[h]
\vspace{1.5ex}
\centering
\scalebox{0.75}{
\addtolength{\tabcolsep}{-3pt}
\begin{tabular}{lll|rrrrrrrrrrrrr}
\toprule

\textbf{Fine-tuned} &
  \textbf{Train Data} &
  \textbf{LLM} &
  \multicolumn{1}{l}{\textbf{AVG}} &
  \multicolumn{1}{l}{\textbf{EN}} &
  \multicolumn{1}{l}{\textbf{ET}} &
  \multicolumn{1}{l}{\textbf{HT}} &
  \multicolumn{1}{l}{\textbf{ID}} &
  \multicolumn{1}{l}{\textbf{IT}} &
  \multicolumn{1}{l}{\textbf{QU}} &
  \multicolumn{1}{l}{\textbf{SW}} &
  \multicolumn{1}{l}{\textbf{TA}} &
  \multicolumn{1}{l}{\textbf{TH}} &
  \multicolumn{1}{l}{\textbf{TR}} &
  \multicolumn{1}{l}{\textbf{VI}} &
  \multicolumn{1}{l}{\textbf{ZH}} \\
  \midrule
\multirow{8}{*}{mBERT} &
  \multirow{4}{*}{GEN} &
  Dolly-v2 &
  54.0 &
  63.4 &
  52.0 &
  52.2 &
  54.0 &
  53.8 &
  47.6 &
  48.6 &
  53.4 &
  \textbf{53.4} &
  52.8 &
  50.4 &
  58.2 \\
 &
   &
  StableVicuna &
  53.5 &
  62.4 &
  51.6 &
  49.2 &
  55.8 &
  55.8 &
  50.0 &
  \textbf{50.2} &
  50.2 &
  52.6 &
  51.0 &
  50.4 &
  56.0 \\
 &
   &
  ChatGPT &
  56.0 &
  64.8 &
  54.8 &
  52.6 &
  58.0 &
  57.4 &
  49.8 &
  48.4 &
  \textbf{55.6} &
  52.8 &
  53.2 &
  53.0 &
  59.0 \\
 &
   &
  GPT-4 &
  \textbf{58.2} &
  \textbf{69.2} &
  \textbf{59.2} &
  \textbf{54.0} &
  \textbf{60.6} &
  \textbf{59.2} &
  \textbf{50.8} &
  48.2 &
  55.0 &
  48.2 &
  \textbf{53.8} &
  \textbf{57.6} &
  \textbf{61.0} \\
    \cmidrule{2-16} 
 &
  \multirow{4}{*}{GEN+ORI} &
  Dolly-v2 &
  54.4 &
  59.8 &
  52.6 &
  53.2 &
  53.0 &
  56.4 &
  \textbf{53.8} &
  \textbf{52.4} &
  50.4 &
  \textbf{54.8} &
  49.8 &
  52.6 &
  58.8 \\
 &
   &
  StableVicuna &
  55.6 &
  65.2 &
  53.4 &
  50.4 &
  59.0 &
  60.0 &
  51.6 &
  50.4 &
  49.4 &
  52.0 &
  52.4 &
  54.0 &
  58.2 \\
 &
   &
  ChatGPT &
  54.6 &
  59.6 &
  56.4 &
  \textbf{53.6} &
  53.8 &
  51.4 &
  51.4 &
  51.6 &
  \textbf{50.4} &
  52.6 &
  \textbf{54.0} &
  55.0 &
  59.2 \\
 &
   &
  GPT-4 &
  \textbf{59.3} &
  \textbf{72.6} &
  \textbf{58.8} &
  53.0 &
  \textbf{62.0} &
  \textbf{61.0} &
  53.0 &
  50.0 &
  54.0 &
  48.2 &
  52.0 &
  \textbf{57.6} &
  \textbf{64.6} \\
    \cmidrule{1-16} 
\multirow{8}{*}{XLMR-Base} &
  \multirow{4}{*}{GEN} &
  Dolly-v2 &
  59.0 &
  64.4 &
  58.8 &
  52.8 &
  60.8 &
  61.0 &
  50.8 &
  55.6 &
  60.4 &
  58.0 &
  57.2 &
  58.6 &
  59.0 \\
 &
   &
  StableVicuna &
  58.5 &
  60.4 &
  59.4 &
  \textbf{53.6} &
  60.8 &
  56.8 &
  49.2 &
  56.0 &
  61.2 &
  60.4 &
  54.8 &
  59.6 &
  58.6 \\
 &
   &
  ChatGPT &
  58.8 &
  62.4 &
  56.4 &
  52.4 &
  61.4 &
  58.6 &
  \textbf{52.2} &
  52.0 &
  63.4 &
  61.2 &
  56.4 &
  59.6 &
  62.8 \\
 &
   &
  GPT-4 &
  \textbf{63.6} &
  \textbf{67.0} &
  \textbf{62.4} &
  52.0 &
  \textbf{68.6} &
  \textbf{62.6} &
  51.8 &
  \textbf{58.6} &
  \textbf{65.4} &
  \textbf{64.8} &
  \textbf{63.2} &
  \textbf{66.6} &
  \textbf{69.6} \\
   \cmidrule{2-16} 
 &
  \multirow{4}{*}{GEN+ORI} &
  Dolly-v2 &
  58.7 &
  65.6 &
  57.6 &
  \textbf{52.2} &
  60.8 &
  58.4 &
  52.4 &
  58.2 &
  57.4 &
  58.0 &
  58.4 &
  58.0 &
  59.8 \\
 &
   &
  StableVicuna &
  61.1 &
  65.0 &
  62.4 &
  49.4 &
  64.2 &
  62.4 &
  46.2 &
  \textbf{60.4} &
  59.6 &
  58.0 &
  58.0 &
  63.0 &
  63.4 \\
 &
   &
  ChatGPT &
  59.8 &
  63.8 &
  61.6 &
  51.6 &
  62.6 &
  59.8 &
  51.2 &
  51.6 &
  60.4 &
  61.6 &
  61.8 &
  64.8 &
  62.0 \\
 &
   &
  GPT-4 &
  \textbf{63.6} &
  \textbf{69.6} &
  \textbf{63.8} &
  51.2 &
  \textbf{67.2} &
  \textbf{62.4} &
  \textbf{52.6} &
  58.4 &
  \textbf{63.8} &
  \textbf{66.0} &
  \textbf{64.2} &
  \textbf{66.8} &
  \textbf{69.4} \\
   \cmidrule{1-16} 
\multirow{8}{*}{XLMR-Large} &
  \multirow{4}{*}{GEN} &
  Dolly-v2 &
  59.6 &
  62.4 &
  58.6 &
  49.6 &
  64.8 &
  59.2 &
  50.6 &
  56.8 &
  60.8 &
  58.8 &
  57.0 &
  61.0 &
  63.0 \\
 &
   &
  StableVicuna &
  65.7 &
  71.4 &
  66.2 &
  50.4 &
  71.4 &
  70.2 &
  50.0 &
  60.0 &
  64.0 &
  63.6 &
  68.0 &
  68.2 &
  69.8 \\
 &
   &
  ChatGPT &
  65.2 &
  71.2 &
  64.6 &
  51.6 &
  70.8 &
  66.6 &
  \textbf{51.0} &
  58.8 &
  66.0 &
  68.2 &
  69.0 &
  68.8 &
  68.8 \\
 &
   &
  GPT-4 &
  \textbf{73.6} &
  \textbf{83.2} &
  \textbf{71.2} &
  \textbf{52.0} &
  \textbf{81.2} &
  \textbf{78.2} &
  \textbf{51.0} &
  \textbf{62.2} &
  \textbf{76.6} &
  \textbf{77.4} &
  \textbf{75.0} &
  \textbf{78.4} &
  \textbf{79.0} \\

  \cmidrule{2-16} 
 &
  \multirow{4}{*}{GEN+ORI} &
  Dolly-v2 &
  66.4 &
  74.2 &
  62.8 &
  \textbf{53.0} &
  72.0 &
  70.4 &
  46.2 &
  61.6 &
  65.6 &
  66.2 &
  69.6 &
  67.6 &
  70.6 \\
 &
   &
  StableVicuna &
  69.9 &
  76.0 &
  69.8 &
  51.2 &
  75.0 &
  74.2 &
  51.2 &
  64.4 &
  70.2 &
  71.6 &
  72.2 &
  72.6 &
  75.4 \\
 &
   &
  ChatGPT &
  69.5 &
  76.4 &
  69.8 &
  48.2 &
  76.0 &
  72.8 &
  50.8 &
  63.4 &
  67.8 &
  70.8 &
  70.2 &
  73.4 &
  77.8 \\
 &
   &
  GPT-4 &
  \textbf{73.7} &
  \textbf{84.6} &
  \textbf{70.4} &
  50.0 &
  \textbf{80.8} &
  \textbf{80.2} &
  \textbf{51.8} &
  \textbf{65.8} &
  \textbf{72.8} &
  \textbf{76.0} &
  \textbf{74.8} &
  \textbf{78.4} &
  \textbf{80.4}
\\
\bottomrule
\end{tabular}
}
\caption{Accuracy on XCOPA with different LLM-generated English data.
}
 \label{tab:all-copa}
\end{table*}
\begin{table*}[!ht]
\centering
\scalebox{0.85}{
\addtolength{\tabcolsep}{-1pt}
\begin{tabular}{lll|rrrrrrr}
\toprule

\textbf{Fine-tuned} &
  \textbf{Training data} &
  \textbf{LLM} &
  \multicolumn{1}{l}{\textbf{AVG}} &
  \multicolumn{1}{l}{\textbf{EN}} &
  \multicolumn{1}{l}{\textbf{FR}} &
  \multicolumn{1}{l}{\textbf{JA}} &
  \multicolumn{1}{l}{\textbf{PT}} &
  \multicolumn{1}{l}{\textbf{RU}} &
  \multicolumn{1}{l}{\textbf{ZH}} \\
  \midrule
\multirow{8}{*}{mBERT} &
  \multirow{4}{*}{GEN} &
  Dolly-v2 &
  \textbf{56.47} &
  \textbf{71.24} &
  53.01 &
  52.45 &
  \textbf{53.23} &
  54.92 &
  53.97 \\
 &
   &
  StableVicuna &
  53.73 &
  54.94 &
  \textbf{56.63} &
  50.26 &
  50.57 &
  52.06 &
  57.94 \\
 &
   &
  ChatGPT &
  56.00 &
  54.94 &
  54.22 &
  \textbf{54.01} &
  52.09 &
  \textbf{55.87} &
  \textbf{64.88} \\
 &
   &
  GPT-4 &
  54.90 &
  56.22 &
  \textbf{56.63} &
  52.55 &
  51.71 &
  52.38 &
  59.92 \\
  \cmidrule{2-10}
 &
  \multirow{4}{*}{GEN+ORI} &
  Dolly-v2 &
  \textbf{59.32} &
  \textbf{71.24} &
  57.83 &
  \textbf{53.81} &
  56.65 &
  59.05 &
  57.34 \\
 &
   &
  StableVicuna &
  58.46 &
  57.94 &
  63.86 &
  \textbf{53.81} &
  \textbf{57.41} &
  58.41 &
  59.33 \\
 &
   &
  ChatGPT &
  58.26 &
  56.65 &
  \textbf{66.27} &
  53.60 &
  56.27 &
  \textbf{60.00} &
  56.75 \\
 &
   &
  GPT-4 &
  57.48 &
  53.65 &
  62.65 &
  \textbf{54.43} &
  55.89 &
  57.14 &
  \textbf{61.11} \\
   \cmidrule{1-10}
\multirow{8}{*}{XLMR-Base} &
  \multirow{4}{*}{GEN} &
  Dolly-v2 &
  59.63 &
  \textbf{71.24} &
  57.83 &
  55.79 &
  \textbf{57.03} &
  57.78 &
  58.13 \\
 &
   &
  StableVicuna &
  58.95 &
  60.09 &
  55.42 &
  57.35 &
  52.47 &
  58.73 &
  69.64 \\
 &
   &
  ChatGPT &
  62.69 &
  69.10 &
  60.24 &
  61.42 &
  \textbf{57.03} &
  \textbf{61.27} &
  67.06 \\
 &
   &
  GPT-4 &
  \textbf{63.32} &
  69.10 &
  \textbf{61.45} &
  \textbf{61.52} &
  56.65 &
  60.95 &
  \textbf{70.24} \\
   \cmidrule{2-10}
 &
  \multirow{4}{*}{GEN+ORI} &
  Dolly-v2 &
  66.33 &
  \textbf{75.54} &
  63.86 &
  65.80 &
  \textbf{64.26} &
  62.86 &
  65.67 \\
 &
   &
  StableVicuna &
  65.97 &
  64.38 &
  66.27 &
  67.15 &
  63.88 &
  \textbf{65.71} &
  68.45 \\
 &
   &
  ChatGPT &
  65.94 &
  65.24 &
  60.24 &
  \textbf{68.93} &
  70.72 &
  62.86 &
  67.66 \\
   \cmidrule{1-10}
 &
   &
  GPT-4 &
  \textbf{66.88} &
  68.24 &
  \textbf{67.47} &
  66.94 &
  63.88 &
  63.49 &
  \textbf{71.23} \\
\multirow{8}{*}{XLMR-Large} &
  \multirow{4}{*}{GEN} &
  Dolly-v2 &
  \textbf{76.86} &
  \textbf{87.55} &
  67.47 &
  \textbf{81.02} &
  \textbf{76.43} &
  74.29 &
  74.40 \\
 &
   &
  StableVicuna &
  68.22 &
  74.25 &
  63.86 &
  68.20 &
  66.16 &
  63.81 &
  73.02 \\
 &
   &
  ChatGPT &
  73.20 &
  81.97 &
  66.27 &
  73.10 &
  66.92 &
  72.38 &
  78.57 \\
 &
   &
  GPT-4 &
  76.37 &
  81.55 &
  \textbf{74.70} &
  75.91 &
  71.86 &
  \textbf{75.24} &
  \textbf{78.97} \\
   \cmidrule{2-10}
 &
  \multirow{4}{*}{GEN+ORI} &
  Dolly-v2 &
  83.10 &
  \textbf{90.56} &
  79.52 &
  85.19 &
  84.03 &
  80.95 &
  78.37 \\
 &
   &
  StableVicuna &
  82.02 &
  83.26 &
  80.72 &
  83.84 &
  86.31 &
  \textbf{82.22} &
  75.79 \\
 &
   &
  ChatGPT &
  83.22 &
  85.84 &
  80.72 &
  \textbf{87.38} &
  85.93 &
  80.95 &
  78.50 \\
 &
   &
  GPT-4 &
  \textbf{83.52} &
  85.41 &
  \textbf{81.93} &
  85.92 &
  \textbf{86.69} &
  80.63 &
  \textbf{80.56}\\
  \bottomrule
\end{tabular}}
\caption{Accuracy on XWinograd with different LLM-generated English data.
}
 \label{tab:xwino}
\end{table*}

\begin{table*}[!ht]
\centering
\vspace{1ex}
\scalebox{0.72}{
\addtolength{\tabcolsep}{-2pt}
\begin{tabular}{lll|rrrrrrrrrrrr}
\toprule

\textbf{Fine-tuned} &
  \textbf{Training data} &
  \textbf{LLM} &
  \textbf{AVG} &
  \textbf{EN} &
  \textbf{RU} &
  \textbf{ZH} &
  \textbf{ES} &
  \textbf{AR} &
  \textbf{HI} &
  \textbf{ID} &
  \textbf{TE} &
  \textbf{SW} &
  \textbf{EU} &
  \textbf{MY} \\
  \midrule
 &
   &
  Dolly-v2 &
  \textbf{68.7} &
  \textbf{78.8} &
  \textbf{71.3} &
  \textbf{73.6} &
  \textbf{74.2} &
  67.4 &
  66.9 &
  69.0 &
  \textbf{65.0} &
  60.9 &
  \textbf{66.8} &
  62.0 \\
 &
   &
  StableVicuna &
  64.6 &
  71.4 &
  66.8 &
  68.8 &
  68.1 &
  64.3 &
  63.6 &
  66.1 &
  61.2 &
  58.6 &
  63.6 &
  58.4 \\
 &
   &
  ChatGPT &
  64.3 &
  69.7 &
  66.4 &
  68.1 &
  68.0 &
  64.6 &
  64.5 &
  66.6 &
  59.8 &
  59.2 &
  62.3 &
  58.4 \\
 &
  \multirow{-4}{*}{GEN} &
  GPT-4 &
  68.0 &
  75.5 &
  70.8 &
  73.3 &
  70.4 &
  \textbf{67.6} &
  \textbf{68.2} &
  \textbf{69.6} &
  63.1 &
  \textbf{62.3} &
  65.4 &
  \textbf{62.2} \\
  \cmidrule{2-15}
 &
   &
  Dolly-v2 &
  68.1 &
  75.7 &
  71.2 &
  72.4 &
  73.2 &
  66.4 &
  67.1 &
  68.9 &
  \textbf{64.5} &
  61.4 &
  67.1 &
  61.0 \\
 &
   &
  StableVicuna &
  67.3 &
  77.0 &
  71.0 &
  70.2 &
  71.4 &
  67.2 &
  66.5 &
  68.4 &
  62.4 &
  60.5 &
  64.3 &
  61.4 \\
 &
   &
  ChatGPT &
  68.3 &
  76.4 &
  68.5 &
  72.9 &
  73.0 &
  66.3 &
  68.6 &
  71.1 &
  62.0 &
  \textbf{62.0} &
  67.4 &
  \textbf{63.4} \\
\multirow{-8}{*}{mBERT} &
  \multirow{-4}{*}{GEN+ORI} &
  GPT-4 &
  \textbf{69.8} &
  \textbf{79.5} &
  \textbf{73.1} &
  \textbf{75.3} &
  \textbf{73.4} &
  \textbf{68.1} &
  \textbf{69.8} &
  \textbf{71.9} &
  64.1 &
  \textbf{62.0} &
  \textbf{68.9} &
  61.6 \\
   \cmidrule{1-15}
 &
   &
  Dolly-v2 &
  75.8 &
  81.4 &
  79.2 &
  80.3 &
  78.0 &
  73.6 &
  74.7 &
  80.7 &
  73.0 &
  68.8 &
  72.2 &
  71.7 \\
 &
   &
  StableVicuna &
  69.6 &
  72.3 &
  71.1 &
  71.5 &
  70.4 &
  68.3 &
  70.4 &
  72.1 &
  68.4 &
  65.7 &
  68.0 &
  67.7 \\
 &
   &
  ChatGPT &
  67.4 &
  69.7 &
  68.9 &
  68.5 &
  68.7 &
  66.1 &
  68.2 &
  68.7 &
  67.0 &
  63.7 &
  65.6 &
  66.6 \\
 &
  \multirow{-4}{*}{GEN} &
  GPT-4 &
  \textbf{74.6} &
  \textbf{78.2} &
  \textbf{78.0} &
  \textbf{78.1} &
  \textbf{77.0} &
  \textbf{73.5} &
  \textbf{75.7} &
  \textbf{77.6} &
  \textbf{71.7} &
  \textbf{68.4} &
  \textbf{73.6} &
  \textbf{69.2} \\
   \cmidrule{2-15}
 &
   &
  Dolly-v2 &
  76.5 &
  81.5 &
  80.0 &
  80.5 &
  79.4 &
  75.1 &
  75.0 &
  79.6 &
  74.5 &
  71.5 &
  72.3 &
  72.6 \\
 &
   &
  StableVicuna &
  74.2 &
  79.2 &
  77.4 &
  77.8 &
  76.4 &
  74.0 &
  74.5 &
  78.2 &
  70.2 &
  67.6 &
  71.7 &
  69.6 \\
 &
   &
  ChatGPT &
  74.5 &
  78.0 &
  76.6 &
  78.8 &
  76.2 &
  72.9 &
  73.9 &
  78.9 &
  71.5 &
  69.6 &
  72.3 &
  71.0 \\
\multirow{-8}{*}{XLMR-Base} &
  \multirow{-4}{*}{GEN+ORI} &
  GPT-4 &
  \textbf{79.3} &
  \textbf{85.4} &
  \textbf{83.2} &
  \textbf{82.6} &
  \textbf{83.0} &
  \textbf{78.0} &
  \textbf{79.9} &
  \textbf{82.7} &
  \textbf{75.9} &
  \textbf{72.9} &
  \textbf{74.9} &
  \textbf{74.3} \\
   \cmidrule{1-15}
 &
   &
  Dolly-v2 &
  84.8 &
  87.4 &
  87.3 &
  87.8 &
  86.6 &
  83.0 &
  84.4 &
  87.1 &
  \textbf{84.1} &
  81.0 &
  82.9 &
  81.4 \\
 &
   &
  StableVicuna &
  74.6 &
  76.7 &
  75.9 &
  77.4 &
  76.2 &
  72.9 &
  74.5 &
  76.2 &
  74.3 &
  70.8 &
  73.5 &
  72.5 \\
 &
   &
  ChatGPT &
  77.3 &
  78.6 &
  79.9 &
  78.0 &
  77.9 &
  75.8 &
  77.4 &
  78.0 &
  76.4 &
  73.5 &
  77.1 &
  77.7 \\
 &
  \multirow{-4}{*}{GEN} &
  GPT-4 &
  \textbf{86.0} &
  \textbf{88.5} &
  \textbf{88.2} &
  \textbf{88.2} &
  \textbf{88.0} &
  \textbf{84.9} &
  \textbf{85.7} &
  \textbf{87.8} &
  83.7 &
  \textbf{81.3} &
  \textbf{85.6} &
  \textbf{84.3} \\
   \cmidrule{2-15}
 &
 
   &
  Dolly-v2 &
  86.4 &
  89.2 &
  87.2 &
  89.5 &
  87.1 &
  85.2 &
  86.7 &
  87.7 &
  \textbf{85.0} &
  83.0 &
  85.7 &
  83.8 \\
 &
   &
  StableVicuna &
  84.8 &
  88.4 &
  87.6 &
  87.8 &
  86.6 &
  82.9 &
  83.3 &
  87.4 &
  83.7 &
  81.3 &
  83.7 &
  80.0 \\
 &
   &
  ChatGPT &
  85.8 &
  88.5 &
  88.0 &
  88.3 &
  87.3 &
  83.7 &
  85.9 &
  87.2 &
  83.7 &
  81.6 &
  85.4 &
  83.8 \\
\multirow{-8}{*}{XLMR-Large} &
  \multirow{-4}{*}{GEN+ORI} &
  GPT-4 &
  \cellcolor[HTML]{FFFFFF}\textbf{88.4} &
  \textbf{92.3} &
  \textbf{91.5} &
  \textbf{91.5} &
  \textbf{90.5} &
  \textbf{86.4} &
  \textbf{88.4} &
  \textbf{91.1} &
  84.8 &
  \textbf{83.1} &
  \textbf{87.4} &
  \textbf{85.2} \\
  \bottomrule
\end{tabular}}
\caption{Accuracy on XStoryCloze with different LLM-generated English data.
}
 \label{tab:xstory}
\end{table*}

\begin{table*}[!ht]
\centering
\scalebox{0.73}{
\addtolength{\tabcolsep}{-3.5pt}
\begin{tabular}{llr|ccccccccccccc}
\toprule
\multicolumn{1}{l}{\cellcolor[HTML]{FFFFFF}\textbf{Model}} &
  \multicolumn{1}{l}{\cellcolor[HTML]{FFFFFF}\textbf{Training Data}} &
  \multicolumn{1}{r|}{\cellcolor[HTML]{FFFFFF}\textbf{|Data|}} &
  \multicolumn{1}{c}{\cellcolor[HTML]{FFFFFF}\textbf{AVG}} &
  \multicolumn{1}{c}{\cellcolor[HTML]{FFFFFF}\textbf{EN}} &
  \multicolumn{1}{c}{\cellcolor[HTML]{FFFFFF}\textbf{ET}} &
  \multicolumn{1}{c}{\cellcolor[HTML]{FFFFFF}\textbf{HT}} &
  \multicolumn{1}{c}{\cellcolor[HTML]{FFFFFF}\textbf{ID}} &
  \multicolumn{1}{c}{\cellcolor[HTML]{FFFFFF}\textbf{IT}} &
  \multicolumn{1}{c}{\cellcolor[HTML]{FFFFFF}\textbf{QU}} &
  \multicolumn{1}{c}{\cellcolor[HTML]{FFFFFF}\textbf{SW}} &
  \multicolumn{1}{c}{\cellcolor[HTML]{FFFFFF}\textbf{TA}} &
  \multicolumn{1}{c}{\cellcolor[HTML]{FFFFFF}\textbf{TH}} &
  \multicolumn{1}{c}{\cellcolor[HTML]{FFFFFF}\textbf{TR}} &
  \multicolumn{1}{c}{\cellcolor[HTML]{FFFFFF}\textbf{VI}} &
  \multicolumn{1}{c}{\cellcolor[HTML]{FFFFFF}\textbf{ZH}} \\\midrule
\cellcolor[HTML]{FFFFFF} &
 \small{$ORI$ \sc({baseline)}}  &
  400 &
  \cellcolor[HTML]{FFFFFF}47.2 &
  \cellcolor[HTML]{FFFFFF}53.8 &
  \cellcolor[HTML]{FFFFFF}44.2 &
  \cellcolor[HTML]{FFFFFF}48.6 &
  \cellcolor[HTML]{FFFFFF}47.2 &
  \cellcolor[HTML]{FFFFFF}46.2 &
  \cellcolor[HTML]{FFFFFF}50.6 &
  \cellcolor[HTML]{FFFFFF}45.4 &
  \cellcolor[HTML]{FFFFFF}48.4 &
  \cellcolor[HTML]{FFFFFF}49.8 &
  \cellcolor[HTML]{FFFFFF}49.8 &
  \cellcolor[HTML]{FFFFFF}43.6 &
  \cellcolor[HTML]{FFFFFF}47.4 \\
\cellcolor[HTML]{FFFFFF} &
  \small{$GEN_{EN}$ } &
  3.7k &
  \cellcolor[HTML]{96D5B4}56.0 &
  \cellcolor[HTML]{8ACFAA}64.8 &
  \cellcolor[HTML]{8CD0AC}54.8 &
  \cellcolor[HTML]{BCE4D1}52.6 &
  \cellcolor[HTML]{8BD0AB}58.0 &
  \cellcolor[HTML]{89CFA9}57.4 &
  \cellcolor[HTML]{FBEDEC}49.8 &
  \cellcolor[HTML]{CDEBDD}48.4 &
  \cellcolor[HTML]{9FD8BB}55.6 &
  \cellcolor[HTML]{CDEBDD}52.8 &
  \cellcolor[HTML]{C6E8D8}53.2 &
  \cellcolor[HTML]{93D3B2}53.0 &
  \cellcolor[HTML]{87CEA7}59.0 \\
\cellcolor[HTML]{FFFFFF} &
  \small{$GEN_{EN}$  + $ORI$}  &
  4.1k &
  \cellcolor[HTML]{9ED8BB}54.6 &
  \cellcolor[HTML]{A7DCC2}59.6 &
  \cellcolor[HTML]{84CCA5}56.4 &
  \cellcolor[HTML]{ABDDC5}53.6 &
  \cellcolor[HTML]{A3DABE}53.8 &
  \cellcolor[HTML]{AADDC5}51.4 &
  \cellcolor[HTML]{F2FAF6}51.4 &
  \cellcolor[HTML]{A5DBC0}51.6 &
  \cellcolor[HTML]{DEF2E8}50.4 &
  \cellcolor[HTML]{D0ECDF}52.6 &
  \cellcolor[HTML]{B9E3CF}54.0 &
  \cellcolor[HTML]{88CEA8}55.0 &
  \cellcolor[HTML]{86CDA7}59.2 \\
\cellcolor[HTML]{FFFFFF} &
  \small{$GEN_{EN}$ + $ORI$ \sc{(tlv)}}  &
  4.1k &
  \cellcolor[HTML]{8ED1AD}57.6 &
  \cellcolor[HTML]{78C89C}68.0 &
  \cellcolor[HTML]{89CFA9}55.4 &
  \cellcolor[HTML]{A9DDC4}54.0 &
  \cellcolor[HTML]{7AC89D}61.2 &
  \cellcolor[HTML]{7CC99E}59.8 &
  \cellcolor[HTML]{EBF7F2}51.8 &
  \cellcolor[HTML]{A7DCC2}51.2 &
  \cellcolor[HTML]{9ED8BB}55.8 &
  \cellcolor[HTML]{B2E0CA}54.4 &
  \cellcolor[HTML]{D7EFE4}52.2 &
  \cellcolor[HTML]{91D2B0}53.4 &
  \cellcolor[HTML]{86CDA7}59.2 \\
\cellcolor[HTML]{FFFFFF} &
  \small{$GEN_{EN}$ } &
  28.6k &
  \cellcolor[HTML]{90D2AF}57.2 &
  \cellcolor[HTML]{82CCA4}66.2 &
  \cellcolor[HTML]{87CEA8}55.8 &
  \cellcolor[HTML]{DBF1E6}50.8 &
  \cellcolor[HTML]{88CEA8}58.6 &
  \cellcolor[HTML]{85CDA6}58.2 &
  \cellcolor[HTML]{D4EEE1}53.2 &
  \cellcolor[HTML]{A7DCC2}51.2 &
  \cellcolor[HTML]{96D5B4}57.2 &
  \cellcolor[HTML]{C6E8D8}53.2 &
  \cellcolor[HTML]{DBF1E6}52.0 &
  \cellcolor[HTML]{82CCA4}56.0 &
  \cellcolor[HTML]{7CC99E}61.0 \\
\cellcolor[HTML]{FFFFFF} &
  \small{$GEN_{EN}$  + $ORI$}  &
  29k &
  \cellcolor[HTML]{91D2B0}57.0 &
  \cellcolor[HTML]{80CBA2}66.6 &
  \cellcolor[HTML]{89CFA9}55.4 &
  \cellcolor[HTML]{D0ECDF}51.4 &
  \cellcolor[HTML]{85CDA6}59.2 &
  \cellcolor[HTML]{82CCA4}58.6 &
  \cellcolor[HTML]{E1F3EB}52.4 &
  \cellcolor[HTML]{A9DDC4}50.8 &
  \cellcolor[HTML]{AADDC5}53.6 &
  \cellcolor[HTML]{C6E8D8}53.2 &
  \cellcolor[HTML]{FCFEFD}50.0 &
  \cellcolor[HTML]{89CFA9}54.8 &
  \cellcolor[HTML]{72C596}62.8 \\
\cellcolor[HTML]{FFFFFF} &
  \small{$GEN_{EN}$ + $ORI$ \sc{(tlv)}}  &
29k &
  \cellcolor[HTML]{91D2B0}57.0 &
  \cellcolor[HTML]{80CBA2}66.6 &
  \cellcolor[HTML]{89CFA9}55.4 &
  \cellcolor[HTML]{D0ECDF}51.4 &
  \cellcolor[HTML]{85CDA6}59.2 &
  \cellcolor[HTML]{82CCA4}58.6 &
  \cellcolor[HTML]{E1F3EB}52.4 &
  \cellcolor[HTML]{A9DDC4}50.8 &
  \cellcolor[HTML]{AADDC5}53.6 &
  \cellcolor[HTML]{C6E8D8}53.2 &
  \cellcolor[HTML]{FCFEFD}50.0 &
  \cellcolor[HTML]{89CFA9}54.8 &
  \cellcolor[HTML]{72C596}62.8 \\
\cellcolor[HTML]{FFFFFF} &
  \small{$GEN_{XX}$ }&
  3.6k/lang &
  \cellcolor[HTML]{8ED1AD}57.5 &
  \cellcolor[HTML]{8ACFAA}64.8 &
  \cellcolor[HTML]{7CC99E}57.8 &
  \cellcolor[HTML]{96D5B4}57.4 &
  \cellcolor[HTML]{8BD0AB}58.0 &
  \cellcolor[HTML]{7AC89D}60.2 &
  \cellcolor[HTML]{BCE4D1}54.6 &
  \cellcolor[HTML]{A6DBC1}51.4 &
  \cellcolor[HTML]{B2E0CA}53.0 &
  \cellcolor[HTML]{FFFFFF}-- &
  \cellcolor[HTML]{FFFFFF}-- &
  \cellcolor[HTML]{93D3B2}53.0 &
  \cellcolor[HTML]{76C79A}62.0 \\
\cellcolor[HTML]{FFFFFF} &
  \small{\small{$GEN_{XX}$ } + $ORI$} &
4k &
  \cellcolor[HTML]{92D3B1}56.8 &
  \cellcolor[HTML]{A7DCC2}59.6 &
  \cellcolor[HTML]{76C79A}58.8 &
  \cellcolor[HTML]{A6DBC1}54.6 &
  \cellcolor[HTML]{95D4B3}56.2 &
  \cellcolor[HTML]{74C698}61.2 &
  \cellcolor[HTML]{CDEBDD}53.6 &
  \cellcolor[HTML]{94D4B2}54.6 &
  \cellcolor[HTML]{AADDC5}53.6 &
  \cellcolor[HTML]{FFFFFF}-- &
  \cellcolor[HTML]{FFFFFF}-- &
  \cellcolor[HTML]{99D5B6}52.0 &
  \cellcolor[HTML]{80CBA2}60.2 \\
\cellcolor[HTML]{FFFFFF} &
  \small{$GEN_{EN}^{Trans} + ORI$} &
4k &
  \cellcolor[HTML]{87CEA8}58.7 &
  \cellcolor[HTML]{A7DCC2}59.6 &
  \cellcolor[HTML]{71C495}59.8 &
  \cellcolor[HTML]{89CFA9}59.8 &
  \cellcolor[HTML]{71C495}62.8 &
  \cellcolor[HTML]{75C699}61.0 &
  \cellcolor[HTML]{FFFFFF}-- &
  \cellcolor[HTML]{9FD8BB}52.6 &
  \cellcolor[HTML]{99D5B6}56.8 &
  \cellcolor[HTML]{C3E7D6}53.4 &
  \cellcolor[HTML]{A4DABF}56.2 &
  \cellcolor[HTML]{76C79A}58.2 &
  \cellcolor[HTML]{85CDA6}59.4 \\
\multirow{-11}{*}{\cellcolor[HTML]{FFFFFF}\small{mBERT}} &
  \small{$GEN_{EN}^{Trans} + ORI$} &
  29k/lang &
  \cellcolor[HTML]{7DCA9F}\textbf{60.6} &
  \cellcolor[HTML]{80CBA2}66.6 &
  \cellcolor[HTML]{6EC393}61.8 &
  \cellcolor[HTML]{94D4B2}57.8 &
  \cellcolor[HTML]{7CC99E}60.8 &
  \cellcolor[HTML]{6EC393}62.2 &
  \cellcolor[HTML]{FFFFFF}-- &
  \cellcolor[HTML]{9CD7B9}53.2 &
  \cellcolor[HTML]{90D2AF}58.4 &
  \cellcolor[HTML]{C6E8D8}53.2 &
  \cellcolor[HTML]{7ECAA0}63.0 &
  \cellcolor[HTML]{6EC393}60.6 &
  \cellcolor[HTML]{6EC393}63.8 \\\midrule
\cellcolor[HTML]{FFFFFF} &
 \small{$ORI$ \sc({baseline)}}  &
  400 &
  \cellcolor[HTML]{FFFFFF}55.6 &
  \cellcolor[HTML]{FFFFFF}57.6 &
  \cellcolor[HTML]{FFFFFF}54.6 &
  \cellcolor[HTML]{FFFFFF}50.6 &
  \cellcolor[HTML]{FFFFFF}59.6 &
  \cellcolor[HTML]{FFFFFF}54.8 &
  \cellcolor[HTML]{FFFFFF}46.0 &
  \cellcolor[HTML]{FFFFFF}55.0 &
  \cellcolor[HTML]{FFFFFF}53.4 &
  \cellcolor[HTML]{FFFFFF}56.2 &
  \cellcolor[HTML]{FFFFFF}55.2 &
  \cellcolor[HTML]{FFFFFF}54.8 &
  \cellcolor[HTML]{FFFFFF}59.6 \\
\cellcolor[HTML]{FFFFFF} &
  \small{$GEN_{EN}$ } &
  3.7k &
  \cellcolor[HTML]{CAEADA}58.8 &
  \cellcolor[HTML]{AFDFC8}62.4 &
  \cellcolor[HTML]{E1F3EB}56.4 &
  \cellcolor[HTML]{E1F3EB}52.4 &
  \cellcolor[HTML]{E1F3EB}61.4 &
  \cellcolor[HTML]{C0E6D3}58.6 &
  \cellcolor[HTML]{A5DBC0}52.2 &
  \cellcolor[HTML]{F3BEB9}52.0 &
  \cellcolor[HTML]{90D2AF}63.4 &
  \cellcolor[HTML]{ABDDC5}61.2 &
  \cellcolor[HTML]{EBF7F2}56.4 &
  \cellcolor[HTML]{AFDFC8}59.6 &
  \cellcolor[HTML]{CAEADA}62.8 \\
\cellcolor[HTML]{FFFFFF} &
  \small{$GEN_{EN}$  + $ORI$}  &
  4.1k &
  \cellcolor[HTML]{B9E3CF}59.8 &
  \cellcolor[HTML]{A5DBC0}63.8 &
  \cellcolor[HTML]{A0D9BC}61.6 &
  \cellcolor[HTML]{EFF9F4}51.6 &
  \cellcolor[HTML]{CDEBDD}62.6 &
  \cellcolor[HTML]{ABDDC5}59.8 &
  \cellcolor[HTML]{AADDC5}51.2 &
  \cellcolor[HTML]{F2BCB7}51.6 &
  \cellcolor[HTML]{A0D9BC}60.4 &
  \cellcolor[HTML]{A9DDC4}61.6 &
  \cellcolor[HTML]{A3DABE}61.8 &
  \cellcolor[HTML]{90D2AF}64.8 &
  \cellcolor[HTML]{D7EFE4}62.0 \\
\cellcolor[HTML]{FFFFFF} &
  \small{$GEN_{EN}$ + $ORI$ \sc{(tlv)}}  &
  4.1k &
  \cellcolor[HTML]{ABDDC5}60.7 &
  \cellcolor[HTML]{A8DCC3}63.2 &
  \cellcolor[HTML]{A0D9BC}61.6 &
  \cellcolor[HTML]{F2FAF6}51.4 &
  \cellcolor[HTML]{AADDC5}64.8 &
  \cellcolor[HTML]{A4DABF}61.2 &
  \cellcolor[HTML]{AADDC5}51.2 &
  \cellcolor[HTML]{F9E0DE}53.6 &
  \cellcolor[HTML]{94D4B2}62.6 &
  \cellcolor[HTML]{A2D9BD}63.0 &
  \cellcolor[HTML]{CDEBDD}58.2 &
  \cellcolor[HTML]{A5DBC0}61.0 &
  \cellcolor[HTML]{A0D9BC}66.6 \\
\cellcolor[HTML]{FFFFFF} &
  \small{$GEN_{EN}$ } &
  28.6k &
  \cellcolor[HTML]{AADDC5}60.8 &
  \cellcolor[HTML]{96D5B4}66.4 &
  \cellcolor[HTML]{D4EEE1}57.2 &
  \cellcolor[HTML]{A9DDC4}56.0 &
  \cellcolor[HTML]{A2D9BD}66.4 &
  \cellcolor[HTML]{A4DABF}61.2 &
  \cellcolor[HTML]{A0D9BC}53.0 &
  \cellcolor[HTML]{FAE5E3}53.8 &
  \cellcolor[HTML]{A3DABE}60.0 &
  \cellcolor[HTML]{A9DDC4}61.6 &
  \cellcolor[HTML]{E8F6EF}56.6 &
  \cellcolor[HTML]{A3DABE}61.4 &
  \cellcolor[HTML]{ACDEC6}64.6 \\
\cellcolor[HTML]{FFFFFF} &
  \small{$GEN_{EN}$  + $ORI$}  &
  29k &
  \cellcolor[HTML]{A3DABF}62.1 &
  \cellcolor[HTML]{A0D9BC}64.6 &
  \cellcolor[HTML]{9FD8BC}61.8 &
  \cellcolor[HTML]{FFFFFF}50.6 &
  \cellcolor[HTML]{9FD8BC}66.8 &
  \cellcolor[HTML]{96D5B4}63.6 &
  \cellcolor[HTML]{DEF2E8}48.0 &
  \cellcolor[HTML]{F5FBF9}55.6 &
  \cellcolor[HTML]{82CCA4}65.8 &
  \cellcolor[HTML]{9ED8BB}63.6 &
  \cellcolor[HTML]{DEF2E8}57.2 &
  \cellcolor[HTML]{99D5B6}63.2 &
  \cellcolor[HTML]{9FD8BC}66.8 \\
\cellcolor[HTML]{FFFFFF} &
  \small{$GEN_{EN}$ + $ORI$ \sc{(tlv)}}  &
29k &
  \cellcolor[HTML]{AADDC4}60.9 &
  \cellcolor[HTML]{96D5B4}66.4 &
  \cellcolor[HTML]{9FD8BC}61.8 &
  \cellcolor[HTML]{FFFFFF}49.8 &
  \cellcolor[HTML]{A3DABE}66.2 &
  \cellcolor[HTML]{ABDDC5}59.8 &
  \cellcolor[HTML]{98D5B5}54.6 &
  \cellcolor[HTML]{F8DCD9}53.4 &
  \cellcolor[HTML]{95D4B3}62.4 &
  \cellcolor[HTML]{9DD7BA}63.8 &
  \cellcolor[HTML]{CDEBDD}58.2 &
  \cellcolor[HTML]{9BD6B8}62.8 &
  \cellcolor[HTML]{A5DBC0}65.8 \\
\cellcolor[HTML]{FFFFFF} &
  \small{$GEN_{XX}$ }&
  3.6k/lang &
  \cellcolor[HTML]{CAEADA}58.8 &
  \cellcolor[HTML]{AFDFC8}62.4 &
  \cellcolor[HTML]{D7EFE4}57.0 &
  \cellcolor[HTML]{ABDDC5}55.6 &
  \cellcolor[HTML]{E1F3EB}61.4 &
  \cellcolor[HTML]{B9E3CF}59.0 &
  \cellcolor[HTML]{92D3B1}55.6 &
  \cellcolor[HTML]{FCF2F1}54.4 &
  \cellcolor[HTML]{C6E8D8}56.8 &
  \cellcolor[HTML]{FFFFFF}-- &
  \cellcolor[HTML]{FFFFFF}-- &
  \cellcolor[HTML]{A7DCC2}60.6 &
  \cellcolor[HTML]{D7EFE4}62.0 \\
\cellcolor[HTML]{FFFFFF} &
  \small{\small{$GEN_{XX}$ } + $ORI$} &
4k &
  \cellcolor[HTML]{B7E2CE}59.9 &
  \cellcolor[HTML]{A5DBC0}63.8 &
  \cellcolor[HTML]{A6DBC1}60.6 &
  \cellcolor[HTML]{B6E2CC}55.0 &
  \cellcolor[HTML]{ACDEC6}64.6 &
  \cellcolor[HTML]{AFDFC8}59.6 &
  \cellcolor[HTML]{A3DABE}52.6 &
  \cellcolor[HTML]{FDF6F5}54.6 &
  \cellcolor[HTML]{CDEBDD}56.4 &
  \cellcolor[HTML]{FFFFFF}-- &
  \cellcolor[HTML]{FFFFFF}-- &
  \cellcolor[HTML]{AFDFC8}59.6 &
  \cellcolor[HTML]{AADDC5}64.8 \\
\cellcolor[HTML]{FFFFFF} &
  \small{$GEN_{EN}^{Trans} + ORI$} &
4k &
  \cellcolor[HTML]{A9DCC3}61.1 &
  \cellcolor[HTML]{A5DBC0}63.8 &
  \cellcolor[HTML]{A9DDC4}60.0 &
  \cellcolor[HTML]{9ED8BB}58.0 &
  \cellcolor[HTML]{A9DDC4}65.0 &
  \cellcolor[HTML]{A6DBC1}60.8 &
  \cellcolor[HTML]{FFFFFF}-- &
  \cellcolor[HTML]{FAE5E3}53.8 &
  \cellcolor[HTML]{A2D9BD}60.2 &
  \cellcolor[HTML]{90D2AF}66.2 &
  \cellcolor[HTML]{E8F6EF}56.6 &
  \cellcolor[HTML]{9CD7B9}62.6 &
  \cellcolor[HTML]{A4DABF}66.0 \\
\multirow{-11}{*}{\cellcolor[HTML]{FFFFFF}\small{XLMR-Base}} &
  \small{$GEN_{EN}^{Trans} + ORI$} &
  29k/lang &
  \cellcolor[HTML]{A3DABE}\textbf{62.2} &
  \cellcolor[HTML]{A0D9BC}64.6 &
  \cellcolor[HTML]{98D5B5}63.2 &
  \cellcolor[HTML]{A3DABE}57.2 &
  \cellcolor[HTML]{AADDC5}64.8 &
  \cellcolor[HTML]{A4DABF}61.2 &
  \cellcolor[HTML]{FFFFFF}-- &
  \cellcolor[HTML]{FFFFFF}55.0 &
  \cellcolor[HTML]{9CD7B9}61.2 &
  \cellcolor[HTML]{CDEBDD}59.2 &
  \cellcolor[HTML]{B7E2CE}59.5 &
  \cellcolor[HTML]{93D3B1}64.2 &
  \cellcolor[HTML]{96D5B4}68.4 \\\midrule
\cellcolor[HTML]{FFFFFF} &
\small{$ORI$ \sc({baseline)}} &
  400 &
  \cellcolor[HTML]{FFFFFF}64.4 &
  \cellcolor[HTML]{FFFFFF}71.4 &
  \cellcolor[HTML]{FFFFFF}62.8 &
  \cellcolor[HTML]{FFFFFF}51.4 &
  \cellcolor[HTML]{FFFFFF}69.0 &
  \cellcolor[HTML]{FFFFFF}65.8 &
  \cellcolor[HTML]{FFFFFF}52.0 &
  \cellcolor[HTML]{FFFFFF}60.6 &
  \cellcolor[HTML]{FFFFFF}62.0 &
  \cellcolor[HTML]{FFFFFF}64.0 &
  \cellcolor[HTML]{FFFFFF}61.2 &
  \cellcolor[HTML]{FFFFFF}69.4 &
  \cellcolor[HTML]{FFFFFF}66.8 \\
\cellcolor[HTML]{FFFFFF} &
  \small{$GEN_{EN}$ } &
  3.7k &
  \cellcolor[HTML]{F2FAF6}65.2 &
  \cellcolor[HTML]{FFFFFF}71.2 &
  \cellcolor[HTML]{E1F3EB}64.6 &
  \cellcolor[HTML]{FCFEFD}51.6 &
  \cellcolor[HTML]{E1F3EB}70.8 &
  \cellcolor[HTML]{F2FAF6}66.6 &
  \cellcolor[HTML]{FBE9E7}51.0 &
  \cellcolor[HTML]{F7D8D5}58.8 &
  \cellcolor[HTML]{BCE4D1}66.0 &
  \cellcolor[HTML]{B9E3CF}68.2 &
  \cellcolor[HTML]{9CD7B9}69.0 &
  \cellcolor[HTML]{FCF2F1}68.8 &
  \cellcolor[HTML]{DEF2E8}68.8 \\
\cellcolor[HTML]{FFFFFF} &
  \small{$GEN_{EN}$  + $ORI$}  &
  4.1k &
  \cellcolor[HTML]{ABDDC5}69.5 &
  \cellcolor[HTML]{ABDDC5}76.4 &
  \cellcolor[HTML]{A0D9BC}69.8 &
  \cellcolor[HTML]{F2BDB8}48.2 &
  \cellcolor[HTML]{A0D9BC}76.0 &
  \cellcolor[HTML]{A0D9BC}72.8 &
  \cellcolor[HTML]{FAE5E3}50.8 &
  \cellcolor[HTML]{D0ECDF}63.4 &
  \cellcolor[HTML]{A7DCC2}67.8 &
  \cellcolor[HTML]{A2D9BD}70.8 &
  \cellcolor[HTML]{95D4B3}70.2 &
  \cellcolor[HTML]{BCE4D1}73.4 &
  \cellcolor[HTML]{8ACFAA}77.8 \\
\cellcolor[HTML]{FFFFFF} &
  \small{$GEN_{EN}$ + $ORI$ \sc{(tlv)}}  &
  4.1k &
  \cellcolor[HTML]{9ED8BA}71.9 &
  \cellcolor[HTML]{94D4B2}80.6 &
  \cellcolor[HTML]{96D5B4}71.6 &
  \cellcolor[HTML]{FCF2F1}50.8 &
  \cellcolor[HTML]{92D3B1}78.6 &
  \cellcolor[HTML]{88CEA8}77.2 &
  \cellcolor[HTML]{FEFAFA}51.8 &
  \cellcolor[HTML]{D7EFE4}63.0 &
  \cellcolor[HTML]{9FD8BB}69.2 &
  \cellcolor[HTML]{9FD8BB}71.2 &
  \cellcolor[HTML]{87CEA8}72.8 &
  \cellcolor[HTML]{9CD7B9}77.2 &
  \cellcolor[HTML]{85CDA6}78.8 \\
\cellcolor[HTML]{FFFFFF} &
  \small{$GEN_{EN}$ } &
  28.6k &
  \cellcolor[HTML]{9ED8BB}71.8 &
  \cellcolor[HTML]{94D4B2}80.6 &
  \cellcolor[HTML]{87CEA7}74.4 &
  \cellcolor[HTML]{FDF6F5}51.0 &
  \cellcolor[HTML]{93D3B1}78.4 &
  \cellcolor[HTML]{93D3B1}75.2 &
  \cellcolor[HTML]{FBEDEC}51.2 &
  \cellcolor[HTML]{D0ECDF}63.4 &
  \cellcolor[HTML]{9CD7B9}69.8 &
  \cellcolor[HTML]{A3DABE}70.6 &
  \cellcolor[HTML]{98D5B5}69.8 &
  \cellcolor[HTML]{A5DBC0}75.6 &
  \cellcolor[HTML]{8CD0AC}77.4 \\
\cellcolor[HTML]{FFFFFF} &
  \small{$GEN_{EN}$  + $ORI$}  &
  29k &
  \cellcolor[HTML]{9BD6B8}\textbf{72.4} &
  \cellcolor[HTML]{92D3B1}81.0 &
  \cellcolor[HTML]{8ACFAA}73.8 &
  \cellcolor[HTML]{CDEBDD}54.4 &
  \cellcolor[HTML]{89CFA9}80.2 &
  \cellcolor[HTML]{93D3B1}75.2 &
  \cellcolor[HTML]{F2BDB8}48.8 &
  \cellcolor[HTML]{F2FAF6}61.4 &
  \cellcolor[HTML]{99D5B6}70.4 &
  \cellcolor[HTML]{91D2B0}73.8 &
  \cellcolor[HTML]{94D4B2}70.4 &
  \cellcolor[HTML]{A5DBC0}75.6 &
  \cellcolor[HTML]{7FCBA1}79.8 \\
\cellcolor[HTML]{FFFFFF} &
  \small{$GEN_{EN}$ + $ORI$ \sc{(tlv)}}  &
29k &
  \cellcolor[HTML]{9BD6B8}\textbf{72.4} &
  \cellcolor[HTML]{92D3B1}81.0 &
  \cellcolor[HTML]{8ACFAA}73.8 &
  \cellcolor[HTML]{CDEBDD}54.4 &
  \cellcolor[HTML]{89CFA9}80.2 &
  \cellcolor[HTML]{93D3B1}75.2 &
  \cellcolor[HTML]{F2BDB8}48.8 &
  \cellcolor[HTML]{F9FDFB}61.0 &
  \cellcolor[HTML]{99D5B6}70.4 &
  \cellcolor[HTML]{91D2B0}73.8 &
  \cellcolor[HTML]{94D4B2}70.4 &
  \cellcolor[HTML]{A5DBC0}75.6 &
  \cellcolor[HTML]{7FCBA1}79.8 \\
\cellcolor[HTML]{FFFFFF} &
  \small{$GEN_{XX}$ }&
  3.6k/lang &
  \cellcolor[HTML]{FBE9E7}63.4 &
  \cellcolor[HTML]{FEFAFA}71.2 &
  \cellcolor[HTML]{FEFAFA}62.6 &
  \cellcolor[HTML]{D0ECDF}54.2 &
  \cellcolor[HTML]{DEF2E8}71.0 &
  \cellcolor[HTML]{FFFFFF}65.8 &
  \cellcolor[HTML]{F4C6C2}49.4 &
  \cellcolor[HTML]{F0AFA9}53.8 &
  \cellcolor[HTML]{F1B3AE}56.4 &
  \cellcolor[HTML]{FFFFFF}-- &
  \cellcolor[HTML]{FFFFFF}-- &
  \cellcolor[HTML]{F1B4AF}64.0 &
  \cellcolor[HTML]{AFDFC8}71.6 \\
\cellcolor[HTML]{FFFFFF} &
  \small{\small{$GEN_{XX}$ } + $ORI$} &
4k &
  \cellcolor[HTML]{F2FAF6}65.2 &
  \cellcolor[HTML]{ABDDC5}76.4 &
  \cellcolor[HTML]{FDF6F5}62.4 &
  \cellcolor[HTML]{C0E6D3}55.2 &
  \cellcolor[HTML]{A6DBC1}75.0 &
  \cellcolor[HTML]{F2BBB6}62.2 &
  \cellcolor[HTML]{DEF2E8}54.0 &
  \cellcolor[HTML]{F5CBC7}58.2 &
  \cellcolor[HTML]{F0B0AA}55.4 &
  \cellcolor[HTML]{FFFFFF}-- &
  \cellcolor[HTML]{FFFFFF}-- &
  \cellcolor[HTML]{F2BDB8}66.2 &
  \cellcolor[HTML]{93D3B1}76.2 \\
\cellcolor[HTML]{FFFFFF} &
  \small{$GEN_{EN}^{Trans} + ORI$} &
4k &
  \cellcolor[HTML]{D4EEE1}67.0 &
  \cellcolor[HTML]{ABDDC5}76.4 &
  \cellcolor[HTML]{F3C2BD}60.0 &
  \cellcolor[HTML]{9AD6B7}59.6 &
  \cellcolor[HTML]{F3C2BD}66.2 &
  \cellcolor[HTML]{F2FAF6}66.6 &
  \cellcolor[HTML]{FFFFFF}-- &
  \cellcolor[HTML]{F8DCD9}59.0 &
  \cellcolor[HTML]{D0ECDF}64.8 &
  \cellcolor[HTML]{9FD8BB}71.2 &
  \cellcolor[HTML]{BCE4D1}65.2 &
  \cellcolor[HTML]{A9DDC4}74.8 &
  \cellcolor[HTML]{96D5B4}75.6 \\
\multirow{-11}{*}{\cellcolor[HTML]{FFFFFF}\small{XLMR-Large}} 

&
  \small{$GEN_{EN}^{Trans} + ORI$} &
  29k/lang &
  \cellcolor[HTML]{A0D9BC}71.5 &
  \cellcolor[HTML]{92D3B1}81.0 &
  \cellcolor[HTML]{95D4B3}71.8 &
  \cellcolor[HTML]{A7DCC2}57.2 &
  \cellcolor[HTML]{8BD0AB}79.8 &
  \cellcolor[HTML]{98D5B5}74.4 &
  \cellcolor[HTML]{FFFFFF}-- &
  \cellcolor[HTML]{F0B3AD}54.8 &
  \cellcolor[HTML]{93D3B1}71.4 &
  \cellcolor[HTML]{98D5B5}72.6 &
  \cellcolor[HTML]{96D5B4}70.0 &
  \cellcolor[HTML]{9CD7B9}77.2 &
  \cellcolor[HTML]{96D5B4}75.6 \\
\bottomrule
\end{tabular}
}
\caption{Full results on XCOPA (with ChatGPT-generated data).
\textsc{+tlv} corresponds to including the original validation set in all \textsc{\textbf{t}}arget \textsc{\textbf{l}}anguages in the \textsc{\textbf{v}}alidation set.
Rows are sorted by the number of instances used in training.
\textsc{avg} shows average results for languages that are available in all settings (excl. QU, TH, TR).
}
 \label{tab:xcopa_full}
\end{table*}

\begin{table*}[t]
\centering
\scalebox{0.75}{
\addtolength{\tabcolsep}{-2pt}

\begin{tabular}{ll|ccccccccccccc}
\toprule
\textbf{Model} &
  \textbf{Training Data} &
  \multicolumn{1}{l}{\textbf{AVG}} &
  \multicolumn{1}{l}{\textbf{EN}} &
  \multicolumn{1}{l}{\textbf{ET}} &
  \multicolumn{1}{l}{\textbf{HT}} &
  \multicolumn{1}{l}{\textbf{ID}} &
  \multicolumn{1}{l}{\textbf{IT}} &
  \multicolumn{1}{l}{\textbf{QU}} &
  \multicolumn{1}{l}{\textbf{SW}} &
  \multicolumn{1}{l}{\textbf{TA}} &
  \multicolumn{1}{l}{\textbf{TH}} &
  \multicolumn{1}{l}{\textbf{TR}} &
  \multicolumn{1}{l}{\textbf{VI}} &
  \multicolumn{1}{l}{\textbf{ZH}} \\
  \midrule
\multirow{7}{*}{mBERT} &
  \small{$ORI$} &
  47.2 &
  53.8 &
  44.2 &
  48.6 &
  47.2 &
  46.2 &
  50.6 &
  45.4 &
  48.4 &
  49.8 &
  49.8 &
  43.6 &
  47.4 \\
 &
  \small{$GEN_{EN}$} &
  58.2 &
  69.2 &
  59.2 &
  54.0 &
  60.6 &
  59.2 &
  50.8 &
  48.2 &
  55.0 &
  48.2 &
  53.8 &
  57.6 &
  61.0 \\
 &
  \small{$GEN_{EN}+ORI$} &
  59.3 &
  72.6 &
  58.8 &
  53.0 &
  62.0 &
  61.0 &
  53.0 &
  50.0 &
  54.0 &
  48.2 &
  52.0 &
  57.6 &
  64.6 \\
 &
  \small{$GEN_{XX}$} &
  60.2 &
  69.2 &
  59.4 &
  56.2 &
  60.2 &
  63.8 &
  54.4 &
  55.2 &
   54.0 &
  \multicolumn{1}{c}{--} &
  \multicolumn{1}{c}{--} &
  61.2 &
  62.2 \\
 &
  \small{$GEN_{XX}+ORI$} &
  61.8 &
  72.6 &
  61.2 &
  58.2 &
  62.2 &
  66.4 &
  54.4 &
  57.4 &
  53.4 &
  \multicolumn{1}{c}{--} &
  \multicolumn{1}{c}{--} &
  63.0 &
  61.8 \\
 &
  \small{$GEN_{EN}^{Trans}$} &
  61.4 &
  69.2 &
  59.2 &
  56.8 &
  65.4 &
  65.2 &
  \multicolumn{1}{c}{--} &
  53.4 &
  56.8 &
  52.6 &
  59.6 &
  61.8 &
  65.0 \\
 &
  \small{$GEN_{EN}^{Trans}+ORI$} &
  62.6 &
  72.6 &
  58.6 &
  55.2 &
  65.6 &
  65.4 &
  \multicolumn{1}{c}{--} &
  53.8 &
  62.6 &
  53.2 &
  58.8 &
  64.6 &
  65.4 \\
  \midrule
\multirow{7}{*}{XLMR-Base} &
  \small{$ORI$} &
  55.6 &
  57.6 &
  54.6 &
  50.6 &
  59.6 &
  54.8 &
  46.0 &
  55.0 &
  53.4 &
  56.2 &
  55.2 &
  54.8 &
  59.6 \\
 &
  \small{$GEN_{EN}$} &
  63.6 &
  67.0 &
  62.4 &
  52.0 &
  68.6 &
  62.6 &
  51.8 &
  58.6 &
  65.4 &
  64.8 &
  63.2 &
  66.6 &
  69.6 \\
 &
  \small{$GEN_{EN}+ORI$} &
  63.6 &
  69.6 &
  63.8 &
  51.2 &
  67.2 &
  62.4 &
  52.6 &
  58.4 &
  63.8 &
  66.0 &
  64.2 &
  66.8 &
  69.4 \\
 &
  \small{$GEN_{XX}$} &
  63.2 &
  67.0 &
  60.8 &
  56.4 &
  68.6 &
  62.4 &
  57.4 &
  58.2 &
  60.2 &
  \multicolumn{1}{c}{--} &
  \multicolumn{1}{c}{--} &
  64.6 &
  70.4 \\
 &
  \small{$GEN_{XX}+ORI$} &
  64.0 &
  69.6 &
  62.2 &
  56.2 &
  68.6 &
  63.8 &
  56.8 &
  57.8 &
  61.2 &
  \multicolumn{1}{c}{--} &
  \multicolumn{1}{c}{--} &
  66.8 &
  70.0 \\
 &
  \small{$GEN_{EN}^{Trans}$} &
  62.5 &
  67.0 &
  60.0 &
  55.6 &
  66.0 &
  62.4 &
  \multicolumn{1}{c}{--} &
  58.0 &
  60.4 &
  64.4 &
  64.6 &
  64.0 &
  68.8 \\
 &
  \small{$GEN_{EN}^{Trans}+ORI$} &
  63.9 &
  69.6 &
  61.6 &
  56.6 &
  68.4 &
  65.2 &
  \multicolumn{1}{c}{--} &
  58.2 &
  60.2 &
  68.0 &
  62.6 &
  66.0 &
  69.6 \\
  \midrule
\multirow{7}{*}{XLMR-Large} &
  \small{$ORI$} &
  64.4 &
  71.4 &
  62.8 &
  51.4 &
  69.0 &
  65.8 &
  52.0 &
  60.6 &
  62.0 &
  64.0 &
  61.2 &
  69.4 &
  66.8 \\
 &
  \small{$GEN_{EN}$} &
  73.6 &
  83.2 &
  71.2 &
  52.0 &
  81.2 &
  78.2 &
  51.0 &
  62.2 &
  76.6 &
  77.4 &
  75.0 &
  78.4 &
  79.0 \\
 &
  \small{$GEN_{EN}+ORI$} &
  73.7 &
  84.6 &
  70.4 &
  50.0 &
  80.8 &
  80.2 &
  51.8 &
  65.8 &
  72.8 &
  76.0 &
  74.8 &
  78.4 &
  80.4 \\
 &
  \small{$GEN_{XX}$} &
  72.8 &
  83.2 &
  75.2 &
  55.2 &
  78.4 &
  76.0 &
  52.4 &
  63.0 &
  68.2 &
  \multicolumn{1}{c}{--} &
  \multicolumn{1}{c}{--} &
  77.8 &
  78.6 \\
 &
  \small{$GEN_{XX}+ORI$} &
  74.6 &
  84.6 &
  77.0 &
  56.0 &
  82.2 &
  77.0 &
  56.0 &
  65.0 &
  73.8 &
  \multicolumn{1}{c}{--} &
  \multicolumn{1}{c}{--} &
  76.2 &
  80.0 \\
 &
  \small{$GEN_{EN}^{Trans}$} &
  71.0 &
  83.2 &
  72.4 &
  55.6 &
  79.4 &
  78.2 &
  \multicolumn{1}{c}{--} &
  60.6 &
  67.8 &
  77.8 &
  72.6 &
  64.0 &
  77.4 \\
 &
  \small{$GEN_{EN}^{Trans}+ORI$} &
  74.1 &
  84.6 &
  74.2 &
  57.2 &
  82.0 &
  77.4 &
  \multicolumn{1}{c}{--} &
  62.2 &
  75.0 &
  75.2 &
  72.8 &
  74.4 &
  79.6 \\
  \bottomrule
\end{tabular}
}
\caption{Accuracy on XCOPA. {\footnotesize{$GEN_{EN}$}} and {\footnotesize{$GEN_{XX}$}} represents 3.7K and 3.6K data in English and target languages generated by GPT-4.
\textsc{avg} shows average results for languages that are available in all settings (excl. QU, TH, TR).}
 \label{tab:gpt4-copa}
\end{table*}


\printbibliography[heading=bibintoc]


\end{document}